%% file: barrier_compose_aiaa.tex
\documentclass[journal]{new-aiaa}
\usepackage[utf8]{inputenc}
\usepackage{textcomp}

\usepackage{graphicx}
\usepackage{amsmath}
\usepackage[version=4]{mhchem}
\usepackage{siunitx}
\usepackage{longtable,tabularx}

\usepackage[dvipsnames]{xcolor}
\usepackage[normalem]{ulem}

\newtoggle{use_aiaa}
\toggletrue{use_aiaa}
\input{custom_preamble.tex}

\title{Composition of Safety Constraints For
Fixed-Wing Collision Avoidance Amidst Limited Communications}

\author{Eric Squires\footnote{Research Engineer, Aerospace, Transportation and Advanced Systems Laboratory of the Georgia Tech Research Institute; eric.squires@gtri.gatech.edu.} and Pietro Pierpaoli\footnote{Postdoc, Electrical and Computer Engineering; pietro.pierpaoli@gatech.edu.}}
\affil{Georgia Institute of Technology, Atlanta, GA 30318}
\author{Rohit Konda\footnote{PhD student, Electrical and Computer Engineering.}}
\affil{University of California Santa Barbara, Santa Barbara, California 93106}
\author{Samuel Coogan\footnote{Assistant Professor, School of Electrical and Computer Engineering as well as the School of Civil and Environmental Engineering; sam.coogan@gatech.edu.}}
\affil{Georgia Institute of Technology, Atlanta, GA 30318}
\author{Magnus Egerstedt\footnote{Dean, Samueli School of Engineering, magnus@uci.edu}}
\affil{University of California, Irvine 92697}

\begin{document}

\renewcommand{\onlyinsubfile}[1]{}
\renewcommand{\notinsubfile}[1]{#1}

\maketitle

\input{abstract.tex}
\input{intro.tex}
\input{background.tex}
\input{construction.tex}
\input{compose.tex}
\input{conclusion.tex}
\input{appendix.tex}

\bibliography{barrier_compose_aiaa}

\end{document}

%% file: custom_preamble.tex
\usepackage{amsthm}

\usepackage{IEEEtrantools}
\usepackage{subcaption}
\usepackage{tikz}

\newtoggle{print_figs}
\toggletrue{print_figs}

\iftoggle{use_aiaa}{
    \newcommand{\firstletter}[2]{\lettrine{#1}{#2}}
    \newcommand{\appendixSection}[1]{\section*{#1}}
    \newenvironment{appendix_env}{\section*{Appendix}}{}
}{
    \newcommand{\firstletter[2]}{\IEEEPARstart{#1}{#2}}
    \newcommand{\appendixSection}[1]{\section{#1}}
    \newenvironment{appendix_env}{
        \begin{appendices}
        \addtocontents{toc}{\protect}
        \addtocontents{toc}{\protect}
    }{
        \end{appendices}
    }
}

\usepackage{subfiles}
\newcommand{\onlyinsubfile}[1]{#1}
\newcommand{\notinsubfile}[1]{}

\newcommand{\Db}{\Delta b_{0}}
\newcommand{\Dc}{\Delta c_{0}}
\newcommand{\VW}{r}
\newcommand{\VWS}{r^2}
\newcommand{\SO}{\sin(\omega \tau + \theta_{1,0})}
\newcommand{\ST}{\sin(\omega \tau + \theta_{2,0})}
\newcommand{\CO}{\cos(\omega \tau + \theta_{1,0})}
\newcommand{\CT}{\cos(\omega \tau + \theta_{2,0})}

\newcommand{\pf}[2]{\frac{\partial #1}{\partial #2}}
\newcommand{\INF}{\inf_{\tau\in[0,\infty)}}
\newcommand{\norm}[1]{\left\lVert #1\right\rVert}
\newcommand{\set}[2]{\{#1\: : \: #2\}}
\newcommand{\R}{\mathbb{R}}
\newcommand{\ra}{\rightarrow}

\newtheorem{proposition}{Proposition}
\newtheorem{theorem}{Theorem}

\newtheorem{corollary}{Corollary}
\theoremstyle{definition}
\newtheorem{definition}{Definition}
\theoremstyle{remark}
\newtheorem{remark}{Remark}
\theoremstyle{definition}
\newtheorem{example}{Example}
\newtheorem{case}{Case}
\newtheorem{assumption}{Assumption}

%% file: abstract.tex
\begin{abstract}
This paper considers how
to ensure that a system of fixed wing Unmanned Aerial 
Vehicles (UAVs) can avoid collisions.
To do so we develop a novel method for creating a barrier function,
which is similar to a Lyapunov function and can
be used to ensure that a system can stay safe
for all future times.
After introducing the general approach,
it is shown how to ensure that collision
avoidance for two vehicles can be guaranteed for all future times.
The construction is then extended to the case of 
arbitrarily many vehicles
by addressing how to satisfy multiple safety objectives simultaneously.
We do this
while ensuring output actuator commands are within specified limits.
Because this formulation requires communication
of control values and may therefore reduce throughput of other important messages, we
then show how to reformulate the solution without this significant communication
overhead while still ensuring safety is maintained and actuator limits are respected.
We validate the
theoretical developments of this paper in the simulator SCRIMMAGE with a simulation of 20 UAVs that maintain safe
distances from each other even though their nominal paths would otherwise
cause a collision.
\end{abstract}

%% file: intro.tex
\section{Introduction}
\firstletter{A}{s}
low-cost, unmanned aerial vehicles (UAVs) find civilian uses,
the low-altitude airspace is increasingly congested, leading to
large-scale UAV operation limitations including
concerns for privacy, the environment, national security, and
safe-flight validation
\cite{prevot2016uas}.
A key challenge for safe-flight validation
in congested environments is ensuring collision avoidance
while enabling vehicles to accomplish their designed missions. Thus,
in this paper
we propose an algorithm that minimally alters a vehicle's nominal 
control input while still ensuring safe
operations.

A variety of approaches to
fixed-wing collision avoidance have been
proposed.
Partially observable Markov decision processes are used in
\cite{temizer2010collision, wolf2011aircraft} to achieve
safe flight distances.
The dynamic window approach, originally introduced in \cite{fox1997dynamic}
for static obstacles and adapted to moving obstacles in \cite{seder2007dynamic},
uses circular arcs for trajectories and limits the set of allowable
velocities to enable a quick optimization of the control input.
In \cite{lalish2008decentralized}, the authors develop a
first-order look-ahead algorithm that can be applied
to vehicles with unicycle dynamics in a decentralized way while
guaranteeing that collisions amongst $k$ vehicles are avoided.
Potential functions \cite{mastellone2008formation, rodriguez2014decentralized} have also been applied to fixed-wing collision avoidance,
where it can be shown that vehicles can safely avoid each other even when 
their sensing range is limited.
Similarly, \cite{panyakeow2010decentralized} discusses how
to combine potential functions with trajectory goals into a navigation function
in order to
provide criteria under which collision avoidance can be guaranteed.
Navigation functions have also been combined with Model Predictive Control (MPC)
by making inter-agent distance requirements implicit in the cost function \cite{di2015potential}.
MPC has additionally been applied to UAV collision avoidance for vehicles with limited sensing \cite{defoort2009motion} and communication constraints \cite{shin2009nonlinear}.
While MPC provides a flexible framework for distributed collision avoidance, its limited horizon 
can make safety guarantees difficult.
In a more general
case, the optimal control formulation in \cite{tomlin1998conflict} allows for
collision avoidance guarantees, but it is computationally intensive as it requires numerically solving the
Hamilton-Jacobi-Bellman equations over an infinite horizon.

Trajectory generation was analyzed in \cite{lai2011board}
where a nonlinear program is developed to find a safe reference trajectory constructed
from polynomials. In \cite{lin2014path} and \cite{lin2015collision}, the authors discuss trajectory generation
using a RRT with dynamics constraints provided by Dubins paths
and a waypoint generation algorithm, respectively.
Reference governors \cite{kolmanovsky2014reference},
where the input reference signal for a nominal closed
loop controller is overridden in order to ensure that safety and performance
constraints are maintained, have 
also been applied to collision avoidance in \cite{tedesco2014collision}.
In \cite{tedesco2014collision} the authors
show how to ensure collision avoidance
for a distributed set of linear systems
via a sequential mixed-integer programming
optimization. The approach considers a finite horizon
in the optimization because it is shown that 
a constant reference can then keep the system safe after
that point. Reference governors are similar
to the approach of this paper in that given
a nominal controller the approach seeks a minimal adjustment
in order to improve safety characteristics. 
However, they differ in how the minimal adjustment
occurs. A reference governors adjusts the setpoint
that a nominal system is designed to achieve.
On the other hand, the approach of this paper does not
require a reference input to the nominal
system and instead allows a nominal controller
to calculate a control input as it normally would.
The approach of this paper
then minimally adjusts the control input to ensure safety.
Finally,
in \cite{althoff2015online}, the authors also consider a trajectory based approach to avoid static obstacles. Similar to evasive maneuvers, traffic rules \cite{pallottino2007decentralized, krontiris2011using}
are a method for encoding hybrid behaviors that can include 
collision avoidance trajectories. In \cite{pallottino2007decentralized},
the authors show that a two vehicle system with
limited sensing range can avoid collisions while reaching
position goals. While in general this may result in conservative behaviors,
they demonstrate in simulation that the decentralized
algorithm continues to allow vehicles to reach their target configuration while
avoiding collisions for as many as 70 vehicles.
Reactive methods are useful because they can often be calculated online
while evasive maneuvers benefit from a lookahead into the future.
In this paper we leverage the merits of both approaches within the framework of control barrier functions.

Motivated by the importance of formal guarantees of collision avoidance that are
computationally feasible and minimally invasive we discuss in this paper how to apply barrier
functions (e.g.,
\cite{prajna2006barrier},
\cite{ames2017control}%
) to the 
UAV collision avoidance problem,
where the system is subject to actuator constraints, nonlinear dynamics, and nonlinear safety constraints.
Barrier functions are similar to Lyapunov functions
and allow for 
guarantees that a system will stay safe (i.e., vehicles will maintain safe distances from each other) for all future times.
Further, 
under some assumptions detailed in Section \ref{sec_background},
a Quadratic Program (QP) can be used to calculate a safe 
control input implied by a barrier function so that
the calculation can be done online
\cite{ames2017control}.
Given such safety guarantees,
barrier functions have been applied to a set of problems including
collision avoidance for autonomous agents (\cite{borrmann2015control,
wang2017safe}), bipedal robots (\cite{nguyen2015safety,
hsu2015control}), adaptive cruise control and lane following
(\cite{xu2017correctness, ames2017control, xu2015robustness, xu2017realizing}),
and in mobile communication networks \cite{wang2016multi}.

However, barrier functions rely on 
being able to find a function for safety set invariance to be guaranteed.
For systems like a fixed wing UAV with actuator
constraints, nonlinear dynamics, and nonlinear safety constraints, generating
such a function can be difficult.
In this respect they are similar to Lyapunov functions. They provide
guarantees when a system designer can find appropriate functions but they may be difficult
to construct.

Nevertheless, there are a variety of approaches to finding
a barrier function given a system and safety constraints.
One approach discussed for instance in
(\cite{xu2017correctness, prajna2004safety, prajna2006barrier, wang2018permissive}),
uses a sum of squares
decomposition \cite{parrilo2003semidefinite}.
In this approach an initially conservative estimate for a barrier function is
found and the associated safe set is iteratively enlarged.
Iterative approaches have also been developed 
when the system has relative degree greater than one.
The conditions for calculating
a safe control input for higher order
systems are given in \cite{nguyen2016exponential}.
In \cite{hsu2015control}, a backstepping approach
is developed that ensures a control barrier function
can be constructed and a similar approach is discussed
in \cite{xu2018constrained}. 
The approach discussed in this paper is most similar to 
\cite{gurriet2018online} where a barrier function
is formulated by calculating the distance to a backup set
after applying a backup controller. In this
paper we develop an alternative approach that does not require the specification of a backup
set.

System-specific arguments have also been applied
to the development of a barrier function.
For instance, geometric insights are exploited
in \cite{nguyen2015safety}, where the authors
develop a barrier function for precise
foot placement by ensuring that the foot
is within the intersection of two circles.
Similarly, in \cite{borrmann2015control, wang2017safe},
the authors develop a barrier function
that ensures a circle and ellipsoid, respectively, around each robot will
not overlap in order to ensure there will be no collisions
for double integrator and quadrotor robots, respectively.
Barrier functions have also been
developed for unicycle dynamics in \cite{xu2017realizing},
where the dynamics
are simplified by considering a point slightly in front
of the vehicle.

Previous work on barrier functions has shown how, given the current
state, a safe control input can be selected to ensure the system
is safe for all times. In this paper, we also ensure
system safety but do so by integrating the dynamics into the future using a known evasive maneuver
that is always available to keep the system safe. In this respect
the system is more predictable since it is known that a particular
control input will be safe. Further, we ensure that actuator
limits are respected which is a significant constraint in the case
of UAVs where the system has non-zero minimum velocity.

Aside from ensuring a barrier function constraint can be
satisfied given actuator limitations, UAV collision avoidance also motivates
the consideration of multiple safety constraints
that must be satisfied at all times. In particular,
because collision avoidance can be viewed
as a constraint for each pairwise combination of vehicles \cite{wang2016multi, glotfelter2017nonsmooth},
we briefly review how barrier functions have been applied to systems with multiple constraints.
A contract-based approach is presented in \cite{xu2017correctness}.
A sum of squares decomposition is presented in \cite{wang2018permissive}
where additional safety constraints map to additional constraints
in the optimization problem.
In \cite{xu2018constrained}, necessary and sufficient conditions 
are given for the existence of a control input that satisfies multiple
barrier function constraints. The approach generalizes to
high order and time-varying systems but requires that actuator constraints
be unbounded.
Barrier function composition has also been addressed in
\cite{xu2017correctness,wang2016multi,glotfelter2017nonsmooth}. In
\cite{xu2017correctness}, the authors partition the state space into regions
for which a single barrier function is active in each component of the
partition. In
\cite{wang2016multi} and \cite{glotfelter2017nonsmooth}
non-smooth barrier functions are discussed,
where the result allows for combining barrier functions
using boolean primitives. One drawback of the boolean
composition approaches is that it is not guaranteed that the composition of
barrier functions will result in a barrier function. 

The high level contribution of the paper is a method for constructing a barrier function
given a safety constraint and system dynamics. In particular,
after a safety engineer specifies an evasive maneuver we then show
how to construct a barrier function.
This paper makes the following technical contributions. First, it generalizes a method
discussed in \cite{borrmann2015control, wang2017safe} for constructing a barrier function that can be used to make safety
guarantees for a system. Second, it examines how to ensure that multiple
safety constraints can be satisfied simultaneously when using this constructive method. Third, it presents an
algorithm for ensuring safety in the context of multi-agent
systems that does not require communication of low level actuator commands. Fourth, it shows how to apply the above theory to a scenario
involving fixed wing UAVs where vehicles must ensure minimum separation
distances are maintained at all times.
This paper expands on the conference version 
\cite{squires2018constructive} which did not consider multiple
constraints and did not consider limited communications.
Finally,
this paper expands on the simulation study presented in \cite{squires2018constructive}
by considering a scenario with 20 vehicles to demonstrate that
all pairwise distances between vehicles
can be kept above a minimum safety distance throughout a scenario.

This paper is organized as follows. Section~\ref{sec_background} discusses
background information for barrier functions. Section~\ref{sec_construction}
discusses a general method for constructing a barrier function
and shows how to apply it to fixed
wing collision avoidance.
Section~\ref{sec_composition} generalizes the results of Section~\ref{sec_construction} by showing how to satisfy multiple constraints
simultaneously.
Section~\ref{sec_limited_communication_control} relaxes the amount
of information required to share between vehicles while still
guaranteeing safety. Section~\ref{sec_simulation} presents
a simulation verification of the approach. Section~\ref{sec_conclusion}
concludes.

%% file: background.tex
\section{BARRIER FUNCTIONS BACKGROUND}

\label{sec_background}

We summarize the necessary background for barrier functions here.
See \cite{ames2017control} for further discussion.
Consider a control affine system 
\begin{equation}
\dot{x} = f(x) + g(x)u
\label{eq_system}
\end{equation}
where $f$ and $g$ are locally Lipschitz functions, $x\in\R^n$, $u\in U\subset\R^m$,
and solutions are forward complete, meaning the system has a unique solution
for all time $t \ge 0$ given a starting condition $x(0)$.

We expand this formulation to a set of $k$ vehicles by considering each
vehicle's state $x_i$ and dynamics $\dot{x}_i = f_i(x_i) + g_i(x_i)u_i$ where
$x_i \in \R^{n_i}, u_i\in U_i\subseteq \R^{m_i}$ and $i \in \{1,\ldots,k\}$.
The overall state for the system is described by
$x = \begin{bmatrix} x_1^T & x_2^T & \cdots & x_k^T\end{bmatrix}^T\in \R^{n}$
where $n = \sum_{i=1}^k n_i$ and
$u = \begin{bmatrix} u_1^T & u_2^T & \cdots & u_k^T \end{bmatrix}^T
\in U_1 \times U_2 \times \cdots \times U_k =U\subset \R^m$,
where $m = m_1 + \cdots + m_k$.
In this case, (\ref{eq_system}) can be represented as
\[
\dot{x}
=
\begin{bmatrix} f_1(x_1) \\ f_2(x_2) \\ \vdots \\ f_k(x_k) \end{bmatrix}
+
\begin{bmatrix} g_1(x_1) & 0 & \cdots & 0 \\
                0 & g_2(x_2) & \cdots & 0 \\
                \vdots & \vdots & \ddots & \vdots \\
                0 & 0 & \cdots & g_k(x_k) \end{bmatrix}
\begin{bmatrix} u_1 \\ u_2 \\ \vdots \\ u_k \end{bmatrix}.
\]
We model the state and control input to the single vehicle by
$x_i = \begin{bmatrix} p_{i,x} & p_{i,y} & \theta_i & p_{i,z} \end{bmatrix}^T$ and
$u_i = \begin{bmatrix} v_i & \omega_i & \zeta_i \end{bmatrix}^T$, respectively.
Further, we assume in this paper
that the vehicles experience
small bank and pitch angles so that single vehicle dynamics
can be approximated by 
\begin{equation}
\dot{x}_i 
=
\begin{bmatrix}
\cos(\theta_i) & 0 & 0\\
\sin(\theta_i) & 0 & 0\\
0 & 1 & 0 \\
0 & 0 & 1 \\
\end{bmatrix}
\begin{bmatrix}
v_i \\
\omega_i \\
\zeta_i
\end{bmatrix},
\label{eq_unicycle_system}
\end{equation}
where the sets of bounded control inputs are $v_i \in [v_{min},v_{max}]$ with $v_{min} > 0$,
$|\omega_i| \le \omega_{max}$, $|\zeta_i| \le \zeta_{max}$. 
Similar approaches to modeling fixed wing aircraft can
be found in \cite{%
    lalish2008decentralized,
    rodriguez2014decentralized,
    panyakeow2010decentralized,
    di2015potential,
    pallottino2007decentralized,
    krontiris2011using}.
We also assume that each vehicle can sense the state of every other
vehicle.

Suppose the set of desired configurations is described by the superlevel set of an output function $h:\R^n\to \R$
defined on an open set $\mathcal{D}\subseteq \R^n$.
The superlevel set of $h$ is then called the safe set and is defined as
\begin{equation}
\mathcal{C} = \set{x\in\mathcal{D}}{h(x) \ge 0}.
\label{eq_safe}
\end{equation}
The objective is now to establish the condition under which the state system belongs to $\mathcal{C}$ for all times.

\begin{definition}\cite{ames2017control}
Given a set $\mathcal{C}\subset \R^n$ defined in
(\ref{eq_safe}) for a continuously
differentiable function $h:\R^n\ra\R$, the function
$h$ is called a \emph{zeroing control barrier function (ZCBF)}
defined on an open set $\mathcal{D}$ with $\mathcal{C}\subset \mathcal{D}\subset \R^n$,
if there exists a Lipschitz continuous extended class $\mathcal{K}$ function $\alpha$
such that
\begin{equation}
\sup_{u\in U}[L_fh(x) + L_gh(x)u + \alpha(h(x))] \ge 0, \:\forall x \in \mathcal{D}.
\label{eq_zcbf_condition}
\end{equation}
\label{def_zcbf}
\end{definition}
In the above definition $L_fh(x) = \pf{h(x)}{x}f(x)$ and $L_gh(x) = \pf{h(x)}{x}g(x)$
denote the Lie derivatives.
From Definition~\ref{def_zcbf}, it follows that the admissible control space is defined as
\begin{equation}
K(x) = \{u \in U\: : \: L_fh(x) + L_gh(x)u + \alpha(h(x)) \ge 0\}.
\label{eq_admissible_control_space}
\end{equation}

\begin{theorem}\cite{ames2017control}
Given a set $\mathcal{C}\subset\R^n$ defined in (\ref{eq_safe})
for a continuously differentiable function $h$, if $h$ is a ZCBF on $\mathcal{D}$,
then any Lipschitz continuous controller $u:\mathcal{D}\ra U$ such that
$u(x)\in K(x)$ will render the set $\mathcal{C}$ forward invariant.
\label{th_fwd_invariance}
\end{theorem}

In \cite{ames2017control} it is also shown  how to calculate $u(x)\in K(x)$ using a Quadratic Program (QP) to support fast, online calculations. In particular, assume there is some nominal control input $\hat{u}\in \mathbb{R}^m$ available that is designed to achieve some performance goal (e.g., path-following)
that has not necessarily been designed to satisfy safety constraints.
Additionally, we assume $U$
can be expressed as the set of all $u$ satisfying the linear inequality $Au \ge b$.
A safe control input can then be calculated using the following Quadratic Program (QP)

\begin{IEEEeqnarray}{Cl}
\label{eq_general_qp}
u^* =
\min_{u\in \R^m}
& \frac{1}{2}\norm{u - \hat{u}}^2 \IEEEyesnumber\IEEEyessubnumber\\
\text{s.t.} & L_fh(x) + L_gh(x) u + \alpha(h(x)) \ge 0 \IEEEyessubnumber\label{eq_general_bc_constraint}\\
& Au \ge b \IEEEyessubnumber\label{eq_general_actuator}.
\end{IEEEeqnarray}

Note that by property (\ref{eq_zcbf_condition}), when $h$ is a ZCBF,
(\ref{eq_general_qp}) is guaranteed to be feasible when $x\in \mathcal{D}$.

%% file: construction.tex
\section{BARRIER FUNCTION CONSTRUCTION}
\label{sec_construction}

In general, in order to apply the barrier function framework discussed in the previous section, one needs to define an appropriate barrier function $h(x)$ representative of the collision avoidance constraints the UAVs must satisfy. To this end, as shown in the following example, the design of a suitable barrier function for fixed wing vehicles is not trivial.

\subsection{Motivating Example}
\label{sec_motivating_example}

In this section we discuss some difficulties with applying barrier functions
to the fixed-wing collision avoidance problem via a concrete example.
Consider a candidate ZCBF, $h$, that encodes a collision avoidance safety constraint in a system of two vehicles with state $x = \begin{bmatrix} x_1^T & x_2^T\end{bmatrix}^T$ and
\begin{equation}
h(x) = d_{1,2}(x) - D_s^2,
\label{eq_euclidean_h}
\end{equation}
where 
\begin{equation*}
d_{1,2}(x) =
(p_{1,x} - p_{2,x})^{2}
+ (p_{1,y} - p_{2,y})^{2}
+ (p_{1,z} - p_{2,z})^{2}
\end{equation*}
is the squared distance between vehicles $1$ and $2$
and $D_s$ is a positive minimum safety distance.
To show why $h$ defined in (\ref{eq_euclidean_h}) is not a ZCBF,
we present an example where, even though the configuration of the aircraft is safe since 
$x\in \mathcal{C}$, $h(x)$ does not satisfy constraint (\ref{eq_zcbf_condition}).
Let $x_1 = \begin{bmatrix} -D_s/2 & 0 & 0 & \epsilon \end{bmatrix}^T$
and $x_2 = \begin{bmatrix} D_s/2 & 0 & \pi & -\epsilon \end{bmatrix}^T$
for some $\epsilon \ge 0$.
First, we note that for $x = \begin{bmatrix} x_1^T & x_2^T\end{bmatrix}^T \in \mathcal{C}$,
$h(x) \ge 0$. Further,
\begin{IEEEeqnarray*}{rCl}
\sup_{u\in U}[L_fh(x) + L_gh(x)u + \alpha(h(x))]  
&=&
\sup_{u\in U} \big[
    2(p_{1,x} - p_{2,x})(v_1\cos\theta_1 - v_2\cos\theta_2) \\
&& \quad \quad + \:
    2(p_{1,y} - p_{2,y})(v_1\sin\theta_1 - v_2\sin\theta_2) \\
&& \quad \quad + \:
    2(p_{1,z} - p_{2,z})(\zeta_1 - \zeta_2)
    \big]\\
&=&
\sup_{u\in U}[-2D_s(v_1 + v_2) + 2\epsilon(\zeta_1 - \zeta_2)] \\
&= &
-4D_sv_{min} + 2\epsilon\zeta_{max}.
\end{IEEEeqnarray*}
Since $v_{min} > 0$ and $D_s > 0$,
if the two vehicles' initial positions satisfy $0 \le \epsilon < 2D_sv_{min} / \zeta_{max}$ we observe that the quantity above does not satisfy constraint (\ref{eq_zcbf_condition}), i.e., 
$\sup_{u\in U}[L_fh(x) + L_gh(x)u + \alpha(h(x))] < 0$.
Therefore, we conclude that $h(x)$ defined in (\ref{eq_euclidean_h}) is not a ZCBF. The problem with this candidate ZCBF is that it does not account for the
fact that by the time the vehicles are close to colliding, it may be too late to avoid each
other due to the limited turning radius and positive minimum velocity.

\subsection{Constructing a Barrier Function via Evading Maneuvers}
\label{sec_theory}

In order to overcome the difficulties demonstrated in the example of Section \ref{sec_motivating_example},
we introduce a method to systematically construct a ZCBF from a safety constraint.
Let 
$\rho:\mathcal{D}\ra \R$ be a \emph{safety function} that represents the safety objective we want
to satisfy at all times so that $\rho(x) \ge 0$ indicates
that the system is safe. In the example
from Section \ref{sec_motivating_example} for vehicles $i$ and $j$,
\begin{equation}
\rho(x) = d_{i,j}(x) - D_s^2.
\label{eq_rho}
\end{equation}
Second, let $\gamma:\mathcal{D}\ra U$ be a \emph{nominal evading maneuver}.
Section \ref{sec_closed_form} 
discusses specific examples of $\gamma$ for the UAV collision avoidance problem.
For now, assuming $\gamma$ has
been selected, let
\begin{IEEEeqnarray}{rCl}
h(x;\rho,\gamma)
&=& \INF \rho(\hat{x}(\tau)),
\label{eq_h}
\end{IEEEeqnarray}
be a candidate ZCBF where $\hat{x}$ and $\dot{\hat{x}}$ are given by
\begin{equation}
\hat{x}(\tau) = x + \int_0^\tau \dot{\hat{x}}(\eta)d\eta,
\label{eq_h_int}
\end{equation}
\begin{equation}
\dot{\hat{x}}(\tau) = f(\hat{x}(\tau)) +
g(\hat{x}(\tau))\gamma(\hat{x}(\tau)).
\label{eq_h_dyn}
\end{equation}
For ease of notation, we will omit the time dependencies whenever
the time is clear from the context.
We assume in this paper that the solution (\ref{eq_h_int}) is well defined and contained in $\mathcal{D}$ for all $\tau \ge 0$
so that $\rho(\hat{x}(\tau))$ is well defined.
This choice of a candidate ZCBF $h$ is motivated by the fact that in (\ref{eq_h}),
$h$ measures how close the state will get to the boundary of the safe set
assuming $\gamma$ is used as the control input for all future time.

In Section \ref{sec_motivating_example} we saw that we could not use the Euclidean
distance for a ZCBF because when a candidate ZCBF $h$ is defined as in (\ref{eq_euclidean_h}),
$K(x)$ could be empty for some $x\in\mathcal{D}$.
In other words, although $x\in\mathcal{D}$ 
there was no control input available to keep the system safe.
With $h$ defined in (\ref{eq_h}), this problem is alleviated.

\begin{theorem}
Given a dynamical system (\ref{eq_system})
and a set $\mathcal{C}\subset \mathcal{D}$ defined in (\ref{eq_safe}) for
a continuously differentiable $h$ defined in (\ref{eq_h})
with a safety function $\rho$ and locally Lipschitz evading maneuver $\gamma$,
$h$ satisfies (\ref{eq_zcbf_condition}) for all $x\in \mathcal{C}$.
If in addition, $L_gh(x)$ is non-zero for all $x\in \partial \mathcal{C}$ 
and $\gamma$ maps to values in the interior of $U$,
then $h$ is a ZCBF on an open set $\mathcal{D}$ where $\mathcal{C}\subset \mathcal{D}$.
\label{th_h_zcbf} 
\end{theorem}

\begin{proof}
We start by assuming $x\in\mathcal{C}$ and show that
$h$ satisfies (\ref{eq_zcbf_condition}).
Because $x\in \mathcal{C}$, $h(x) \ge0$ so $\alpha(h(x)) \ge 0$.
Further, note that $L_fh(x) + L_gh(x)\gamma(x)$
is the derivative along the trajectory of $\hat{x}$.
In other words,
\begin{IEEEeqnarray}{rCl}
L_fh(x) + L_gh(x)\gamma(x)
&=& 
    \lim_{a\ra 0^+}
        \frac{1}{a}\left(\inf_{\tau\in[a,\infty)} \rho(\hat{x}(\tau))
        - \INF\rho(\hat{x}(\tau))\right).
\label{eq_lie}
\end{IEEEeqnarray}
Consider the term inside the parenthesis in (\ref{eq_lie}), namely
\begin{equation*}
\inf_{\tau\in[a,\infty)} \rho(\hat{x}(\tau))
    - \INF\rho(\hat{x}(\tau))
\label{eq_lie_paren}
\end{equation*}
and notice that it is the subtraction of an infimum of the same function
$\rho$ evaluated on two different intervals. Further,
note that the first interval is a subset of the second interval
since $a$
approaches $0$ from above. Thus, the term inside the parenthesis on the right
hand side of (\ref{eq_lie}) is non-negative so
$L_fh(x) + L_gh(x)\gamma(x) \ge 0$.  We can then
conclude that
$L_fh(x) + L_gh(x)\gamma(x) + \alpha(h(x)) \ge 0$
so $\gamma(x)\in K(x)$.

Now assume that 
$L_gh(x)$ is non-zero for some $x\in \partial \mathcal{C}$ 
and $\gamma$ maps to values in the interior of $U$.
We will show that there is an open set $\mathcal{D}$ that is a strict
superset of $\mathcal{C}$ for which (\ref{eq_zcbf_condition})
holds.
Let $x\in\partial\mathcal{C}$ be such that
$L_gh(x)$ is non-zero and
$B(x,\mu)$ be a ball of radius $\mu> 0$
such that for all $z\in B(x,\mu) \setminus \mathcal{C}$,
$L_gh(z)$ is non-zero.
Such a ball exists such that $B(x,\mu) \setminus \mathcal{C}$ is nonempty because $L_gh(x)$ is continuous.
Let $d(z)$ be a non-zero vector such that $d(z) + \gamma(x)\in U$ where $d(z)$ is a non-zero vector in the direction of $L_gh(z)$.
Note that such a vector exists because $\gamma$ maps to the interior of $U$. Also
note that $L_gh(z) d(z) > 0$.
Further restrict $\mu$ so that $L_gh(z) d(z) + \alpha(h(z)) \ge 0$
for all $z\in B(x;\mu) \setminus \mathcal{C}$.
Note that for similar reasons discussed earlier in the proof,
$L_fh(z) + L_gh(z)\gamma(z) \ge 0$.
Then 
\[
L_fh(z) + L_gh(z) (\gamma(z) + d(z)) + \alpha(h(z))
\ge L_gh(x) d(z) + \alpha(h(z)) \\
\ge 0.
\]
\end{proof}

\begin{figure}
\begin{center}
\def\svgwidth{0.3\textwidth}
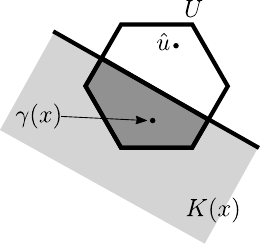
\end{center}
\caption{%
$h$
defined in (\ref{eq_h}) can be a barrier function
because $\gamma$ or a small offset from $\gamma$ is in the intersection of $U$ and $K(x)$ for $x\in\mathcal{D}$.
}
\label{fig_geometry_single}
\end{figure}

\begin{remark}
The intuitive reason why $h$ is a ZCBF is that whenever $h(x)$ is
non-negative, we have by definition a control input $\gamma$ available
to keep the system safe.
A geometric view is presented in Figure \ref{fig_geometry_single}.
Note that $\gamma$ is not
the output of the Quadratic Program (\ref{eq_general_qp}).
Instead, the role of $\gamma$ is to allow $h$ to be evaluated via (\ref{eq_h}).
\end{remark}
\begin{remark}
Theorem \ref{th_h_zcbf} holds for any class $\mathcal{K}$ function $\alpha$. 
When $\alpha(h(x)) = 0$, (\ref{eq_zcbf_condition}) becomes $\dot{h}(x) \ge 0$.
In other words, Theorem (\ref{th_h_zcbf}) can also be used to prove Lyapunov stability properties
of a set by flipping the inequality.
\end{remark}
\begin{remark}
We have found an error in Lemma 1 of the conference version of this paper \cite{squires2018constructive}
and Theorem 2 of that paper was based on that Lemma.
Therefore Theorem~\ref{th_h_zcbf} of this
paper is reformulated so as not to require that Lemma.
\end{remark}

\subsection{Deriving a Barrier Function for UAV Collision Avoidance}
\label{sec_closed_form}

We now consider how to calculate $h$ defined in (\ref{eq_h}) for the UAV collision avoidance
problem.
From Theorem \ref{th_h_zcbf} the only restriction on $\gamma$ 
and $\rho$ is
that $\gamma$ is locally Lipschitz and that $h$
is
continuously differentiable
so there is some flexibility in choosing $\gamma$ and $\rho$.
In this section we discuss two cases where we can choose $\gamma$ and $\rho$ so that $h$ can be calculated in closed form.
Let the initial state for vehicle $i$ ($i = 1,2$) be given
by $\begin{bmatrix} p_{i,x_0} & p_{i,y_0} & \theta_{i,0} & p_{i,z_0} \end{bmatrix}^T$.
For these examples we can calculate $h$ in (\ref{eq_h})
for arbitrary initial states in closed form.
Section \ref{sec_composition} generalizes the results
from Section \ref{sec_theory}
by showing how to calculate $k(k-1)/2$
barrier functions to ensure that the $k(k-1)/2$
pairwise distance constraints are always satisfied.
Because the examples in this section calculate $h$ in (\ref{eq_h}) using pairwise
distance constraints,
the calculations in these examples 
will also apply to the case of more than two vehicles.
In other words, with the result of this section
we can calculate barrier functions in closed 
form from arbitrary initial states and numbers of vehicles.
Note that the solutions in this section solve
for $h$ in (\ref{eq_h}) in closed form where $\tau$
approaches infinity.

We emphasize that the specification of an evasive maneuver
$\gamma$ is necessary to evaluate $h$ in (\ref{eq_h}). In other
words, without a safety engineer specifying $\gamma$ there cannot
be a barrier function $h$. However, $\gamma$ is never
actually directly applied to the actuators. Instead,
its role is to specify $h$ so that the final 
actuator command $u$ calculated in (\ref{eq_general_qp})
can actually be applied to the aircraft.
In this section
we give two examples where for a given $\gamma$,
$h$ can be calculated in closed form even though
it is an integration over an infinite horizon.
While we provide two examples of an evasive maneuver to calculate a continuously
differentiable $h$ from (\ref{eq_h})
in closed form, we note that it is a system specific derivation and
have not identified a general method for finding a $\gamma$
for an arbitrary system that allows $h$ to be calculated
    in closed form. However, we emphasize that any
    Lipschitz continuous $\gamma$
    resulting in an $h$ such that (\ref{eq_h})
    can be calculated in closed form and where $h$
    is continuously differentiable can be used to ensure
    safety via Theorem~\ref{th_h_zcbf}. 
    While for any such $\gamma$ the safety characteristics
    of the system are guaranteed via Theorem~\ref{th_h_zcbf},
    the performance characteristics (e.g., deviation from the nominal
    path) may be different for different evasive maneuvers.

\begin{example}
\label{ex_hturn}
In the first case,
let
\begin{equation}
\gamma_{turn} = \begin{bmatrix} \sigma v & \omega & 0 & v & \omega & 0 \end{bmatrix}^T
\label{eq_gamma_turn}
\end{equation}
with $0 < \sigma \le 1$, $\omega \ne 0$. In other words,
$\gamma_{turn}$ is defined by the same turn rate for both vehicles but possibly different translational velocities.
See Fig. \ref{fig_three_veh_a} for an example.
Define $r = \frac{v}{\omega}$ to be the turn radius of the evasive maneuver when traveling at speed $v$, $b_{1,0} = p_{1,x_0} - \sigma\VW\sin(\theta_{1,0})$,
$b_{2,0} = p_{2,x_0} - \VW\sin(\theta_{2,0})$,
$c_{1,0} = p_{1,y_0} + \sigma\VW\cos(\theta_{1,0})$,
$c_{2,0} = p_{2,y_0} + \VW\cos(\theta_{2,0})$,
$\Delta b_0 = b_{1,0} - b_{2,0}$,
$\Delta c_0 = c_{1,0} - c_{2,0}$,
and $\delta > 0$.
Let
\begin{equation}
\rho(x) = 
d_{1,2}(x) - 2\delta + \delta\sin(\theta_1) - \delta \cos(\theta_1) - D_s^2,
\label{eq_rho_adjusted}
\end{equation}
where the $\delta$ terms
are introduced to affect the smoothness of $h$. See the Appendix for details.
Then
\begin{IEEEeqnarray*}{rCCl}
h(x)
&=&
\INF&
\left(\Db + \sigma\VW\SO - \VW\ST)\right)^2
+
\left(\Dc - \sigma\VW\CO + \VW\CT\right)^2 \\
&&& +
\left(p_{1,z_0} - p_{2,z_0}\right)^2
-2\delta + \delta\sin(\omega\tau +\theta_{1,0}) - \delta\cos(\omega \tau + \theta_{1,0})- D_s^2.
\end{IEEEeqnarray*}
By expanding the square terms and applying two trigonometric identities,%
\footnote{The identities are
$\sin^2(\alpha) + \cos^2(\alpha) = 1$ and
$\cos(\alpha - \beta) = \cos(\alpha)\cos(\beta) + \sin(\alpha)\sin(\beta)$.}
we get
\begin{IEEEeqnarray}{rCCl}
h(x) &=&
\INF&
\Db^2 + \Dc^2 + (1 + \sigma^2)\VWS - 2\sigma\VWS\cos(\theta_{1,0} - \theta_{2,0})
+ 2\sigma\Db\VW\SO - 2\Db\VW\ST\nonumber\\
&&& - 2\sigma\Dc\VW\CO + 2\Dc\VW\CT \nonumber\\
&&&+ \left(p_{1,z_0} - p_{2,z_0}\right)^2 
-2\delta + \delta\sin(\omega \tau + \theta_{1,0}) - \delta\cos(\omega \tau + \theta_{1,0})- D_s^2.
\label{eq_h_turn_expanded}
\end{IEEEeqnarray}
Grouping constant terms and applying phasor addition yields
\begin{equation}
h(x)
= \INF
A_1 + A_2\cos(\omega\tau + \Theta) - D_s^2,
\label{eq_min_h_turn}
\end{equation}
where $A_1$ results from grouping constant terms, while $A_2$ and $\Theta$
are the amplitude and phase resulting from the phasor addition
so that 
$A_1$ and $A_2$ are functions of $x$.
By convention
$A_1$ and $A_2$ are nonnegative
with appropriate calculation of $\Theta$.
The minimum in
(\ref{eq_min_h_turn}) then occurs
at $\tau = (\pi - \Theta + l 2\pi) / \omega $ for
integers $l$ resulting in nonnegative $t$ so that
$h(x) = A_1 - A_2 - D_s^2$.
Note that
for the case where
\begin{equation}
\rho(x) = 
\sqrt{d_{1,2}(x) - 2\delta + \delta \sin(\theta_{1}) - \delta \cos(\theta_1)} - D_s,
\label{eq_rho_adjusted_sqrt}
\end{equation}
the same reasoning yields $h(x) = \sqrt{A_1 - A_2} - D_s$
for $\rho$ defined in (\ref{eq_rho_adjusted_sqrt}).
To ensure that the square root is well defined,
we must then require that 
$A_1 - A_2 \ge 0$ which occurs when the vehicles do not get more
than $2 \delta$ from each other along the trajectory defined by
(\ref{eq_h_int}) using $\gamma_{turn}$ in (\ref{eq_gamma_turn}).
Since $\delta$ can be chosen to be arbitrarily small,
it can be chosen so that $\delta \ll D_s$ so the vehicles
are very far outside the safe set before this condition occurs.
\end{example}

\begin{example}
\label{ex_hstraight}
For a second case, let
$\rho$ be given in (\ref{eq_rho})
and
\begin{equation}
\gamma_{straight} = \begin{bmatrix} v_1 & 0 & \zeta_1 & v_2 & 0 & \zeta_2 \end{bmatrix}^T,
\label{eq_gamma_straight}
\end{equation}
where $v_1\ne v_2$.
In other words, $\gamma_{straight}$ uses a 0 turn rate while allowing the vehicles to have different speeds.
In this case we have
\begin{IEEEeqnarray}{rCCl}
h(x) &=&
\INF& \left(p_{1, x_0} + \tau v_1\cos(\theta_{1,0}) - p_{2, x_0} - \tau v_2\cos(\theta_{2,0})\right)^2 
+
\left(p_{1,y_0} + \tau v_1\sin(\theta_{1,0}) - p_{2,y_0} - \tau v_2\sin(\theta_{2,0})\right)^2 \nonumber \\
&&& + \left(p_{1,z_0} + \tau \zeta_1 - p_{2,z_0} - \tau \zeta_2\right)^2
- D_s^2,
\label{eq_min_h_straight}
\end{IEEEeqnarray}
which is quadratic in $t$ so the minimum can be calculated in closed form.
See the Appendix for an analysis of the differentiability of $h$ in this case.
\end{example}

The evasive maneuvers in (\ref{eq_gamma_turn}) and (\ref{eq_gamma_straight}) 
(when $\zeta_1 = \zeta_2 = 0$ in (\ref{eq_gamma_straight}))
both encode
trajectories where the vehicles maintain the same altitude
for all times and therefore appear to not be exploiting an
important evasive capability of the aircraft, namely the ability
to change altitudes. However,
this is not actually the case. Although $\gamma_{turn}$ and $\gamma_{straight}$ (for $\zeta_1 = \zeta_2 = 0$)
are purely planar maneuvers, they nevertheless can induce
behaviors that exploit altitude changes. To see this,
note that for $h$ in (\ref{eq_min_h_turn}) and (\ref{eq_min_h_straight}),
\begin{equation}
\pf{h(x)}{p_{1,z_0}} = 2(p_{1,z_0} - p_{2,z_0}),
\label{eq_h_der_z}
\end{equation}
which is not equal to zero for $p_{1,z_0} \ne p_{2,z_0}$. A similar calculation also
holds for $\pf{h(x)}{p_{2,z_0}}$. In other words,
$h$ changes as a function of initial altitude.
Specifically, this means that the QP can exploit $\zeta_1$ and $\zeta_2$
because 
the fourth and eighth elements of $L_gh(x(t))$ are non-zero
when $p_{1,z_0} \ne p_{2,z_0}$,
i.e., the QP in (\ref{eq_general_qp})
can exploit the altitude control input even though $\gamma_{turn}$
and $\gamma_{straight}$ do not necessarily include an altitude
changing term in the evasive maneuver.

\subsection{Simulation of Two Vehicles}
\label{sec_simulation2veh}
We demonstrate the theoretical development of this section
in simulation using
SCRIMMAGE \cite{demarco2019simulating}.
SCRIMMAGE is a multi-agent simulator
designed to scale to high numbers of vehicles
and includes a plugin-interface that makes
it easy to experiment with different motion models
and controllers without having to change code.
This makes it simple to swap out nominal
controllers and vary the fidelity of fixed-wing
UAVs from the unicycle dynamics in (\ref{eq_unicycle_system}) used in this section up to
a 6-DOF model. 

For the simulation, let $k$ vehicles be positioned
in a circle of radius 200 around the origin, where $k= 2$ in this simulation. In other words,
vehicle $i$ has initial state
$
x_i =
\begin{bmatrix}
200 \cos\left(i \frac{2\pi}{k} + \pi\right) &
200 \sin\left(i \frac{2\pi}{k} + \pi\right) &
i \frac{2\pi}{k} + \psi &
\epsilon_i
\end{bmatrix}^T,
\label{eq_state_beg}
$ where $\psi$ is an additional offset so that vehicles are not necessarily starting with orientation pointing at the origin.
The goal position for vehicle $i$ is on the other side of the origin:
$
x_{i,g} =
\begin{bmatrix}
200 \cos\left(i \frac{2\pi}{k}\right) &
200 \sin\left(i \frac{2\pi}{k}\pi\right)
\end{bmatrix}^T.
$

This setup is selected so that the vehicles are on a collision course.
The nominal controller is that described in \cite{olfati2002near}
with constant $\lambda = 1$. Additionally, we let $v_{min}=15$ meters/second,
$v_{max} = 25$ meters/second, $\zeta_{max} = 3.9$ meters/second, $\omega_{max} = 13$ degrees/second,
$D_s = 5$ meters, and $\delta = 0.01$ meters$^2$.
The
choice of $\zeta_{max}$ results from assuming a maximum pitch
of 15 degrees while traveling at $v_{min}$.
$\omega_{max}$ is chosen to be consistent with a constant rate turn \cite{clancy1975aerodynamics} with a $30$ degree bank with a speed of $v_{max}$.
We note that while the experiments do not consider dynamics or sensor noise,
the robustness of barrier functions to noise was previously discussed in \cite{xu2015robustness}.
Each vehicle evaluates (\ref{eq_general_qp})
at each timestep where we use OSQP \cite{osqp} to evaluate
the QP. 
We investigate the performance of the vehicles
when $h$ defined in (\ref{eq_h}) is constructed from
$\gamma_{turn}$ in (\ref{eq_gamma_turn}) and $\gamma_{straight}$ (\ref{eq_gamma_straight}), respectively, where
$\gamma_{turn} = \begin{bmatrix} v & \omega & 0 & v & \omega & 0 \end{bmatrix}^T$,
$\gamma_{straight} = \begin{bmatrix}v & 0 & 0 & v & 0 & 0 \end{bmatrix}^T$,
and $v = 0.9 v_{min} + 0.1 v_{max}$ and $\omega = 0.9\omega_{max}$.
For the scenario with $\gamma_{turn}$, we let $\psi = 0$ so that the vehicles
start with orientation pointing at the origin. For the scenario
with $\gamma_{straight}$, we let $\psi = 2^\circ$
because if the vehicles pointed at the origin they would not start in the safe set.
Additionally, for the $\gamma_{turn}$ case we use $\rho$ in (\ref{eq_rho_adjusted_sqrt}).
Similarly, for the $\gamma_{straight}$ case we use $\rho(x) = \sqrt{d_{1,2}(x)} - D_s$.
Details of the distance between the vehicles and control
signals are shown in Figure \ref{fig_constraints_satisfied_2veh}.
Note that the resulting trajectory can be different depending
on which $\gamma$ is used as shown in Figure \ref{fig_2path}.
Nevertheless, in both cases the vehicles are able to maintain safe distances
from each other and satisfy actuator constraints throughout the simulation regardless of which $\gamma$
is used to construct a $h$. 

In the second experiment, we examine the effect of altitude
control on the evasive behavior of the aircraft. Because (\ref{eq_h_der_z})
predicts that $\pf{h(x)}{p_{i,z_0}}\ne 0$ (for $i=1,2$) only when the vehicles are not at
the same altitude, we start the vehicles at an altitude of
$-1$ and $1$, respectively. This offset is
small enough to ensure that the nominal path of the vehicles
still involves a collision. As was done in the previous experiment,
we set $\psi = 0^\circ$ and $\psi = 2^\circ$ degrees
when using $\gamma_{turn}$ and
$\gamma_{straight}$, respectively. In Figure 2 we show
the output of $\zeta_{1}$, where overriding behavior peaks
around 8.2 seconds. Notice that the actuator output is within
the limits of $\pm \zeta_{max}$. Further, the vehicles maintain safe
distances at all times. This occurs even though the evading maneuver does
not explicitly encode altitude changes.

\begin{figure}
\centering
\begin{subfigure}{\columnwidth}
\centering
\input{imgs/sim/2_veh1_actuator0.pgf}
\caption{}
\end{subfigure}

\begin{subfigure}{\columnwidth}
\centering
\input{imgs/sim/2_veh1_actuator1.pgf}
\caption{}
\end{subfigure}

\begin{subfigure}{\columnwidth}
\centering
\input{imgs/sim/2_veh_min_d.pgf}
\caption{}
\end{subfigure}

\begin{subfigure}{\columnwidth}
\centering
\input{imgs/sim/2_veh1_path.pgf}
\caption{}
\label{fig_2path}
\end{subfigure}

\caption[Control outputs for the 2 vehicle case]{%
2 vehicle scenario statistics.
(a) vehicle 1 velocity, (b) vehicle 1 turn rate,
(c) intervehicle distance, 
(d) vehicle 1 path.
Adapted with permission from \cite{squires2018constructive} \copyright 2018 IEEE.
}{}
\label{fig_constraints_satisfied_2veh}
\end{figure}

\begin{figure}
\centering

\begin{subfigure}{\columnwidth}
\centering
\input{imgs/sim/alt_2_veh1_actuator2.pgf}
\caption{ }
\end{subfigure}

\begin{subfigure}{\columnwidth}
\centering
\input{imgs/sim/alt_2_veh1_actuator2_zoom_in.pgf}
\caption{ }
\end{subfigure}

\caption{%
(a) $\zeta_{1}$ output for barrier functions based on
$\gamma_{turn}$ and $\gamma_{straight}$
where an override is apparent around $8.2$ seconds.
(b) A zoomed in plot of the override.
}
\label{fig_altitude_actuator}

\end{figure}

%% file: imgs/svg/geometry.pdf_tex
\begingroup%
  \makeatletter%
  \providecommand\color[2][]{%
    \errmessage{(Inkscape) Color is used for the text in Inkscape, but the package 'color.sty' is not loaded}%
    \renewcommand\color[2][]{}%
  }%
  \providecommand\transparent[1]{%
    \errmessage{(Inkscape) Transparency is used (non-zero) for the text in Inkscape, but the package 'transparent.sty' is not loaded}%
    \renewcommand\transparent[1]{}%
  }%
  \providecommand\rotatebox[2]{#2}%
  \newcommand*\fsize{\dimexpr\f@size pt\relax}%
  \newcommand*\lineheight[1]{\fontsize{\fsize}{#1\fsize}\selectfont}%
  \ifx\svgwidth\undefined%
    \setlength{\unitlength}{124.76651917bp}%
    \ifx\svgscale\undefined%
      \relax%
    \else%
      \setlength{\unitlength}{\unitlength * \real{\svgscale}}%
    \fi%
  \else%
    \setlength{\unitlength}{\svgwidth}%
  \fi%
  \global\let\svgwidth\undefined%
  \global\let\svgscale\undefined%
  \makeatother%
  \begin{picture}(1,0.93745894)%
    \lineheight{1}%
    \setlength\tabcolsep{0pt}%
    \put(0,0){\includegraphics[width=\unitlength,page=1]{imgs/svg/geometry.pdf}}%
  \end{picture}%
\endgroup%

%% file: imgs/sim/2_veh1_actuator0.pgf
\begingroup%
\makeatletter%
\begin{pgfpicture}%
\pgfpathrectangle{\pgfpointorigin}{\pgfqpoint{3.800000in}{1.600000in}}%
\pgfusepath{use as bounding box, clip}%
\begin{pgfscope}%
\pgfsetbuttcap%
\pgfsetmiterjoin%
\definecolor{currentfill}{rgb}{1.000000,1.000000,1.000000}%
\pgfsetfillcolor{currentfill}%
\pgfsetlinewidth{0.000000pt}%
\definecolor{currentstroke}{rgb}{1.000000,1.000000,1.000000}%
\pgfsetstrokecolor{currentstroke}%
\pgfsetdash{}{0pt}%
\pgfpathmoveto{\pgfqpoint{0.000000in}{0.000000in}}%
\pgfpathlineto{\pgfqpoint{3.800000in}{0.000000in}}%
\pgfpathlineto{\pgfqpoint{3.800000in}{1.600000in}}%
\pgfpathlineto{\pgfqpoint{0.000000in}{1.600000in}}%
\pgfpathclose%
\pgfusepath{fill}%
\end{pgfscope}%
\begin{pgfscope}%
\pgfsetbuttcap%
\pgfsetmiterjoin%
\definecolor{currentfill}{rgb}{1.000000,1.000000,1.000000}%
\pgfsetfillcolor{currentfill}%
\pgfsetlinewidth{0.000000pt}%
\definecolor{currentstroke}{rgb}{0.000000,0.000000,0.000000}%
\pgfsetstrokecolor{currentstroke}%
\pgfsetstrokeopacity{0.000000}%
\pgfsetdash{}{0pt}%
\pgfpathmoveto{\pgfqpoint{0.430556in}{0.415000in}}%
\pgfpathlineto{\pgfqpoint{3.800000in}{0.415000in}}%
\pgfpathlineto{\pgfqpoint{3.800000in}{1.600000in}}%
\pgfpathlineto{\pgfqpoint{0.430556in}{1.600000in}}%
\pgfpathclose%
\pgfusepath{fill}%
\end{pgfscope}%
\begin{pgfscope}%
\pgfsetbuttcap%
\pgfsetroundjoin%
\definecolor{currentfill}{rgb}{0.000000,0.000000,0.000000}%
\pgfsetfillcolor{currentfill}%
\pgfsetlinewidth{0.803000pt}%
\definecolor{currentstroke}{rgb}{0.000000,0.000000,0.000000}%
\pgfsetstrokecolor{currentstroke}%
\pgfsetdash{}{0pt}%
\pgfsys@defobject{currentmarker}{\pgfqpoint{0.000000in}{-0.048611in}}{\pgfqpoint{0.000000in}{0.000000in}}{%
\pgfpathmoveto{\pgfqpoint{0.000000in}{0.000000in}}%
\pgfpathlineto{\pgfqpoint{0.000000in}{-0.048611in}}%
\pgfusepath{stroke,fill}%
}%
\begin{pgfscope}%
\pgfsys@transformshift{0.582946in}{0.415000in}%
\pgfsys@useobject{currentmarker}{}%
\end{pgfscope}%
\end{pgfscope}%
\begin{pgfscope}%
\definecolor{textcolor}{rgb}{0.000000,0.000000,0.000000}%
\pgfsetstrokecolor{textcolor}%
\pgfsetfillcolor{textcolor}%
\pgftext[x=0.582946in,y=0.317777in,,top]{\color{textcolor}\rmfamily\fontsize{10.000000}{12.000000}\selectfont \(\displaystyle {0}\)}%
\end{pgfscope}%
\begin{pgfscope}%
\pgfsetbuttcap%
\pgfsetroundjoin%
\definecolor{currentfill}{rgb}{0.000000,0.000000,0.000000}%
\pgfsetfillcolor{currentfill}%
\pgfsetlinewidth{0.803000pt}%
\definecolor{currentstroke}{rgb}{0.000000,0.000000,0.000000}%
\pgfsetstrokecolor{currentstroke}%
\pgfsetdash{}{0pt}%
\pgfsys@defobject{currentmarker}{\pgfqpoint{0.000000in}{-0.048611in}}{\pgfqpoint{0.000000in}{0.000000in}}{%
\pgfpathmoveto{\pgfqpoint{0.000000in}{0.000000in}}%
\pgfpathlineto{\pgfqpoint{0.000000in}{-0.048611in}}%
\pgfusepath{stroke,fill}%
}%
\begin{pgfscope}%
\pgfsys@transformshift{1.349112in}{0.415000in}%
\pgfsys@useobject{currentmarker}{}%
\end{pgfscope}%
\end{pgfscope}%
\begin{pgfscope}%
\definecolor{textcolor}{rgb}{0.000000,0.000000,0.000000}%
\pgfsetstrokecolor{textcolor}%
\pgfsetfillcolor{textcolor}%
\pgftext[x=1.349112in,y=0.317777in,,top]{\color{textcolor}\rmfamily\fontsize{10.000000}{12.000000}\selectfont \(\displaystyle {5}\)}%
\end{pgfscope}%
\begin{pgfscope}%
\pgfsetbuttcap%
\pgfsetroundjoin%
\definecolor{currentfill}{rgb}{0.000000,0.000000,0.000000}%
\pgfsetfillcolor{currentfill}%
\pgfsetlinewidth{0.803000pt}%
\definecolor{currentstroke}{rgb}{0.000000,0.000000,0.000000}%
\pgfsetstrokecolor{currentstroke}%
\pgfsetdash{}{0pt}%
\pgfsys@defobject{currentmarker}{\pgfqpoint{0.000000in}{-0.048611in}}{\pgfqpoint{0.000000in}{0.000000in}}{%
\pgfpathmoveto{\pgfqpoint{0.000000in}{0.000000in}}%
\pgfpathlineto{\pgfqpoint{0.000000in}{-0.048611in}}%
\pgfusepath{stroke,fill}%
}%
\begin{pgfscope}%
\pgfsys@transformshift{2.115278in}{0.415000in}%
\pgfsys@useobject{currentmarker}{}%
\end{pgfscope}%
\end{pgfscope}%
\begin{pgfscope}%
\definecolor{textcolor}{rgb}{0.000000,0.000000,0.000000}%
\pgfsetstrokecolor{textcolor}%
\pgfsetfillcolor{textcolor}%
\pgftext[x=2.115278in,y=0.317777in,,top]{\color{textcolor}\rmfamily\fontsize{10.000000}{12.000000}\selectfont \(\displaystyle {10}\)}%
\end{pgfscope}%
\begin{pgfscope}%
\pgfsetbuttcap%
\pgfsetroundjoin%
\definecolor{currentfill}{rgb}{0.000000,0.000000,0.000000}%
\pgfsetfillcolor{currentfill}%
\pgfsetlinewidth{0.803000pt}%
\definecolor{currentstroke}{rgb}{0.000000,0.000000,0.000000}%
\pgfsetstrokecolor{currentstroke}%
\pgfsetdash{}{0pt}%
\pgfsys@defobject{currentmarker}{\pgfqpoint{0.000000in}{-0.048611in}}{\pgfqpoint{0.000000in}{0.000000in}}{%
\pgfpathmoveto{\pgfqpoint{0.000000in}{0.000000in}}%
\pgfpathlineto{\pgfqpoint{0.000000in}{-0.048611in}}%
\pgfusepath{stroke,fill}%
}%
\begin{pgfscope}%
\pgfsys@transformshift{2.881444in}{0.415000in}%
\pgfsys@useobject{currentmarker}{}%
\end{pgfscope}%
\end{pgfscope}%
\begin{pgfscope}%
\definecolor{textcolor}{rgb}{0.000000,0.000000,0.000000}%
\pgfsetstrokecolor{textcolor}%
\pgfsetfillcolor{textcolor}%
\pgftext[x=2.881444in,y=0.317777in,,top]{\color{textcolor}\rmfamily\fontsize{10.000000}{12.000000}\selectfont \(\displaystyle {15}\)}%
\end{pgfscope}%
\begin{pgfscope}%
\pgfsetbuttcap%
\pgfsetroundjoin%
\definecolor{currentfill}{rgb}{0.000000,0.000000,0.000000}%
\pgfsetfillcolor{currentfill}%
\pgfsetlinewidth{0.803000pt}%
\definecolor{currentstroke}{rgb}{0.000000,0.000000,0.000000}%
\pgfsetstrokecolor{currentstroke}%
\pgfsetdash{}{0pt}%
\pgfsys@defobject{currentmarker}{\pgfqpoint{0.000000in}{-0.048611in}}{\pgfqpoint{0.000000in}{0.000000in}}{%
\pgfpathmoveto{\pgfqpoint{0.000000in}{0.000000in}}%
\pgfpathlineto{\pgfqpoint{0.000000in}{-0.048611in}}%
\pgfusepath{stroke,fill}%
}%
\begin{pgfscope}%
\pgfsys@transformshift{3.647610in}{0.415000in}%
\pgfsys@useobject{currentmarker}{}%
\end{pgfscope}%
\end{pgfscope}%
\begin{pgfscope}%
\definecolor{textcolor}{rgb}{0.000000,0.000000,0.000000}%
\pgfsetstrokecolor{textcolor}%
\pgfsetfillcolor{textcolor}%
\pgftext[x=3.647610in,y=0.317777in,,top]{\color{textcolor}\rmfamily\fontsize{10.000000}{12.000000}\selectfont \(\displaystyle {20}\)}%
\end{pgfscope}%
\begin{pgfscope}%
\definecolor{textcolor}{rgb}{0.000000,0.000000,0.000000}%
\pgfsetstrokecolor{textcolor}%
\pgfsetfillcolor{textcolor}%
\pgftext[x=2.115278in,y=0.138889in,,top]{\color{textcolor}\rmfamily\fontsize{10.000000}{12.000000}\selectfont time (seconds)}%
\end{pgfscope}%
\begin{pgfscope}%
\pgfsetbuttcap%
\pgfsetroundjoin%
\definecolor{currentfill}{rgb}{0.000000,0.000000,0.000000}%
\pgfsetfillcolor{currentfill}%
\pgfsetlinewidth{0.803000pt}%
\definecolor{currentstroke}{rgb}{0.000000,0.000000,0.000000}%
\pgfsetstrokecolor{currentstroke}%
\pgfsetdash{}{0pt}%
\pgfsys@defobject{currentmarker}{\pgfqpoint{-0.048611in}{0.000000in}}{\pgfqpoint{-0.000000in}{0.000000in}}{%
\pgfpathmoveto{\pgfqpoint{-0.000000in}{0.000000in}}%
\pgfpathlineto{\pgfqpoint{-0.048611in}{0.000000in}}%
\pgfusepath{stroke,fill}%
}%
\begin{pgfscope}%
\pgfsys@transformshift{0.430556in}{0.468863in}%
\pgfsys@useobject{currentmarker}{}%
\end{pgfscope}%
\end{pgfscope}%
\begin{pgfscope}%
\definecolor{textcolor}{rgb}{0.000000,0.000000,0.000000}%
\pgfsetstrokecolor{textcolor}%
\pgfsetfillcolor{textcolor}%
\pgftext[x=0.194444in, y=0.420669in, left, base]{\color{textcolor}\rmfamily\fontsize{10.000000}{12.000000}\selectfont \(\displaystyle {15}\)}%
\end{pgfscope}%
\begin{pgfscope}%
\pgfsetbuttcap%
\pgfsetroundjoin%
\definecolor{currentfill}{rgb}{0.000000,0.000000,0.000000}%
\pgfsetfillcolor{currentfill}%
\pgfsetlinewidth{0.803000pt}%
\definecolor{currentstroke}{rgb}{0.000000,0.000000,0.000000}%
\pgfsetstrokecolor{currentstroke}%
\pgfsetdash{}{0pt}%
\pgfsys@defobject{currentmarker}{\pgfqpoint{-0.048611in}{0.000000in}}{\pgfqpoint{-0.000000in}{0.000000in}}{%
\pgfpathmoveto{\pgfqpoint{-0.000000in}{0.000000in}}%
\pgfpathlineto{\pgfqpoint{-0.048611in}{0.000000in}}%
\pgfusepath{stroke,fill}%
}%
\begin{pgfscope}%
\pgfsys@transformshift{0.430556in}{1.007500in}%
\pgfsys@useobject{currentmarker}{}%
\end{pgfscope}%
\end{pgfscope}%
\begin{pgfscope}%
\definecolor{textcolor}{rgb}{0.000000,0.000000,0.000000}%
\pgfsetstrokecolor{textcolor}%
\pgfsetfillcolor{textcolor}%
\pgftext[x=0.194444in, y=0.959305in, left, base]{\color{textcolor}\rmfamily\fontsize{10.000000}{12.000000}\selectfont \(\displaystyle {20}\)}%
\end{pgfscope}%
\begin{pgfscope}%
\pgfsetbuttcap%
\pgfsetroundjoin%
\definecolor{currentfill}{rgb}{0.000000,0.000000,0.000000}%
\pgfsetfillcolor{currentfill}%
\pgfsetlinewidth{0.803000pt}%
\definecolor{currentstroke}{rgb}{0.000000,0.000000,0.000000}%
\pgfsetstrokecolor{currentstroke}%
\pgfsetdash{}{0pt}%
\pgfsys@defobject{currentmarker}{\pgfqpoint{-0.048611in}{0.000000in}}{\pgfqpoint{-0.000000in}{0.000000in}}{%
\pgfpathmoveto{\pgfqpoint{-0.000000in}{0.000000in}}%
\pgfpathlineto{\pgfqpoint{-0.048611in}{0.000000in}}%
\pgfusepath{stroke,fill}%
}%
\begin{pgfscope}%
\pgfsys@transformshift{0.430556in}{1.546136in}%
\pgfsys@useobject{currentmarker}{}%
\end{pgfscope}%
\end{pgfscope}%
\begin{pgfscope}%
\definecolor{textcolor}{rgb}{0.000000,0.000000,0.000000}%
\pgfsetstrokecolor{textcolor}%
\pgfsetfillcolor{textcolor}%
\pgftext[x=0.194444in, y=1.497942in, left, base]{\color{textcolor}\rmfamily\fontsize{10.000000}{12.000000}\selectfont \(\displaystyle {25}\)}%
\end{pgfscope}%
\begin{pgfscope}%
\definecolor{textcolor}{rgb}{0.000000,0.000000,0.000000}%
\pgfsetstrokecolor{textcolor}%
\pgfsetfillcolor{textcolor}%
\pgftext[x=0.138889in,y=1.007500in,,bottom,rotate=90.000000]{\color{textcolor}\rmfamily\fontsize{10.000000}{12.000000}\selectfont \(\displaystyle v\) (m/s)}%
\end{pgfscope}%
\begin{pgfscope}%
\pgfpathrectangle{\pgfqpoint{0.430556in}{0.415000in}}{\pgfqpoint{3.369444in}{1.185000in}}%
\pgfusepath{clip}%
\pgfsetbuttcap%
\pgfsetroundjoin%
\pgfsetlinewidth{1.505625pt}%
\definecolor{currentstroke}{rgb}{0.121569,0.466667,0.705882}%
\pgfsetstrokecolor{currentstroke}%
\pgfsetdash{{5.550000pt}{2.400000pt}}{0.000000pt}%
\pgfpathmoveto{\pgfqpoint{0.583712in}{1.546136in}}%
\pgfpathlineto{\pgfqpoint{1.502345in}{1.546136in}}%
\pgfpathlineto{\pgfqpoint{1.510007in}{1.532915in}}%
\pgfpathlineto{\pgfqpoint{1.572832in}{1.425198in}}%
\pgfpathlineto{\pgfqpoint{1.636424in}{1.320579in}}%
\pgfpathlineto{\pgfqpoint{1.700016in}{1.220222in}}%
\pgfpathlineto{\pgfqpoint{1.764374in}{1.122825in}}%
\pgfpathlineto{\pgfqpoint{1.827966in}{1.030580in}}%
\pgfpathlineto{\pgfqpoint{1.886960in}{0.948829in}}%
\pgfpathlineto{\pgfqpoint{1.952085in}{0.861556in}}%
\pgfpathlineto{\pgfqpoint{2.017975in}{0.776955in}}%
\pgfpathlineto{\pgfqpoint{2.084631in}{0.694995in}}%
\pgfpathlineto{\pgfqpoint{2.151288in}{0.616526in}}%
\pgfpathlineto{\pgfqpoint{2.218710in}{0.540556in}}%
\pgfpathlineto{\pgfqpoint{2.284601in}{0.469474in}}%
\pgfpathlineto{\pgfqpoint{2.286899in}{0.468863in}}%
\pgfpathlineto{\pgfqpoint{3.646843in}{0.468863in}}%
\pgfpathlineto{\pgfqpoint{3.646843in}{0.468863in}}%
\pgfusepath{stroke}%
\end{pgfscope}%
\begin{pgfscope}%
\pgfpathrectangle{\pgfqpoint{0.430556in}{0.415000in}}{\pgfqpoint{3.369444in}{1.185000in}}%
\pgfusepath{clip}%
\pgfsetrectcap%
\pgfsetroundjoin%
\pgfsetlinewidth{1.505625pt}%
\definecolor{currentstroke}{rgb}{1.000000,0.498039,0.054902}%
\pgfsetstrokecolor{currentstroke}%
\pgfsetdash{}{0pt}%
\pgfpathmoveto{\pgfqpoint{0.583712in}{1.546136in}}%
\pgfpathlineto{\pgfqpoint{1.503111in}{1.545328in}}%
\pgfpathlineto{\pgfqpoint{1.566703in}{1.435801in}}%
\pgfpathlineto{\pgfqpoint{1.630295in}{1.330730in}}%
\pgfpathlineto{\pgfqpoint{1.680096in}{1.248079in}}%
\pgfpathlineto{\pgfqpoint{1.713807in}{1.187585in}}%
\pgfpathlineto{\pgfqpoint{1.751349in}{1.115418in}}%
\pgfpathlineto{\pgfqpoint{1.831030in}{0.959830in}}%
\pgfpathlineto{\pgfqpoint{1.836393in}{0.953262in}}%
\pgfpathlineto{\pgfqpoint{1.837160in}{0.959065in}}%
\pgfpathlineto{\pgfqpoint{1.837926in}{0.958497in}}%
\pgfpathlineto{\pgfqpoint{1.862443in}{0.942602in}}%
\pgfpathlineto{\pgfqpoint{1.888493in}{0.922996in}}%
\pgfpathlineto{\pgfqpoint{1.915309in}{0.899967in}}%
\pgfpathlineto{\pgfqpoint{1.942891in}{0.873359in}}%
\pgfpathlineto{\pgfqpoint{1.972005in}{0.842165in}}%
\pgfpathlineto{\pgfqpoint{2.002651in}{0.806007in}}%
\pgfpathlineto{\pgfqpoint{2.132133in}{0.647827in}}%
\pgfpathlineto{\pgfqpoint{2.199556in}{0.570508in}}%
\pgfpathlineto{\pgfqpoint{2.267745in}{0.495698in}}%
\pgfpathlineto{\pgfqpoint{2.293794in}{0.468863in}}%
\pgfpathlineto{\pgfqpoint{3.646843in}{0.468863in}}%
\pgfpathlineto{\pgfqpoint{3.646843in}{0.468863in}}%
\pgfusepath{stroke}%
\end{pgfscope}%
\begin{pgfscope}%
\pgfpathrectangle{\pgfqpoint{0.430556in}{0.415000in}}{\pgfqpoint{3.369444in}{1.185000in}}%
\pgfusepath{clip}%
\pgfsetrectcap%
\pgfsetroundjoin%
\pgfsetlinewidth{1.505625pt}%
\definecolor{currentstroke}{rgb}{0.000000,0.000000,0.000000}%
\pgfsetstrokecolor{currentstroke}%
\pgfsetdash{}{0pt}%
\pgfpathmoveto{\pgfqpoint{0.583712in}{0.468863in}}%
\pgfpathlineto{\pgfqpoint{3.646843in}{0.468863in}}%
\pgfpathlineto{\pgfqpoint{3.646843in}{0.468863in}}%
\pgfusepath{stroke}%
\end{pgfscope}%
\begin{pgfscope}%
\pgfpathrectangle{\pgfqpoint{0.430556in}{0.415000in}}{\pgfqpoint{3.369444in}{1.185000in}}%
\pgfusepath{clip}%
\pgfsetrectcap%
\pgfsetroundjoin%
\pgfsetlinewidth{1.505625pt}%
\definecolor{currentstroke}{rgb}{0.000000,0.000000,0.000000}%
\pgfsetstrokecolor{currentstroke}%
\pgfsetdash{}{0pt}%
\pgfpathmoveto{\pgfqpoint{0.583712in}{1.546136in}}%
\pgfpathlineto{\pgfqpoint{3.646843in}{1.546136in}}%
\pgfpathlineto{\pgfqpoint{3.646843in}{1.546136in}}%
\pgfusepath{stroke}%
\end{pgfscope}%
\begin{pgfscope}%
\pgfsetrectcap%
\pgfsetmiterjoin%
\pgfsetlinewidth{0.803000pt}%
\definecolor{currentstroke}{rgb}{0.000000,0.000000,0.000000}%
\pgfsetstrokecolor{currentstroke}%
\pgfsetdash{}{0pt}%
\pgfpathmoveto{\pgfqpoint{0.430556in}{0.415000in}}%
\pgfpathlineto{\pgfqpoint{0.430556in}{1.600000in}}%
\pgfusepath{stroke}%
\end{pgfscope}%
\begin{pgfscope}%
\pgfsetrectcap%
\pgfsetmiterjoin%
\pgfsetlinewidth{0.803000pt}%
\definecolor{currentstroke}{rgb}{0.000000,0.000000,0.000000}%
\pgfsetstrokecolor{currentstroke}%
\pgfsetdash{}{0pt}%
\pgfpathmoveto{\pgfqpoint{3.800000in}{0.415000in}}%
\pgfpathlineto{\pgfqpoint{3.800000in}{1.600000in}}%
\pgfusepath{stroke}%
\end{pgfscope}%
\begin{pgfscope}%
\pgfsetrectcap%
\pgfsetmiterjoin%
\pgfsetlinewidth{0.803000pt}%
\definecolor{currentstroke}{rgb}{0.000000,0.000000,0.000000}%
\pgfsetstrokecolor{currentstroke}%
\pgfsetdash{}{0pt}%
\pgfpathmoveto{\pgfqpoint{0.430556in}{0.415000in}}%
\pgfpathlineto{\pgfqpoint{3.800000in}{0.415000in}}%
\pgfusepath{stroke}%
\end{pgfscope}%
\begin{pgfscope}%
\pgfsetrectcap%
\pgfsetmiterjoin%
\pgfsetlinewidth{0.803000pt}%
\definecolor{currentstroke}{rgb}{0.000000,0.000000,0.000000}%
\pgfsetstrokecolor{currentstroke}%
\pgfsetdash{}{0pt}%
\pgfpathmoveto{\pgfqpoint{0.430556in}{1.600000in}}%
\pgfpathlineto{\pgfqpoint{3.800000in}{1.600000in}}%
\pgfusepath{stroke}%
\end{pgfscope}%
\begin{pgfscope}%
\pgfsetbuttcap%
\pgfsetmiterjoin%
\definecolor{currentfill}{rgb}{1.000000,1.000000,1.000000}%
\pgfsetfillcolor{currentfill}%
\pgfsetfillopacity{0.800000}%
\pgfsetlinewidth{1.003750pt}%
\definecolor{currentstroke}{rgb}{0.800000,0.800000,0.800000}%
\pgfsetstrokecolor{currentstroke}%
\pgfsetstrokeopacity{0.800000}%
\pgfsetdash{}{0pt}%
\pgfpathmoveto{\pgfqpoint{2.766351in}{1.088812in}}%
\pgfpathlineto{\pgfqpoint{3.702778in}{1.088812in}}%
\pgfpathquadraticcurveto{\pgfqpoint{3.730556in}{1.088812in}}{\pgfqpoint{3.730556in}{1.116590in}}%
\pgfpathlineto{\pgfqpoint{3.730556in}{1.502778in}}%
\pgfpathquadraticcurveto{\pgfqpoint{3.730556in}{1.530556in}}{\pgfqpoint{3.702778in}{1.530556in}}%
\pgfpathlineto{\pgfqpoint{2.766351in}{1.530556in}}%
\pgfpathquadraticcurveto{\pgfqpoint{2.738574in}{1.530556in}}{\pgfqpoint{2.738574in}{1.502778in}}%
\pgfpathlineto{\pgfqpoint{2.738574in}{1.116590in}}%
\pgfpathquadraticcurveto{\pgfqpoint{2.738574in}{1.088812in}}{\pgfqpoint{2.766351in}{1.088812in}}%
\pgfpathclose%
\pgfusepath{stroke,fill}%
\end{pgfscope}%
\begin{pgfscope}%
\pgfsetbuttcap%
\pgfsetroundjoin%
\pgfsetlinewidth{1.505625pt}%
\definecolor{currentstroke}{rgb}{0.121569,0.466667,0.705882}%
\pgfsetstrokecolor{currentstroke}%
\pgfsetdash{{5.550000pt}{2.400000pt}}{0.000000pt}%
\pgfpathmoveto{\pgfqpoint{2.794129in}{1.426389in}}%
\pgfpathlineto{\pgfqpoint{3.071907in}{1.426389in}}%
\pgfusepath{stroke}%
\end{pgfscope}%
\begin{pgfscope}%
\definecolor{textcolor}{rgb}{0.000000,0.000000,0.000000}%
\pgfsetstrokecolor{textcolor}%
\pgfsetfillcolor{textcolor}%
\pgftext[x=3.183018in,y=1.377778in,left,base]{\color{textcolor}\rmfamily\fontsize{10.000000}{12.000000}\selectfont \(\displaystyle \gamma_{straight}\)}%
\end{pgfscope}%
\begin{pgfscope}%
\pgfsetrectcap%
\pgfsetroundjoin%
\pgfsetlinewidth{1.505625pt}%
\definecolor{currentstroke}{rgb}{1.000000,0.498039,0.054902}%
\pgfsetstrokecolor{currentstroke}%
\pgfsetdash{}{0pt}%
\pgfpathmoveto{\pgfqpoint{2.794129in}{1.219985in}}%
\pgfpathlineto{\pgfqpoint{3.071907in}{1.219985in}}%
\pgfusepath{stroke}%
\end{pgfscope}%
\begin{pgfscope}%
\definecolor{textcolor}{rgb}{0.000000,0.000000,0.000000}%
\pgfsetstrokecolor{textcolor}%
\pgfsetfillcolor{textcolor}%
\pgftext[x=3.183018in,y=1.171374in,left,base]{\color{textcolor}\rmfamily\fontsize{10.000000}{12.000000}\selectfont \(\displaystyle \gamma_{turn}\)}%
\end{pgfscope}%
\end{pgfpicture}%
\makeatother%
\endgroup%

%% file: imgs/sim/2_veh1_actuator1.pgf
\begingroup%
\makeatletter%
\begin{pgfpicture}%
\pgfpathrectangle{\pgfpointorigin}{\pgfqpoint{3.800000in}{1.600000in}}%
\pgfusepath{use as bounding box, clip}%
\begin{pgfscope}%
\pgfsetbuttcap%
\pgfsetmiterjoin%
\definecolor{currentfill}{rgb}{1.000000,1.000000,1.000000}%
\pgfsetfillcolor{currentfill}%
\pgfsetlinewidth{0.000000pt}%
\definecolor{currentstroke}{rgb}{1.000000,1.000000,1.000000}%
\pgfsetstrokecolor{currentstroke}%
\pgfsetdash{}{0pt}%
\pgfpathmoveto{\pgfqpoint{0.000000in}{0.000000in}}%
\pgfpathlineto{\pgfqpoint{3.800000in}{0.000000in}}%
\pgfpathlineto{\pgfqpoint{3.800000in}{1.600000in}}%
\pgfpathlineto{\pgfqpoint{0.000000in}{1.600000in}}%
\pgfpathclose%
\pgfusepath{fill}%
\end{pgfscope}%
\begin{pgfscope}%
\pgfsetbuttcap%
\pgfsetmiterjoin%
\definecolor{currentfill}{rgb}{1.000000,1.000000,1.000000}%
\pgfsetfillcolor{currentfill}%
\pgfsetlinewidth{0.000000pt}%
\definecolor{currentstroke}{rgb}{0.000000,0.000000,0.000000}%
\pgfsetstrokecolor{currentstroke}%
\pgfsetstrokeopacity{0.000000}%
\pgfsetdash{}{0pt}%
\pgfpathmoveto{\pgfqpoint{0.538581in}{0.415000in}}%
\pgfpathlineto{\pgfqpoint{3.800000in}{0.415000in}}%
\pgfpathlineto{\pgfqpoint{3.800000in}{1.600000in}}%
\pgfpathlineto{\pgfqpoint{0.538581in}{1.600000in}}%
\pgfpathclose%
\pgfusepath{fill}%
\end{pgfscope}%
\begin{pgfscope}%
\pgfsetbuttcap%
\pgfsetroundjoin%
\definecolor{currentfill}{rgb}{0.000000,0.000000,0.000000}%
\pgfsetfillcolor{currentfill}%
\pgfsetlinewidth{0.803000pt}%
\definecolor{currentstroke}{rgb}{0.000000,0.000000,0.000000}%
\pgfsetstrokecolor{currentstroke}%
\pgfsetdash{}{0pt}%
\pgfsys@defobject{currentmarker}{\pgfqpoint{0.000000in}{-0.048611in}}{\pgfqpoint{0.000000in}{0.000000in}}{%
\pgfpathmoveto{\pgfqpoint{0.000000in}{0.000000in}}%
\pgfpathlineto{\pgfqpoint{0.000000in}{-0.048611in}}%
\pgfusepath{stroke,fill}%
}%
\begin{pgfscope}%
\pgfsys@transformshift{0.686085in}{0.415000in}%
\pgfsys@useobject{currentmarker}{}%
\end{pgfscope}%
\end{pgfscope}%
\begin{pgfscope}%
\definecolor{textcolor}{rgb}{0.000000,0.000000,0.000000}%
\pgfsetstrokecolor{textcolor}%
\pgfsetfillcolor{textcolor}%
\pgftext[x=0.686085in,y=0.317777in,,top]{\color{textcolor}\rmfamily\fontsize{10.000000}{12.000000}\selectfont \(\displaystyle {0}\)}%
\end{pgfscope}%
\begin{pgfscope}%
\pgfsetbuttcap%
\pgfsetroundjoin%
\definecolor{currentfill}{rgb}{0.000000,0.000000,0.000000}%
\pgfsetfillcolor{currentfill}%
\pgfsetlinewidth{0.803000pt}%
\definecolor{currentstroke}{rgb}{0.000000,0.000000,0.000000}%
\pgfsetstrokecolor{currentstroke}%
\pgfsetdash{}{0pt}%
\pgfsys@defobject{currentmarker}{\pgfqpoint{0.000000in}{-0.048611in}}{\pgfqpoint{0.000000in}{0.000000in}}{%
\pgfpathmoveto{\pgfqpoint{0.000000in}{0.000000in}}%
\pgfpathlineto{\pgfqpoint{0.000000in}{-0.048611in}}%
\pgfusepath{stroke,fill}%
}%
\begin{pgfscope}%
\pgfsys@transformshift{1.427688in}{0.415000in}%
\pgfsys@useobject{currentmarker}{}%
\end{pgfscope}%
\end{pgfscope}%
\begin{pgfscope}%
\definecolor{textcolor}{rgb}{0.000000,0.000000,0.000000}%
\pgfsetstrokecolor{textcolor}%
\pgfsetfillcolor{textcolor}%
\pgftext[x=1.427688in,y=0.317777in,,top]{\color{textcolor}\rmfamily\fontsize{10.000000}{12.000000}\selectfont \(\displaystyle {5}\)}%
\end{pgfscope}%
\begin{pgfscope}%
\pgfsetbuttcap%
\pgfsetroundjoin%
\definecolor{currentfill}{rgb}{0.000000,0.000000,0.000000}%
\pgfsetfillcolor{currentfill}%
\pgfsetlinewidth{0.803000pt}%
\definecolor{currentstroke}{rgb}{0.000000,0.000000,0.000000}%
\pgfsetstrokecolor{currentstroke}%
\pgfsetdash{}{0pt}%
\pgfsys@defobject{currentmarker}{\pgfqpoint{0.000000in}{-0.048611in}}{\pgfqpoint{0.000000in}{0.000000in}}{%
\pgfpathmoveto{\pgfqpoint{0.000000in}{0.000000in}}%
\pgfpathlineto{\pgfqpoint{0.000000in}{-0.048611in}}%
\pgfusepath{stroke,fill}%
}%
\begin{pgfscope}%
\pgfsys@transformshift{2.169290in}{0.415000in}%
\pgfsys@useobject{currentmarker}{}%
\end{pgfscope}%
\end{pgfscope}%
\begin{pgfscope}%
\definecolor{textcolor}{rgb}{0.000000,0.000000,0.000000}%
\pgfsetstrokecolor{textcolor}%
\pgfsetfillcolor{textcolor}%
\pgftext[x=2.169290in,y=0.317777in,,top]{\color{textcolor}\rmfamily\fontsize{10.000000}{12.000000}\selectfont \(\displaystyle {10}\)}%
\end{pgfscope}%
\begin{pgfscope}%
\pgfsetbuttcap%
\pgfsetroundjoin%
\definecolor{currentfill}{rgb}{0.000000,0.000000,0.000000}%
\pgfsetfillcolor{currentfill}%
\pgfsetlinewidth{0.803000pt}%
\definecolor{currentstroke}{rgb}{0.000000,0.000000,0.000000}%
\pgfsetstrokecolor{currentstroke}%
\pgfsetdash{}{0pt}%
\pgfsys@defobject{currentmarker}{\pgfqpoint{0.000000in}{-0.048611in}}{\pgfqpoint{0.000000in}{0.000000in}}{%
\pgfpathmoveto{\pgfqpoint{0.000000in}{0.000000in}}%
\pgfpathlineto{\pgfqpoint{0.000000in}{-0.048611in}}%
\pgfusepath{stroke,fill}%
}%
\begin{pgfscope}%
\pgfsys@transformshift{2.910893in}{0.415000in}%
\pgfsys@useobject{currentmarker}{}%
\end{pgfscope}%
\end{pgfscope}%
\begin{pgfscope}%
\definecolor{textcolor}{rgb}{0.000000,0.000000,0.000000}%
\pgfsetstrokecolor{textcolor}%
\pgfsetfillcolor{textcolor}%
\pgftext[x=2.910893in,y=0.317777in,,top]{\color{textcolor}\rmfamily\fontsize{10.000000}{12.000000}\selectfont \(\displaystyle {15}\)}%
\end{pgfscope}%
\begin{pgfscope}%
\pgfsetbuttcap%
\pgfsetroundjoin%
\definecolor{currentfill}{rgb}{0.000000,0.000000,0.000000}%
\pgfsetfillcolor{currentfill}%
\pgfsetlinewidth{0.803000pt}%
\definecolor{currentstroke}{rgb}{0.000000,0.000000,0.000000}%
\pgfsetstrokecolor{currentstroke}%
\pgfsetdash{}{0pt}%
\pgfsys@defobject{currentmarker}{\pgfqpoint{0.000000in}{-0.048611in}}{\pgfqpoint{0.000000in}{0.000000in}}{%
\pgfpathmoveto{\pgfqpoint{0.000000in}{0.000000in}}%
\pgfpathlineto{\pgfqpoint{0.000000in}{-0.048611in}}%
\pgfusepath{stroke,fill}%
}%
\begin{pgfscope}%
\pgfsys@transformshift{3.652495in}{0.415000in}%
\pgfsys@useobject{currentmarker}{}%
\end{pgfscope}%
\end{pgfscope}%
\begin{pgfscope}%
\definecolor{textcolor}{rgb}{0.000000,0.000000,0.000000}%
\pgfsetstrokecolor{textcolor}%
\pgfsetfillcolor{textcolor}%
\pgftext[x=3.652495in,y=0.317777in,,top]{\color{textcolor}\rmfamily\fontsize{10.000000}{12.000000}\selectfont \(\displaystyle {20}\)}%
\end{pgfscope}%
\begin{pgfscope}%
\definecolor{textcolor}{rgb}{0.000000,0.000000,0.000000}%
\pgfsetstrokecolor{textcolor}%
\pgfsetfillcolor{textcolor}%
\pgftext[x=2.169290in,y=0.138889in,,top]{\color{textcolor}\rmfamily\fontsize{10.000000}{12.000000}\selectfont time (seconds)}%
\end{pgfscope}%
\begin{pgfscope}%
\pgfsetbuttcap%
\pgfsetroundjoin%
\definecolor{currentfill}{rgb}{0.000000,0.000000,0.000000}%
\pgfsetfillcolor{currentfill}%
\pgfsetlinewidth{0.803000pt}%
\definecolor{currentstroke}{rgb}{0.000000,0.000000,0.000000}%
\pgfsetstrokecolor{currentstroke}%
\pgfsetdash{}{0pt}%
\pgfsys@defobject{currentmarker}{\pgfqpoint{-0.048611in}{0.000000in}}{\pgfqpoint{-0.000000in}{0.000000in}}{%
\pgfpathmoveto{\pgfqpoint{-0.000000in}{0.000000in}}%
\pgfpathlineto{\pgfqpoint{-0.048611in}{0.000000in}}%
\pgfusepath{stroke,fill}%
}%
\begin{pgfscope}%
\pgfsys@transformshift{0.538581in}{0.593164in}%
\pgfsys@useobject{currentmarker}{}%
\end{pgfscope}%
\end{pgfscope}%
\begin{pgfscope}%
\definecolor{textcolor}{rgb}{0.000000,0.000000,0.000000}%
\pgfsetstrokecolor{textcolor}%
\pgfsetfillcolor{textcolor}%
\pgftext[x=0.194444in, y=0.544970in, left, base]{\color{textcolor}\rmfamily\fontsize{10.000000}{12.000000}\selectfont \(\displaystyle {-10}\)}%
\end{pgfscope}%
\begin{pgfscope}%
\pgfsetbuttcap%
\pgfsetroundjoin%
\definecolor{currentfill}{rgb}{0.000000,0.000000,0.000000}%
\pgfsetfillcolor{currentfill}%
\pgfsetlinewidth{0.803000pt}%
\definecolor{currentstroke}{rgb}{0.000000,0.000000,0.000000}%
\pgfsetstrokecolor{currentstroke}%
\pgfsetdash{}{0pt}%
\pgfsys@defobject{currentmarker}{\pgfqpoint{-0.048611in}{0.000000in}}{\pgfqpoint{-0.000000in}{0.000000in}}{%
\pgfpathmoveto{\pgfqpoint{-0.000000in}{0.000000in}}%
\pgfpathlineto{\pgfqpoint{-0.048611in}{0.000000in}}%
\pgfusepath{stroke,fill}%
}%
\begin{pgfscope}%
\pgfsys@transformshift{0.538581in}{1.007500in}%
\pgfsys@useobject{currentmarker}{}%
\end{pgfscope}%
\end{pgfscope}%
\begin{pgfscope}%
\definecolor{textcolor}{rgb}{0.000000,0.000000,0.000000}%
\pgfsetstrokecolor{textcolor}%
\pgfsetfillcolor{textcolor}%
\pgftext[x=0.371914in, y=0.959305in, left, base]{\color{textcolor}\rmfamily\fontsize{10.000000}{12.000000}\selectfont \(\displaystyle {0}\)}%
\end{pgfscope}%
\begin{pgfscope}%
\pgfsetbuttcap%
\pgfsetroundjoin%
\definecolor{currentfill}{rgb}{0.000000,0.000000,0.000000}%
\pgfsetfillcolor{currentfill}%
\pgfsetlinewidth{0.803000pt}%
\definecolor{currentstroke}{rgb}{0.000000,0.000000,0.000000}%
\pgfsetstrokecolor{currentstroke}%
\pgfsetdash{}{0pt}%
\pgfsys@defobject{currentmarker}{\pgfqpoint{-0.048611in}{0.000000in}}{\pgfqpoint{-0.000000in}{0.000000in}}{%
\pgfpathmoveto{\pgfqpoint{-0.000000in}{0.000000in}}%
\pgfpathlineto{\pgfqpoint{-0.048611in}{0.000000in}}%
\pgfusepath{stroke,fill}%
}%
\begin{pgfscope}%
\pgfsys@transformshift{0.538581in}{1.421836in}%
\pgfsys@useobject{currentmarker}{}%
\end{pgfscope}%
\end{pgfscope}%
\begin{pgfscope}%
\definecolor{textcolor}{rgb}{0.000000,0.000000,0.000000}%
\pgfsetstrokecolor{textcolor}%
\pgfsetfillcolor{textcolor}%
\pgftext[x=0.302469in, y=1.373641in, left, base]{\color{textcolor}\rmfamily\fontsize{10.000000}{12.000000}\selectfont \(\displaystyle {10}\)}%
\end{pgfscope}%
\begin{pgfscope}%
\definecolor{textcolor}{rgb}{0.000000,0.000000,0.000000}%
\pgfsetstrokecolor{textcolor}%
\pgfsetfillcolor{textcolor}%
\pgftext[x=0.138889in,y=1.007500in,,bottom,rotate=90.000000]{\color{textcolor}\rmfamily\fontsize{10.000000}{12.000000}\selectfont \(\displaystyle \omega\) (deg/sec)}%
\end{pgfscope}%
\begin{pgfscope}%
\pgfpathrectangle{\pgfqpoint{0.538581in}{0.415000in}}{\pgfqpoint{3.261419in}{1.185000in}}%
\pgfusepath{clip}%
\pgfsetbuttcap%
\pgfsetroundjoin%
\pgfsetlinewidth{1.505625pt}%
\definecolor{currentstroke}{rgb}{0.121569,0.466667,0.705882}%
\pgfsetstrokecolor{currentstroke}%
\pgfsetdash{{5.550000pt}{2.400000pt}}{0.000000pt}%
\pgfpathmoveto{\pgfqpoint{0.686827in}{0.614840in}}%
\pgfpathlineto{\pgfqpoint{0.690535in}{0.772999in}}%
\pgfpathlineto{\pgfqpoint{0.694243in}{0.846240in}}%
\pgfpathlineto{\pgfqpoint{0.698693in}{0.893170in}}%
\pgfpathlineto{\pgfqpoint{0.703884in}{0.924167in}}%
\pgfpathlineto{\pgfqpoint{0.709817in}{0.945237in}}%
\pgfpathlineto{\pgfqpoint{0.715750in}{0.958601in}}%
\pgfpathlineto{\pgfqpoint{0.722424in}{0.968648in}}%
\pgfpathlineto{\pgfqpoint{0.730582in}{0.976873in}}%
\pgfpathlineto{\pgfqpoint{0.740222in}{0.983335in}}%
\pgfpathlineto{\pgfqpoint{0.752088in}{0.988581in}}%
\pgfpathlineto{\pgfqpoint{0.768403in}{0.993163in}}%
\pgfpathlineto{\pgfqpoint{0.790651in}{0.996914in}}%
\pgfpathlineto{\pgfqpoint{0.823282in}{1.000002in}}%
\pgfpathlineto{\pgfqpoint{0.875194in}{1.002481in}}%
\pgfpathlineto{\pgfqpoint{0.967153in}{1.004368in}}%
\pgfpathlineto{\pgfqpoint{1.154778in}{1.005624in}}%
\pgfpathlineto{\pgfqpoint{1.551535in}{1.005800in}}%
\pgfpathlineto{\pgfqpoint{1.765117in}{1.004183in}}%
\pgfpathlineto{\pgfqpoint{1.829636in}{1.001770in}}%
\pgfpathlineto{\pgfqpoint{1.857076in}{0.998801in}}%
\pgfpathlineto{\pgfqpoint{1.871908in}{0.995217in}}%
\pgfpathlineto{\pgfqpoint{1.880807in}{0.991061in}}%
\pgfpathlineto{\pgfqpoint{1.886740in}{0.986190in}}%
\pgfpathlineto{\pgfqpoint{1.891189in}{0.980029in}}%
\pgfpathlineto{\pgfqpoint{1.894897in}{0.971230in}}%
\pgfpathlineto{\pgfqpoint{1.897864in}{0.958645in}}%
\pgfpathlineto{\pgfqpoint{1.900089in}{0.941374in}}%
\pgfpathlineto{\pgfqpoint{1.902313in}{0.905077in}}%
\pgfpathlineto{\pgfqpoint{1.903797in}{0.845895in}}%
\pgfpathlineto{\pgfqpoint{1.905280in}{0.626076in}}%
\pgfpathlineto{\pgfqpoint{1.906763in}{0.468863in}}%
\pgfpathlineto{\pgfqpoint{1.911954in}{0.468863in}}%
\pgfpathlineto{\pgfqpoint{1.917146in}{0.768800in}}%
\pgfpathlineto{\pgfqpoint{1.921595in}{0.890200in}}%
\pgfpathlineto{\pgfqpoint{1.926045in}{0.949709in}}%
\pgfpathlineto{\pgfqpoint{1.930494in}{0.978954in}}%
\pgfpathlineto{\pgfqpoint{1.934944in}{0.993363in}}%
\pgfpathlineto{\pgfqpoint{1.939394in}{1.000481in}}%
\pgfpathlineto{\pgfqpoint{1.944585in}{1.004389in}}%
\pgfpathlineto{\pgfqpoint{1.951259in}{1.006402in}}%
\pgfpathlineto{\pgfqpoint{1.963866in}{1.007344in}}%
\pgfpathlineto{\pgfqpoint{2.019487in}{1.007500in}}%
\pgfpathlineto{\pgfqpoint{3.651754in}{1.007500in}}%
\pgfpathlineto{\pgfqpoint{3.651754in}{1.007500in}}%
\pgfusepath{stroke}%
\end{pgfscope}%
\begin{pgfscope}%
\pgfpathrectangle{\pgfqpoint{0.538581in}{0.415000in}}{\pgfqpoint{3.261419in}{1.185000in}}%
\pgfusepath{clip}%
\pgfsetrectcap%
\pgfsetroundjoin%
\pgfsetlinewidth{1.505625pt}%
\definecolor{currentstroke}{rgb}{1.000000,0.498039,0.054902}%
\pgfsetstrokecolor{currentstroke}%
\pgfsetdash{}{0pt}%
\pgfpathmoveto{\pgfqpoint{0.686827in}{1.007500in}}%
\pgfpathlineto{\pgfqpoint{1.705047in}{1.007500in}}%
\pgfpathlineto{\pgfqpoint{1.717654in}{1.292886in}}%
\pgfpathlineto{\pgfqpoint{1.723587in}{1.359478in}}%
\pgfpathlineto{\pgfqpoint{1.730262in}{1.410346in}}%
\pgfpathlineto{\pgfqpoint{1.736936in}{1.445507in}}%
\pgfpathlineto{\pgfqpoint{1.744352in}{1.472827in}}%
\pgfpathlineto{\pgfqpoint{1.751768in}{1.491899in}}%
\pgfpathlineto{\pgfqpoint{1.759184in}{1.505259in}}%
\pgfpathlineto{\pgfqpoint{1.766600in}{1.514473in}}%
\pgfpathlineto{\pgfqpoint{1.774758in}{1.521011in}}%
\pgfpathlineto{\pgfqpoint{1.782915in}{1.524615in}}%
\pgfpathlineto{\pgfqpoint{1.791073in}{1.525842in}}%
\pgfpathlineto{\pgfqpoint{1.799972in}{1.524873in}}%
\pgfpathlineto{\pgfqpoint{1.808872in}{1.521725in}}%
\pgfpathlineto{\pgfqpoint{1.818512in}{1.515922in}}%
\pgfpathlineto{\pgfqpoint{1.828153in}{1.507459in}}%
\pgfpathlineto{\pgfqpoint{1.837794in}{1.495838in}}%
\pgfpathlineto{\pgfqpoint{1.846693in}{1.481436in}}%
\pgfpathlineto{\pgfqpoint{1.854851in}{1.463832in}}%
\pgfpathlineto{\pgfqpoint{1.862267in}{1.442320in}}%
\pgfpathlineto{\pgfqpoint{1.868941in}{1.415867in}}%
\pgfpathlineto{\pgfqpoint{1.874874in}{1.383070in}}%
\pgfpathlineto{\pgfqpoint{1.880065in}{1.342151in}}%
\pgfpathlineto{\pgfqpoint{1.884515in}{1.291138in}}%
\pgfpathlineto{\pgfqpoint{1.888965in}{1.212559in}}%
\pgfpathlineto{\pgfqpoint{1.892673in}{1.104737in}}%
\pgfpathlineto{\pgfqpoint{1.895639in}{0.958624in}}%
\pgfpathlineto{\pgfqpoint{1.898605in}{0.683218in}}%
\pgfpathlineto{\pgfqpoint{1.900830in}{0.468863in}}%
\pgfpathlineto{\pgfqpoint{2.061016in}{0.468863in}}%
\pgfpathlineto{\pgfqpoint{2.061758in}{0.486394in}}%
\pgfpathlineto{\pgfqpoint{2.066949in}{0.757044in}}%
\pgfpathlineto{\pgfqpoint{2.071399in}{0.874530in}}%
\pgfpathlineto{\pgfqpoint{2.075848in}{0.936746in}}%
\pgfpathlineto{\pgfqpoint{2.080298in}{0.969767in}}%
\pgfpathlineto{\pgfqpoint{2.084748in}{0.987332in}}%
\pgfpathlineto{\pgfqpoint{2.089197in}{0.996696in}}%
\pgfpathlineto{\pgfqpoint{2.094388in}{1.002270in}}%
\pgfpathlineto{\pgfqpoint{2.100321in}{1.005209in}}%
\pgfpathlineto{\pgfqpoint{2.109221in}{1.006831in}}%
\pgfpathlineto{\pgfqpoint{2.128502in}{1.007452in}}%
\pgfpathlineto{\pgfqpoint{2.315386in}{1.007500in}}%
\pgfpathlineto{\pgfqpoint{3.651754in}{1.007500in}}%
\pgfpathlineto{\pgfqpoint{3.651754in}{1.007500in}}%
\pgfusepath{stroke}%
\end{pgfscope}%
\begin{pgfscope}%
\pgfpathrectangle{\pgfqpoint{0.538581in}{0.415000in}}{\pgfqpoint{3.261419in}{1.185000in}}%
\pgfusepath{clip}%
\pgfsetrectcap%
\pgfsetroundjoin%
\pgfsetlinewidth{1.505625pt}%
\definecolor{currentstroke}{rgb}{0.000000,0.000000,0.000000}%
\pgfsetstrokecolor{currentstroke}%
\pgfsetdash{}{0pt}%
\pgfpathmoveto{\pgfqpoint{0.686827in}{0.468863in}}%
\pgfpathlineto{\pgfqpoint{3.651754in}{0.468863in}}%
\pgfpathlineto{\pgfqpoint{3.651754in}{0.468863in}}%
\pgfusepath{stroke}%
\end{pgfscope}%
\begin{pgfscope}%
\pgfpathrectangle{\pgfqpoint{0.538581in}{0.415000in}}{\pgfqpoint{3.261419in}{1.185000in}}%
\pgfusepath{clip}%
\pgfsetrectcap%
\pgfsetroundjoin%
\pgfsetlinewidth{1.505625pt}%
\definecolor{currentstroke}{rgb}{0.000000,0.000000,0.000000}%
\pgfsetstrokecolor{currentstroke}%
\pgfsetdash{}{0pt}%
\pgfpathmoveto{\pgfqpoint{0.686827in}{1.546136in}}%
\pgfpathlineto{\pgfqpoint{3.651754in}{1.546136in}}%
\pgfpathlineto{\pgfqpoint{3.651754in}{1.546136in}}%
\pgfusepath{stroke}%
\end{pgfscope}%
\begin{pgfscope}%
\pgfsetrectcap%
\pgfsetmiterjoin%
\pgfsetlinewidth{0.803000pt}%
\definecolor{currentstroke}{rgb}{0.000000,0.000000,0.000000}%
\pgfsetstrokecolor{currentstroke}%
\pgfsetdash{}{0pt}%
\pgfpathmoveto{\pgfqpoint{0.538581in}{0.415000in}}%
\pgfpathlineto{\pgfqpoint{0.538581in}{1.600000in}}%
\pgfusepath{stroke}%
\end{pgfscope}%
\begin{pgfscope}%
\pgfsetrectcap%
\pgfsetmiterjoin%
\pgfsetlinewidth{0.803000pt}%
\definecolor{currentstroke}{rgb}{0.000000,0.000000,0.000000}%
\pgfsetstrokecolor{currentstroke}%
\pgfsetdash{}{0pt}%
\pgfpathmoveto{\pgfqpoint{3.800000in}{0.415000in}}%
\pgfpathlineto{\pgfqpoint{3.800000in}{1.600000in}}%
\pgfusepath{stroke}%
\end{pgfscope}%
\begin{pgfscope}%
\pgfsetrectcap%
\pgfsetmiterjoin%
\pgfsetlinewidth{0.803000pt}%
\definecolor{currentstroke}{rgb}{0.000000,0.000000,0.000000}%
\pgfsetstrokecolor{currentstroke}%
\pgfsetdash{}{0pt}%
\pgfpathmoveto{\pgfqpoint{0.538581in}{0.415000in}}%
\pgfpathlineto{\pgfqpoint{3.800000in}{0.415000in}}%
\pgfusepath{stroke}%
\end{pgfscope}%
\begin{pgfscope}%
\pgfsetrectcap%
\pgfsetmiterjoin%
\pgfsetlinewidth{0.803000pt}%
\definecolor{currentstroke}{rgb}{0.000000,0.000000,0.000000}%
\pgfsetstrokecolor{currentstroke}%
\pgfsetdash{}{0pt}%
\pgfpathmoveto{\pgfqpoint{0.538581in}{1.600000in}}%
\pgfpathlineto{\pgfqpoint{3.800000in}{1.600000in}}%
\pgfusepath{stroke}%
\end{pgfscope}%
\end{pgfpicture}%
\makeatother%
\endgroup%

%% file: imgs/sim/2_veh_min_d.pgf
\begingroup%
\makeatletter%
\begin{pgfpicture}%
\pgfpathrectangle{\pgfpointorigin}{\pgfqpoint{3.800000in}{1.600000in}}%
\pgfusepath{use as bounding box, clip}%
\begin{pgfscope}%
\pgfsetbuttcap%
\pgfsetmiterjoin%
\definecolor{currentfill}{rgb}{1.000000,1.000000,1.000000}%
\pgfsetfillcolor{currentfill}%
\pgfsetlinewidth{0.000000pt}%
\definecolor{currentstroke}{rgb}{1.000000,1.000000,1.000000}%
\pgfsetstrokecolor{currentstroke}%
\pgfsetdash{}{0pt}%
\pgfpathmoveto{\pgfqpoint{0.000000in}{0.000000in}}%
\pgfpathlineto{\pgfqpoint{3.800000in}{0.000000in}}%
\pgfpathlineto{\pgfqpoint{3.800000in}{1.600000in}}%
\pgfpathlineto{\pgfqpoint{0.000000in}{1.600000in}}%
\pgfpathclose%
\pgfusepath{fill}%
\end{pgfscope}%
\begin{pgfscope}%
\pgfsetbuttcap%
\pgfsetmiterjoin%
\definecolor{currentfill}{rgb}{1.000000,1.000000,1.000000}%
\pgfsetfillcolor{currentfill}%
\pgfsetlinewidth{0.000000pt}%
\definecolor{currentstroke}{rgb}{0.000000,0.000000,0.000000}%
\pgfsetstrokecolor{currentstroke}%
\pgfsetstrokeopacity{0.000000}%
\pgfsetdash{}{0pt}%
\pgfpathmoveto{\pgfqpoint{0.500000in}{0.415000in}}%
\pgfpathlineto{\pgfqpoint{3.800000in}{0.415000in}}%
\pgfpathlineto{\pgfqpoint{3.800000in}{1.600000in}}%
\pgfpathlineto{\pgfqpoint{0.500000in}{1.600000in}}%
\pgfpathclose%
\pgfusepath{fill}%
\end{pgfscope}%
\begin{pgfscope}%
\pgfsetbuttcap%
\pgfsetroundjoin%
\definecolor{currentfill}{rgb}{0.000000,0.000000,0.000000}%
\pgfsetfillcolor{currentfill}%
\pgfsetlinewidth{0.803000pt}%
\definecolor{currentstroke}{rgb}{0.000000,0.000000,0.000000}%
\pgfsetstrokecolor{currentstroke}%
\pgfsetdash{}{0pt}%
\pgfsys@defobject{currentmarker}{\pgfqpoint{0.000000in}{-0.048611in}}{\pgfqpoint{0.000000in}{0.000000in}}{%
\pgfpathmoveto{\pgfqpoint{0.000000in}{0.000000in}}%
\pgfpathlineto{\pgfqpoint{0.000000in}{-0.048611in}}%
\pgfusepath{stroke,fill}%
}%
\begin{pgfscope}%
\pgfsys@transformshift{0.649250in}{0.415000in}%
\pgfsys@useobject{currentmarker}{}%
\end{pgfscope}%
\end{pgfscope}%
\begin{pgfscope}%
\definecolor{textcolor}{rgb}{0.000000,0.000000,0.000000}%
\pgfsetstrokecolor{textcolor}%
\pgfsetfillcolor{textcolor}%
\pgftext[x=0.649250in,y=0.317777in,,top]{\color{textcolor}\rmfamily\fontsize{10.000000}{12.000000}\selectfont \(\displaystyle {0}\)}%
\end{pgfscope}%
\begin{pgfscope}%
\pgfsetbuttcap%
\pgfsetroundjoin%
\definecolor{currentfill}{rgb}{0.000000,0.000000,0.000000}%
\pgfsetfillcolor{currentfill}%
\pgfsetlinewidth{0.803000pt}%
\definecolor{currentstroke}{rgb}{0.000000,0.000000,0.000000}%
\pgfsetstrokecolor{currentstroke}%
\pgfsetdash{}{0pt}%
\pgfsys@defobject{currentmarker}{\pgfqpoint{0.000000in}{-0.048611in}}{\pgfqpoint{0.000000in}{0.000000in}}{%
\pgfpathmoveto{\pgfqpoint{0.000000in}{0.000000in}}%
\pgfpathlineto{\pgfqpoint{0.000000in}{-0.048611in}}%
\pgfusepath{stroke,fill}%
}%
\begin{pgfscope}%
\pgfsys@transformshift{1.399625in}{0.415000in}%
\pgfsys@useobject{currentmarker}{}%
\end{pgfscope}%
\end{pgfscope}%
\begin{pgfscope}%
\definecolor{textcolor}{rgb}{0.000000,0.000000,0.000000}%
\pgfsetstrokecolor{textcolor}%
\pgfsetfillcolor{textcolor}%
\pgftext[x=1.399625in,y=0.317777in,,top]{\color{textcolor}\rmfamily\fontsize{10.000000}{12.000000}\selectfont \(\displaystyle {5}\)}%
\end{pgfscope}%
\begin{pgfscope}%
\pgfsetbuttcap%
\pgfsetroundjoin%
\definecolor{currentfill}{rgb}{0.000000,0.000000,0.000000}%
\pgfsetfillcolor{currentfill}%
\pgfsetlinewidth{0.803000pt}%
\definecolor{currentstroke}{rgb}{0.000000,0.000000,0.000000}%
\pgfsetstrokecolor{currentstroke}%
\pgfsetdash{}{0pt}%
\pgfsys@defobject{currentmarker}{\pgfqpoint{0.000000in}{-0.048611in}}{\pgfqpoint{0.000000in}{0.000000in}}{%
\pgfpathmoveto{\pgfqpoint{0.000000in}{0.000000in}}%
\pgfpathlineto{\pgfqpoint{0.000000in}{-0.048611in}}%
\pgfusepath{stroke,fill}%
}%
\begin{pgfscope}%
\pgfsys@transformshift{2.150000in}{0.415000in}%
\pgfsys@useobject{currentmarker}{}%
\end{pgfscope}%
\end{pgfscope}%
\begin{pgfscope}%
\definecolor{textcolor}{rgb}{0.000000,0.000000,0.000000}%
\pgfsetstrokecolor{textcolor}%
\pgfsetfillcolor{textcolor}%
\pgftext[x=2.150000in,y=0.317777in,,top]{\color{textcolor}\rmfamily\fontsize{10.000000}{12.000000}\selectfont \(\displaystyle {10}\)}%
\end{pgfscope}%
\begin{pgfscope}%
\pgfsetbuttcap%
\pgfsetroundjoin%
\definecolor{currentfill}{rgb}{0.000000,0.000000,0.000000}%
\pgfsetfillcolor{currentfill}%
\pgfsetlinewidth{0.803000pt}%
\definecolor{currentstroke}{rgb}{0.000000,0.000000,0.000000}%
\pgfsetstrokecolor{currentstroke}%
\pgfsetdash{}{0pt}%
\pgfsys@defobject{currentmarker}{\pgfqpoint{0.000000in}{-0.048611in}}{\pgfqpoint{0.000000in}{0.000000in}}{%
\pgfpathmoveto{\pgfqpoint{0.000000in}{0.000000in}}%
\pgfpathlineto{\pgfqpoint{0.000000in}{-0.048611in}}%
\pgfusepath{stroke,fill}%
}%
\begin{pgfscope}%
\pgfsys@transformshift{2.900375in}{0.415000in}%
\pgfsys@useobject{currentmarker}{}%
\end{pgfscope}%
\end{pgfscope}%
\begin{pgfscope}%
\definecolor{textcolor}{rgb}{0.000000,0.000000,0.000000}%
\pgfsetstrokecolor{textcolor}%
\pgfsetfillcolor{textcolor}%
\pgftext[x=2.900375in,y=0.317777in,,top]{\color{textcolor}\rmfamily\fontsize{10.000000}{12.000000}\selectfont \(\displaystyle {15}\)}%
\end{pgfscope}%
\begin{pgfscope}%
\pgfsetbuttcap%
\pgfsetroundjoin%
\definecolor{currentfill}{rgb}{0.000000,0.000000,0.000000}%
\pgfsetfillcolor{currentfill}%
\pgfsetlinewidth{0.803000pt}%
\definecolor{currentstroke}{rgb}{0.000000,0.000000,0.000000}%
\pgfsetstrokecolor{currentstroke}%
\pgfsetdash{}{0pt}%
\pgfsys@defobject{currentmarker}{\pgfqpoint{0.000000in}{-0.048611in}}{\pgfqpoint{0.000000in}{0.000000in}}{%
\pgfpathmoveto{\pgfqpoint{0.000000in}{0.000000in}}%
\pgfpathlineto{\pgfqpoint{0.000000in}{-0.048611in}}%
\pgfusepath{stroke,fill}%
}%
\begin{pgfscope}%
\pgfsys@transformshift{3.650750in}{0.415000in}%
\pgfsys@useobject{currentmarker}{}%
\end{pgfscope}%
\end{pgfscope}%
\begin{pgfscope}%
\definecolor{textcolor}{rgb}{0.000000,0.000000,0.000000}%
\pgfsetstrokecolor{textcolor}%
\pgfsetfillcolor{textcolor}%
\pgftext[x=3.650750in,y=0.317777in,,top]{\color{textcolor}\rmfamily\fontsize{10.000000}{12.000000}\selectfont \(\displaystyle {20}\)}%
\end{pgfscope}%
\begin{pgfscope}%
\definecolor{textcolor}{rgb}{0.000000,0.000000,0.000000}%
\pgfsetstrokecolor{textcolor}%
\pgfsetfillcolor{textcolor}%
\pgftext[x=2.150000in,y=0.138889in,,top]{\color{textcolor}\rmfamily\fontsize{10.000000}{12.000000}\selectfont time (seconds)}%
\end{pgfscope}%
\begin{pgfscope}%
\pgfsetbuttcap%
\pgfsetroundjoin%
\definecolor{currentfill}{rgb}{0.000000,0.000000,0.000000}%
\pgfsetfillcolor{currentfill}%
\pgfsetlinewidth{0.803000pt}%
\definecolor{currentstroke}{rgb}{0.000000,0.000000,0.000000}%
\pgfsetstrokecolor{currentstroke}%
\pgfsetdash{}{0pt}%
\pgfsys@defobject{currentmarker}{\pgfqpoint{-0.048611in}{0.000000in}}{\pgfqpoint{-0.000000in}{0.000000in}}{%
\pgfpathmoveto{\pgfqpoint{-0.000000in}{0.000000in}}%
\pgfpathlineto{\pgfqpoint{-0.048611in}{0.000000in}}%
\pgfusepath{stroke,fill}%
}%
\begin{pgfscope}%
\pgfsys@transformshift{0.500000in}{0.455213in}%
\pgfsys@useobject{currentmarker}{}%
\end{pgfscope}%
\end{pgfscope}%
\begin{pgfscope}%
\definecolor{textcolor}{rgb}{0.000000,0.000000,0.000000}%
\pgfsetstrokecolor{textcolor}%
\pgfsetfillcolor{textcolor}%
\pgftext[x=0.333333in, y=0.407019in, left, base]{\color{textcolor}\rmfamily\fontsize{10.000000}{12.000000}\selectfont \(\displaystyle {0}\)}%
\end{pgfscope}%
\begin{pgfscope}%
\pgfsetbuttcap%
\pgfsetroundjoin%
\definecolor{currentfill}{rgb}{0.000000,0.000000,0.000000}%
\pgfsetfillcolor{currentfill}%
\pgfsetlinewidth{0.803000pt}%
\definecolor{currentstroke}{rgb}{0.000000,0.000000,0.000000}%
\pgfsetstrokecolor{currentstroke}%
\pgfsetdash{}{0pt}%
\pgfsys@defobject{currentmarker}{\pgfqpoint{-0.048611in}{0.000000in}}{\pgfqpoint{-0.000000in}{0.000000in}}{%
\pgfpathmoveto{\pgfqpoint{-0.000000in}{0.000000in}}%
\pgfpathlineto{\pgfqpoint{-0.048611in}{0.000000in}}%
\pgfusepath{stroke,fill}%
}%
\begin{pgfscope}%
\pgfsys@transformshift{0.500000in}{1.001220in}%
\pgfsys@useobject{currentmarker}{}%
\end{pgfscope}%
\end{pgfscope}%
\begin{pgfscope}%
\definecolor{textcolor}{rgb}{0.000000,0.000000,0.000000}%
\pgfsetstrokecolor{textcolor}%
\pgfsetfillcolor{textcolor}%
\pgftext[x=0.194444in, y=0.953026in, left, base]{\color{textcolor}\rmfamily\fontsize{10.000000}{12.000000}\selectfont \(\displaystyle {200}\)}%
\end{pgfscope}%
\begin{pgfscope}%
\pgfsetbuttcap%
\pgfsetroundjoin%
\definecolor{currentfill}{rgb}{0.000000,0.000000,0.000000}%
\pgfsetfillcolor{currentfill}%
\pgfsetlinewidth{0.803000pt}%
\definecolor{currentstroke}{rgb}{0.000000,0.000000,0.000000}%
\pgfsetstrokecolor{currentstroke}%
\pgfsetdash{}{0pt}%
\pgfsys@defobject{currentmarker}{\pgfqpoint{-0.048611in}{0.000000in}}{\pgfqpoint{-0.000000in}{0.000000in}}{%
\pgfpathmoveto{\pgfqpoint{-0.000000in}{0.000000in}}%
\pgfpathlineto{\pgfqpoint{-0.048611in}{0.000000in}}%
\pgfusepath{stroke,fill}%
}%
\begin{pgfscope}%
\pgfsys@transformshift{0.500000in}{1.547228in}%
\pgfsys@useobject{currentmarker}{}%
\end{pgfscope}%
\end{pgfscope}%
\begin{pgfscope}%
\definecolor{textcolor}{rgb}{0.000000,0.000000,0.000000}%
\pgfsetstrokecolor{textcolor}%
\pgfsetfillcolor{textcolor}%
\pgftext[x=0.194444in, y=1.499033in, left, base]{\color{textcolor}\rmfamily\fontsize{10.000000}{12.000000}\selectfont \(\displaystyle {400}\)}%
\end{pgfscope}%
\begin{pgfscope}%
\definecolor{textcolor}{rgb}{0.000000,0.000000,0.000000}%
\pgfsetstrokecolor{textcolor}%
\pgfsetfillcolor{textcolor}%
\pgftext[x=0.138889in,y=1.007500in,,bottom,rotate=90.000000]{\color{textcolor}\rmfamily\fontsize{10.000000}{12.000000}\selectfont vehicle dist (m)}%
\end{pgfscope}%
\begin{pgfscope}%
\pgfpathrectangle{\pgfqpoint{0.500000in}{0.415000in}}{\pgfqpoint{3.300000in}{1.185000in}}%
\pgfusepath{clip}%
\pgfsetbuttcap%
\pgfsetroundjoin%
\pgfsetlinewidth{1.505625pt}%
\definecolor{currentstroke}{rgb}{0.121569,0.466667,0.705882}%
\pgfsetstrokecolor{currentstroke}%
\pgfsetdash{{5.550000pt}{2.400000pt}}{0.000000pt}%
\pgfpathmoveto{\pgfqpoint{0.650000in}{1.546136in}}%
\pgfpathlineto{\pgfqpoint{1.619485in}{0.666305in}}%
\pgfpathlineto{\pgfqpoint{1.690020in}{0.606704in}}%
\pgfpathlineto{\pgfqpoint{1.758304in}{0.551873in}}%
\pgfpathlineto{\pgfqpoint{1.816834in}{0.507574in}}%
\pgfpathlineto{\pgfqpoint{1.851351in}{0.483898in}}%
\pgfpathlineto{\pgfqpoint{1.867859in}{0.474883in}}%
\pgfpathlineto{\pgfqpoint{1.878364in}{0.471500in}}%
\pgfpathlineto{\pgfqpoint{1.886619in}{0.471096in}}%
\pgfpathlineto{\pgfqpoint{1.895623in}{0.473039in}}%
\pgfpathlineto{\pgfqpoint{1.906879in}{0.477927in}}%
\pgfpathlineto{\pgfqpoint{1.925638in}{0.488819in}}%
\pgfpathlineto{\pgfqpoint{1.970661in}{0.518282in}}%
\pgfpathlineto{\pgfqpoint{2.079465in}{0.589338in}}%
\pgfpathlineto{\pgfqpoint{2.162006in}{0.640420in}}%
\pgfpathlineto{\pgfqpoint{2.244547in}{0.688872in}}%
\pgfpathlineto{\pgfqpoint{2.328589in}{0.735611in}}%
\pgfpathlineto{\pgfqpoint{3.650000in}{1.456426in}}%
\pgfpathlineto{\pgfqpoint{3.650000in}{1.456426in}}%
\pgfusepath{stroke}%
\end{pgfscope}%
\begin{pgfscope}%
\pgfpathrectangle{\pgfqpoint{0.500000in}{0.415000in}}{\pgfqpoint{3.300000in}{1.185000in}}%
\pgfusepath{clip}%
\pgfsetrectcap%
\pgfsetroundjoin%
\pgfsetlinewidth{1.505625pt}%
\definecolor{currentstroke}{rgb}{1.000000,0.498039,0.054902}%
\pgfsetstrokecolor{currentstroke}%
\pgfsetdash{}{0pt}%
\pgfpathmoveto{\pgfqpoint{0.650000in}{1.546136in}}%
\pgfpathlineto{\pgfqpoint{1.620986in}{0.664457in}}%
\pgfpathlineto{\pgfqpoint{1.693022in}{0.603428in}}%
\pgfpathlineto{\pgfqpoint{1.760556in}{0.549031in}}%
\pgfpathlineto{\pgfqpoint{1.811581in}{0.510593in}}%
\pgfpathlineto{\pgfqpoint{1.843847in}{0.488805in}}%
\pgfpathlineto{\pgfqpoint{1.861856in}{0.479026in}}%
\pgfpathlineto{\pgfqpoint{1.873862in}{0.474879in}}%
\pgfpathlineto{\pgfqpoint{1.883617in}{0.473856in}}%
\pgfpathlineto{\pgfqpoint{1.893372in}{0.475227in}}%
\pgfpathlineto{\pgfqpoint{1.904628in}{0.479192in}}%
\pgfpathlineto{\pgfqpoint{1.921136in}{0.487578in}}%
\pgfpathlineto{\pgfqpoint{1.951151in}{0.505713in}}%
\pgfpathlineto{\pgfqpoint{2.174763in}{0.645261in}}%
\pgfpathlineto{\pgfqpoint{2.261056in}{0.695241in}}%
\pgfpathlineto{\pgfqpoint{2.350350in}{0.744336in}}%
\pgfpathlineto{\pgfqpoint{3.012932in}{1.104915in}}%
\pgfpathlineto{\pgfqpoint{3.650000in}{1.452303in}}%
\pgfpathlineto{\pgfqpoint{3.650000in}{1.452303in}}%
\pgfusepath{stroke}%
\end{pgfscope}%
\begin{pgfscope}%
\pgfpathrectangle{\pgfqpoint{0.500000in}{0.415000in}}{\pgfqpoint{3.300000in}{1.185000in}}%
\pgfusepath{clip}%
\pgfsetrectcap%
\pgfsetroundjoin%
\pgfsetlinewidth{1.505625pt}%
\definecolor{currentstroke}{rgb}{0.000000,0.000000,0.000000}%
\pgfsetstrokecolor{currentstroke}%
\pgfsetdash{}{0pt}%
\pgfpathmoveto{\pgfqpoint{0.650000in}{0.468863in}}%
\pgfpathlineto{\pgfqpoint{3.650000in}{0.468863in}}%
\pgfpathlineto{\pgfqpoint{3.650000in}{0.468863in}}%
\pgfusepath{stroke}%
\end{pgfscope}%
\begin{pgfscope}%
\pgfsetrectcap%
\pgfsetmiterjoin%
\pgfsetlinewidth{0.803000pt}%
\definecolor{currentstroke}{rgb}{0.000000,0.000000,0.000000}%
\pgfsetstrokecolor{currentstroke}%
\pgfsetdash{}{0pt}%
\pgfpathmoveto{\pgfqpoint{0.500000in}{0.415000in}}%
\pgfpathlineto{\pgfqpoint{0.500000in}{1.600000in}}%
\pgfusepath{stroke}%
\end{pgfscope}%
\begin{pgfscope}%
\pgfsetrectcap%
\pgfsetmiterjoin%
\pgfsetlinewidth{0.803000pt}%
\definecolor{currentstroke}{rgb}{0.000000,0.000000,0.000000}%
\pgfsetstrokecolor{currentstroke}%
\pgfsetdash{}{0pt}%
\pgfpathmoveto{\pgfqpoint{3.800000in}{0.415000in}}%
\pgfpathlineto{\pgfqpoint{3.800000in}{1.600000in}}%
\pgfusepath{stroke}%
\end{pgfscope}%
\begin{pgfscope}%
\pgfsetrectcap%
\pgfsetmiterjoin%
\pgfsetlinewidth{0.803000pt}%
\definecolor{currentstroke}{rgb}{0.000000,0.000000,0.000000}%
\pgfsetstrokecolor{currentstroke}%
\pgfsetdash{}{0pt}%
\pgfpathmoveto{\pgfqpoint{0.500000in}{0.415000in}}%
\pgfpathlineto{\pgfqpoint{3.800000in}{0.415000in}}%
\pgfusepath{stroke}%
\end{pgfscope}%
\begin{pgfscope}%
\pgfsetrectcap%
\pgfsetmiterjoin%
\pgfsetlinewidth{0.803000pt}%
\definecolor{currentstroke}{rgb}{0.000000,0.000000,0.000000}%
\pgfsetstrokecolor{currentstroke}%
\pgfsetdash{}{0pt}%
\pgfpathmoveto{\pgfqpoint{0.500000in}{1.600000in}}%
\pgfpathlineto{\pgfqpoint{3.800000in}{1.600000in}}%
\pgfusepath{stroke}%
\end{pgfscope}%
\end{pgfpicture}%
\makeatother%
\endgroup%

%% file: imgs/sim/2_veh1_path.pgf
\begingroup%
\makeatletter%
\begin{pgfpicture}%
\pgfpathrectangle{\pgfpointorigin}{\pgfqpoint{3.800000in}{1.600000in}}%
\pgfusepath{use as bounding box, clip}%
\begin{pgfscope}%
\pgfsetbuttcap%
\pgfsetmiterjoin%
\definecolor{currentfill}{rgb}{1.000000,1.000000,1.000000}%
\pgfsetfillcolor{currentfill}%
\pgfsetlinewidth{0.000000pt}%
\definecolor{currentstroke}{rgb}{1.000000,1.000000,1.000000}%
\pgfsetstrokecolor{currentstroke}%
\pgfsetdash{}{0pt}%
\pgfpathmoveto{\pgfqpoint{0.000000in}{0.000000in}}%
\pgfpathlineto{\pgfqpoint{3.800000in}{0.000000in}}%
\pgfpathlineto{\pgfqpoint{3.800000in}{1.600000in}}%
\pgfpathlineto{\pgfqpoint{0.000000in}{1.600000in}}%
\pgfpathclose%
\pgfusepath{fill}%
\end{pgfscope}%
\begin{pgfscope}%
\pgfsetbuttcap%
\pgfsetmiterjoin%
\definecolor{currentfill}{rgb}{1.000000,1.000000,1.000000}%
\pgfsetfillcolor{currentfill}%
\pgfsetlinewidth{0.000000pt}%
\definecolor{currentstroke}{rgb}{0.000000,0.000000,0.000000}%
\pgfsetstrokecolor{currentstroke}%
\pgfsetstrokeopacity{0.000000}%
\pgfsetdash{}{0pt}%
\pgfpathmoveto{\pgfqpoint{0.361111in}{0.415000in}}%
\pgfpathlineto{\pgfqpoint{3.716014in}{0.415000in}}%
\pgfpathlineto{\pgfqpoint{3.716014in}{1.600000in}}%
\pgfpathlineto{\pgfqpoint{0.361111in}{1.600000in}}%
\pgfpathclose%
\pgfusepath{fill}%
\end{pgfscope}%
\begin{pgfscope}%
\pgfsetbuttcap%
\pgfsetroundjoin%
\definecolor{currentfill}{rgb}{0.000000,0.000000,0.000000}%
\pgfsetfillcolor{currentfill}%
\pgfsetlinewidth{0.803000pt}%
\definecolor{currentstroke}{rgb}{0.000000,0.000000,0.000000}%
\pgfsetstrokecolor{currentstroke}%
\pgfsetdash{}{0pt}%
\pgfsys@defobject{currentmarker}{\pgfqpoint{0.000000in}{-0.048611in}}{\pgfqpoint{0.000000in}{0.000000in}}{%
\pgfpathmoveto{\pgfqpoint{0.000000in}{0.000000in}}%
\pgfpathlineto{\pgfqpoint{0.000000in}{-0.048611in}}%
\pgfusepath{stroke,fill}%
}%
\begin{pgfscope}%
\pgfsys@transformshift{0.513607in}{0.415000in}%
\pgfsys@useobject{currentmarker}{}%
\end{pgfscope}%
\end{pgfscope}%
\begin{pgfscope}%
\definecolor{textcolor}{rgb}{0.000000,0.000000,0.000000}%
\pgfsetstrokecolor{textcolor}%
\pgfsetfillcolor{textcolor}%
\pgftext[x=0.513607in,y=0.317777in,,top]{\color{textcolor}\rmfamily\fontsize{10.000000}{12.000000}\selectfont \(\displaystyle {-200}\)}%
\end{pgfscope}%
\begin{pgfscope}%
\pgfsetbuttcap%
\pgfsetroundjoin%
\definecolor{currentfill}{rgb}{0.000000,0.000000,0.000000}%
\pgfsetfillcolor{currentfill}%
\pgfsetlinewidth{0.803000pt}%
\definecolor{currentstroke}{rgb}{0.000000,0.000000,0.000000}%
\pgfsetstrokecolor{currentstroke}%
\pgfsetdash{}{0pt}%
\pgfsys@defobject{currentmarker}{\pgfqpoint{0.000000in}{-0.048611in}}{\pgfqpoint{0.000000in}{0.000000in}}{%
\pgfpathmoveto{\pgfqpoint{0.000000in}{0.000000in}}%
\pgfpathlineto{\pgfqpoint{0.000000in}{-0.048611in}}%
\pgfusepath{stroke,fill}%
}%
\begin{pgfscope}%
\pgfsys@transformshift{1.309160in}{0.415000in}%
\pgfsys@useobject{currentmarker}{}%
\end{pgfscope}%
\end{pgfscope}%
\begin{pgfscope}%
\definecolor{textcolor}{rgb}{0.000000,0.000000,0.000000}%
\pgfsetstrokecolor{textcolor}%
\pgfsetfillcolor{textcolor}%
\pgftext[x=1.309160in,y=0.317777in,,top]{\color{textcolor}\rmfamily\fontsize{10.000000}{12.000000}\selectfont \(\displaystyle {-100}\)}%
\end{pgfscope}%
\begin{pgfscope}%
\pgfsetbuttcap%
\pgfsetroundjoin%
\definecolor{currentfill}{rgb}{0.000000,0.000000,0.000000}%
\pgfsetfillcolor{currentfill}%
\pgfsetlinewidth{0.803000pt}%
\definecolor{currentstroke}{rgb}{0.000000,0.000000,0.000000}%
\pgfsetstrokecolor{currentstroke}%
\pgfsetdash{}{0pt}%
\pgfsys@defobject{currentmarker}{\pgfqpoint{0.000000in}{-0.048611in}}{\pgfqpoint{0.000000in}{0.000000in}}{%
\pgfpathmoveto{\pgfqpoint{0.000000in}{0.000000in}}%
\pgfpathlineto{\pgfqpoint{0.000000in}{-0.048611in}}%
\pgfusepath{stroke,fill}%
}%
\begin{pgfscope}%
\pgfsys@transformshift{2.104712in}{0.415000in}%
\pgfsys@useobject{currentmarker}{}%
\end{pgfscope}%
\end{pgfscope}%
\begin{pgfscope}%
\definecolor{textcolor}{rgb}{0.000000,0.000000,0.000000}%
\pgfsetstrokecolor{textcolor}%
\pgfsetfillcolor{textcolor}%
\pgftext[x=2.104712in,y=0.317777in,,top]{\color{textcolor}\rmfamily\fontsize{10.000000}{12.000000}\selectfont \(\displaystyle {0}\)}%
\end{pgfscope}%
\begin{pgfscope}%
\pgfsetbuttcap%
\pgfsetroundjoin%
\definecolor{currentfill}{rgb}{0.000000,0.000000,0.000000}%
\pgfsetfillcolor{currentfill}%
\pgfsetlinewidth{0.803000pt}%
\definecolor{currentstroke}{rgb}{0.000000,0.000000,0.000000}%
\pgfsetstrokecolor{currentstroke}%
\pgfsetdash{}{0pt}%
\pgfsys@defobject{currentmarker}{\pgfqpoint{0.000000in}{-0.048611in}}{\pgfqpoint{0.000000in}{0.000000in}}{%
\pgfpathmoveto{\pgfqpoint{0.000000in}{0.000000in}}%
\pgfpathlineto{\pgfqpoint{0.000000in}{-0.048611in}}%
\pgfusepath{stroke,fill}%
}%
\begin{pgfscope}%
\pgfsys@transformshift{2.900265in}{0.415000in}%
\pgfsys@useobject{currentmarker}{}%
\end{pgfscope}%
\end{pgfscope}%
\begin{pgfscope}%
\definecolor{textcolor}{rgb}{0.000000,0.000000,0.000000}%
\pgfsetstrokecolor{textcolor}%
\pgfsetfillcolor{textcolor}%
\pgftext[x=2.900265in,y=0.317777in,,top]{\color{textcolor}\rmfamily\fontsize{10.000000}{12.000000}\selectfont \(\displaystyle {100}\)}%
\end{pgfscope}%
\begin{pgfscope}%
\pgfsetbuttcap%
\pgfsetroundjoin%
\definecolor{currentfill}{rgb}{0.000000,0.000000,0.000000}%
\pgfsetfillcolor{currentfill}%
\pgfsetlinewidth{0.803000pt}%
\definecolor{currentstroke}{rgb}{0.000000,0.000000,0.000000}%
\pgfsetstrokecolor{currentstroke}%
\pgfsetdash{}{0pt}%
\pgfsys@defobject{currentmarker}{\pgfqpoint{0.000000in}{-0.048611in}}{\pgfqpoint{0.000000in}{0.000000in}}{%
\pgfpathmoveto{\pgfqpoint{0.000000in}{0.000000in}}%
\pgfpathlineto{\pgfqpoint{0.000000in}{-0.048611in}}%
\pgfusepath{stroke,fill}%
}%
\begin{pgfscope}%
\pgfsys@transformshift{3.695818in}{0.415000in}%
\pgfsys@useobject{currentmarker}{}%
\end{pgfscope}%
\end{pgfscope}%
\begin{pgfscope}%
\definecolor{textcolor}{rgb}{0.000000,0.000000,0.000000}%
\pgfsetstrokecolor{textcolor}%
\pgfsetfillcolor{textcolor}%
\pgftext[x=3.695818in,y=0.317777in,,top]{\color{textcolor}\rmfamily\fontsize{10.000000}{12.000000}\selectfont \(\displaystyle {200}\)}%
\end{pgfscope}%
\begin{pgfscope}%
\definecolor{textcolor}{rgb}{0.000000,0.000000,0.000000}%
\pgfsetstrokecolor{textcolor}%
\pgfsetfillcolor{textcolor}%
\pgftext[x=2.038563in,y=0.138889in,,top]{\color{textcolor}\rmfamily\fontsize{10.000000}{12.000000}\selectfont \(\displaystyle x\) (meters)}%
\end{pgfscope}%
\begin{pgfscope}%
\pgfsetbuttcap%
\pgfsetroundjoin%
\definecolor{currentfill}{rgb}{0.000000,0.000000,0.000000}%
\pgfsetfillcolor{currentfill}%
\pgfsetlinewidth{0.803000pt}%
\definecolor{currentstroke}{rgb}{0.000000,0.000000,0.000000}%
\pgfsetstrokecolor{currentstroke}%
\pgfsetdash{}{0pt}%
\pgfsys@defobject{currentmarker}{\pgfqpoint{-0.048611in}{0.000000in}}{\pgfqpoint{-0.000000in}{0.000000in}}{%
\pgfpathmoveto{\pgfqpoint{-0.000000in}{0.000000in}}%
\pgfpathlineto{\pgfqpoint{-0.048611in}{0.000000in}}%
\pgfusepath{stroke,fill}%
}%
\begin{pgfscope}%
\pgfsys@transformshift{0.361111in}{0.468863in}%
\pgfsys@useobject{currentmarker}{}%
\end{pgfscope}%
\end{pgfscope}%
\begin{pgfscope}%
\definecolor{textcolor}{rgb}{0.000000,0.000000,0.000000}%
\pgfsetstrokecolor{textcolor}%
\pgfsetfillcolor{textcolor}%
\pgftext[x=0.194444in, y=0.420669in, left, base]{\color{textcolor}\rmfamily\fontsize{10.000000}{12.000000}\selectfont \(\displaystyle {0}\)}%
\end{pgfscope}%
\begin{pgfscope}%
\pgfsetbuttcap%
\pgfsetroundjoin%
\definecolor{currentfill}{rgb}{0.000000,0.000000,0.000000}%
\pgfsetfillcolor{currentfill}%
\pgfsetlinewidth{0.803000pt}%
\definecolor{currentstroke}{rgb}{0.000000,0.000000,0.000000}%
\pgfsetstrokecolor{currentstroke}%
\pgfsetdash{}{0pt}%
\pgfsys@defobject{currentmarker}{\pgfqpoint{-0.048611in}{0.000000in}}{\pgfqpoint{-0.000000in}{0.000000in}}{%
\pgfpathmoveto{\pgfqpoint{-0.000000in}{0.000000in}}%
\pgfpathlineto{\pgfqpoint{-0.048611in}{0.000000in}}%
\pgfusepath{stroke,fill}%
}%
\begin{pgfscope}%
\pgfsys@transformshift{0.361111in}{0.869368in}%
\pgfsys@useobject{currentmarker}{}%
\end{pgfscope}%
\end{pgfscope}%
\begin{pgfscope}%
\definecolor{textcolor}{rgb}{0.000000,0.000000,0.000000}%
\pgfsetstrokecolor{textcolor}%
\pgfsetfillcolor{textcolor}%
\pgftext[x=0.194444in, y=0.821174in, left, base]{\color{textcolor}\rmfamily\fontsize{10.000000}{12.000000}\selectfont \(\displaystyle {2}\)}%
\end{pgfscope}%
\begin{pgfscope}%
\pgfsetbuttcap%
\pgfsetroundjoin%
\definecolor{currentfill}{rgb}{0.000000,0.000000,0.000000}%
\pgfsetfillcolor{currentfill}%
\pgfsetlinewidth{0.803000pt}%
\definecolor{currentstroke}{rgb}{0.000000,0.000000,0.000000}%
\pgfsetstrokecolor{currentstroke}%
\pgfsetdash{}{0pt}%
\pgfsys@defobject{currentmarker}{\pgfqpoint{-0.048611in}{0.000000in}}{\pgfqpoint{-0.000000in}{0.000000in}}{%
\pgfpathmoveto{\pgfqpoint{-0.000000in}{0.000000in}}%
\pgfpathlineto{\pgfqpoint{-0.048611in}{0.000000in}}%
\pgfusepath{stroke,fill}%
}%
\begin{pgfscope}%
\pgfsys@transformshift{0.361111in}{1.269873in}%
\pgfsys@useobject{currentmarker}{}%
\end{pgfscope}%
\end{pgfscope}%
\begin{pgfscope}%
\definecolor{textcolor}{rgb}{0.000000,0.000000,0.000000}%
\pgfsetstrokecolor{textcolor}%
\pgfsetfillcolor{textcolor}%
\pgftext[x=0.194444in, y=1.221679in, left, base]{\color{textcolor}\rmfamily\fontsize{10.000000}{12.000000}\selectfont \(\displaystyle {4}\)}%
\end{pgfscope}%
\begin{pgfscope}%
\definecolor{textcolor}{rgb}{0.000000,0.000000,0.000000}%
\pgfsetstrokecolor{textcolor}%
\pgfsetfillcolor{textcolor}%
\pgftext[x=0.138889in,y=1.007500in,,bottom,rotate=90.000000]{\color{textcolor}\rmfamily\fontsize{10.000000}{12.000000}\selectfont \(\displaystyle y\) (meters)}%
\end{pgfscope}%
\begin{pgfscope}%
\pgfpathrectangle{\pgfqpoint{0.361111in}{0.415000in}}{\pgfqpoint{3.354903in}{1.185000in}}%
\pgfusepath{clip}%
\pgfsetbuttcap%
\pgfsetroundjoin%
\pgfsetlinewidth{1.505625pt}%
\definecolor{currentstroke}{rgb}{0.121569,0.466667,0.705882}%
\pgfsetstrokecolor{currentstroke}%
\pgfsetdash{{5.550000pt}{2.400000pt}}{0.000000pt}%
\pgfpathmoveto{\pgfqpoint{0.513607in}{0.468863in}}%
\pgfpathlineto{\pgfqpoint{0.527125in}{0.479374in}}%
\pgfpathlineto{\pgfqpoint{0.549989in}{0.493963in}}%
\pgfpathlineto{\pgfqpoint{0.583792in}{0.512704in}}%
\pgfpathlineto{\pgfqpoint{0.632510in}{0.536999in}}%
\pgfpathlineto{\pgfqpoint{0.698132in}{0.567099in}}%
\pgfpathlineto{\pgfqpoint{0.782648in}{0.603326in}}%
\pgfpathlineto{\pgfqpoint{0.888045in}{0.646006in}}%
\pgfpathlineto{\pgfqpoint{1.014324in}{0.694670in}}%
\pgfpathlineto{\pgfqpoint{1.159497in}{0.748175in}}%
\pgfpathlineto{\pgfqpoint{1.319586in}{0.804770in}}%
\pgfpathlineto{\pgfqpoint{1.487631in}{0.861809in}}%
\pgfpathlineto{\pgfqpoint{1.655678in}{0.916513in}}%
\pgfpathlineto{\pgfqpoint{1.811961in}{0.965097in}}%
\pgfpathlineto{\pgfqpoint{1.942015in}{1.003295in}}%
\pgfpathlineto{\pgfqpoint{2.038404in}{1.029434in}}%
\pgfpathlineto{\pgfqpoint{2.095440in}{1.042845in}}%
\pgfpathlineto{\pgfqpoint{2.108994in}{1.044558in}}%
\pgfpathlineto{\pgfqpoint{2.117704in}{1.042985in}}%
\pgfpathlineto{\pgfqpoint{2.137323in}{1.036411in}}%
\pgfpathlineto{\pgfqpoint{2.403316in}{0.939573in}}%
\pgfpathlineto{\pgfqpoint{3.563519in}{0.517045in}}%
\pgfpathlineto{\pgfqpoint{3.563519in}{0.517045in}}%
\pgfusepath{stroke}%
\end{pgfscope}%
\begin{pgfscope}%
\pgfpathrectangle{\pgfqpoint{0.361111in}{0.415000in}}{\pgfqpoint{3.354903in}{1.185000in}}%
\pgfusepath{clip}%
\pgfsetrectcap%
\pgfsetroundjoin%
\pgfsetlinewidth{1.505625pt}%
\definecolor{currentstroke}{rgb}{1.000000,0.498039,0.054902}%
\pgfsetstrokecolor{currentstroke}%
\pgfsetdash{}{0pt}%
\pgfpathmoveto{\pgfqpoint{0.513607in}{0.468863in}}%
\pgfpathlineto{\pgfqpoint{1.890565in}{0.469884in}}%
\pgfpathlineto{\pgfqpoint{1.901368in}{0.472871in}}%
\pgfpathlineto{\pgfqpoint{1.911211in}{0.477879in}}%
\pgfpathlineto{\pgfqpoint{1.920993in}{0.485321in}}%
\pgfpathlineto{\pgfqpoint{1.931594in}{0.496387in}}%
\pgfpathlineto{\pgfqpoint{1.942117in}{0.510633in}}%
\pgfpathlineto{\pgfqpoint{1.953426in}{0.529714in}}%
\pgfpathlineto{\pgfqpoint{1.965494in}{0.554514in}}%
\pgfpathlineto{\pgfqpoint{1.978293in}{0.585935in}}%
\pgfpathlineto{\pgfqpoint{1.991787in}{0.624872in}}%
\pgfpathlineto{\pgfqpoint{2.006767in}{0.675159in}}%
\pgfpathlineto{\pgfqpoint{2.022339in}{0.735354in}}%
\pgfpathlineto{\pgfqpoint{2.039256in}{0.809851in}}%
\pgfpathlineto{\pgfqpoint{2.057425in}{0.900151in}}%
\pgfpathlineto{\pgfqpoint{2.078287in}{1.015816in}}%
\pgfpathlineto{\pgfqpoint{2.113920in}{1.216809in}}%
\pgfpathlineto{\pgfqpoint{2.131340in}{1.298867in}}%
\pgfpathlineto{\pgfqpoint{2.147218in}{1.363382in}}%
\pgfpathlineto{\pgfqpoint{2.162299in}{1.415633in}}%
\pgfpathlineto{\pgfqpoint{2.175825in}{1.455038in}}%
\pgfpathlineto{\pgfqpoint{2.188549in}{1.485676in}}%
\pgfpathlineto{\pgfqpoint{2.200471in}{1.508723in}}%
\pgfpathlineto{\pgfqpoint{2.210853in}{1.524327in}}%
\pgfpathlineto{\pgfqpoint{2.220449in}{1.535039in}}%
\pgfpathlineto{\pgfqpoint{2.229265in}{1.541728in}}%
\pgfpathlineto{\pgfqpoint{2.237308in}{1.545188in}}%
\pgfpathlineto{\pgfqpoint{2.245312in}{1.546117in}}%
\pgfpathlineto{\pgfqpoint{2.252552in}{1.544786in}}%
\pgfpathlineto{\pgfqpoint{2.260476in}{1.541014in}}%
\pgfpathlineto{\pgfqpoint{2.276191in}{1.530175in}}%
\pgfpathlineto{\pgfqpoint{2.380818in}{1.452050in}}%
\pgfpathlineto{\pgfqpoint{3.557504in}{0.572277in}}%
\pgfpathlineto{\pgfqpoint{3.557504in}{0.572277in}}%
\pgfusepath{stroke}%
\end{pgfscope}%
\begin{pgfscope}%
\pgfsetrectcap%
\pgfsetmiterjoin%
\pgfsetlinewidth{0.803000pt}%
\definecolor{currentstroke}{rgb}{0.000000,0.000000,0.000000}%
\pgfsetstrokecolor{currentstroke}%
\pgfsetdash{}{0pt}%
\pgfpathmoveto{\pgfqpoint{0.361111in}{0.415000in}}%
\pgfpathlineto{\pgfqpoint{0.361111in}{1.600000in}}%
\pgfusepath{stroke}%
\end{pgfscope}%
\begin{pgfscope}%
\pgfsetrectcap%
\pgfsetmiterjoin%
\pgfsetlinewidth{0.803000pt}%
\definecolor{currentstroke}{rgb}{0.000000,0.000000,0.000000}%
\pgfsetstrokecolor{currentstroke}%
\pgfsetdash{}{0pt}%
\pgfpathmoveto{\pgfqpoint{3.716014in}{0.415000in}}%
\pgfpathlineto{\pgfqpoint{3.716014in}{1.600000in}}%
\pgfusepath{stroke}%
\end{pgfscope}%
\begin{pgfscope}%
\pgfsetrectcap%
\pgfsetmiterjoin%
\pgfsetlinewidth{0.803000pt}%
\definecolor{currentstroke}{rgb}{0.000000,0.000000,0.000000}%
\pgfsetstrokecolor{currentstroke}%
\pgfsetdash{}{0pt}%
\pgfpathmoveto{\pgfqpoint{0.361111in}{0.415000in}}%
\pgfpathlineto{\pgfqpoint{3.716014in}{0.415000in}}%
\pgfusepath{stroke}%
\end{pgfscope}%
\begin{pgfscope}%
\pgfsetrectcap%
\pgfsetmiterjoin%
\pgfsetlinewidth{0.803000pt}%
\definecolor{currentstroke}{rgb}{0.000000,0.000000,0.000000}%
\pgfsetstrokecolor{currentstroke}%
\pgfsetdash{}{0pt}%
\pgfpathmoveto{\pgfqpoint{0.361111in}{1.600000in}}%
\pgfpathlineto{\pgfqpoint{3.716014in}{1.600000in}}%
\pgfusepath{stroke}%
\end{pgfscope}%
\end{pgfpicture}%
\makeatother%
\endgroup%

%% file: imgs/sim/alt_2_veh1_actuator2.pgf
\begingroup%
\makeatletter%
\begin{pgfpicture}%
\pgfpathrectangle{\pgfpointorigin}{\pgfqpoint{3.800000in}{1.600000in}}%
\pgfusepath{use as bounding box, clip}%
\begin{pgfscope}%
\pgfsetbuttcap%
\pgfsetmiterjoin%
\definecolor{currentfill}{rgb}{1.000000,1.000000,1.000000}%
\pgfsetfillcolor{currentfill}%
\pgfsetlinewidth{0.000000pt}%
\definecolor{currentstroke}{rgb}{1.000000,1.000000,1.000000}%
\pgfsetstrokecolor{currentstroke}%
\pgfsetdash{}{0pt}%
\pgfpathmoveto{\pgfqpoint{0.000000in}{0.000000in}}%
\pgfpathlineto{\pgfqpoint{3.800000in}{0.000000in}}%
\pgfpathlineto{\pgfqpoint{3.800000in}{1.600000in}}%
\pgfpathlineto{\pgfqpoint{0.000000in}{1.600000in}}%
\pgfpathclose%
\pgfusepath{fill}%
\end{pgfscope}%
\begin{pgfscope}%
\pgfsetbuttcap%
\pgfsetmiterjoin%
\definecolor{currentfill}{rgb}{1.000000,1.000000,1.000000}%
\pgfsetfillcolor{currentfill}%
\pgfsetlinewidth{0.000000pt}%
\definecolor{currentstroke}{rgb}{0.000000,0.000000,0.000000}%
\pgfsetstrokecolor{currentstroke}%
\pgfsetstrokeopacity{0.000000}%
\pgfsetdash{}{0pt}%
\pgfpathmoveto{\pgfqpoint{0.577161in}{0.415000in}}%
\pgfpathlineto{\pgfqpoint{3.800000in}{0.415000in}}%
\pgfpathlineto{\pgfqpoint{3.800000in}{1.600000in}}%
\pgfpathlineto{\pgfqpoint{0.577161in}{1.600000in}}%
\pgfpathclose%
\pgfusepath{fill}%
\end{pgfscope}%
\begin{pgfscope}%
\pgfsetbuttcap%
\pgfsetroundjoin%
\definecolor{currentfill}{rgb}{0.000000,0.000000,0.000000}%
\pgfsetfillcolor{currentfill}%
\pgfsetlinewidth{0.803000pt}%
\definecolor{currentstroke}{rgb}{0.000000,0.000000,0.000000}%
\pgfsetstrokecolor{currentstroke}%
\pgfsetdash{}{0pt}%
\pgfsys@defobject{currentmarker}{\pgfqpoint{0.000000in}{-0.048611in}}{\pgfqpoint{0.000000in}{0.000000in}}{%
\pgfpathmoveto{\pgfqpoint{0.000000in}{0.000000in}}%
\pgfpathlineto{\pgfqpoint{0.000000in}{-0.048611in}}%
\pgfusepath{stroke,fill}%
}%
\begin{pgfscope}%
\pgfsys@transformshift{0.722921in}{0.415000in}%
\pgfsys@useobject{currentmarker}{}%
\end{pgfscope}%
\end{pgfscope}%
\begin{pgfscope}%
\definecolor{textcolor}{rgb}{0.000000,0.000000,0.000000}%
\pgfsetstrokecolor{textcolor}%
\pgfsetfillcolor{textcolor}%
\pgftext[x=0.722921in,y=0.317777in,,top]{\color{textcolor}\rmfamily\fontsize{10.000000}{12.000000}\selectfont \(\displaystyle {0}\)}%
\end{pgfscope}%
\begin{pgfscope}%
\pgfsetbuttcap%
\pgfsetroundjoin%
\definecolor{currentfill}{rgb}{0.000000,0.000000,0.000000}%
\pgfsetfillcolor{currentfill}%
\pgfsetlinewidth{0.803000pt}%
\definecolor{currentstroke}{rgb}{0.000000,0.000000,0.000000}%
\pgfsetstrokecolor{currentstroke}%
\pgfsetdash{}{0pt}%
\pgfsys@defobject{currentmarker}{\pgfqpoint{0.000000in}{-0.048611in}}{\pgfqpoint{0.000000in}{0.000000in}}{%
\pgfpathmoveto{\pgfqpoint{0.000000in}{0.000000in}}%
\pgfpathlineto{\pgfqpoint{0.000000in}{-0.048611in}}%
\pgfusepath{stroke,fill}%
}%
\begin{pgfscope}%
\pgfsys@transformshift{1.455751in}{0.415000in}%
\pgfsys@useobject{currentmarker}{}%
\end{pgfscope}%
\end{pgfscope}%
\begin{pgfscope}%
\definecolor{textcolor}{rgb}{0.000000,0.000000,0.000000}%
\pgfsetstrokecolor{textcolor}%
\pgfsetfillcolor{textcolor}%
\pgftext[x=1.455751in,y=0.317777in,,top]{\color{textcolor}\rmfamily\fontsize{10.000000}{12.000000}\selectfont \(\displaystyle {5}\)}%
\end{pgfscope}%
\begin{pgfscope}%
\pgfsetbuttcap%
\pgfsetroundjoin%
\definecolor{currentfill}{rgb}{0.000000,0.000000,0.000000}%
\pgfsetfillcolor{currentfill}%
\pgfsetlinewidth{0.803000pt}%
\definecolor{currentstroke}{rgb}{0.000000,0.000000,0.000000}%
\pgfsetstrokecolor{currentstroke}%
\pgfsetdash{}{0pt}%
\pgfsys@defobject{currentmarker}{\pgfqpoint{0.000000in}{-0.048611in}}{\pgfqpoint{0.000000in}{0.000000in}}{%
\pgfpathmoveto{\pgfqpoint{0.000000in}{0.000000in}}%
\pgfpathlineto{\pgfqpoint{0.000000in}{-0.048611in}}%
\pgfusepath{stroke,fill}%
}%
\begin{pgfscope}%
\pgfsys@transformshift{2.188581in}{0.415000in}%
\pgfsys@useobject{currentmarker}{}%
\end{pgfscope}%
\end{pgfscope}%
\begin{pgfscope}%
\definecolor{textcolor}{rgb}{0.000000,0.000000,0.000000}%
\pgfsetstrokecolor{textcolor}%
\pgfsetfillcolor{textcolor}%
\pgftext[x=2.188581in,y=0.317777in,,top]{\color{textcolor}\rmfamily\fontsize{10.000000}{12.000000}\selectfont \(\displaystyle {10}\)}%
\end{pgfscope}%
\begin{pgfscope}%
\pgfsetbuttcap%
\pgfsetroundjoin%
\definecolor{currentfill}{rgb}{0.000000,0.000000,0.000000}%
\pgfsetfillcolor{currentfill}%
\pgfsetlinewidth{0.803000pt}%
\definecolor{currentstroke}{rgb}{0.000000,0.000000,0.000000}%
\pgfsetstrokecolor{currentstroke}%
\pgfsetdash{}{0pt}%
\pgfsys@defobject{currentmarker}{\pgfqpoint{0.000000in}{-0.048611in}}{\pgfqpoint{0.000000in}{0.000000in}}{%
\pgfpathmoveto{\pgfqpoint{0.000000in}{0.000000in}}%
\pgfpathlineto{\pgfqpoint{0.000000in}{-0.048611in}}%
\pgfusepath{stroke,fill}%
}%
\begin{pgfscope}%
\pgfsys@transformshift{2.921410in}{0.415000in}%
\pgfsys@useobject{currentmarker}{}%
\end{pgfscope}%
\end{pgfscope}%
\begin{pgfscope}%
\definecolor{textcolor}{rgb}{0.000000,0.000000,0.000000}%
\pgfsetstrokecolor{textcolor}%
\pgfsetfillcolor{textcolor}%
\pgftext[x=2.921410in,y=0.317777in,,top]{\color{textcolor}\rmfamily\fontsize{10.000000}{12.000000}\selectfont \(\displaystyle {15}\)}%
\end{pgfscope}%
\begin{pgfscope}%
\pgfsetbuttcap%
\pgfsetroundjoin%
\definecolor{currentfill}{rgb}{0.000000,0.000000,0.000000}%
\pgfsetfillcolor{currentfill}%
\pgfsetlinewidth{0.803000pt}%
\definecolor{currentstroke}{rgb}{0.000000,0.000000,0.000000}%
\pgfsetstrokecolor{currentstroke}%
\pgfsetdash{}{0pt}%
\pgfsys@defobject{currentmarker}{\pgfqpoint{0.000000in}{-0.048611in}}{\pgfqpoint{0.000000in}{0.000000in}}{%
\pgfpathmoveto{\pgfqpoint{0.000000in}{0.000000in}}%
\pgfpathlineto{\pgfqpoint{0.000000in}{-0.048611in}}%
\pgfusepath{stroke,fill}%
}%
\begin{pgfscope}%
\pgfsys@transformshift{3.654240in}{0.415000in}%
\pgfsys@useobject{currentmarker}{}%
\end{pgfscope}%
\end{pgfscope}%
\begin{pgfscope}%
\definecolor{textcolor}{rgb}{0.000000,0.000000,0.000000}%
\pgfsetstrokecolor{textcolor}%
\pgfsetfillcolor{textcolor}%
\pgftext[x=3.654240in,y=0.317777in,,top]{\color{textcolor}\rmfamily\fontsize{10.000000}{12.000000}\selectfont \(\displaystyle {20}\)}%
\end{pgfscope}%
\begin{pgfscope}%
\definecolor{textcolor}{rgb}{0.000000,0.000000,0.000000}%
\pgfsetstrokecolor{textcolor}%
\pgfsetfillcolor{textcolor}%
\pgftext[x=2.188581in,y=0.138889in,,top]{\color{textcolor}\rmfamily\fontsize{10.000000}{12.000000}\selectfont time (seconds)}%
\end{pgfscope}%
\begin{pgfscope}%
\pgfsetbuttcap%
\pgfsetroundjoin%
\definecolor{currentfill}{rgb}{0.000000,0.000000,0.000000}%
\pgfsetfillcolor{currentfill}%
\pgfsetlinewidth{0.803000pt}%
\definecolor{currentstroke}{rgb}{0.000000,0.000000,0.000000}%
\pgfsetstrokecolor{currentstroke}%
\pgfsetdash{}{0pt}%
\pgfsys@defobject{currentmarker}{\pgfqpoint{-0.048611in}{0.000000in}}{\pgfqpoint{-0.000000in}{0.000000in}}{%
\pgfpathmoveto{\pgfqpoint{-0.000000in}{0.000000in}}%
\pgfpathlineto{\pgfqpoint{-0.048611in}{0.000000in}}%
\pgfusepath{stroke,fill}%
}%
\begin{pgfscope}%
\pgfsys@transformshift{0.577161in}{0.662220in}%
\pgfsys@useobject{currentmarker}{}%
\end{pgfscope}%
\end{pgfscope}%
\begin{pgfscope}%
\definecolor{textcolor}{rgb}{0.000000,0.000000,0.000000}%
\pgfsetstrokecolor{textcolor}%
\pgfsetfillcolor{textcolor}%
\pgftext[x=0.194444in, y=0.614026in, left, base]{\color{textcolor}\rmfamily\fontsize{10.000000}{12.000000}\selectfont \(\displaystyle {-2.5}\)}%
\end{pgfscope}%
\begin{pgfscope}%
\pgfsetbuttcap%
\pgfsetroundjoin%
\definecolor{currentfill}{rgb}{0.000000,0.000000,0.000000}%
\pgfsetfillcolor{currentfill}%
\pgfsetlinewidth{0.803000pt}%
\definecolor{currentstroke}{rgb}{0.000000,0.000000,0.000000}%
\pgfsetstrokecolor{currentstroke}%
\pgfsetdash{}{0pt}%
\pgfsys@defobject{currentmarker}{\pgfqpoint{-0.048611in}{0.000000in}}{\pgfqpoint{-0.000000in}{0.000000in}}{%
\pgfpathmoveto{\pgfqpoint{-0.000000in}{0.000000in}}%
\pgfpathlineto{\pgfqpoint{-0.048611in}{0.000000in}}%
\pgfusepath{stroke,fill}%
}%
\begin{pgfscope}%
\pgfsys@transformshift{0.577161in}{1.007500in}%
\pgfsys@useobject{currentmarker}{}%
\end{pgfscope}%
\end{pgfscope}%
\begin{pgfscope}%
\definecolor{textcolor}{rgb}{0.000000,0.000000,0.000000}%
\pgfsetstrokecolor{textcolor}%
\pgfsetfillcolor{textcolor}%
\pgftext[x=0.302469in, y=0.959305in, left, base]{\color{textcolor}\rmfamily\fontsize{10.000000}{12.000000}\selectfont \(\displaystyle {0.0}\)}%
\end{pgfscope}%
\begin{pgfscope}%
\pgfsetbuttcap%
\pgfsetroundjoin%
\definecolor{currentfill}{rgb}{0.000000,0.000000,0.000000}%
\pgfsetfillcolor{currentfill}%
\pgfsetlinewidth{0.803000pt}%
\definecolor{currentstroke}{rgb}{0.000000,0.000000,0.000000}%
\pgfsetstrokecolor{currentstroke}%
\pgfsetdash{}{0pt}%
\pgfsys@defobject{currentmarker}{\pgfqpoint{-0.048611in}{0.000000in}}{\pgfqpoint{-0.000000in}{0.000000in}}{%
\pgfpathmoveto{\pgfqpoint{-0.000000in}{0.000000in}}%
\pgfpathlineto{\pgfqpoint{-0.048611in}{0.000000in}}%
\pgfusepath{stroke,fill}%
}%
\begin{pgfscope}%
\pgfsys@transformshift{0.577161in}{1.352780in}%
\pgfsys@useobject{currentmarker}{}%
\end{pgfscope}%
\end{pgfscope}%
\begin{pgfscope}%
\definecolor{textcolor}{rgb}{0.000000,0.000000,0.000000}%
\pgfsetstrokecolor{textcolor}%
\pgfsetfillcolor{textcolor}%
\pgftext[x=0.302469in, y=1.304585in, left, base]{\color{textcolor}\rmfamily\fontsize{10.000000}{12.000000}\selectfont \(\displaystyle {2.5}\)}%
\end{pgfscope}%
\begin{pgfscope}%
\definecolor{textcolor}{rgb}{0.000000,0.000000,0.000000}%
\pgfsetstrokecolor{textcolor}%
\pgfsetfillcolor{textcolor}%
\pgftext[x=0.138889in,y=1.007500in,,bottom,rotate=90.000000]{\color{textcolor}\rmfamily\fontsize{10.000000}{12.000000}\selectfont \(\displaystyle \zeta\) (m/s)}%
\end{pgfscope}%
\begin{pgfscope}%
\pgfpathrectangle{\pgfqpoint{0.577161in}{0.415000in}}{\pgfqpoint{3.222839in}{1.185000in}}%
\pgfusepath{clip}%
\pgfsetbuttcap%
\pgfsetroundjoin%
\pgfsetlinewidth{1.505625pt}%
\definecolor{currentstroke}{rgb}{0.121569,0.466667,0.705882}%
\pgfsetstrokecolor{currentstroke}%
\pgfsetdash{{5.550000pt}{2.400000pt}}{0.000000pt}%
\pgfpathmoveto{\pgfqpoint{0.723654in}{1.021186in}}%
\pgfpathlineto{\pgfqpoint{1.139901in}{1.017773in}}%
\pgfpathlineto{\pgfqpoint{1.898380in}{1.011294in}}%
\pgfpathlineto{\pgfqpoint{1.913037in}{1.009027in}}%
\pgfpathlineto{\pgfqpoint{1.919632in}{1.005894in}}%
\pgfpathlineto{\pgfqpoint{1.923296in}{1.001486in}}%
\pgfpathlineto{\pgfqpoint{1.926227in}{0.993291in}}%
\pgfpathlineto{\pgfqpoint{1.926960in}{0.991092in}}%
\pgfpathlineto{\pgfqpoint{1.929892in}{1.013893in}}%
\pgfpathlineto{\pgfqpoint{2.887700in}{1.010824in}}%
\pgfpathlineto{\pgfqpoint{3.653507in}{1.009470in}}%
\pgfpathlineto{\pgfqpoint{3.653507in}{1.009470in}}%
\pgfusepath{stroke}%
\end{pgfscope}%
\begin{pgfscope}%
\pgfpathrectangle{\pgfqpoint{0.577161in}{0.415000in}}{\pgfqpoint{3.222839in}{1.185000in}}%
\pgfusepath{clip}%
\pgfsetrectcap%
\pgfsetroundjoin%
\pgfsetlinewidth{1.505625pt}%
\definecolor{currentstroke}{rgb}{1.000000,0.498039,0.054902}%
\pgfsetstrokecolor{currentstroke}%
\pgfsetdash{}{0pt}%
\pgfpathmoveto{\pgfqpoint{0.723654in}{1.021311in}}%
\pgfpathlineto{\pgfqpoint{1.353887in}{1.016481in}}%
\pgfpathlineto{\pgfqpoint{1.790654in}{1.012695in}}%
\pgfpathlineto{\pgfqpoint{1.837555in}{1.009398in}}%
\pgfpathlineto{\pgfqpoint{1.863937in}{1.005472in}}%
\pgfpathlineto{\pgfqpoint{1.880059in}{1.000930in}}%
\pgfpathlineto{\pgfqpoint{1.890319in}{0.995971in}}%
\pgfpathlineto{\pgfqpoint{1.898380in}{0.989600in}}%
\pgfpathlineto{\pgfqpoint{1.904243in}{0.982189in}}%
\pgfpathlineto{\pgfqpoint{1.909372in}{0.971832in}}%
\pgfpathlineto{\pgfqpoint{1.913769in}{0.957068in}}%
\pgfpathlineto{\pgfqpoint{1.917434in}{0.935874in}}%
\pgfpathlineto{\pgfqpoint{1.920365in}{0.906307in}}%
\pgfpathlineto{\pgfqpoint{1.923296in}{1.014959in}}%
\pgfpathlineto{\pgfqpoint{3.653507in}{1.009789in}}%
\pgfpathlineto{\pgfqpoint{3.653507in}{1.009789in}}%
\pgfusepath{stroke}%
\end{pgfscope}%
\begin{pgfscope}%
\pgfpathrectangle{\pgfqpoint{0.577161in}{0.415000in}}{\pgfqpoint{3.222839in}{1.185000in}}%
\pgfusepath{clip}%
\pgfsetrectcap%
\pgfsetroundjoin%
\pgfsetlinewidth{1.505625pt}%
\definecolor{currentstroke}{rgb}{0.000000,0.000000,0.000000}%
\pgfsetstrokecolor{currentstroke}%
\pgfsetdash{}{0pt}%
\pgfpathmoveto{\pgfqpoint{0.723654in}{0.468863in}}%
\pgfpathlineto{\pgfqpoint{3.653507in}{0.468863in}}%
\pgfpathlineto{\pgfqpoint{3.653507in}{0.468863in}}%
\pgfusepath{stroke}%
\end{pgfscope}%
\begin{pgfscope}%
\pgfpathrectangle{\pgfqpoint{0.577161in}{0.415000in}}{\pgfqpoint{3.222839in}{1.185000in}}%
\pgfusepath{clip}%
\pgfsetrectcap%
\pgfsetroundjoin%
\pgfsetlinewidth{1.505625pt}%
\definecolor{currentstroke}{rgb}{0.000000,0.000000,0.000000}%
\pgfsetstrokecolor{currentstroke}%
\pgfsetdash{}{0pt}%
\pgfpathmoveto{\pgfqpoint{0.723654in}{1.546136in}}%
\pgfpathlineto{\pgfqpoint{3.653507in}{1.546136in}}%
\pgfpathlineto{\pgfqpoint{3.653507in}{1.546136in}}%
\pgfusepath{stroke}%
\end{pgfscope}%
\begin{pgfscope}%
\pgfsetrectcap%
\pgfsetmiterjoin%
\pgfsetlinewidth{0.803000pt}%
\definecolor{currentstroke}{rgb}{0.000000,0.000000,0.000000}%
\pgfsetstrokecolor{currentstroke}%
\pgfsetdash{}{0pt}%
\pgfpathmoveto{\pgfqpoint{0.577161in}{0.415000in}}%
\pgfpathlineto{\pgfqpoint{0.577161in}{1.600000in}}%
\pgfusepath{stroke}%
\end{pgfscope}%
\begin{pgfscope}%
\pgfsetrectcap%
\pgfsetmiterjoin%
\pgfsetlinewidth{0.803000pt}%
\definecolor{currentstroke}{rgb}{0.000000,0.000000,0.000000}%
\pgfsetstrokecolor{currentstroke}%
\pgfsetdash{}{0pt}%
\pgfpathmoveto{\pgfqpoint{3.800000in}{0.415000in}}%
\pgfpathlineto{\pgfqpoint{3.800000in}{1.600000in}}%
\pgfusepath{stroke}%
\end{pgfscope}%
\begin{pgfscope}%
\pgfsetrectcap%
\pgfsetmiterjoin%
\pgfsetlinewidth{0.803000pt}%
\definecolor{currentstroke}{rgb}{0.000000,0.000000,0.000000}%
\pgfsetstrokecolor{currentstroke}%
\pgfsetdash{}{0pt}%
\pgfpathmoveto{\pgfqpoint{0.577161in}{0.415000in}}%
\pgfpathlineto{\pgfqpoint{3.800000in}{0.415000in}}%
\pgfusepath{stroke}%
\end{pgfscope}%
\begin{pgfscope}%
\pgfsetrectcap%
\pgfsetmiterjoin%
\pgfsetlinewidth{0.803000pt}%
\definecolor{currentstroke}{rgb}{0.000000,0.000000,0.000000}%
\pgfsetstrokecolor{currentstroke}%
\pgfsetdash{}{0pt}%
\pgfpathmoveto{\pgfqpoint{0.577161in}{1.600000in}}%
\pgfpathlineto{\pgfqpoint{3.800000in}{1.600000in}}%
\pgfusepath{stroke}%
\end{pgfscope}%
\begin{pgfscope}%
\pgfsetbuttcap%
\pgfsetmiterjoin%
\definecolor{currentfill}{rgb}{1.000000,1.000000,1.000000}%
\pgfsetfillcolor{currentfill}%
\pgfsetfillopacity{0.800000}%
\pgfsetlinewidth{1.003750pt}%
\definecolor{currentstroke}{rgb}{0.800000,0.800000,0.800000}%
\pgfsetstrokecolor{currentstroke}%
\pgfsetstrokeopacity{0.800000}%
\pgfsetdash{}{0pt}%
\pgfpathmoveto{\pgfqpoint{2.766351in}{1.088812in}}%
\pgfpathlineto{\pgfqpoint{3.702778in}{1.088812in}}%
\pgfpathquadraticcurveto{\pgfqpoint{3.730556in}{1.088812in}}{\pgfqpoint{3.730556in}{1.116590in}}%
\pgfpathlineto{\pgfqpoint{3.730556in}{1.502778in}}%
\pgfpathquadraticcurveto{\pgfqpoint{3.730556in}{1.530556in}}{\pgfqpoint{3.702778in}{1.530556in}}%
\pgfpathlineto{\pgfqpoint{2.766351in}{1.530556in}}%
\pgfpathquadraticcurveto{\pgfqpoint{2.738574in}{1.530556in}}{\pgfqpoint{2.738574in}{1.502778in}}%
\pgfpathlineto{\pgfqpoint{2.738574in}{1.116590in}}%
\pgfpathquadraticcurveto{\pgfqpoint{2.738574in}{1.088812in}}{\pgfqpoint{2.766351in}{1.088812in}}%
\pgfpathclose%
\pgfusepath{stroke,fill}%
\end{pgfscope}%
\begin{pgfscope}%
\pgfsetbuttcap%
\pgfsetroundjoin%
\pgfsetlinewidth{1.505625pt}%
\definecolor{currentstroke}{rgb}{0.121569,0.466667,0.705882}%
\pgfsetstrokecolor{currentstroke}%
\pgfsetdash{{5.550000pt}{2.400000pt}}{0.000000pt}%
\pgfpathmoveto{\pgfqpoint{2.794129in}{1.426389in}}%
\pgfpathlineto{\pgfqpoint{3.071907in}{1.426389in}}%
\pgfusepath{stroke}%
\end{pgfscope}%
\begin{pgfscope}%
\definecolor{textcolor}{rgb}{0.000000,0.000000,0.000000}%
\pgfsetstrokecolor{textcolor}%
\pgfsetfillcolor{textcolor}%
\pgftext[x=3.183018in,y=1.377778in,left,base]{\color{textcolor}\rmfamily\fontsize{10.000000}{12.000000}\selectfont \(\displaystyle \gamma_{straight}\)}%
\end{pgfscope}%
\begin{pgfscope}%
\pgfsetrectcap%
\pgfsetroundjoin%
\pgfsetlinewidth{1.505625pt}%
\definecolor{currentstroke}{rgb}{1.000000,0.498039,0.054902}%
\pgfsetstrokecolor{currentstroke}%
\pgfsetdash{}{0pt}%
\pgfpathmoveto{\pgfqpoint{2.794129in}{1.219985in}}%
\pgfpathlineto{\pgfqpoint{3.071907in}{1.219985in}}%
\pgfusepath{stroke}%
\end{pgfscope}%
\begin{pgfscope}%
\definecolor{textcolor}{rgb}{0.000000,0.000000,0.000000}%
\pgfsetstrokecolor{textcolor}%
\pgfsetfillcolor{textcolor}%
\pgftext[x=3.183018in,y=1.171374in,left,base]{\color{textcolor}\rmfamily\fontsize{10.000000}{12.000000}\selectfont \(\displaystyle \gamma_{turn}\)}%
\end{pgfscope}%
\end{pgfpicture}%
\makeatother%
\endgroup%

%% file: imgs/sim/alt_2_veh1_actuator2_zoom_in.pgf
\begingroup%
\makeatletter%
\begin{pgfpicture}%
\pgfpathrectangle{\pgfpointorigin}{\pgfqpoint{3.800000in}{1.600000in}}%
\pgfusepath{use as bounding box, clip}%
\begin{pgfscope}%
\pgfsetbuttcap%
\pgfsetmiterjoin%
\definecolor{currentfill}{rgb}{1.000000,1.000000,1.000000}%
\pgfsetfillcolor{currentfill}%
\pgfsetlinewidth{0.000000pt}%
\definecolor{currentstroke}{rgb}{1.000000,1.000000,1.000000}%
\pgfsetstrokecolor{currentstroke}%
\pgfsetdash{}{0pt}%
\pgfpathmoveto{\pgfqpoint{0.000000in}{0.000000in}}%
\pgfpathlineto{\pgfqpoint{3.800000in}{0.000000in}}%
\pgfpathlineto{\pgfqpoint{3.800000in}{1.600000in}}%
\pgfpathlineto{\pgfqpoint{0.000000in}{1.600000in}}%
\pgfpathclose%
\pgfusepath{fill}%
\end{pgfscope}%
\begin{pgfscope}%
\pgfsetbuttcap%
\pgfsetmiterjoin%
\definecolor{currentfill}{rgb}{1.000000,1.000000,1.000000}%
\pgfsetfillcolor{currentfill}%
\pgfsetlinewidth{0.000000pt}%
\definecolor{currentstroke}{rgb}{0.000000,0.000000,0.000000}%
\pgfsetstrokecolor{currentstroke}%
\pgfsetstrokeopacity{0.000000}%
\pgfsetdash{}{0pt}%
\pgfpathmoveto{\pgfqpoint{0.646606in}{0.415000in}}%
\pgfpathlineto{\pgfqpoint{3.784051in}{0.415000in}}%
\pgfpathlineto{\pgfqpoint{3.784051in}{1.600000in}}%
\pgfpathlineto{\pgfqpoint{0.646606in}{1.600000in}}%
\pgfpathclose%
\pgfusepath{fill}%
\end{pgfscope}%
\begin{pgfscope}%
\pgfsetbuttcap%
\pgfsetroundjoin%
\definecolor{currentfill}{rgb}{0.000000,0.000000,0.000000}%
\pgfsetfillcolor{currentfill}%
\pgfsetlinewidth{0.803000pt}%
\definecolor{currentstroke}{rgb}{0.000000,0.000000,0.000000}%
\pgfsetstrokecolor{currentstroke}%
\pgfsetdash{}{0pt}%
\pgfsys@defobject{currentmarker}{\pgfqpoint{0.000000in}{-0.048611in}}{\pgfqpoint{0.000000in}{0.000000in}}{%
\pgfpathmoveto{\pgfqpoint{0.000000in}{0.000000in}}%
\pgfpathlineto{\pgfqpoint{0.000000in}{-0.048611in}}%
\pgfusepath{stroke,fill}%
}%
\begin{pgfscope}%
\pgfsys@transformshift{0.690864in}{0.415000in}%
\pgfsys@useobject{currentmarker}{}%
\end{pgfscope}%
\end{pgfscope}%
\begin{pgfscope}%
\definecolor{textcolor}{rgb}{0.000000,0.000000,0.000000}%
\pgfsetstrokecolor{textcolor}%
\pgfsetfillcolor{textcolor}%
\pgftext[x=0.690864in,y=0.317777in,,top]{\color{textcolor}\rmfamily\fontsize{10.000000}{12.000000}\selectfont \(\displaystyle {8.100}\)}%
\end{pgfscope}%
\begin{pgfscope}%
\pgfsetbuttcap%
\pgfsetroundjoin%
\definecolor{currentfill}{rgb}{0.000000,0.000000,0.000000}%
\pgfsetfillcolor{currentfill}%
\pgfsetlinewidth{0.803000pt}%
\definecolor{currentstroke}{rgb}{0.000000,0.000000,0.000000}%
\pgfsetstrokecolor{currentstroke}%
\pgfsetdash{}{0pt}%
\pgfsys@defobject{currentmarker}{\pgfqpoint{0.000000in}{-0.048611in}}{\pgfqpoint{0.000000in}{0.000000in}}{%
\pgfpathmoveto{\pgfqpoint{0.000000in}{0.000000in}}%
\pgfpathlineto{\pgfqpoint{0.000000in}{-0.048611in}}%
\pgfusepath{stroke,fill}%
}%
\begin{pgfscope}%
\pgfsys@transformshift{1.182627in}{0.415000in}%
\pgfsys@useobject{currentmarker}{}%
\end{pgfscope}%
\end{pgfscope}%
\begin{pgfscope}%
\definecolor{textcolor}{rgb}{0.000000,0.000000,0.000000}%
\pgfsetstrokecolor{textcolor}%
\pgfsetfillcolor{textcolor}%
\pgftext[x=1.182627in,y=0.317777in,,top]{\color{textcolor}\rmfamily\fontsize{10.000000}{12.000000}\selectfont \(\displaystyle {8.125}\)}%
\end{pgfscope}%
\begin{pgfscope}%
\pgfsetbuttcap%
\pgfsetroundjoin%
\definecolor{currentfill}{rgb}{0.000000,0.000000,0.000000}%
\pgfsetfillcolor{currentfill}%
\pgfsetlinewidth{0.803000pt}%
\definecolor{currentstroke}{rgb}{0.000000,0.000000,0.000000}%
\pgfsetstrokecolor{currentstroke}%
\pgfsetdash{}{0pt}%
\pgfsys@defobject{currentmarker}{\pgfqpoint{0.000000in}{-0.048611in}}{\pgfqpoint{0.000000in}{0.000000in}}{%
\pgfpathmoveto{\pgfqpoint{0.000000in}{0.000000in}}%
\pgfpathlineto{\pgfqpoint{0.000000in}{-0.048611in}}%
\pgfusepath{stroke,fill}%
}%
\begin{pgfscope}%
\pgfsys@transformshift{1.674390in}{0.415000in}%
\pgfsys@useobject{currentmarker}{}%
\end{pgfscope}%
\end{pgfscope}%
\begin{pgfscope}%
\definecolor{textcolor}{rgb}{0.000000,0.000000,0.000000}%
\pgfsetstrokecolor{textcolor}%
\pgfsetfillcolor{textcolor}%
\pgftext[x=1.674390in,y=0.317777in,,top]{\color{textcolor}\rmfamily\fontsize{10.000000}{12.000000}\selectfont \(\displaystyle {8.150}\)}%
\end{pgfscope}%
\begin{pgfscope}%
\pgfsetbuttcap%
\pgfsetroundjoin%
\definecolor{currentfill}{rgb}{0.000000,0.000000,0.000000}%
\pgfsetfillcolor{currentfill}%
\pgfsetlinewidth{0.803000pt}%
\definecolor{currentstroke}{rgb}{0.000000,0.000000,0.000000}%
\pgfsetstrokecolor{currentstroke}%
\pgfsetdash{}{0pt}%
\pgfsys@defobject{currentmarker}{\pgfqpoint{0.000000in}{-0.048611in}}{\pgfqpoint{0.000000in}{0.000000in}}{%
\pgfpathmoveto{\pgfqpoint{0.000000in}{0.000000in}}%
\pgfpathlineto{\pgfqpoint{0.000000in}{-0.048611in}}%
\pgfusepath{stroke,fill}%
}%
\begin{pgfscope}%
\pgfsys@transformshift{2.166152in}{0.415000in}%
\pgfsys@useobject{currentmarker}{}%
\end{pgfscope}%
\end{pgfscope}%
\begin{pgfscope}%
\definecolor{textcolor}{rgb}{0.000000,0.000000,0.000000}%
\pgfsetstrokecolor{textcolor}%
\pgfsetfillcolor{textcolor}%
\pgftext[x=2.166152in,y=0.317777in,,top]{\color{textcolor}\rmfamily\fontsize{10.000000}{12.000000}\selectfont \(\displaystyle {8.175}\)}%
\end{pgfscope}%
\begin{pgfscope}%
\pgfsetbuttcap%
\pgfsetroundjoin%
\definecolor{currentfill}{rgb}{0.000000,0.000000,0.000000}%
\pgfsetfillcolor{currentfill}%
\pgfsetlinewidth{0.803000pt}%
\definecolor{currentstroke}{rgb}{0.000000,0.000000,0.000000}%
\pgfsetstrokecolor{currentstroke}%
\pgfsetdash{}{0pt}%
\pgfsys@defobject{currentmarker}{\pgfqpoint{0.000000in}{-0.048611in}}{\pgfqpoint{0.000000in}{0.000000in}}{%
\pgfpathmoveto{\pgfqpoint{0.000000in}{0.000000in}}%
\pgfpathlineto{\pgfqpoint{0.000000in}{-0.048611in}}%
\pgfusepath{stroke,fill}%
}%
\begin{pgfscope}%
\pgfsys@transformshift{2.657915in}{0.415000in}%
\pgfsys@useobject{currentmarker}{}%
\end{pgfscope}%
\end{pgfscope}%
\begin{pgfscope}%
\definecolor{textcolor}{rgb}{0.000000,0.000000,0.000000}%
\pgfsetstrokecolor{textcolor}%
\pgfsetfillcolor{textcolor}%
\pgftext[x=2.657915in,y=0.317777in,,top]{\color{textcolor}\rmfamily\fontsize{10.000000}{12.000000}\selectfont \(\displaystyle {8.200}\)}%
\end{pgfscope}%
\begin{pgfscope}%
\pgfsetbuttcap%
\pgfsetroundjoin%
\definecolor{currentfill}{rgb}{0.000000,0.000000,0.000000}%
\pgfsetfillcolor{currentfill}%
\pgfsetlinewidth{0.803000pt}%
\definecolor{currentstroke}{rgb}{0.000000,0.000000,0.000000}%
\pgfsetstrokecolor{currentstroke}%
\pgfsetdash{}{0pt}%
\pgfsys@defobject{currentmarker}{\pgfqpoint{0.000000in}{-0.048611in}}{\pgfqpoint{0.000000in}{0.000000in}}{%
\pgfpathmoveto{\pgfqpoint{0.000000in}{0.000000in}}%
\pgfpathlineto{\pgfqpoint{0.000000in}{-0.048611in}}%
\pgfusepath{stroke,fill}%
}%
\begin{pgfscope}%
\pgfsys@transformshift{3.149678in}{0.415000in}%
\pgfsys@useobject{currentmarker}{}%
\end{pgfscope}%
\end{pgfscope}%
\begin{pgfscope}%
\definecolor{textcolor}{rgb}{0.000000,0.000000,0.000000}%
\pgfsetstrokecolor{textcolor}%
\pgfsetfillcolor{textcolor}%
\pgftext[x=3.149678in,y=0.317777in,,top]{\color{textcolor}\rmfamily\fontsize{10.000000}{12.000000}\selectfont \(\displaystyle {8.225}\)}%
\end{pgfscope}%
\begin{pgfscope}%
\pgfsetbuttcap%
\pgfsetroundjoin%
\definecolor{currentfill}{rgb}{0.000000,0.000000,0.000000}%
\pgfsetfillcolor{currentfill}%
\pgfsetlinewidth{0.803000pt}%
\definecolor{currentstroke}{rgb}{0.000000,0.000000,0.000000}%
\pgfsetstrokecolor{currentstroke}%
\pgfsetdash{}{0pt}%
\pgfsys@defobject{currentmarker}{\pgfqpoint{0.000000in}{-0.048611in}}{\pgfqpoint{0.000000in}{0.000000in}}{%
\pgfpathmoveto{\pgfqpoint{0.000000in}{0.000000in}}%
\pgfpathlineto{\pgfqpoint{0.000000in}{-0.048611in}}%
\pgfusepath{stroke,fill}%
}%
\begin{pgfscope}%
\pgfsys@transformshift{3.641440in}{0.415000in}%
\pgfsys@useobject{currentmarker}{}%
\end{pgfscope}%
\end{pgfscope}%
\begin{pgfscope}%
\definecolor{textcolor}{rgb}{0.000000,0.000000,0.000000}%
\pgfsetstrokecolor{textcolor}%
\pgfsetfillcolor{textcolor}%
\pgftext[x=3.641440in,y=0.317777in,,top]{\color{textcolor}\rmfamily\fontsize{10.000000}{12.000000}\selectfont \(\displaystyle {8.250}\)}%
\end{pgfscope}%
\begin{pgfscope}%
\definecolor{textcolor}{rgb}{0.000000,0.000000,0.000000}%
\pgfsetstrokecolor{textcolor}%
\pgfsetfillcolor{textcolor}%
\pgftext[x=2.215329in,y=0.138889in,,top]{\color{textcolor}\rmfamily\fontsize{10.000000}{12.000000}\selectfont time (seconds)}%
\end{pgfscope}%
\begin{pgfscope}%
\pgfsetbuttcap%
\pgfsetroundjoin%
\definecolor{currentfill}{rgb}{0.000000,0.000000,0.000000}%
\pgfsetfillcolor{currentfill}%
\pgfsetlinewidth{0.803000pt}%
\definecolor{currentstroke}{rgb}{0.000000,0.000000,0.000000}%
\pgfsetstrokecolor{currentstroke}%
\pgfsetdash{}{0pt}%
\pgfsys@defobject{currentmarker}{\pgfqpoint{-0.048611in}{0.000000in}}{\pgfqpoint{-0.000000in}{0.000000in}}{%
\pgfpathmoveto{\pgfqpoint{-0.000000in}{0.000000in}}%
\pgfpathlineto{\pgfqpoint{-0.048611in}{0.000000in}}%
\pgfusepath{stroke,fill}%
}%
\begin{pgfscope}%
\pgfsys@transformshift{0.646606in}{0.445153in}%
\pgfsys@useobject{currentmarker}{}%
\end{pgfscope}%
\end{pgfscope}%
\begin{pgfscope}%
\definecolor{textcolor}{rgb}{0.000000,0.000000,0.000000}%
\pgfsetstrokecolor{textcolor}%
\pgfsetfillcolor{textcolor}%
\pgftext[x=0.194444in, y=0.396958in, left, base]{\color{textcolor}\rmfamily\fontsize{10.000000}{12.000000}\selectfont \(\displaystyle {-0.75}\)}%
\end{pgfscope}%
\begin{pgfscope}%
\pgfsetbuttcap%
\pgfsetroundjoin%
\definecolor{currentfill}{rgb}{0.000000,0.000000,0.000000}%
\pgfsetfillcolor{currentfill}%
\pgfsetlinewidth{0.803000pt}%
\definecolor{currentstroke}{rgb}{0.000000,0.000000,0.000000}%
\pgfsetstrokecolor{currentstroke}%
\pgfsetdash{}{0pt}%
\pgfsys@defobject{currentmarker}{\pgfqpoint{-0.048611in}{0.000000in}}{\pgfqpoint{-0.000000in}{0.000000in}}{%
\pgfpathmoveto{\pgfqpoint{-0.000000in}{0.000000in}}%
\pgfpathlineto{\pgfqpoint{-0.048611in}{0.000000in}}%
\pgfusepath{stroke,fill}%
}%
\begin{pgfscope}%
\pgfsys@transformshift{0.646606in}{0.787493in}%
\pgfsys@useobject{currentmarker}{}%
\end{pgfscope}%
\end{pgfscope}%
\begin{pgfscope}%
\definecolor{textcolor}{rgb}{0.000000,0.000000,0.000000}%
\pgfsetstrokecolor{textcolor}%
\pgfsetfillcolor{textcolor}%
\pgftext[x=0.194444in, y=0.739299in, left, base]{\color{textcolor}\rmfamily\fontsize{10.000000}{12.000000}\selectfont \(\displaystyle {-0.50}\)}%
\end{pgfscope}%
\begin{pgfscope}%
\pgfsetbuttcap%
\pgfsetroundjoin%
\definecolor{currentfill}{rgb}{0.000000,0.000000,0.000000}%
\pgfsetfillcolor{currentfill}%
\pgfsetlinewidth{0.803000pt}%
\definecolor{currentstroke}{rgb}{0.000000,0.000000,0.000000}%
\pgfsetstrokecolor{currentstroke}%
\pgfsetdash{}{0pt}%
\pgfsys@defobject{currentmarker}{\pgfqpoint{-0.048611in}{0.000000in}}{\pgfqpoint{-0.000000in}{0.000000in}}{%
\pgfpathmoveto{\pgfqpoint{-0.000000in}{0.000000in}}%
\pgfpathlineto{\pgfqpoint{-0.048611in}{0.000000in}}%
\pgfusepath{stroke,fill}%
}%
\begin{pgfscope}%
\pgfsys@transformshift{0.646606in}{1.129834in}%
\pgfsys@useobject{currentmarker}{}%
\end{pgfscope}%
\end{pgfscope}%
\begin{pgfscope}%
\definecolor{textcolor}{rgb}{0.000000,0.000000,0.000000}%
\pgfsetstrokecolor{textcolor}%
\pgfsetfillcolor{textcolor}%
\pgftext[x=0.194444in, y=1.081640in, left, base]{\color{textcolor}\rmfamily\fontsize{10.000000}{12.000000}\selectfont \(\displaystyle {-0.25}\)}%
\end{pgfscope}%
\begin{pgfscope}%
\pgfsetbuttcap%
\pgfsetroundjoin%
\definecolor{currentfill}{rgb}{0.000000,0.000000,0.000000}%
\pgfsetfillcolor{currentfill}%
\pgfsetlinewidth{0.803000pt}%
\definecolor{currentstroke}{rgb}{0.000000,0.000000,0.000000}%
\pgfsetstrokecolor{currentstroke}%
\pgfsetdash{}{0pt}%
\pgfsys@defobject{currentmarker}{\pgfqpoint{-0.048611in}{0.000000in}}{\pgfqpoint{-0.000000in}{0.000000in}}{%
\pgfpathmoveto{\pgfqpoint{-0.000000in}{0.000000in}}%
\pgfpathlineto{\pgfqpoint{-0.048611in}{0.000000in}}%
\pgfusepath{stroke,fill}%
}%
\begin{pgfscope}%
\pgfsys@transformshift{0.646606in}{1.472175in}%
\pgfsys@useobject{currentmarker}{}%
\end{pgfscope}%
\end{pgfscope}%
\begin{pgfscope}%
\definecolor{textcolor}{rgb}{0.000000,0.000000,0.000000}%
\pgfsetstrokecolor{textcolor}%
\pgfsetfillcolor{textcolor}%
\pgftext[x=0.302469in, y=1.423980in, left, base]{\color{textcolor}\rmfamily\fontsize{10.000000}{12.000000}\selectfont \(\displaystyle {0.00}\)}%
\end{pgfscope}%
\begin{pgfscope}%
\definecolor{textcolor}{rgb}{0.000000,0.000000,0.000000}%
\pgfsetstrokecolor{textcolor}%
\pgfsetfillcolor{textcolor}%
\pgftext[x=0.138889in,y=1.007500in,,bottom,rotate=90.000000]{\color{textcolor}\rmfamily\fontsize{10.000000}{12.000000}\selectfont \(\displaystyle \zeta\) (m/s)}%
\end{pgfscope}%
\begin{pgfscope}%
\pgfpathrectangle{\pgfqpoint{0.646606in}{0.415000in}}{\pgfqpoint{3.137446in}{1.185000in}}%
\pgfusepath{clip}%
\pgfsetbuttcap%
\pgfsetroundjoin%
\definecolor{currentfill}{rgb}{0.121569,0.466667,0.705882}%
\pgfsetfillcolor{currentfill}%
\pgfsetlinewidth{1.003750pt}%
\definecolor{currentstroke}{rgb}{0.121569,0.466667,0.705882}%
\pgfsetstrokecolor{currentstroke}%
\pgfsetdash{}{0pt}%
\pgfsys@defobject{currentmarker}{\pgfqpoint{-0.041667in}{-0.041667in}}{\pgfqpoint{0.041667in}{0.041667in}}{%
\pgfpathmoveto{\pgfqpoint{0.000000in}{-0.041667in}}%
\pgfpathcurveto{\pgfqpoint{0.011050in}{-0.041667in}}{\pgfqpoint{0.021649in}{-0.037276in}}{\pgfqpoint{0.029463in}{-0.029463in}}%
\pgfpathcurveto{\pgfqpoint{0.037276in}{-0.021649in}}{\pgfqpoint{0.041667in}{-0.011050in}}{\pgfqpoint{0.041667in}{0.000000in}}%
\pgfpathcurveto{\pgfqpoint{0.041667in}{0.011050in}}{\pgfqpoint{0.037276in}{0.021649in}}{\pgfqpoint{0.029463in}{0.029463in}}%
\pgfpathcurveto{\pgfqpoint{0.021649in}{0.037276in}}{\pgfqpoint{0.011050in}{0.041667in}}{\pgfqpoint{0.000000in}{0.041667in}}%
\pgfpathcurveto{\pgfqpoint{-0.011050in}{0.041667in}}{\pgfqpoint{-0.021649in}{0.037276in}}{\pgfqpoint{-0.029463in}{0.029463in}}%
\pgfpathcurveto{\pgfqpoint{-0.037276in}{0.021649in}}{\pgfqpoint{-0.041667in}{0.011050in}}{\pgfqpoint{-0.041667in}{0.000000in}}%
\pgfpathcurveto{\pgfqpoint{-0.041667in}{-0.011050in}}{\pgfqpoint{-0.037276in}{-0.021649in}}{\pgfqpoint{-0.029463in}{-0.029463in}}%
\pgfpathcurveto{\pgfqpoint{-0.021649in}{-0.037276in}}{\pgfqpoint{-0.011050in}{-0.041667in}}{\pgfqpoint{0.000000in}{-0.041667in}}%
\pgfpathclose%
\pgfusepath{stroke,fill}%
}%
\begin{pgfscope}%
\pgfsys@transformshift{0.789217in}{1.492875in}%
\pgfsys@useobject{currentmarker}{}%
\end{pgfscope}%
\begin{pgfscope}%
\pgfsys@transformshift{0.887569in}{1.491170in}%
\pgfsys@useobject{currentmarker}{}%
\end{pgfscope}%
\begin{pgfscope}%
\pgfsys@transformshift{0.985922in}{1.489324in}%
\pgfsys@useobject{currentmarker}{}%
\end{pgfscope}%
\begin{pgfscope}%
\pgfsys@transformshift{1.084274in}{1.487316in}%
\pgfsys@useobject{currentmarker}{}%
\end{pgfscope}%
\begin{pgfscope}%
\pgfsys@transformshift{1.182627in}{1.485125in}%
\pgfsys@useobject{currentmarker}{}%
\end{pgfscope}%
\begin{pgfscope}%
\pgfsys@transformshift{1.280979in}{1.482724in}%
\pgfsys@useobject{currentmarker}{}%
\end{pgfscope}%
\begin{pgfscope}%
\pgfsys@transformshift{1.379332in}{1.480082in}%
\pgfsys@useobject{currentmarker}{}%
\end{pgfscope}%
\begin{pgfscope}%
\pgfsys@transformshift{1.477685in}{1.477161in}%
\pgfsys@useobject{currentmarker}{}%
\end{pgfscope}%
\begin{pgfscope}%
\pgfsys@transformshift{1.576037in}{1.473912in}%
\pgfsys@useobject{currentmarker}{}%
\end{pgfscope}%
\begin{pgfscope}%
\pgfsys@transformshift{1.674390in}{1.470279in}%
\pgfsys@useobject{currentmarker}{}%
\end{pgfscope}%
\begin{pgfscope}%
\pgfsys@transformshift{1.772742in}{1.466189in}%
\pgfsys@useobject{currentmarker}{}%
\end{pgfscope}%
\begin{pgfscope}%
\pgfsys@transformshift{1.871095in}{1.461552in}%
\pgfsys@useobject{currentmarker}{}%
\end{pgfscope}%
\begin{pgfscope}%
\pgfsys@transformshift{1.969447in}{1.456251in}%
\pgfsys@useobject{currentmarker}{}%
\end{pgfscope}%
\begin{pgfscope}%
\pgfsys@transformshift{2.067800in}{1.450139in}%
\pgfsys@useobject{currentmarker}{}%
\end{pgfscope}%
\begin{pgfscope}%
\pgfsys@transformshift{2.166152in}{1.443018in}%
\pgfsys@useobject{currentmarker}{}%
\end{pgfscope}%
\begin{pgfscope}%
\pgfsys@transformshift{2.264505in}{1.434630in}%
\pgfsys@useobject{currentmarker}{}%
\end{pgfscope}%
\begin{pgfscope}%
\pgfsys@transformshift{2.362857in}{1.424629in}%
\pgfsys@useobject{currentmarker}{}%
\end{pgfscope}%
\begin{pgfscope}%
\pgfsys@transformshift{2.461210in}{1.412547in}%
\pgfsys@useobject{currentmarker}{}%
\end{pgfscope}%
\begin{pgfscope}%
\pgfsys@transformshift{2.559562in}{1.397769in}%
\pgfsys@useobject{currentmarker}{}%
\end{pgfscope}%
\begin{pgfscope}%
\pgfsys@transformshift{2.657915in}{1.379537in}%
\pgfsys@useobject{currentmarker}{}%
\end{pgfscope}%
\begin{pgfscope}%
\pgfsys@transformshift{2.756267in}{1.357173in}%
\pgfsys@useobject{currentmarker}{}%
\end{pgfscope}%
\begin{pgfscope}%
\pgfsys@transformshift{2.854620in}{1.331295in}%
\pgfsys@useobject{currentmarker}{}%
\end{pgfscope}%
\begin{pgfscope}%
\pgfsys@transformshift{2.952972in}{1.309489in}%
\pgfsys@useobject{currentmarker}{}%
\end{pgfscope}%
\begin{pgfscope}%
\pgfsys@transformshift{3.051325in}{1.393421in}%
\pgfsys@useobject{currentmarker}{}%
\end{pgfscope}%
\begin{pgfscope}%
\pgfsys@transformshift{3.149678in}{1.535555in}%
\pgfsys@useobject{currentmarker}{}%
\end{pgfscope}%
\begin{pgfscope}%
\pgfsys@transformshift{3.248030in}{1.535595in}%
\pgfsys@useobject{currentmarker}{}%
\end{pgfscope}%
\begin{pgfscope}%
\pgfsys@transformshift{3.346383in}{1.535563in}%
\pgfsys@useobject{currentmarker}{}%
\end{pgfscope}%
\begin{pgfscope}%
\pgfsys@transformshift{3.444735in}{1.535531in}%
\pgfsys@useobject{currentmarker}{}%
\end{pgfscope}%
\begin{pgfscope}%
\pgfsys@transformshift{3.543088in}{1.535500in}%
\pgfsys@useobject{currentmarker}{}%
\end{pgfscope}%
\begin{pgfscope}%
\pgfsys@transformshift{3.641440in}{1.535468in}%
\pgfsys@useobject{currentmarker}{}%
\end{pgfscope}%
\end{pgfscope}%
\begin{pgfscope}%
\pgfpathrectangle{\pgfqpoint{0.646606in}{0.415000in}}{\pgfqpoint{3.137446in}{1.185000in}}%
\pgfusepath{clip}%
\pgfsetbuttcap%
\pgfsetroundjoin%
\definecolor{currentfill}{rgb}{1.000000,0.498039,0.054902}%
\pgfsetfillcolor{currentfill}%
\pgfsetlinewidth{1.003750pt}%
\definecolor{currentstroke}{rgb}{1.000000,0.498039,0.054902}%
\pgfsetstrokecolor{currentstroke}%
\pgfsetdash{}{0pt}%
\pgfsys@defobject{currentmarker}{\pgfqpoint{-0.041667in}{-0.041667in}}{\pgfqpoint{0.041667in}{0.041667in}}{%
\pgfpathmoveto{\pgfqpoint{-0.041667in}{-0.041667in}}%
\pgfpathlineto{\pgfqpoint{0.041667in}{0.041667in}}%
\pgfpathmoveto{\pgfqpoint{-0.041667in}{0.041667in}}%
\pgfpathlineto{\pgfqpoint{0.041667in}{-0.041667in}}%
\pgfusepath{stroke,fill}%
}%
\begin{pgfscope}%
\pgfsys@transformshift{0.789217in}{1.077936in}%
\pgfsys@useobject{currentmarker}{}%
\end{pgfscope}%
\begin{pgfscope}%
\pgfsys@transformshift{0.887569in}{1.054954in}%
\pgfsys@useobject{currentmarker}{}%
\end{pgfscope}%
\begin{pgfscope}%
\pgfsys@transformshift{0.985922in}{1.029865in}%
\pgfsys@useobject{currentmarker}{}%
\end{pgfscope}%
\begin{pgfscope}%
\pgfsys@transformshift{1.084274in}{1.002377in}%
\pgfsys@useobject{currentmarker}{}%
\end{pgfscope}%
\begin{pgfscope}%
\pgfsys@transformshift{1.182627in}{0.972146in}%
\pgfsys@useobject{currentmarker}{}%
\end{pgfscope}%
\begin{pgfscope}%
\pgfsys@transformshift{1.280979in}{0.938757in}%
\pgfsys@useobject{currentmarker}{}%
\end{pgfscope}%
\begin{pgfscope}%
\pgfsys@transformshift{1.379332in}{0.901715in}%
\pgfsys@useobject{currentmarker}{}%
\end{pgfscope}%
\begin{pgfscope}%
\pgfsys@transformshift{1.477685in}{0.860423in}%
\pgfsys@useobject{currentmarker}{}%
\end{pgfscope}%
\begin{pgfscope}%
\pgfsys@transformshift{1.576037in}{0.814153in}%
\pgfsys@useobject{currentmarker}{}%
\end{pgfscope}%
\begin{pgfscope}%
\pgfsys@transformshift{1.674390in}{0.762017in}%
\pgfsys@useobject{currentmarker}{}%
\end{pgfscope}%
\begin{pgfscope}%
\pgfsys@transformshift{1.772742in}{0.702925in}%
\pgfsys@useobject{currentmarker}{}%
\end{pgfscope}%
\begin{pgfscope}%
\pgfsys@transformshift{1.871095in}{0.635532in}%
\pgfsys@useobject{currentmarker}{}%
\end{pgfscope}%
\begin{pgfscope}%
\pgfsys@transformshift{1.969447in}{0.558185in}%
\pgfsys@useobject{currentmarker}{}%
\end{pgfscope}%
\begin{pgfscope}%
\pgfsys@transformshift{2.067800in}{0.468863in}%
\pgfsys@useobject{currentmarker}{}%
\end{pgfscope}%
\begin{pgfscope}%
\pgfsys@transformshift{2.166152in}{0.801826in}%
\pgfsys@useobject{currentmarker}{}%
\end{pgfscope}%
\begin{pgfscope}%
\pgfsys@transformshift{2.264505in}{1.159667in}%
\pgfsys@useobject{currentmarker}{}%
\end{pgfscope}%
\begin{pgfscope}%
\pgfsys@transformshift{2.362857in}{1.514821in}%
\pgfsys@useobject{currentmarker}{}%
\end{pgfscope}%
\begin{pgfscope}%
\pgfsys@transformshift{2.461210in}{1.546136in}%
\pgfsys@useobject{currentmarker}{}%
\end{pgfscope}%
\begin{pgfscope}%
\pgfsys@transformshift{2.559562in}{1.546115in}%
\pgfsys@useobject{currentmarker}{}%
\end{pgfscope}%
\begin{pgfscope}%
\pgfsys@transformshift{2.657915in}{1.546078in}%
\pgfsys@useobject{currentmarker}{}%
\end{pgfscope}%
\begin{pgfscope}%
\pgfsys@transformshift{2.756267in}{1.546041in}%
\pgfsys@useobject{currentmarker}{}%
\end{pgfscope}%
\begin{pgfscope}%
\pgfsys@transformshift{2.854620in}{1.546004in}%
\pgfsys@useobject{currentmarker}{}%
\end{pgfscope}%
\begin{pgfscope}%
\pgfsys@transformshift{2.952972in}{1.545967in}%
\pgfsys@useobject{currentmarker}{}%
\end{pgfscope}%
\begin{pgfscope}%
\pgfsys@transformshift{3.051325in}{1.545930in}%
\pgfsys@useobject{currentmarker}{}%
\end{pgfscope}%
\begin{pgfscope}%
\pgfsys@transformshift{3.149678in}{1.545893in}%
\pgfsys@useobject{currentmarker}{}%
\end{pgfscope}%
\begin{pgfscope}%
\pgfsys@transformshift{3.248030in}{1.545856in}%
\pgfsys@useobject{currentmarker}{}%
\end{pgfscope}%
\begin{pgfscope}%
\pgfsys@transformshift{3.346383in}{1.545820in}%
\pgfsys@useobject{currentmarker}{}%
\end{pgfscope}%
\begin{pgfscope}%
\pgfsys@transformshift{3.444735in}{1.545783in}%
\pgfsys@useobject{currentmarker}{}%
\end{pgfscope}%
\begin{pgfscope}%
\pgfsys@transformshift{3.543088in}{1.545746in}%
\pgfsys@useobject{currentmarker}{}%
\end{pgfscope}%
\begin{pgfscope}%
\pgfsys@transformshift{3.641440in}{1.545709in}%
\pgfsys@useobject{currentmarker}{}%
\end{pgfscope}%
\end{pgfscope}%
\begin{pgfscope}%
\pgfsetrectcap%
\pgfsetmiterjoin%
\pgfsetlinewidth{0.803000pt}%
\definecolor{currentstroke}{rgb}{0.000000,0.000000,0.000000}%
\pgfsetstrokecolor{currentstroke}%
\pgfsetdash{}{0pt}%
\pgfpathmoveto{\pgfqpoint{0.646606in}{0.415000in}}%
\pgfpathlineto{\pgfqpoint{0.646606in}{1.600000in}}%
\pgfusepath{stroke}%
\end{pgfscope}%
\begin{pgfscope}%
\pgfsetrectcap%
\pgfsetmiterjoin%
\pgfsetlinewidth{0.803000pt}%
\definecolor{currentstroke}{rgb}{0.000000,0.000000,0.000000}%
\pgfsetstrokecolor{currentstroke}%
\pgfsetdash{}{0pt}%
\pgfpathmoveto{\pgfqpoint{3.784051in}{0.415000in}}%
\pgfpathlineto{\pgfqpoint{3.784051in}{1.600000in}}%
\pgfusepath{stroke}%
\end{pgfscope}%
\begin{pgfscope}%
\pgfsetrectcap%
\pgfsetmiterjoin%
\pgfsetlinewidth{0.803000pt}%
\definecolor{currentstroke}{rgb}{0.000000,0.000000,0.000000}%
\pgfsetstrokecolor{currentstroke}%
\pgfsetdash{}{0pt}%
\pgfpathmoveto{\pgfqpoint{0.646606in}{0.415000in}}%
\pgfpathlineto{\pgfqpoint{3.784051in}{0.415000in}}%
\pgfusepath{stroke}%
\end{pgfscope}%
\begin{pgfscope}%
\pgfsetrectcap%
\pgfsetmiterjoin%
\pgfsetlinewidth{0.803000pt}%
\definecolor{currentstroke}{rgb}{0.000000,0.000000,0.000000}%
\pgfsetstrokecolor{currentstroke}%
\pgfsetdash{}{0pt}%
\pgfpathmoveto{\pgfqpoint{0.646606in}{1.600000in}}%
\pgfpathlineto{\pgfqpoint{3.784051in}{1.600000in}}%
\pgfusepath{stroke}%
\end{pgfscope}%
\end{pgfpicture}%
\makeatother%
\endgroup%

%% file: compose.tex
\section{COMPOSITION OF MULTIPLE SAFETY CONSTRAINTS}
\label{sec_composition}

\subsection{Motivating Example}
\label{sec_composition_motivating_example}

\begin{figure*}
\begin{center}
\begin{subfigure}{0.3\textwidth}
\begin{center}
\def\svgwidth{0.99\textwidth}
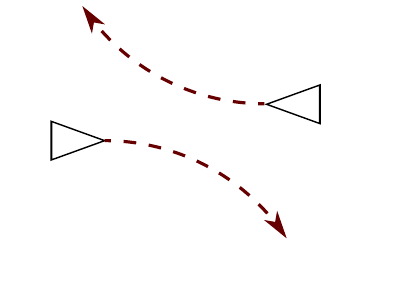
\caption{ }
\label{fig_three_veh_a}
\end{center}
\end{subfigure}
\begin{subfigure}{0.3\textwidth}
\begin{center}
\def\svgwidth{0.99\textwidth}
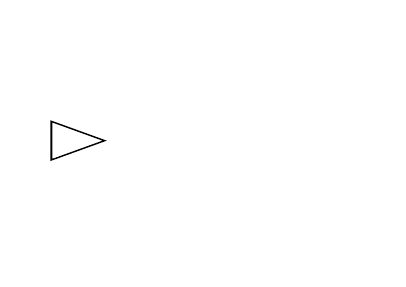
\caption{ }
\end{center}
\end{subfigure}
\begin{subfigure}{0.3\textwidth}
\begin{center}
\def\svgwidth{0.99\textwidth}
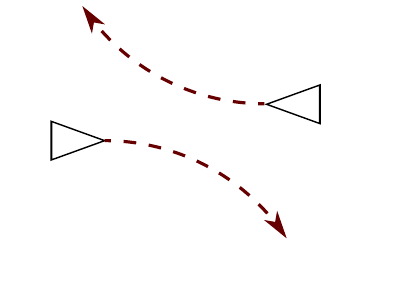
\caption{ }
\label{fig_three_veh_c}
\end{center}
\end{subfigure}
\end{center}
\caption{%
A geometric view of the example given in Section~\ref{sec_composition_motivating_example}.
}
\label{fig_three_veh}
\end{figure*}

\begin{figure*}
\begin{center}
\begin{subfigure}{0.3\textwidth}
\def\svgwidth{0.99\textwidth}
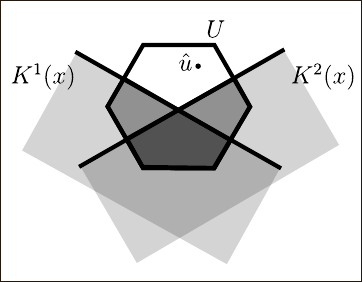
\caption{ }
\end{subfigure}
\begin{subfigure}{0.3\textwidth}
\def\svgwidth{0.99\textwidth}
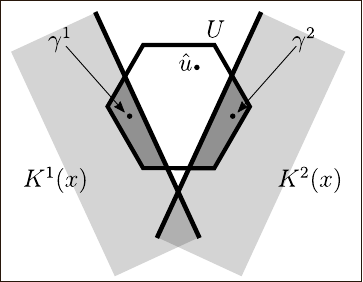
\caption{ }
\end{subfigure}
\begin{subfigure}{0.3\textwidth}
\def\svgwidth{0.99\textwidth}
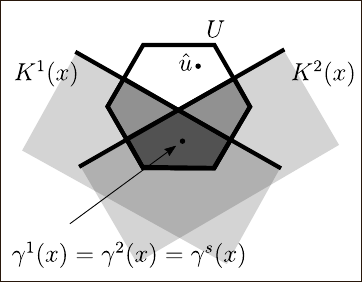
\caption{ }
\end{subfigure}
\end{center}
\caption{%
The shared evading maneuver ensures that
the same safe control input satisfies each safety constraint.
}
\label{fig_geometry_multiple}
\end{figure*}
Although the constructive method introduced in (\ref{eq_h}) can produce
a barrier function in the presence of actuator constraints
that ensures two vehicles do not collide, the formulation
does not extend immediately to collision avoidance for systems with more than
two vehicles.
To see this, 
we present a specific example where three UAVs with a collision avoidance
safety objective cannot use the results from Section \ref{sec_theory}
to ensure safety. A plot of this scenario is shown in Figure \ref{fig_three_veh}.
We index the vehicles by $i = 1, 2, 3$.
To ensure collision-free trajectories, 
and considering the safety function defined in (\ref{eq_rho_adjusted}),
three pairwise constraints must be nonnegative at all times:
\begin{IEEEeqnarray*}{rCl}
\rho^1(x)
&=& d_{1,2}(x) - 2\delta + \delta\sin(\theta_1) - \delta\cos(\theta_1)  - D_s^2, \\
\rho^2(x)
&=& d_{1,3}(x) - 2\delta + \delta\sin(\theta_1) - \delta\cos(\theta_1)  - D_s^2, \\
\rho^3(x)
&=& d_{2,3}(x) - 2\delta + \delta\sin(\theta_2) - \delta\cos(\theta_2)  - D_s^2.
\end{IEEEeqnarray*}
We now apply the results of Section \ref{sec_construction} to these constraints
and for simplicity, let $\delta$ be approximately $0$.
For each constraint, define an arbitrarily chosen nominal evading maneuver
\begin{IEEEeqnarray}{l}
\label{eq_evasive_maneuvers_not_shared}
\gamma^1(x) =
\begin{bmatrix} 1 & -1 & 0& 1 & -1 & 0& 1 & -1& 0 \end{bmatrix}^T \IEEEyessubnumber\\
\gamma^2(x) = \gamma^3(x) =
\begin{bmatrix} 1 & 1& 0 & 1 & 1& 0 & 1 & 1 & 0\end{bmatrix}^T.\IEEEyessubnumber
\label{eq_evasive_maneuvers_not_shared2}
\end{IEEEeqnarray}
In other words, $\gamma^1$ encodes an evasive maneuver where all the vehicles turn right while $\gamma^2$
and $\gamma^3$ encode a maneuver where all the vehicles turn left.
We note that $h^j$ ($j=1,\ldots,3$) defined in (\ref{eq_h}) and constructed from $\rho^j$ and $\gamma^j$ are ZCBFs.
In this example we let $v_{min} = 1$, $v_{max} = 2$, $\omega_{max} = 1$, and $D_s = 0.5$
so that 
the vehicles follow a circular trajectory with radius $r=1$
when applying $v_{min}$ and $\omega_{max}$.
Assume now that the vehicles have the following initial states
\begin{IEEEeqnarray*}{rCl}
x_1 &=& \begin{bmatrix} 0 & 0 & 0 & 0\end{bmatrix}^T, \\
x_2 &=& \begin{bmatrix} (2r + D_s)\sin\psi & (2r + D_s)\cos\psi - 2r & \pi& 0\end{bmatrix}^T, \\
x_3 &=& \begin{bmatrix} (2r + D_s)\sin\psi & 2r - (2r + D_s)\cos\psi & \pi& 0\end{bmatrix}^T,
\end{IEEEeqnarray*}
where $\psi = \arccos\left(\frac{D_s/2 + 2r}{2r + D_s}\right)$.
Then
$h^{1}(x) = h^{2}(x) = h^{3}(x) = 0$ and the barrier constraints in (\ref{eq_zcbf_condition})
for $h^{1}(x)$ and $h^{2}(x)$ become
\begin{IEEEeqnarray}{rCl}
-0.4(v_1 + \omega_1 + v_2 + \omega_2) &\ge& 0 \label{eq_ab_constraint}\\
0.4(-v_1 + \omega_1 - v_3 + \omega_3) &\ge& 0\label{eq_ac_constraint}.
\end{IEEEeqnarray}
Although $h^{1}$ and $h^{2}$
are ZCBFs, these two constraints cannot be simultaneously satisfied for
$v_i \in [v_{min},v_{max}]$
and $|\omega_i|\le \omega_{max}$.
In particular, after substituting the minimum velocity $v_1 = v_2 = 1$, the first equation dictates that $\omega_1 + \omega_2 \le -2$
(i.e., vehicles $1$ and $2$ must turn right).
Similarly, the second equation dictates that vehicle $1$ and $3$
must turn left.
The problem with
this scenario is that
vehicle 1 cannot simultaneously execute both nominal evading maneuvers
(i.e., turn both left and right at the same time).
To solve this problem, we will make sure that the
evasive maneuver applied by a vehicle is the same for every 
barrier function. A geometric view of the general
problem and its solution are shown in Figure~\ref{fig_geometry_multiple}.

\subsection{Sufficient Conditions for Satisfying Multiple Safety Constraints}

In order to solve the issues arising when vehicles have to simultaneously respect
multiple constraints, we now extend the use of the constructive 
technique introduced in (\ref{eq_h}).  
In this section we extend the reasoning of \cite{ames2017control}
to the case of $q$ constraints. Consider a nonlinear autonomous system
\begin{equation}
\dot{x} = f(x) 
\label{eq_nonlinear_system}
\end{equation}
where $f$ is locally Lipschitz. Then we have a similar definition to Definition~\ref{def_zcbf}
for autonomous systems.
\begin{definition}\cite{ames2017control}
Given a set $\mathcal{C}\subset \R^n$ defined in
(\ref{eq_safe}) for a continuously
differentiable function $h:\R^n\ra\R$, the function
$h$ is called a \emph{zeroing barrier function (ZBF)}
defined on an open set $\mathcal{D}$ with $\mathcal{C}\subset \mathcal{D}\subset \R^n$,
if there exists a Lipschitz continuous extended class $\mathcal{K}$ function $\alpha$
such that
\begin{equation}
L_fh(x) \ge -\alpha(h(x)), \:\forall x \in \mathcal{D}.
\label{eq_zbf_condition}
\end{equation}
\label{def_zbf}
\end{definition}

When there are $q$ constraints, we consider the case of $q$ barrier
functions where each barrier function is denoted $h^j$ on $\mathcal{D}^j$ with associated safe set $\mathcal{C}^j$
and admissible control space $K^j(x)$ for $x\in\mathcal{D}^j$
for $j\in\{1,\ldots,q\}$.
We are interested in the conditions under which all
safety constraints can be satisfied for all future times.
In other words, under the assumption that 
$x(0)\in \mathcal{C}^j$ we want to show that $x(t)\in\mathcal{C}^j$
for all $t \ge 0$.
Hence, we are interested in the forward invariance of
the intersection of all the safe sets, which motivates 
the following definitions
\begin{IEEEeqnarray}{rCl}
\mathcal{C}_{\cap} &=& \mathcal{C}^1\cap \mathcal{C}^2 \cap \cdots \cap \mathcal{C}^q, \label{eq_overall_safe_set} \\
K_\cap(x) &=& \set{u\in U}{u\in K^1(x)\cap K^2(x) \cap \cdots \cap K^q(x)}. \label{eq_kcap}
\end{IEEEeqnarray}
where $\mathcal{D}_\cap$ is an open superset of $\mathcal{C}_\cap$ and 
$x\in\mathcal{D}_\cap$.
We can now present a multiple constraint analogue of Theorem~\ref{th_fwd_invariance}
by following the same reasoning as \cite{ames2017control}.

\begin{proposition}
Given a dynamical system (\ref{eq_nonlinear_system}) and a set $\mathcal{C}_\cap$
defined by (\ref{eq_overall_safe_set}) for continuously
differentiable functions $h^j: R^n \to R$ where $h^j$ is a ZBF on $D^j$ with $\mathcal{C}^j
\subset D^j \subset R$ and $\pf{h^j(x)}{x} \ne 0$ for any $x\in\partial C_\cap$ where $h^j(x) = 0$,
then $\mathcal{C}_\cap$ is
forward invariant. 
\label{prop_multiple_fwd_inv_zbf}
\end{proposition}
\begin{proof}
The proof is the same as that for Proposition 1 of \cite{ames2017control},
namely $\dot{h}^j(x) = -\alpha(x) \ge 0$ for 
any $j$ such that $h^j(x) = 0$ so the result follows by Nagumo's Theorem \cite{blanchini2008set}.
We add the assumption that $\pf{h^j(x)}{x}$ is non-zero for all $x \in \partial \mathcal{C}_\cap$
such that $h^j(x) = 0$ to ensure that the tangent cone 
in Nagumo's Theorem is non-empty.
\end{proof}
Then for autonomous systems with dynamics (\ref{eq_system}), we have the following corollary of Theorem~\ref{th_fwd_invariance}.
\begin{corollary}
Given a dynamical system (\ref{eq_system}) and a set $\mathcal{C}_\cap$
defined by (\ref{eq_overall_safe_set}) for continuously
differentiable functions $h^j: R^n \to R$ where $h^j$ is a ZCBF on $D^j$
and $\pf{h^j(x)}{x} \ne 0$ for any $x\in\partial\mathcal{C}_\cap$ where $h^j(x) = 0$, then
any Lipschitz continuous controller $u:\mathcal{D}_\cap\ra U$
such that $u(x)\in K_\cap(x)$ 
will render the set $\mathcal{C}_{\cap}$ forward invariant.
\label{cor_multiple_fwd_inv}
\end{corollary}

\subsection{The Shared Nominal Evading Maneuver Assumption}

Suppose there are $q$ constraints $\rho^j:\mathcal{D}^j \ra \R$ ($j=1,\ldots,q$)
that must be greater than or equal to 0 at all times.
For the $k$ agents with pairwise constraints $q = k(k-1)/2$.
We assume that for each constraint $j=1,\ldots,q$,
a locally Lipschitz nominal evading maneuver $\gamma^j$
has been selected using the framework
in (\ref{eq_h}).
An example for fixed-wing UAVs with collision avoidance 
safety constraints is given in (\ref{eq_gamma_turn}).
Given $q$ safety functions $\rho^j$
and evading maneuvers $\gamma^j$ for $j\in\{1,\ldots,q\}$,
we construct $q$ output functions $h^j$ defined on $\mathcal{D}^j$ similarly to
(\ref{eq_h}) where
\begin{IEEEeqnarray}{rCl}
h^j(x;\rho,\gamma) &=& \INF \rho^j(\hat{x}^j(\tau)), \label{eq_h_j}\\
\hat{x}^j(\tau) &=& x + \int_0^\tau \dot{\hat{x}}^j(\eta)d\eta, \\
\dot{\hat{x}}^j(\tau) &=& f(\hat{x}^j(\tau)) + g(\hat{x}^j(\tau))\gamma^j(\hat{x}^j(\tau)).
\end{IEEEeqnarray}

Section \ref{sec_composition_motivating_example}
showed an example where $K_\cap$ could be empty
for some $x\in\mathcal{C}_\cap$. As a result,
the assumptions of Corollary \ref{cor_multiple_fwd_inv}
could not be satisfied.
In order to address the issue discussed in Section \ref{sec_composition_motivating_example},
we introduce an additional constraint on $\gamma^j$
($j=1,\ldots,q$) that all $h^j$ are constructed from
the same nominal evading maneuver.
\begin{assumption}
Given a dynamical system (\ref{eq_system})
and $q$ output functions $h^j$ defined in (\ref{eq_h_j})
for given safety functions $\rho^j$ and evading maneuvers $\gamma^j$
for $j\in\{1,\ldots,q\}$,
the
\emph{shared evading maneuver assumption} holds
if 
$\gamma^1(x) = \cdots = \gamma^q(x)$
for all $x\in \mathcal{D}_\cap$.
The \emph{shared evading maneuver} is denoted $\gamma^s$ 
so that 
\begin{equation}
\gamma^s(x) = \gamma^1(x) = \cdots = \gamma^q(x)
\label{eq_gamma_s}
\end{equation}
for all $x\in \mathcal{D}_\cap$.
\label{as_shared_evading_assumption}
\end{assumption}
\begin{remark}
This assumption requires
that each $h^j$ ($j=1,\ldots,q$)
be constructed from the same nominal evading maneuver.
Note, however, that this does not imply that each $h^j$
must be constructed from the same safety function.
\end{remark}
The example in 
Section \ref{sec_composition_motivating_example}
does not satisfy Assumption~\ref{as_shared_evading_assumption}
because $\gamma^1(x)$ and $\gamma^2(x)$
defined in (\ref{eq_evasive_maneuvers_not_shared})
are not the same. To enforce that the shared
evasive maneuver assumption holds, one option is to change $\gamma^1$
so that
\begin{equation}
\gamma^1(x) =
\begin{bmatrix} 1 & 1 & 0& 1 & 1& 0 & 1 & 1& 0 \end{bmatrix}^T.
\label{eq_evasive_maneuvers_fixed}
\end{equation}
In other words, using $\gamma^1$ defined in (\ref{eq_evasive_maneuvers_fixed})
and $\gamma^2$ and $\gamma^3$ in (\ref{eq_evasive_maneuvers_not_shared2})
implies an evasive maneuver where all vehicles turn left
for each constraint. Another example where the shared nominal evading maneuver assumption holds is as follows:
\begin{IEEEeqnarray*}{l}
\gamma^s(x) = \gamma^1(x) = \gamma^2(x) = \gamma^3(x) = \\
\qquad \qquad \begin{bmatrix} 1 & 1 & 0& 1.5 & 0 & 0& 2 & -1 & 0\end{bmatrix}^T.
\end{IEEEeqnarray*}
In this case,
$\gamma^s(x)$
encodes an evasive maneuver where vehicle $1$ turns
left with a linear velocity of $1$, vehicle $2$ stays
straight with a linear velocity of $1.5$,
and vehicle $3$ turns right with a linear velocity of $2$.
These three nominal evading maneuvers satisfy the
shared evasive maneuver assumption
because for all $x\in\mathcal{D}_\cap$, $\gamma^1(x) = \gamma^2(x) = \gamma^3(x)$.

To see the purpose of Assumption~\ref{as_shared_evading_assumption}, we first examine
the case of a single constraint.
In particular, let $h$ be defined
in (\ref{eq_h}) and consider the role
of $\gamma$ in establishing that $h$ is a ZCBF.
From Definition \ref{def_zcbf},
for $h$ to be used for a barrier function,
$K(x)$ must be nonempty for all $x\in\mathcal{D}$.
With $h$ defined as in (\ref{eq_h}),
this property is satisfied 
by $\gamma(x)$ or a perturbation of $\gamma(x)$ for all $x\in\mathcal{D}$ (see Theorem \ref{th_h_zcbf}).
The analogue condition for multiple constraints
is that $K_\cap(x)$
is non-empty 
for all $x\in\mathcal{D}_\cap$.
If each $h^j$ defined in (\ref{eq_h}) is a ZCBF and is constructed from
$\gamma^j$ then by similar reasoning to Theorem \ref{th_h_zcbf},
$\gamma^j(x)$ or a perturbation of $\gamma(x)$ is in $K^j(x)$ for all $x\in \mathcal{D}_\cap$.
This allows us to 
state a multiple constraint analogue to Theorem \ref{th_h_zcbf}.
In the following, we denote 
the inner product as $\langle L_gh^{j_1}(x), L_gh^{j_2}(x) \rangle$
for $j_1,j_2\in\{1,\ldots,q\}$.

\begin{theorem}
Given a dynamical system (\ref{eq_system})
and a set $\mathcal{C}_\cap \subset \mathcal{D}_\cap$ defined in (\ref{eq_overall_safe_set})
for $q$ continuously differentiable functions $h^j$
defined in (\ref{eq_h_j})
with safety functions $\rho^j$ and evading maneuvers $\gamma^j$ where $k\in\{1,\ldots,q\}$,
if $h^j$ is a ZCBF for $k\in\{1,\ldots,q\}$ and
Assumption~\ref{as_shared_evading_assumption} holds then
$K_\cap(x)$ is non-empty for all $x\in\mathcal{C}_\cap$.
If in addition, 
$\gamma^s$ defined in (\ref{eq_gamma_s}) maps to the interior of $U$
and for all $x\in\partial \mathcal{C}_\cap$,
$\langle L_gh^{j_1}(x), L_gh^{j_2}(x) \rangle > 0$ for $j_1 \ne j_2$ and $j_1,j_2\in \{1,\ldots,q\}$,
then there is an open set that is a superset of $\mathcal{C}_\cap$
for which $K_\cap(x)$ is non-empty for all $x$ in the open set.
\label{th_multiple_h_zcbf} 
\end{theorem}
\begin{proof}
To prove the first statement, note that it was shown in the proof of Theorem
\ref{th_h_zcbf} that $\gamma^s$ is in $K^j(x)$ for $j=1,\ldots,q$
and $x\in\mathcal{C}_\cap$.
To prove the second statement, note that we can use the same 
method as was used in the proof of Theorem \ref{th_h_zcbf}
to find a vector $d(z)$ such that $h^j(z)$
satisfies (\ref{eq_zcbf_condition})
for all $z\in B(x,\mu)$ given $x\in\partial \mathcal{C}_\cap$.
In particular, because 
$\langle L_gh^{j_1}(x), L_gh^{j_2}(x)\rangle > 0$,
$L_gh^{j}(x) \ne 0$ for $j=1,\ldots,q$
there exists a vector $d_{all}(x)$
such that
$\langle d_{all}(x), L_gh^{j}(x)\rangle > 0$.
We choose $d_{all}(x)$ with sufficiently small norm.
Using the notation
of the proof of Theorem \ref{th_h_zcbf},
for sufficiently small $\mu$,
the projection of $d_{all}(x)$ 
onto $L_gh(z)$ will be in the direction of $L_gh(z)$
for $z\in B(x,\mu)$
because $L_gh(x)$ is continuous.
\end{proof}
\begin{remark}
A geometric view of the
problem introduced in Section~\ref{sec_composition_motivating_example}
and its resolution via the shared evading maneuver assumption is shown in Figure~\ref{fig_geometry_multiple}.
\end{remark}

Similar to the QP in (\ref{eq_general_qp}),
we write a QP
with $q$ constraints and let
$\hat{u} = \begin{bmatrix} \hat{u}_1^T & \hat{u}_2^T & \cdots & \hat{u}_k^T \end{bmatrix}^T$
where $\hat{u}_i$ is the nominal input of vehicle $i$ for $i=1,\ldots,k$.
To emphasize that all $h^j$ are constructed from $\gamma^s$, we write
\newcommand{\hj}{h^{j,s}}
\newcommand{\hjx}{\hj(x)}
$\hj$ for each $j=1,\ldots,q$ as follows:
\begin{IEEEeqnarray}{rCl}
\label{eq_stacked_qp_q_constraints}
u^* &=& \min_{u\in \R^m} \frac{1}{2}\norm{u - \hat{u}}^2 \\
\text{ s.t. } &&
    Au \ge b. \nonumber \\
&& L_f\hjx + L_g\hjx u + \alpha(\hjx)  \ge 0  
\qquad j\in\{1,\ldots,q\}.\nonumber
\end{IEEEeqnarray}

\section{CONTROL CALCULATION WITH LIMITED COMMUNICATION}
\label{sec_limited_communication_control}

The QP in (\ref{eq_stacked_qp_q_constraints}) is a centralized
calculation. In particular, it requires
that each vehicle's nominal control input $\hat{u}_i$
be communicated. Frequently communicating
this signal when there are many vehicles
may reduce throughput for other important
messages or introduce communication delays
because a network can only support a limited
number of bits per second through a network.
Thus, 
we show how to ensure
safety constraints can be satisfied by reformulating the QP so that the vehicles
can calculate a safe control signal without
requiring each other's nominal control input.
However, we continue to assume that each vehicle
can sense the state of every other vehicle.

We start by considering the two vehicle case and then generalize
to the $k$ vehicle case.
Let $\gamma^s = \begin{bmatrix} \gamma^{s^T}_1 & \gamma^{s^T}_2\end{bmatrix}^T$
be the shared evading maneuver where 
$\gamma_1^{s^T}$ is the part of $\gamma^s$
that is applied to vehicle 1
and therefore has the same size as $u_1$.
Define $\gamma^s_2$ similarly for vehicle 2.
Similarly decompose $b$ in (\ref{eq_general_actuator})
as
$b = \begin{bmatrix} b_1^T & b_2^T \end{bmatrix}^T$ and
$L_g\hjx$
as
$L_g\hjx
    = \begin{bmatrix} [L_g\hjx]_1^T & [L_g\hjx]_2^T \end{bmatrix}^T$.
Further, let $A$ in (\ref{eq_general_actuator})
be block diagonal with block entries $A_1$ and $A_2$
so that $A_iu_u\ge b_i$ represents the actuator constraint for
vehicle $i$ for $i = 1, 2$.

We want to find a way of calculating $u_1$ and $u_2$
such that $u = \begin{bmatrix} u_1^T & u_2^T \end{bmatrix}^T$
satisfies $Au \ge b$ and $u\in K^j(x)$ for all $x\in \mathcal{D}$
where the calculation for $u_1$ does not require knowledge of
$\hat{u}_2$ or the final value for $u_2$.
Similarly, we want to calculate $u_2$ without knowledge of $\hat{u}_1$ or $u_1$.
This is a trivial requirement for actuator 
constraints since $A_iu_i\ge b_i$ for $i = 1,2$ if and only if $Au\ge b$.
However, the constraint that $u\in K_\cap(x)$ involves both $u_1$
and $u_2$ so we reformulate it as follows:
\begin{IEEEeqnarray*}{rCl}
0 
&\le& L_f\hjx + L_g\hjx u + \alpha(\hjx) \\
&=& \kappa_1(x, u_1) + \kappa_2(x, u_2)
\end{IEEEeqnarray*}
where
\begin{equation*}
\kappa_1(x, u_1)
=
L_f\hjx + [L_g\hjx]_1u_1 + \alpha(\hjx)
 + [L_g\hjx]_2\gamma^s_2 
 - \frac{1}{2}(L_f\hjx + L_g\hjx \gamma^s + \alpha(\hjx))
\end{equation*}
and
\begin{equation*}
\kappa_2(x, u_2)
=
L_f\hjx + [L_g\hjx]_2u_2 + \alpha(\hjx) \\
 + [L_g\hjx]_1\gamma^s_1 \\
 - \frac{1}{2}(L_f\hjx + L_g\hjx \gamma^s + \alpha(\hjx)).
\end{equation*}

Notice that $\kappa_1$ is not a function of $u_2$ and $\kappa_2$ is not a function of $u_1$.
In other words, if we can select $u_1$ and $u_2$ such that $\kappa_1(x,u_1) \ge 0$
and $\kappa_2(x,u_2) \ge 0$ then $u = \begin{bmatrix} u_1^T & u_2^T \end{bmatrix}^T\in K_{\cap}(x)$
$\forall x \in \mathcal{C}$. For $x\in\mathcal{C}_\cap$, this can be done by letting $u_1 = \gamma^s_1(x)$
and $u_2 = \gamma^s_2(x)$ and noting that this implies
\[
\kappa_1(x, \gamma^s_1) + \kappa_2(x, \gamma^s_2)
=  L_f\hjx + L_g\hjx \gamma^s + \alpha(\hjx)
\ge  0.
\]
For $x\notin \mathcal{C}_\cap$, a pertubation of $\gamma^s_1(x)$ and $\gamma^s_2(x)$
using a similar method as shown in the proof of Theorem \ref{th_multiple_h_zcbf}
suffices.
In other words, we can find $u$ without vehicle 1 needing to know $\hat{u}_2$
or $u_2$ and similarly for vehicle 2. Each vehicle $i$ ($i=1,2$) could then calculate
the following QP:
\begin{IEEEeqnarray}{rCl}
\label{eq_two_veh_qp}
u^* &=& \min_{u\in \R^{m_i}} \frac{1}{2}\norm{u - \hat{u}_i}^2 \\
\text{ s.t. } &&
    A_iu_i \ge b_i \nonumber \\
&& \kappa_i(x, u_i) \ge 0.  \nonumber
\end{IEEEeqnarray}
Note that $\kappa_i(x, u_i)$ is linear in $u_i$.

We now generalize the above discussion to $k$ vehicles.
Let
$\gamma^s = \begin{bmatrix} \gamma_1^{s^T} & \cdots & \gamma_k^{s^T} \end{bmatrix}^T$,
where $\gamma^s_i$ maps to vectors of the same size as $u_i$ for $i=1,\ldots,k$
with similar decomposition for 
$b = \begin{bmatrix} b_1^T & \cdots & b_k^T\end{bmatrix}^T$ and
$L_g\hjx = \begin{bmatrix} [L_g\hjx]_1^T & \cdots & [L_g\hjx]_k^T\end{bmatrix}^T$.
Further, assume
$A$ in (\ref{eq_general_actuator}) is block diagonal
with block entries $A_i$ for $i=1,\ldots,k$
where $A_i$ is a $m_i\times m_i$ matrix.
This assumption means that actuator constraints are not coupled between vehicles.
For constraint $j$ for $j=1,\ldots,q$,
let
\[
\mathcal{V}^j =
\{i\in\{1,\ldots,k\}:
\exists x\in\mathcal{D} \text{ s.t. }[L_g\hjx]_i \ne 0_{m_i}\}
\]
where $0_{m_i}$ is the zero vector in $\R^{m_i}$.
$\mathcal{V}^j$ represents the set of vehicles whose control input
affects the time derivative of $h^j$ for some $x\in\mathcal{D}$.
We let $|\mathcal{V}^j|$ denote the cardinality of $\mathcal{V}^j$, and note that
for the case of pairwise collision avoidance,
$|\mathcal{V}^j| = 2$ for all $j=1,\ldots,q$.
In the example with three vehicles in Section \ref{sec_composition},
$\mathcal{V}_1 = \{1,2\}$, $\mathcal{V}_2 = \{1,3\}$, $\mathcal{V}_3 = \{2,3\}$.
Finally, we denote
$u_{\setminus i} = \begin{bmatrix} u_1^T & \cdots & u_{i-1}^T & u_{i+1}^T & \cdots & u_k^T\end{bmatrix}^T$,
with similar definitions for $\gamma^s_{\setminus i}$, $\hat{u}_{\setminus i}$,
and $[L_g\hjx]_{\setminus i}$.

With the above definitions,
we can now state a limited communication analogue for the admissible control space in (\ref{eq_admissible_control_space}).
The limited communication admissible control space
for constraint $j$ ($j=1,\ldots,q$) and vehicle $i$ ($i\in\mathcal{V}^j$)
is defined as
\begin{IEEEeqnarray*}{rCCl}
\mathcal{K}_{i}^{j}(x)
&=&
\Big\{u_i\in U_i\::\: 0\le&
L_f \hjx + [L_g \hjx]_iu_i + \alpha(\hjx)
+ [L_g\hjx]_{\setminus i}\gamma^s_{\setminus i}(x) \nonumber\\
&&& - \frac{|\mathcal{V}^j|-1}{|\mathcal{V}^j|} \Big(L_f\hjx + L_g\hjx\gamma^s(x)
 + \alpha(\hjx)\Big)
\Big\}.
\end{IEEEeqnarray*}
Let $\mathcal{S}_i = \set{j\in\{1,\ldots,q\}}{i\in\mathcal{V}^j}$
so that $\mathcal{S}_i$ is the set of safety constraint indices where $u_i$
has an effect on the time derivative of the associated barrier function
for some $x\in\mathcal{D}$.
For the three vehicle example of Section \ref{sec_composition},
$\mathcal{S}_1 = \{1, 2\}$, $\mathcal{S}_2 = \{1,3\}$, $\mathcal{S}_3 = \{2,3\}$.
The limited communication admissible control space for vehicle $i$ is then
$
\mathcal{K}_i(x) = \bigcap_{l\in\mathcal{S}_i}\mathcal{K}_{i}^{l}(x)
$
and the overall limited communication admissible control space is
\begin{equation*}
\mathcal{K}(x) = 
\left\{u = \begin{bmatrix}u_1^T & \cdots & u_k^T\end{bmatrix}^T
    \in U\::\: u_i\in \mathcal{K}_i(x)\;\forall i\in\{1,\ldots,k\}\right\}.
\end{equation*}

\begin{theorem}
Given a dynamical system (\ref{eq_system})
and a set $\mathcal{C}_\cap \subset \mathcal{D}_\cap$ defined in (\ref{eq_overall_safe_set})
for $q$ continuously differentiable functions $h^j$
defined in (\ref{eq_h_j})
with safety functions $\rho^j$ and evading maneuvers $\gamma^j$ where $k\in\{1,\ldots,q\}$,
if $h^j$ is a ZCBF for $k\in\{1,\ldots,q\}$ and Assumption~\ref{as_shared_evading_assumption} holds
then $\forall x\in \mathcal{D}_\cap$,
$\mathcal{K}(x)\subseteq K_\cap(x)$.
Further, $\mathcal{K}(x)$ is non-empty for all $x\in\mathcal{C}_\cap$.
If in addition, 
$\gamma^s$ maps to the interior of $U$
and for all $x\in \partial \mathcal{C}_\cap$,
$\langle [L_gh^{j_1}(x)]_i, [L_gh^{j_2}(x)]_i\rangle > 0$
for $j=1,\ldots,q$ and $i=1,\ldots,k$
and $j_1\ne j_2$ and $j_1,j_2\in\{1,\ldots,q\}$,
then there is an open set that is a superset of $C_\cap$
for which $\mathcal{K}(x)$ is non-empty for all $x$ in the open set.
\label{thm_limited_communication_control}
\end{theorem}
\begin{proof}
For the first statement, assume $u\in \mathcal{K}(x)$
so that $u_i\in\mathcal{K}_i(x)$ $\forall i\in\{1,\ldots,k\}$.
This means that $A_{i}u_i \ge b_i$ so that,
because $A$ is block diagonal, $Au \ge b$.
Further, it means that for any constraint $j=1,\ldots,q$
and any $i\in\mathcal{V}^j$,
\begin{equation}
0 \le
 L_f \hjx + [L_g \hjx]_iu_i + \alpha(\hjx) 
+ [L_g\hjx]_{\setminus i}\gamma^s_{\setminus i}(x) 
- \frac{|\mathcal{V}^j|-1}{|\mathcal{V}^j|} \Big(L_f\hjx + L_g\hjx\gamma^s(x) 
+ \alpha(\hjx)\Big).
\label{eq_dec_proof_calK_in_K1}
\end{equation}
To simplify (\ref{eq_dec_proof_calK_in_K1}), note
that by definition, $[L_gh^{j,s}(x)]_i = 0_{m_i}$ for $i\ne \mathcal{V}^j$ so that
\begin{IEEEeqnarray}{rCl}
    \sum_{i\in\mathcal{V}^j} [L_g\hjx]_i u_i
&=& 
\sum_{i\in\{1,\ldots,k\}} [L_g\hjx]_i u_i \nonumber\\
&=& L_g\hjx u.
\label{eq_dec_proof_calK_in_K2}
\end{IEEEeqnarray}
Using (\ref{eq_dec_proof_calK_in_K2}) in the following then yields
\begin{IEEEeqnarray}{rCl}
\sum_{i\in\mathcal{V}^j}[L_g\hjx]_{\setminus i} \gamma^s_{\setminus i}(x) 
&=&
\sum_{i\in\mathcal{V}^j}\left(L_g\hjx\gamma^s(x) - [L_g\hjx]_i\gamma^s_i(x)\right) \nonumber\\
&=&
|\mathcal{V}^j|L_g\hjx\gamma^s(x) -
    \sum_{i\in\mathcal{V}^j}[L_g\hjx]_i\gamma^s_i(x) \nonumber\\
&=&
|\mathcal{V}^j|L_g\hjx\gamma^s(x) 
    - 
    L_g\hjx\gamma^s(x) \nonumber\\
&=&
(|\mathcal{V}^j|-1)L_g\hjx\gamma^s(x).
\label{eq_dec_proof_calK_in_K3}
\end{IEEEeqnarray}
Summing (\ref{eq_dec_proof_calK_in_K1}) over $i\in\mathcal{V}^j$ and
using (\ref{eq_dec_proof_calK_in_K2}) and (\ref{eq_dec_proof_calK_in_K3}) yields
\begin{IEEEeqnarray*}{rCl}
0 &\le&
|\mathcal{V}^j|L_f\hjx + L_g\hjx u + |\mathcal{V}^j|\alpha(\hjx )
    + (|\mathcal{V}^j| - 1)L_g\hjx \gamma^s(x) \\
&& - (|\mathcal{V}^j| - 1)\Big(L_f\hjx  + L_g\hjx  \gamma^s(x)  
+ \alpha(\hjx )\Big) \\
&=&
L_f\hjx  + L_g\hjx u + \alpha(\hjx ).
\end{IEEEeqnarray*}
Since this is true for all $j=1,\ldots,q$, $u\in K_\cap(x)$.
Then $\mathcal{K}(x)\subseteq K_\cap(x)$ for all $x\in\mathcal{C}_\cap$.

Consider now the second statement, namely that $\gamma^s\in\mathcal{K}(x)$.
For $j=1,\ldots,q$,
consider any $i\in \mathcal{V}^j$ and let $u_i = \gamma_i^s$. Then
\begin{IEEEeqnarray*}{l}
L_f \hjx + [L_g \hjx]_iu_i + \alpha(\hjx)
+ [L_g\hjx]_{\setminus i}\gamma^s_{\setminus i}(x) 
- \frac{|\mathcal{V}^j|-1}{|\mathcal{V}^j|} \Big(L_f\hjx + L_g\hjx\gamma^s(x)
+ \alpha(\hjx)\Big) \\
\qquad = \frac{1}{|\mathcal{V}^{j}|}\left(L_fh^j(x) + L_gh^j(x)\gamma^s(x) + \alpha(h^j(x))\right)  \\
\qquad \ge 0.
\end{IEEEeqnarray*} 
The inequality is true because $x\in \mathcal{C}_\cap$ implies
$\alpha(\hjx) \ge 0$. See
the proof for Theorem \ref{th_h_zcbf}
for why
$L_f\hjx + L_g\hjx\gamma^s(x) \ge 0$.
Then $\gamma^s_i\in \mathcal{K}_{i}^{j}$ for any $j=1,\ldots,q$
and $i\in\mathcal{V}^j$. Then $\gamma^s_i\in\mathcal{K}_i$.
Then $\gamma^s(x) \in \mathcal{K}(x)$.

Finally, the last statement where $\mathcal{K}(x)$ is nonempty
for all $x$ in an open set that is a superset of $\mathcal{C}_\cap$
follows similarly to the proof of Theorem \ref{th_multiple_h_zcbf}.
\end{proof}

We now write a QP similar to (\ref{eq_stacked_qp_q_constraints})
but without requiring knowledge of other agents' low level control values
as follows:
\begin{IEEEeqnarray}{rCl}
\label{eq_qp_dec}
u_i^* &=& \min_{u_i\in\R^{m_i}} \frac{1}{2}\norm{u_i - \hat{u}_i}^2 \\
\text{s.t.} &&
    A_{i}u_i \ge b_i \nonumber\\
&& L_f \hjx  + [L_g \hjx ]_iu_i + \alpha(\hjx ) 
+ 
    [L_g\hjx ]_{\setminus i}\gamma^s_{\setminus i}(x) \nonumber\\
&&     \quad
      - \frac{|\mathcal{V}^j|-1}{|\mathcal{V}^j|} \Big(L_f\hjx  + L_g\hjx \gamma^s(x)  
        + \alpha(\hjx )\Big) \ge 0
    \qquad j\in\mathcal{S}_i\nonumber.
\end{IEEEeqnarray}.

We note that the solution from the centralized QP
(\ref{eq_stacked_qp_q_constraints}) may be different than the solution from the limited communication QPs
(\ref{eq_qp_dec}) because $\mathcal{K}(x)$ may be a strict subset of $K_\cap(x)$.
To see this, let $k = 2$, $q = 1$,
$L_fh(x) = 0$, $\alpha(h(x)) = 0$, $m_1=m_2=1$,
$[L_gh(x)]_2\gamma^s_2(x) = -1$, and $[L_gh(x)]_1\gamma^s_1(x) = 1$. Then
the barrier function constraint in (\ref{eq_qp_dec})
becomes
$
[L_gh(x)]_1u_1 \ge 1,
$
while the barrier function constraint in (\ref{eq_stacked_qp_q_constraints})
becomes
$
L_gh(x) u \ge 0.
$
Since $u_1 = 0$ is feasible for the latter
but not the former equation,
we do not have that $\mathcal{K}(x) = K_\cap(x)$.
Because $\mathcal{K}(x) \subset K_\cap(x)$,
it may be that the total cost of each vehicle
calculating (\ref{eq_qp_dec}) is higher than
the centralized calculation (\ref{eq_stacked_qp_q_constraints}).
In other words, the calculated safe control
may not be as close to the nominal control signal
in a least squares sense when using (\ref{eq_qp_dec})
as opposed to (\ref{eq_stacked_qp_q_constraints}).
Nevertheless, in either case of (\ref{eq_stacked_qp_q_constraints})
or (\ref{eq_qp_dec}), a solution exists
to the corresponding QP such that
$u\in K_\cap$.

Another difference between the limited communication (\ref{eq_qp_dec}) and the centralized (\ref{eq_stacked_qp_q_constraints}) QPs
is how the size of the optimization variable and number of constraints
vary with the number of vehicles $k$. In the centralized approach (\ref{eq_stacked_qp_q_constraints})
the size of the optimization variable grows linearly with $k$ while 
the number of constraints grows quadratically. On the other hand, in the limited communication QP (\ref{eq_qp_dec}),
the size of the optimization variable and number of constraints are constant and 
linear, respectively.

\section{SIMULATION}

\label{sec_simulation}

In this section we repeat the scenario discussed in Section
\ref{sec_simulation2veh} but consider $k = 20$ vehicles. For the scenario where $h$
is constructed from $\gamma_{turn}$, we use $\begin{bmatrix} v & \omega & 0 & v &
\omega & 0 \end{bmatrix}^T$ where $v = 0.9 v_{min} + 0.1 v_{max}$ and $\omega =
0.9\omega_{max}$.  For the scenario where $h$ is constructed from
$\gamma_{straight}$, we let $\gamma^i = \begin{bmatrix} (1 + 0.01i)v & 0 & 0
\end{bmatrix}^T$ so that each vehicle uses a different translational velocity
as is required to ensure differentiability of $h$ (see Section \ref{sec_closed_form}).
Note that this does not violate the shared evading maneuver assumption because 
$\gamma^s = \begin{bmatrix} (\gamma^1)^T & \cdots & (\gamma^k)^T \end{bmatrix}^T$.
Additionally, we let $\psi = 0$ and $\psi = 25^\circ$ in the scenario where $h$ is
constructed from $\gamma_{turn}$ and $\gamma_{straight}$, respectively. Offsetting the initial orientation $25^\circ$
from pointing at the origin is required so that the vehicles can start in the safe set when using $\gamma_{straight}$.
A video of the resulting behavior is available in \cite{squires2020collisionvideo}.
Quantitative results for both scenarios are shown in Figure
\ref{fig_constraints_satisfied} which shows similar outputs to the results for
the two vehicle simulation shown in Figure
\ref{fig_constraints_satisfied_2veh}.
Small random perturbations, e.g. adding a uniformly sampled angle within $\pm 5^\circ$ to the initial heading,
does not appear to change
qualitative results.
We also compare the approach of this paper to a navigation function
from \cite{panyakeow2010decentralized} in Figure \ref{fig_constraints_satisfied}.
Note that when using a navigation function the vehicles
begin the evasive maneuver earlier
than when the collision avoidance
algorithm is based on a barrier function constructed
from $\gamma_{turn}$.
Using a less aggressive $\alpha$ function, such as a linear function with a small coefficient
instead of a cubic function, may have caused the behavior from the barrier function
override to similarly override earlier.
A comparison of potential and barrier functions can also be found in
\cite{singletary2020comparative}.
Also note that the pairwise distance
between all vehicles are kept above the minimum safety distance $D_s$ while
satisfying actuator constraints.

\begin{figure}
\centering
\begin{subfigure}{\columnwidth}
\centering
\input{imgs/sim/20_veh1_actuator0.pgf}
\caption{}
\end{subfigure}

\begin{subfigure}{\columnwidth}
\centering
\input{imgs/sim/20_veh1_actuator1.pgf}
\caption{}
\end{subfigure}

\begin{subfigure}{\columnwidth}
\centering
\input{imgs/sim/20_veh_min_d.pgf}
\caption{}
\end{subfigure}

\begin{subfigure}{\columnwidth}
\centering
\input{imgs/sim/20_veh1_path.pgf}
\caption{}
\label{fig_constraints_satisfied_path}
\end{subfigure}

\caption[Control outputs for the 20 vehicle case]{
20 vehicle scenario statistics.
(a) vehicle 1 velocity, (b) vehicle 1 turn rate,
(c) intervehicle distance, 
(d) vehicle 1 path.
}{}
\label{fig_constraints_satisfied}
\end{figure}

%% file: imgs/svg/two_aircraft1.pdf_tex
\begingroup%
  \makeatletter%
  \providecommand\color[2][]{%
    \errmessage{(Inkscape) Color is used for the text in Inkscape, but the package 'color.sty' is not loaded}%
    \renewcommand\color[2][]{}%
  }%
  \providecommand\transparent[1]{%
    \errmessage{(Inkscape) Transparency is used (non-zero) for the text in Inkscape, but the package 'transparent.sty' is not loaded}%
    \renewcommand\transparent[1]{}%
  }%
  \providecommand\rotatebox[2]{#2}%
  \newcommand*\fsize{\dimexpr\f@size pt\relax}%
  \newcommand*\lineheight[1]{\fontsize{\fsize}{#1\fsize}\selectfont}%
  \ifx\svgwidth\undefined%
    \setlength{\unitlength}{194.83023168bp}%
    \ifx\svgscale\undefined%
      \relax%
    \else%
      \setlength{\unitlength}{\unitlength * \real{\svgscale}}%
    \fi%
  \else%
    \setlength{\unitlength}{\svgwidth}%
  \fi%
  \global\let\svgwidth\undefined%
  \global\let\svgscale\undefined%
  \makeatother%
  \begin{picture}(1,0.70599922)%
    \lineheight{1}%
    \setlength\tabcolsep{0pt}%
    \put(0,0){\includegraphics[width=\unitlength,page=1]{imgs/svg/two_aircraft1.pdf}}%
    \put(0.05028141,0.34618097){\color[rgb]{0,0,0}\makebox(0,0)[lt]{\lineheight{0}\smash{\begin{tabular}[t]{l}$1$\end{tabular}}}}%
    \put(0.83587598,0.43934777){\color[rgb]{0,0,0}\makebox(0,0)[lt]{\lineheight{0}\smash{\begin{tabular}[t]{l}$2$\end{tabular}}}}%
    \put(0,0){\includegraphics[width=\unitlength,page=2]{imgs/svg/two_aircraft1.pdf}}%
  \end{picture}%
\endgroup%

%% file: imgs/svg/two_aircraft2.pdf_tex
\begingroup%
  \makeatletter%
  \providecommand\color[2][]{%
    \errmessage{(Inkscape) Color is used for the text in Inkscape, but the package 'color.sty' is not loaded}%
    \renewcommand\color[2][]{}%
  }%
  \providecommand\transparent[1]{%
    \errmessage{(Inkscape) Transparency is used (non-zero) for the text in Inkscape, but the package 'transparent.sty' is not loaded}%
    \renewcommand\transparent[1]{}%
  }%
  \providecommand\rotatebox[2]{#2}%
  \newcommand*\fsize{\dimexpr\f@size pt\relax}%
  \newcommand*\lineheight[1]{\fontsize{\fsize}{#1\fsize}\selectfont}%
  \ifx\svgwidth\undefined%
    \setlength{\unitlength}{194.83023168bp}%
    \ifx\svgscale\undefined%
      \relax%
    \else%
      \setlength{\unitlength}{\unitlength * \real{\svgscale}}%
    \fi%
  \else%
    \setlength{\unitlength}{\svgwidth}%
  \fi%
  \global\let\svgwidth\undefined%
  \global\let\svgscale\undefined%
  \makeatother%
  \begin{picture}(1,0.70599922)%
    \lineheight{1}%
    \setlength\tabcolsep{0pt}%
    \put(0,0){\includegraphics[width=\unitlength,page=1]{imgs/svg/two_aircraft2.pdf}}%
    \put(0.05028116,0.34618098){\color[rgb]{0,0,0}\makebox(0,0)[lt]{\lineheight{0}\smash{\begin{tabular}[t]{l}$1$\end{tabular}}}}%
    \put(0,0){\includegraphics[width=\unitlength,page=2]{imgs/svg/two_aircraft2.pdf}}%
    \put(0.83587598,0.26058637){\color[rgb]{0,0,0}\makebox(0,0)[lt]{\lineheight{0}\smash{\begin{tabular}[t]{l}$3$\end{tabular}}}}%
    \put(0,0){\includegraphics[width=\unitlength,page=3]{imgs/svg/two_aircraft2.pdf}}%
  \end{picture}%
\endgroup%

%% file: imgs/svg/two_aircraft3.pdf_tex
\begingroup%
  \makeatletter%
  \providecommand\color[2][]{%
    \errmessage{(Inkscape) Color is used for the text in Inkscape, but the package 'color.sty' is not loaded}%
    \renewcommand\color[2][]{}%
  }%
  \providecommand\transparent[1]{%
    \errmessage{(Inkscape) Transparency is used (non-zero) for the text in Inkscape, but the package 'transparent.sty' is not loaded}%
    \renewcommand\transparent[1]{}%
  }%
  \providecommand\rotatebox[2]{#2}%
  \newcommand*\fsize{\dimexpr\f@size pt\relax}%
  \newcommand*\lineheight[1]{\fontsize{\fsize}{#1\fsize}\selectfont}%
  \ifx\svgwidth\undefined%
    \setlength{\unitlength}{194.83023168bp}%
    \ifx\svgscale\undefined%
      \relax%
    \else%
      \setlength{\unitlength}{\unitlength * \real{\svgscale}}%
    \fi%
  \else%
    \setlength{\unitlength}{\svgwidth}%
  \fi%
  \global\let\svgwidth\undefined%
  \global\let\svgscale\undefined%
  \makeatother%
  \begin{picture}(1,0.70599922)%
    \lineheight{1}%
    \setlength\tabcolsep{0pt}%
    \put(0,0){\includegraphics[width=\unitlength,page=1]{imgs/svg/two_aircraft3.pdf}}%
    \put(0.05028141,0.34618098){\color[rgb]{0,0,0}\makebox(0,0)[lt]{\lineheight{0}\smash{\begin{tabular}[t]{l}$1$\end{tabular}}}}%
    \put(0.83587598,0.43934778){\color[rgb]{0,0,0}\makebox(0,0)[lt]{\lineheight{0}\smash{\begin{tabular}[t]{l}$2$\end{tabular}}}}%
    \put(0,0){\includegraphics[width=\unitlength,page=2]{imgs/svg/two_aircraft3.pdf}}%
    \put(0.83587598,0.26058634){\color[rgb]{0,0,0}\makebox(0,0)[lt]{\lineheight{0}\smash{\begin{tabular}[t]{l}$3$\end{tabular}}}}%
    \put(0,0){\includegraphics[width=\unitlength,page=3]{imgs/svg/two_aircraft3.pdf}}%
  \end{picture}%
\endgroup%

%% file: imgs/svg/geometry_multi.pdf_tex
\begingroup%
  \makeatletter%
  \providecommand\color[2][]{%
    \errmessage{(Inkscape) Color is used for the text in Inkscape, but the package 'color.sty' is not loaded}%
    \renewcommand\color[2][]{}%
  }%
  \providecommand\transparent[1]{%
    \errmessage{(Inkscape) Transparency is used (non-zero) for the text in Inkscape, but the package 'transparent.sty' is not loaded}%
    \renewcommand\transparent[1]{}%
  }%
  \providecommand\rotatebox[2]{#2}%
  \newcommand*\fsize{\dimexpr\f@size pt\relax}%
  \newcommand*\lineheight[1]{\fontsize{\fsize}{#1\fsize}\selectfont}%
  \ifx\svgwidth\undefined%
    \setlength{\unitlength}{173.87049537bp}%
    \ifx\svgscale\undefined%
      \relax%
    \else%
      \setlength{\unitlength}{\unitlength * \real{\svgscale}}%
    \fi%
  \else%
    \setlength{\unitlength}{\svgwidth}%
  \fi%
  \global\let\svgwidth\undefined%
  \global\let\svgscale\undefined%
  \makeatother%
  \begin{picture}(1,0.77761259)%
    \lineheight{1}%
    \setlength\tabcolsep{0pt}%
    \put(0,0){\includegraphics[width=\unitlength,page=1]{imgs/svg/geometry_multi.pdf}}%
  \end{picture}%
\endgroup%

%% file: imgs/svg/geometry_multi_gamma.pdf_tex
\begingroup%
  \makeatletter%
  \providecommand\color[2][]{%
    \errmessage{(Inkscape) Color is used for the text in Inkscape, but the package 'color.sty' is not loaded}%
    \renewcommand\color[2][]{}%
  }%
  \providecommand\transparent[1]{%
    \errmessage{(Inkscape) Transparency is used (non-zero) for the text in Inkscape, but the package 'transparent.sty' is not loaded}%
    \renewcommand\transparent[1]{}%
  }%
  \providecommand\rotatebox[2]{#2}%
  \newcommand*\fsize{\dimexpr\f@size pt\relax}%
  \newcommand*\lineheight[1]{\fontsize{\fsize}{#1\fsize}\selectfont}%
  \ifx\svgwidth\undefined%
    \setlength{\unitlength}{173.87049537bp}%
    \ifx\svgscale\undefined%
      \relax%
    \else%
      \setlength{\unitlength}{\unitlength * \real{\svgscale}}%
    \fi%
  \else%
    \setlength{\unitlength}{\svgwidth}%
  \fi%
  \global\let\svgwidth\undefined%
  \global\let\svgscale\undefined%
  \makeatother%
  \begin{picture}(1,0.77761259)%
    \lineheight{1}%
    \setlength\tabcolsep{0pt}%
    \put(0,0){\includegraphics[width=\unitlength,page=1]{imgs/svg/geometry_multi_gamma.pdf}}%
  \end{picture}%
\endgroup%

%% file: imgs/svg/geometry_multi_shared_gamma.pdf_tex
\begingroup%
  \makeatletter%
  \providecommand\color[2][]{%
    \errmessage{(Inkscape) Color is used for the text in Inkscape, but the package 'color.sty' is not loaded}%
    \renewcommand\color[2][]{}%
  }%
  \providecommand\transparent[1]{%
    \errmessage{(Inkscape) Transparency is used (non-zero) for the text in Inkscape, but the package 'transparent.sty' is not loaded}%
    \renewcommand\transparent[1]{}%
  }%
  \providecommand\rotatebox[2]{#2}%
  \newcommand*\fsize{\dimexpr\f@size pt\relax}%
  \newcommand*\lineheight[1]{\fontsize{\fsize}{#1\fsize}\selectfont}%
  \ifx\svgwidth\undefined%
    \setlength{\unitlength}{173.87049537bp}%
    \ifx\svgscale\undefined%
      \relax%
    \else%
      \setlength{\unitlength}{\unitlength * \real{\svgscale}}%
    \fi%
  \else%
    \setlength{\unitlength}{\svgwidth}%
  \fi%
  \global\let\svgwidth\undefined%
  \global\let\svgscale\undefined%
  \makeatother%
  \begin{picture}(1,0.77761259)%
    \lineheight{1}%
    \setlength\tabcolsep{0pt}%
    \put(0,0){\includegraphics[width=\unitlength,page=1]{imgs/svg/geometry_multi_shared_gamma.pdf}}%
  \end{picture}%
\endgroup%

%% file: imgs/sim/20_veh1_actuator0.pgf
\begingroup%
\makeatletter%
\begin{pgfpicture}%
\pgfpathrectangle{\pgfpointorigin}{\pgfqpoint{3.800000in}{1.600000in}}%
\pgfusepath{use as bounding box, clip}%
\begin{pgfscope}%
\pgfsetbuttcap%
\pgfsetmiterjoin%
\definecolor{currentfill}{rgb}{1.000000,1.000000,1.000000}%
\pgfsetfillcolor{currentfill}%
\pgfsetlinewidth{0.000000pt}%
\definecolor{currentstroke}{rgb}{1.000000,1.000000,1.000000}%
\pgfsetstrokecolor{currentstroke}%
\pgfsetdash{}{0pt}%
\pgfpathmoveto{\pgfqpoint{0.000000in}{0.000000in}}%
\pgfpathlineto{\pgfqpoint{3.800000in}{0.000000in}}%
\pgfpathlineto{\pgfqpoint{3.800000in}{1.600000in}}%
\pgfpathlineto{\pgfqpoint{0.000000in}{1.600000in}}%
\pgfpathclose%
\pgfusepath{fill}%
\end{pgfscope}%
\begin{pgfscope}%
\pgfsetbuttcap%
\pgfsetmiterjoin%
\definecolor{currentfill}{rgb}{1.000000,1.000000,1.000000}%
\pgfsetfillcolor{currentfill}%
\pgfsetlinewidth{0.000000pt}%
\definecolor{currentstroke}{rgb}{0.000000,0.000000,0.000000}%
\pgfsetstrokecolor{currentstroke}%
\pgfsetstrokeopacity{0.000000}%
\pgfsetdash{}{0pt}%
\pgfpathmoveto{\pgfqpoint{0.430556in}{0.415000in}}%
\pgfpathlineto{\pgfqpoint{3.800000in}{0.415000in}}%
\pgfpathlineto{\pgfqpoint{3.800000in}{1.600000in}}%
\pgfpathlineto{\pgfqpoint{0.430556in}{1.600000in}}%
\pgfpathclose%
\pgfusepath{fill}%
\end{pgfscope}%
\begin{pgfscope}%
\pgfsetbuttcap%
\pgfsetroundjoin%
\definecolor{currentfill}{rgb}{0.000000,0.000000,0.000000}%
\pgfsetfillcolor{currentfill}%
\pgfsetlinewidth{0.803000pt}%
\definecolor{currentstroke}{rgb}{0.000000,0.000000,0.000000}%
\pgfsetstrokecolor{currentstroke}%
\pgfsetdash{}{0pt}%
\pgfsys@defobject{currentmarker}{\pgfqpoint{0.000000in}{-0.048611in}}{\pgfqpoint{0.000000in}{0.000000in}}{%
\pgfpathmoveto{\pgfqpoint{0.000000in}{0.000000in}}%
\pgfpathlineto{\pgfqpoint{0.000000in}{-0.048611in}}%
\pgfusepath{stroke,fill}%
}%
\begin{pgfscope}%
\pgfsys@transformshift{0.582946in}{0.415000in}%
\pgfsys@useobject{currentmarker}{}%
\end{pgfscope}%
\end{pgfscope}%
\begin{pgfscope}%
\definecolor{textcolor}{rgb}{0.000000,0.000000,0.000000}%
\pgfsetstrokecolor{textcolor}%
\pgfsetfillcolor{textcolor}%
\pgftext[x=0.582946in,y=0.317777in,,top]{\color{textcolor}\rmfamily\fontsize{10.000000}{12.000000}\selectfont \(\displaystyle {0}\)}%
\end{pgfscope}%
\begin{pgfscope}%
\pgfsetbuttcap%
\pgfsetroundjoin%
\definecolor{currentfill}{rgb}{0.000000,0.000000,0.000000}%
\pgfsetfillcolor{currentfill}%
\pgfsetlinewidth{0.803000pt}%
\definecolor{currentstroke}{rgb}{0.000000,0.000000,0.000000}%
\pgfsetstrokecolor{currentstroke}%
\pgfsetdash{}{0pt}%
\pgfsys@defobject{currentmarker}{\pgfqpoint{0.000000in}{-0.048611in}}{\pgfqpoint{0.000000in}{0.000000in}}{%
\pgfpathmoveto{\pgfqpoint{0.000000in}{0.000000in}}%
\pgfpathlineto{\pgfqpoint{0.000000in}{-0.048611in}}%
\pgfusepath{stroke,fill}%
}%
\begin{pgfscope}%
\pgfsys@transformshift{1.349112in}{0.415000in}%
\pgfsys@useobject{currentmarker}{}%
\end{pgfscope}%
\end{pgfscope}%
\begin{pgfscope}%
\definecolor{textcolor}{rgb}{0.000000,0.000000,0.000000}%
\pgfsetstrokecolor{textcolor}%
\pgfsetfillcolor{textcolor}%
\pgftext[x=1.349112in,y=0.317777in,,top]{\color{textcolor}\rmfamily\fontsize{10.000000}{12.000000}\selectfont \(\displaystyle {5}\)}%
\end{pgfscope}%
\begin{pgfscope}%
\pgfsetbuttcap%
\pgfsetroundjoin%
\definecolor{currentfill}{rgb}{0.000000,0.000000,0.000000}%
\pgfsetfillcolor{currentfill}%
\pgfsetlinewidth{0.803000pt}%
\definecolor{currentstroke}{rgb}{0.000000,0.000000,0.000000}%
\pgfsetstrokecolor{currentstroke}%
\pgfsetdash{}{0pt}%
\pgfsys@defobject{currentmarker}{\pgfqpoint{0.000000in}{-0.048611in}}{\pgfqpoint{0.000000in}{0.000000in}}{%
\pgfpathmoveto{\pgfqpoint{0.000000in}{0.000000in}}%
\pgfpathlineto{\pgfqpoint{0.000000in}{-0.048611in}}%
\pgfusepath{stroke,fill}%
}%
\begin{pgfscope}%
\pgfsys@transformshift{2.115278in}{0.415000in}%
\pgfsys@useobject{currentmarker}{}%
\end{pgfscope}%
\end{pgfscope}%
\begin{pgfscope}%
\definecolor{textcolor}{rgb}{0.000000,0.000000,0.000000}%
\pgfsetstrokecolor{textcolor}%
\pgfsetfillcolor{textcolor}%
\pgftext[x=2.115278in,y=0.317777in,,top]{\color{textcolor}\rmfamily\fontsize{10.000000}{12.000000}\selectfont \(\displaystyle {10}\)}%
\end{pgfscope}%
\begin{pgfscope}%
\pgfsetbuttcap%
\pgfsetroundjoin%
\definecolor{currentfill}{rgb}{0.000000,0.000000,0.000000}%
\pgfsetfillcolor{currentfill}%
\pgfsetlinewidth{0.803000pt}%
\definecolor{currentstroke}{rgb}{0.000000,0.000000,0.000000}%
\pgfsetstrokecolor{currentstroke}%
\pgfsetdash{}{0pt}%
\pgfsys@defobject{currentmarker}{\pgfqpoint{0.000000in}{-0.048611in}}{\pgfqpoint{0.000000in}{0.000000in}}{%
\pgfpathmoveto{\pgfqpoint{0.000000in}{0.000000in}}%
\pgfpathlineto{\pgfqpoint{0.000000in}{-0.048611in}}%
\pgfusepath{stroke,fill}%
}%
\begin{pgfscope}%
\pgfsys@transformshift{2.881444in}{0.415000in}%
\pgfsys@useobject{currentmarker}{}%
\end{pgfscope}%
\end{pgfscope}%
\begin{pgfscope}%
\definecolor{textcolor}{rgb}{0.000000,0.000000,0.000000}%
\pgfsetstrokecolor{textcolor}%
\pgfsetfillcolor{textcolor}%
\pgftext[x=2.881444in,y=0.317777in,,top]{\color{textcolor}\rmfamily\fontsize{10.000000}{12.000000}\selectfont \(\displaystyle {15}\)}%
\end{pgfscope}%
\begin{pgfscope}%
\pgfsetbuttcap%
\pgfsetroundjoin%
\definecolor{currentfill}{rgb}{0.000000,0.000000,0.000000}%
\pgfsetfillcolor{currentfill}%
\pgfsetlinewidth{0.803000pt}%
\definecolor{currentstroke}{rgb}{0.000000,0.000000,0.000000}%
\pgfsetstrokecolor{currentstroke}%
\pgfsetdash{}{0pt}%
\pgfsys@defobject{currentmarker}{\pgfqpoint{0.000000in}{-0.048611in}}{\pgfqpoint{0.000000in}{0.000000in}}{%
\pgfpathmoveto{\pgfqpoint{0.000000in}{0.000000in}}%
\pgfpathlineto{\pgfqpoint{0.000000in}{-0.048611in}}%
\pgfusepath{stroke,fill}%
}%
\begin{pgfscope}%
\pgfsys@transformshift{3.647610in}{0.415000in}%
\pgfsys@useobject{currentmarker}{}%
\end{pgfscope}%
\end{pgfscope}%
\begin{pgfscope}%
\definecolor{textcolor}{rgb}{0.000000,0.000000,0.000000}%
\pgfsetstrokecolor{textcolor}%
\pgfsetfillcolor{textcolor}%
\pgftext[x=3.647610in,y=0.317777in,,top]{\color{textcolor}\rmfamily\fontsize{10.000000}{12.000000}\selectfont \(\displaystyle {20}\)}%
\end{pgfscope}%
\begin{pgfscope}%
\definecolor{textcolor}{rgb}{0.000000,0.000000,0.000000}%
\pgfsetstrokecolor{textcolor}%
\pgfsetfillcolor{textcolor}%
\pgftext[x=2.115278in,y=0.138889in,,top]{\color{textcolor}\rmfamily\fontsize{10.000000}{12.000000}\selectfont time (seconds)}%
\end{pgfscope}%
\begin{pgfscope}%
\pgfsetbuttcap%
\pgfsetroundjoin%
\definecolor{currentfill}{rgb}{0.000000,0.000000,0.000000}%
\pgfsetfillcolor{currentfill}%
\pgfsetlinewidth{0.803000pt}%
\definecolor{currentstroke}{rgb}{0.000000,0.000000,0.000000}%
\pgfsetstrokecolor{currentstroke}%
\pgfsetdash{}{0pt}%
\pgfsys@defobject{currentmarker}{\pgfqpoint{-0.048611in}{0.000000in}}{\pgfqpoint{-0.000000in}{0.000000in}}{%
\pgfpathmoveto{\pgfqpoint{-0.000000in}{0.000000in}}%
\pgfpathlineto{\pgfqpoint{-0.048611in}{0.000000in}}%
\pgfusepath{stroke,fill}%
}%
\begin{pgfscope}%
\pgfsys@transformshift{0.430556in}{0.468863in}%
\pgfsys@useobject{currentmarker}{}%
\end{pgfscope}%
\end{pgfscope}%
\begin{pgfscope}%
\definecolor{textcolor}{rgb}{0.000000,0.000000,0.000000}%
\pgfsetstrokecolor{textcolor}%
\pgfsetfillcolor{textcolor}%
\pgftext[x=0.194444in, y=0.420669in, left, base]{\color{textcolor}\rmfamily\fontsize{10.000000}{12.000000}\selectfont \(\displaystyle {15}\)}%
\end{pgfscope}%
\begin{pgfscope}%
\pgfsetbuttcap%
\pgfsetroundjoin%
\definecolor{currentfill}{rgb}{0.000000,0.000000,0.000000}%
\pgfsetfillcolor{currentfill}%
\pgfsetlinewidth{0.803000pt}%
\definecolor{currentstroke}{rgb}{0.000000,0.000000,0.000000}%
\pgfsetstrokecolor{currentstroke}%
\pgfsetdash{}{0pt}%
\pgfsys@defobject{currentmarker}{\pgfqpoint{-0.048611in}{0.000000in}}{\pgfqpoint{-0.000000in}{0.000000in}}{%
\pgfpathmoveto{\pgfqpoint{-0.000000in}{0.000000in}}%
\pgfpathlineto{\pgfqpoint{-0.048611in}{0.000000in}}%
\pgfusepath{stroke,fill}%
}%
\begin{pgfscope}%
\pgfsys@transformshift{0.430556in}{1.007492in}%
\pgfsys@useobject{currentmarker}{}%
\end{pgfscope}%
\end{pgfscope}%
\begin{pgfscope}%
\definecolor{textcolor}{rgb}{0.000000,0.000000,0.000000}%
\pgfsetstrokecolor{textcolor}%
\pgfsetfillcolor{textcolor}%
\pgftext[x=0.194444in, y=0.959297in, left, base]{\color{textcolor}\rmfamily\fontsize{10.000000}{12.000000}\selectfont \(\displaystyle {20}\)}%
\end{pgfscope}%
\begin{pgfscope}%
\pgfsetbuttcap%
\pgfsetroundjoin%
\definecolor{currentfill}{rgb}{0.000000,0.000000,0.000000}%
\pgfsetfillcolor{currentfill}%
\pgfsetlinewidth{0.803000pt}%
\definecolor{currentstroke}{rgb}{0.000000,0.000000,0.000000}%
\pgfsetstrokecolor{currentstroke}%
\pgfsetdash{}{0pt}%
\pgfsys@defobject{currentmarker}{\pgfqpoint{-0.048611in}{0.000000in}}{\pgfqpoint{-0.000000in}{0.000000in}}{%
\pgfpathmoveto{\pgfqpoint{-0.000000in}{0.000000in}}%
\pgfpathlineto{\pgfqpoint{-0.048611in}{0.000000in}}%
\pgfusepath{stroke,fill}%
}%
\begin{pgfscope}%
\pgfsys@transformshift{0.430556in}{1.546120in}%
\pgfsys@useobject{currentmarker}{}%
\end{pgfscope}%
\end{pgfscope}%
\begin{pgfscope}%
\definecolor{textcolor}{rgb}{0.000000,0.000000,0.000000}%
\pgfsetstrokecolor{textcolor}%
\pgfsetfillcolor{textcolor}%
\pgftext[x=0.194444in, y=1.497926in, left, base]{\color{textcolor}\rmfamily\fontsize{10.000000}{12.000000}\selectfont \(\displaystyle {25}\)}%
\end{pgfscope}%
\begin{pgfscope}%
\definecolor{textcolor}{rgb}{0.000000,0.000000,0.000000}%
\pgfsetstrokecolor{textcolor}%
\pgfsetfillcolor{textcolor}%
\pgftext[x=0.138889in,y=1.007500in,,bottom,rotate=90.000000]{\color{textcolor}\rmfamily\fontsize{10.000000}{12.000000}\selectfont \(\displaystyle v\) (m/s)}%
\end{pgfscope}%
\begin{pgfscope}%
\pgfpathrectangle{\pgfqpoint{0.430556in}{0.415000in}}{\pgfqpoint{3.369444in}{1.185000in}}%
\pgfusepath{clip}%
\pgfsetbuttcap%
\pgfsetroundjoin%
\pgfsetlinewidth{1.505625pt}%
\definecolor{currentstroke}{rgb}{0.121569,0.466667,0.705882}%
\pgfsetstrokecolor{currentstroke}%
\pgfsetdash{{5.550000pt}{2.400000pt}}{0.000000pt}%
\pgfpathmoveto{\pgfqpoint{0.583712in}{1.546120in}}%
\pgfpathlineto{\pgfqpoint{1.513071in}{1.545175in}}%
\pgfpathlineto{\pgfqpoint{1.575897in}{1.437001in}}%
\pgfpathlineto{\pgfqpoint{1.639489in}{1.331933in}}%
\pgfpathlineto{\pgfqpoint{1.703081in}{1.231137in}}%
\pgfpathlineto{\pgfqpoint{1.767439in}{1.133299in}}%
\pgfpathlineto{\pgfqpoint{1.831796in}{1.039486in}}%
\pgfpathlineto{\pgfqpoint{1.858612in}{0.999404in}}%
\pgfpathlineto{\pgfqpoint{1.937527in}{0.876339in}}%
\pgfpathlineto{\pgfqpoint{1.963577in}{0.835371in}}%
\pgfpathlineto{\pgfqpoint{1.977368in}{0.799225in}}%
\pgfpathlineto{\pgfqpoint{1.991925in}{0.767457in}}%
\pgfpathlineto{\pgfqpoint{2.007248in}{0.739348in}}%
\pgfpathlineto{\pgfqpoint{2.023338in}{0.714331in}}%
\pgfpathlineto{\pgfqpoint{2.040194in}{0.692063in}}%
\pgfpathlineto{\pgfqpoint{2.055517in}{0.675108in}}%
\pgfpathlineto{\pgfqpoint{2.069308in}{0.663492in}}%
\pgfpathlineto{\pgfqpoint{2.084631in}{0.653495in}}%
\pgfpathlineto{\pgfqpoint{2.103019in}{0.644246in}}%
\pgfpathlineto{\pgfqpoint{2.118343in}{0.636809in}}%
\pgfpathlineto{\pgfqpoint{2.155118in}{0.595379in}}%
\pgfpathlineto{\pgfqpoint{2.191128in}{0.558122in}}%
\pgfpathlineto{\pgfqpoint{2.227138in}{0.524052in}}%
\pgfpathlineto{\pgfqpoint{2.261616in}{0.494436in}}%
\pgfpathlineto{\pgfqpoint{2.292262in}{0.470872in}}%
\pgfpathlineto{\pgfqpoint{2.296859in}{0.468863in}}%
\pgfpathlineto{\pgfqpoint{3.646843in}{0.468863in}}%
\pgfpathlineto{\pgfqpoint{3.646843in}{0.468863in}}%
\pgfusepath{stroke}%
\end{pgfscope}%
\begin{pgfscope}%
\pgfpathrectangle{\pgfqpoint{0.430556in}{0.415000in}}{\pgfqpoint{3.369444in}{1.185000in}}%
\pgfusepath{clip}%
\pgfsetrectcap%
\pgfsetroundjoin%
\pgfsetlinewidth{1.505625pt}%
\definecolor{currentstroke}{rgb}{1.000000,0.498039,0.054902}%
\pgfsetstrokecolor{currentstroke}%
\pgfsetdash{}{0pt}%
\pgfpathmoveto{\pgfqpoint{0.583712in}{1.546120in}}%
\pgfpathlineto{\pgfqpoint{1.355241in}{1.546120in}}%
\pgfpathlineto{\pgfqpoint{1.356007in}{1.537012in}}%
\pgfpathlineto{\pgfqpoint{1.362903in}{1.343738in}}%
\pgfpathlineto{\pgfqpoint{1.370565in}{1.192926in}}%
\pgfpathlineto{\pgfqpoint{1.378226in}{1.084134in}}%
\pgfpathlineto{\pgfqpoint{1.386654in}{0.995661in}}%
\pgfpathlineto{\pgfqpoint{1.395082in}{0.929006in}}%
\pgfpathlineto{\pgfqpoint{1.404276in}{0.873243in}}%
\pgfpathlineto{\pgfqpoint{1.414236in}{0.826700in}}%
\pgfpathlineto{\pgfqpoint{1.424196in}{0.790332in}}%
\pgfpathlineto{\pgfqpoint{1.434923in}{0.759317in}}%
\pgfpathlineto{\pgfqpoint{1.446415in}{0.732908in}}%
\pgfpathlineto{\pgfqpoint{1.458674in}{0.710426in}}%
\pgfpathlineto{\pgfqpoint{1.471699in}{0.691275in}}%
\pgfpathlineto{\pgfqpoint{1.485489in}{0.674943in}}%
\pgfpathlineto{\pgfqpoint{1.500813in}{0.660333in}}%
\pgfpathlineto{\pgfqpoint{1.517668in}{0.647455in}}%
\pgfpathlineto{\pgfqpoint{1.536056in}{0.636227in}}%
\pgfpathlineto{\pgfqpoint{1.556743in}{0.626184in}}%
\pgfpathlineto{\pgfqpoint{1.580494in}{0.617115in}}%
\pgfpathlineto{\pgfqpoint{1.608842in}{0.608757in}}%
\pgfpathlineto{\pgfqpoint{1.642553in}{0.601256in}}%
\pgfpathlineto{\pgfqpoint{1.683160in}{0.594604in}}%
\pgfpathlineto{\pgfqpoint{1.732961in}{0.588782in}}%
\pgfpathlineto{\pgfqpoint{1.795787in}{0.583753in}}%
\pgfpathlineto{\pgfqpoint{1.876234in}{0.579627in}}%
\pgfpathlineto{\pgfqpoint{1.947488in}{0.577952in}}%
\pgfpathlineto{\pgfqpoint{1.961279in}{0.579151in}}%
\pgfpathlineto{\pgfqpoint{1.987328in}{0.591412in}}%
\pgfpathlineto{\pgfqpoint{2.013378in}{0.606134in}}%
\pgfpathlineto{\pgfqpoint{2.038661in}{0.622953in}}%
\pgfpathlineto{\pgfqpoint{2.063179in}{0.641951in}}%
\pgfpathlineto{\pgfqpoint{2.086164in}{0.662593in}}%
\pgfpathlineto{\pgfqpoint{2.107616in}{0.684864in}}%
\pgfpathlineto{\pgfqpoint{2.127537in}{0.708760in}}%
\pgfpathlineto{\pgfqpoint{2.145924in}{0.734246in}}%
\pgfpathlineto{\pgfqpoint{2.163546in}{0.762532in}}%
\pgfpathlineto{\pgfqpoint{2.177337in}{0.785162in}}%
\pgfpathlineto{\pgfqpoint{2.201088in}{0.778343in}}%
\pgfpathlineto{\pgfqpoint{2.225606in}{0.768936in}}%
\pgfpathlineto{\pgfqpoint{2.250889in}{0.756767in}}%
\pgfpathlineto{\pgfqpoint{2.276939in}{0.741671in}}%
\pgfpathlineto{\pgfqpoint{2.303755in}{0.723487in}}%
\pgfpathlineto{\pgfqpoint{2.331337in}{0.702062in}}%
\pgfpathlineto{\pgfqpoint{2.360451in}{0.676548in}}%
\pgfpathlineto{\pgfqpoint{2.390331in}{0.647371in}}%
\pgfpathlineto{\pgfqpoint{2.421744in}{0.613547in}}%
\pgfpathlineto{\pgfqpoint{2.513684in}{0.511069in}}%
\pgfpathlineto{\pgfqpoint{2.553525in}{0.468863in}}%
\pgfpathlineto{\pgfqpoint{3.646843in}{0.468863in}}%
\pgfpathlineto{\pgfqpoint{3.646843in}{0.468863in}}%
\pgfusepath{stroke}%
\end{pgfscope}%
\begin{pgfscope}%
\pgfpathrectangle{\pgfqpoint{0.430556in}{0.415000in}}{\pgfqpoint{3.369444in}{1.185000in}}%
\pgfusepath{clip}%
\pgfsetbuttcap%
\pgfsetroundjoin%
\pgfsetlinewidth{1.505625pt}%
\definecolor{currentstroke}{rgb}{0.172549,0.627451,0.172549}%
\pgfsetstrokecolor{currentstroke}%
\pgfsetdash{{1.500000pt}{2.475000pt}}{0.000000pt}%
\pgfpathmoveto{\pgfqpoint{0.583712in}{1.007492in}}%
\pgfpathlineto{\pgfqpoint{3.646843in}{1.007492in}}%
\pgfpathlineto{\pgfqpoint{3.646843in}{1.007492in}}%
\pgfusepath{stroke}%
\end{pgfscope}%
\begin{pgfscope}%
\pgfpathrectangle{\pgfqpoint{0.430556in}{0.415000in}}{\pgfqpoint{3.369444in}{1.185000in}}%
\pgfusepath{clip}%
\pgfsetrectcap%
\pgfsetroundjoin%
\pgfsetlinewidth{1.505625pt}%
\definecolor{currentstroke}{rgb}{0.000000,0.000000,0.000000}%
\pgfsetstrokecolor{currentstroke}%
\pgfsetdash{}{0pt}%
\pgfpathmoveto{\pgfqpoint{0.583712in}{0.468863in}}%
\pgfpathlineto{\pgfqpoint{3.646843in}{0.468863in}}%
\pgfpathlineto{\pgfqpoint{3.646843in}{0.468863in}}%
\pgfusepath{stroke}%
\end{pgfscope}%
\begin{pgfscope}%
\pgfpathrectangle{\pgfqpoint{0.430556in}{0.415000in}}{\pgfqpoint{3.369444in}{1.185000in}}%
\pgfusepath{clip}%
\pgfsetrectcap%
\pgfsetroundjoin%
\pgfsetlinewidth{1.505625pt}%
\definecolor{currentstroke}{rgb}{0.000000,0.000000,0.000000}%
\pgfsetstrokecolor{currentstroke}%
\pgfsetdash{}{0pt}%
\pgfpathmoveto{\pgfqpoint{0.583712in}{1.546120in}}%
\pgfpathlineto{\pgfqpoint{3.646843in}{1.546120in}}%
\pgfpathlineto{\pgfqpoint{3.646843in}{1.546120in}}%
\pgfusepath{stroke}%
\end{pgfscope}%
\begin{pgfscope}%
\pgfsetrectcap%
\pgfsetmiterjoin%
\pgfsetlinewidth{0.803000pt}%
\definecolor{currentstroke}{rgb}{0.000000,0.000000,0.000000}%
\pgfsetstrokecolor{currentstroke}%
\pgfsetdash{}{0pt}%
\pgfpathmoveto{\pgfqpoint{0.430556in}{0.415000in}}%
\pgfpathlineto{\pgfqpoint{0.430556in}{1.600000in}}%
\pgfusepath{stroke}%
\end{pgfscope}%
\begin{pgfscope}%
\pgfsetrectcap%
\pgfsetmiterjoin%
\pgfsetlinewidth{0.803000pt}%
\definecolor{currentstroke}{rgb}{0.000000,0.000000,0.000000}%
\pgfsetstrokecolor{currentstroke}%
\pgfsetdash{}{0pt}%
\pgfpathmoveto{\pgfqpoint{3.800000in}{0.415000in}}%
\pgfpathlineto{\pgfqpoint{3.800000in}{1.600000in}}%
\pgfusepath{stroke}%
\end{pgfscope}%
\begin{pgfscope}%
\pgfsetrectcap%
\pgfsetmiterjoin%
\pgfsetlinewidth{0.803000pt}%
\definecolor{currentstroke}{rgb}{0.000000,0.000000,0.000000}%
\pgfsetstrokecolor{currentstroke}%
\pgfsetdash{}{0pt}%
\pgfpathmoveto{\pgfqpoint{0.430556in}{0.415000in}}%
\pgfpathlineto{\pgfqpoint{3.800000in}{0.415000in}}%
\pgfusepath{stroke}%
\end{pgfscope}%
\begin{pgfscope}%
\pgfsetrectcap%
\pgfsetmiterjoin%
\pgfsetlinewidth{0.803000pt}%
\definecolor{currentstroke}{rgb}{0.000000,0.000000,0.000000}%
\pgfsetstrokecolor{currentstroke}%
\pgfsetdash{}{0pt}%
\pgfpathmoveto{\pgfqpoint{0.430556in}{1.600000in}}%
\pgfpathlineto{\pgfqpoint{3.800000in}{1.600000in}}%
\pgfusepath{stroke}%
\end{pgfscope}%
\begin{pgfscope}%
\pgfsetbuttcap%
\pgfsetmiterjoin%
\definecolor{currentfill}{rgb}{1.000000,1.000000,1.000000}%
\pgfsetfillcolor{currentfill}%
\pgfsetfillopacity{0.800000}%
\pgfsetlinewidth{1.003750pt}%
\definecolor{currentstroke}{rgb}{0.800000,0.800000,0.800000}%
\pgfsetstrokecolor{currentstroke}%
\pgfsetstrokeopacity{0.800000}%
\pgfsetdash{}{0pt}%
\pgfpathmoveto{\pgfqpoint{2.742518in}{0.882408in}}%
\pgfpathlineto{\pgfqpoint{3.702778in}{0.882408in}}%
\pgfpathquadraticcurveto{\pgfqpoint{3.730556in}{0.882408in}}{\pgfqpoint{3.730556in}{0.910186in}}%
\pgfpathlineto{\pgfqpoint{3.730556in}{1.502778in}}%
\pgfpathquadraticcurveto{\pgfqpoint{3.730556in}{1.530556in}}{\pgfqpoint{3.702778in}{1.530556in}}%
\pgfpathlineto{\pgfqpoint{2.742518in}{1.530556in}}%
\pgfpathquadraticcurveto{\pgfqpoint{2.714740in}{1.530556in}}{\pgfqpoint{2.714740in}{1.502778in}}%
\pgfpathlineto{\pgfqpoint{2.714740in}{0.910186in}}%
\pgfpathquadraticcurveto{\pgfqpoint{2.714740in}{0.882408in}}{\pgfqpoint{2.742518in}{0.882408in}}%
\pgfpathclose%
\pgfusepath{stroke,fill}%
\end{pgfscope}%
\begin{pgfscope}%
\pgfsetbuttcap%
\pgfsetroundjoin%
\pgfsetlinewidth{1.505625pt}%
\definecolor{currentstroke}{rgb}{0.121569,0.466667,0.705882}%
\pgfsetstrokecolor{currentstroke}%
\pgfsetdash{{5.550000pt}{2.400000pt}}{0.000000pt}%
\pgfpathmoveto{\pgfqpoint{2.770296in}{1.426389in}}%
\pgfpathlineto{\pgfqpoint{3.048074in}{1.426389in}}%
\pgfusepath{stroke}%
\end{pgfscope}%
\begin{pgfscope}%
\definecolor{textcolor}{rgb}{0.000000,0.000000,0.000000}%
\pgfsetstrokecolor{textcolor}%
\pgfsetfillcolor{textcolor}%
\pgftext[x=3.159185in,y=1.377778in,left,base]{\color{textcolor}\rmfamily\fontsize{10.000000}{12.000000}\selectfont \(\displaystyle \gamma_{straight}\)}%
\end{pgfscope}%
\begin{pgfscope}%
\pgfsetrectcap%
\pgfsetroundjoin%
\pgfsetlinewidth{1.505625pt}%
\definecolor{currentstroke}{rgb}{1.000000,0.498039,0.054902}%
\pgfsetstrokecolor{currentstroke}%
\pgfsetdash{}{0pt}%
\pgfpathmoveto{\pgfqpoint{2.770296in}{1.219985in}}%
\pgfpathlineto{\pgfqpoint{3.048074in}{1.219985in}}%
\pgfusepath{stroke}%
\end{pgfscope}%
\begin{pgfscope}%
\definecolor{textcolor}{rgb}{0.000000,0.000000,0.000000}%
\pgfsetstrokecolor{textcolor}%
\pgfsetfillcolor{textcolor}%
\pgftext[x=3.159185in,y=1.171374in,left,base]{\color{textcolor}\rmfamily\fontsize{10.000000}{12.000000}\selectfont \(\displaystyle \gamma_{turn}\)}%
\end{pgfscope}%
\begin{pgfscope}%
\pgfsetbuttcap%
\pgfsetroundjoin%
\pgfsetlinewidth{1.505625pt}%
\definecolor{currentstroke}{rgb}{0.172549,0.627451,0.172549}%
\pgfsetstrokecolor{currentstroke}%
\pgfsetdash{{1.500000pt}{2.475000pt}}{0.000000pt}%
\pgfpathmoveto{\pgfqpoint{2.770296in}{1.026312in}}%
\pgfpathlineto{\pgfqpoint{3.048074in}{1.026312in}}%
\pgfusepath{stroke}%
\end{pgfscope}%
\begin{pgfscope}%
\definecolor{textcolor}{rgb}{0.000000,0.000000,0.000000}%
\pgfsetstrokecolor{textcolor}%
\pgfsetfillcolor{textcolor}%
\pgftext[x=3.159185in,y=0.977701in,left,base]{\color{textcolor}\rmfamily\fontsize{10.000000}{12.000000}\selectfont \(\displaystyle \gamma_{nav func}\)}%
\end{pgfscope}%
\end{pgfpicture}%
\makeatother%
\endgroup%

%% file: imgs/sim/20_veh1_actuator1.pgf
\begingroup%
\makeatletter%
\begin{pgfpicture}%
\pgfpathrectangle{\pgfpointorigin}{\pgfqpoint{3.800000in}{1.600000in}}%
\pgfusepath{use as bounding box, clip}%
\begin{pgfscope}%
\pgfsetbuttcap%
\pgfsetmiterjoin%
\definecolor{currentfill}{rgb}{1.000000,1.000000,1.000000}%
\pgfsetfillcolor{currentfill}%
\pgfsetlinewidth{0.000000pt}%
\definecolor{currentstroke}{rgb}{1.000000,1.000000,1.000000}%
\pgfsetstrokecolor{currentstroke}%
\pgfsetdash{}{0pt}%
\pgfpathmoveto{\pgfqpoint{0.000000in}{0.000000in}}%
\pgfpathlineto{\pgfqpoint{3.800000in}{0.000000in}}%
\pgfpathlineto{\pgfqpoint{3.800000in}{1.600000in}}%
\pgfpathlineto{\pgfqpoint{0.000000in}{1.600000in}}%
\pgfpathclose%
\pgfusepath{fill}%
\end{pgfscope}%
\begin{pgfscope}%
\pgfsetbuttcap%
\pgfsetmiterjoin%
\definecolor{currentfill}{rgb}{1.000000,1.000000,1.000000}%
\pgfsetfillcolor{currentfill}%
\pgfsetlinewidth{0.000000pt}%
\definecolor{currentstroke}{rgb}{0.000000,0.000000,0.000000}%
\pgfsetstrokecolor{currentstroke}%
\pgfsetstrokeopacity{0.000000}%
\pgfsetdash{}{0pt}%
\pgfpathmoveto{\pgfqpoint{0.538581in}{0.415000in}}%
\pgfpathlineto{\pgfqpoint{3.800000in}{0.415000in}}%
\pgfpathlineto{\pgfqpoint{3.800000in}{1.600000in}}%
\pgfpathlineto{\pgfqpoint{0.538581in}{1.600000in}}%
\pgfpathclose%
\pgfusepath{fill}%
\end{pgfscope}%
\begin{pgfscope}%
\pgfsetbuttcap%
\pgfsetroundjoin%
\definecolor{currentfill}{rgb}{0.000000,0.000000,0.000000}%
\pgfsetfillcolor{currentfill}%
\pgfsetlinewidth{0.803000pt}%
\definecolor{currentstroke}{rgb}{0.000000,0.000000,0.000000}%
\pgfsetstrokecolor{currentstroke}%
\pgfsetdash{}{0pt}%
\pgfsys@defobject{currentmarker}{\pgfqpoint{0.000000in}{-0.048611in}}{\pgfqpoint{0.000000in}{0.000000in}}{%
\pgfpathmoveto{\pgfqpoint{0.000000in}{0.000000in}}%
\pgfpathlineto{\pgfqpoint{0.000000in}{-0.048611in}}%
\pgfusepath{stroke,fill}%
}%
\begin{pgfscope}%
\pgfsys@transformshift{0.686085in}{0.415000in}%
\pgfsys@useobject{currentmarker}{}%
\end{pgfscope}%
\end{pgfscope}%
\begin{pgfscope}%
\definecolor{textcolor}{rgb}{0.000000,0.000000,0.000000}%
\pgfsetstrokecolor{textcolor}%
\pgfsetfillcolor{textcolor}%
\pgftext[x=0.686085in,y=0.317777in,,top]{\color{textcolor}\rmfamily\fontsize{10.000000}{12.000000}\selectfont \(\displaystyle {0}\)}%
\end{pgfscope}%
\begin{pgfscope}%
\pgfsetbuttcap%
\pgfsetroundjoin%
\definecolor{currentfill}{rgb}{0.000000,0.000000,0.000000}%
\pgfsetfillcolor{currentfill}%
\pgfsetlinewidth{0.803000pt}%
\definecolor{currentstroke}{rgb}{0.000000,0.000000,0.000000}%
\pgfsetstrokecolor{currentstroke}%
\pgfsetdash{}{0pt}%
\pgfsys@defobject{currentmarker}{\pgfqpoint{0.000000in}{-0.048611in}}{\pgfqpoint{0.000000in}{0.000000in}}{%
\pgfpathmoveto{\pgfqpoint{0.000000in}{0.000000in}}%
\pgfpathlineto{\pgfqpoint{0.000000in}{-0.048611in}}%
\pgfusepath{stroke,fill}%
}%
\begin{pgfscope}%
\pgfsys@transformshift{1.427688in}{0.415000in}%
\pgfsys@useobject{currentmarker}{}%
\end{pgfscope}%
\end{pgfscope}%
\begin{pgfscope}%
\definecolor{textcolor}{rgb}{0.000000,0.000000,0.000000}%
\pgfsetstrokecolor{textcolor}%
\pgfsetfillcolor{textcolor}%
\pgftext[x=1.427688in,y=0.317777in,,top]{\color{textcolor}\rmfamily\fontsize{10.000000}{12.000000}\selectfont \(\displaystyle {5}\)}%
\end{pgfscope}%
\begin{pgfscope}%
\pgfsetbuttcap%
\pgfsetroundjoin%
\definecolor{currentfill}{rgb}{0.000000,0.000000,0.000000}%
\pgfsetfillcolor{currentfill}%
\pgfsetlinewidth{0.803000pt}%
\definecolor{currentstroke}{rgb}{0.000000,0.000000,0.000000}%
\pgfsetstrokecolor{currentstroke}%
\pgfsetdash{}{0pt}%
\pgfsys@defobject{currentmarker}{\pgfqpoint{0.000000in}{-0.048611in}}{\pgfqpoint{0.000000in}{0.000000in}}{%
\pgfpathmoveto{\pgfqpoint{0.000000in}{0.000000in}}%
\pgfpathlineto{\pgfqpoint{0.000000in}{-0.048611in}}%
\pgfusepath{stroke,fill}%
}%
\begin{pgfscope}%
\pgfsys@transformshift{2.169290in}{0.415000in}%
\pgfsys@useobject{currentmarker}{}%
\end{pgfscope}%
\end{pgfscope}%
\begin{pgfscope}%
\definecolor{textcolor}{rgb}{0.000000,0.000000,0.000000}%
\pgfsetstrokecolor{textcolor}%
\pgfsetfillcolor{textcolor}%
\pgftext[x=2.169290in,y=0.317777in,,top]{\color{textcolor}\rmfamily\fontsize{10.000000}{12.000000}\selectfont \(\displaystyle {10}\)}%
\end{pgfscope}%
\begin{pgfscope}%
\pgfsetbuttcap%
\pgfsetroundjoin%
\definecolor{currentfill}{rgb}{0.000000,0.000000,0.000000}%
\pgfsetfillcolor{currentfill}%
\pgfsetlinewidth{0.803000pt}%
\definecolor{currentstroke}{rgb}{0.000000,0.000000,0.000000}%
\pgfsetstrokecolor{currentstroke}%
\pgfsetdash{}{0pt}%
\pgfsys@defobject{currentmarker}{\pgfqpoint{0.000000in}{-0.048611in}}{\pgfqpoint{0.000000in}{0.000000in}}{%
\pgfpathmoveto{\pgfqpoint{0.000000in}{0.000000in}}%
\pgfpathlineto{\pgfqpoint{0.000000in}{-0.048611in}}%
\pgfusepath{stroke,fill}%
}%
\begin{pgfscope}%
\pgfsys@transformshift{2.910893in}{0.415000in}%
\pgfsys@useobject{currentmarker}{}%
\end{pgfscope}%
\end{pgfscope}%
\begin{pgfscope}%
\definecolor{textcolor}{rgb}{0.000000,0.000000,0.000000}%
\pgfsetstrokecolor{textcolor}%
\pgfsetfillcolor{textcolor}%
\pgftext[x=2.910893in,y=0.317777in,,top]{\color{textcolor}\rmfamily\fontsize{10.000000}{12.000000}\selectfont \(\displaystyle {15}\)}%
\end{pgfscope}%
\begin{pgfscope}%
\pgfsetbuttcap%
\pgfsetroundjoin%
\definecolor{currentfill}{rgb}{0.000000,0.000000,0.000000}%
\pgfsetfillcolor{currentfill}%
\pgfsetlinewidth{0.803000pt}%
\definecolor{currentstroke}{rgb}{0.000000,0.000000,0.000000}%
\pgfsetstrokecolor{currentstroke}%
\pgfsetdash{}{0pt}%
\pgfsys@defobject{currentmarker}{\pgfqpoint{0.000000in}{-0.048611in}}{\pgfqpoint{0.000000in}{0.000000in}}{%
\pgfpathmoveto{\pgfqpoint{0.000000in}{0.000000in}}%
\pgfpathlineto{\pgfqpoint{0.000000in}{-0.048611in}}%
\pgfusepath{stroke,fill}%
}%
\begin{pgfscope}%
\pgfsys@transformshift{3.652495in}{0.415000in}%
\pgfsys@useobject{currentmarker}{}%
\end{pgfscope}%
\end{pgfscope}%
\begin{pgfscope}%
\definecolor{textcolor}{rgb}{0.000000,0.000000,0.000000}%
\pgfsetstrokecolor{textcolor}%
\pgfsetfillcolor{textcolor}%
\pgftext[x=3.652495in,y=0.317777in,,top]{\color{textcolor}\rmfamily\fontsize{10.000000}{12.000000}\selectfont \(\displaystyle {20}\)}%
\end{pgfscope}%
\begin{pgfscope}%
\definecolor{textcolor}{rgb}{0.000000,0.000000,0.000000}%
\pgfsetstrokecolor{textcolor}%
\pgfsetfillcolor{textcolor}%
\pgftext[x=2.169290in,y=0.138889in,,top]{\color{textcolor}\rmfamily\fontsize{10.000000}{12.000000}\selectfont time (seconds)}%
\end{pgfscope}%
\begin{pgfscope}%
\pgfsetbuttcap%
\pgfsetroundjoin%
\definecolor{currentfill}{rgb}{0.000000,0.000000,0.000000}%
\pgfsetfillcolor{currentfill}%
\pgfsetlinewidth{0.803000pt}%
\definecolor{currentstroke}{rgb}{0.000000,0.000000,0.000000}%
\pgfsetstrokecolor{currentstroke}%
\pgfsetdash{}{0pt}%
\pgfsys@defobject{currentmarker}{\pgfqpoint{-0.048611in}{0.000000in}}{\pgfqpoint{-0.000000in}{0.000000in}}{%
\pgfpathmoveto{\pgfqpoint{-0.000000in}{0.000000in}}%
\pgfpathlineto{\pgfqpoint{-0.048611in}{0.000000in}}%
\pgfusepath{stroke,fill}%
}%
\begin{pgfscope}%
\pgfsys@transformshift{0.538581in}{0.593355in}%
\pgfsys@useobject{currentmarker}{}%
\end{pgfscope}%
\end{pgfscope}%
\begin{pgfscope}%
\definecolor{textcolor}{rgb}{0.000000,0.000000,0.000000}%
\pgfsetstrokecolor{textcolor}%
\pgfsetfillcolor{textcolor}%
\pgftext[x=0.194444in, y=0.545161in, left, base]{\color{textcolor}\rmfamily\fontsize{10.000000}{12.000000}\selectfont \(\displaystyle {-10}\)}%
\end{pgfscope}%
\begin{pgfscope}%
\pgfsetbuttcap%
\pgfsetroundjoin%
\definecolor{currentfill}{rgb}{0.000000,0.000000,0.000000}%
\pgfsetfillcolor{currentfill}%
\pgfsetlinewidth{0.803000pt}%
\definecolor{currentstroke}{rgb}{0.000000,0.000000,0.000000}%
\pgfsetstrokecolor{currentstroke}%
\pgfsetdash{}{0pt}%
\pgfsys@defobject{currentmarker}{\pgfqpoint{-0.048611in}{0.000000in}}{\pgfqpoint{-0.000000in}{0.000000in}}{%
\pgfpathmoveto{\pgfqpoint{-0.000000in}{0.000000in}}%
\pgfpathlineto{\pgfqpoint{-0.048611in}{0.000000in}}%
\pgfusepath{stroke,fill}%
}%
\begin{pgfscope}%
\pgfsys@transformshift{0.538581in}{1.007608in}%
\pgfsys@useobject{currentmarker}{}%
\end{pgfscope}%
\end{pgfscope}%
\begin{pgfscope}%
\definecolor{textcolor}{rgb}{0.000000,0.000000,0.000000}%
\pgfsetstrokecolor{textcolor}%
\pgfsetfillcolor{textcolor}%
\pgftext[x=0.371914in, y=0.959413in, left, base]{\color{textcolor}\rmfamily\fontsize{10.000000}{12.000000}\selectfont \(\displaystyle {0}\)}%
\end{pgfscope}%
\begin{pgfscope}%
\pgfsetbuttcap%
\pgfsetroundjoin%
\definecolor{currentfill}{rgb}{0.000000,0.000000,0.000000}%
\pgfsetfillcolor{currentfill}%
\pgfsetlinewidth{0.803000pt}%
\definecolor{currentstroke}{rgb}{0.000000,0.000000,0.000000}%
\pgfsetstrokecolor{currentstroke}%
\pgfsetdash{}{0pt}%
\pgfsys@defobject{currentmarker}{\pgfqpoint{-0.048611in}{0.000000in}}{\pgfqpoint{-0.000000in}{0.000000in}}{%
\pgfpathmoveto{\pgfqpoint{-0.000000in}{0.000000in}}%
\pgfpathlineto{\pgfqpoint{-0.048611in}{0.000000in}}%
\pgfusepath{stroke,fill}%
}%
\begin{pgfscope}%
\pgfsys@transformshift{0.538581in}{1.421861in}%
\pgfsys@useobject{currentmarker}{}%
\end{pgfscope}%
\end{pgfscope}%
\begin{pgfscope}%
\definecolor{textcolor}{rgb}{0.000000,0.000000,0.000000}%
\pgfsetstrokecolor{textcolor}%
\pgfsetfillcolor{textcolor}%
\pgftext[x=0.302469in, y=1.373666in, left, base]{\color{textcolor}\rmfamily\fontsize{10.000000}{12.000000}\selectfont \(\displaystyle {10}\)}%
\end{pgfscope}%
\begin{pgfscope}%
\definecolor{textcolor}{rgb}{0.000000,0.000000,0.000000}%
\pgfsetstrokecolor{textcolor}%
\pgfsetfillcolor{textcolor}%
\pgftext[x=0.138889in,y=1.007500in,,bottom,rotate=90.000000]{\color{textcolor}\rmfamily\fontsize{10.000000}{12.000000}\selectfont \(\displaystyle \omega\) (deg/sec)}%
\end{pgfscope}%
\begin{pgfscope}%
\pgfpathrectangle{\pgfqpoint{0.538581in}{0.415000in}}{\pgfqpoint{3.261419in}{1.185000in}}%
\pgfusepath{clip}%
\pgfsetbuttcap%
\pgfsetroundjoin%
\pgfsetlinewidth{1.505625pt}%
\definecolor{currentstroke}{rgb}{0.121569,0.466667,0.705882}%
\pgfsetstrokecolor{currentstroke}%
\pgfsetdash{{5.550000pt}{2.400000pt}}{0.000000pt}%
\pgfpathmoveto{\pgfqpoint{0.686827in}{0.469079in}}%
\pgfpathlineto{\pgfqpoint{0.956770in}{0.469079in}}%
\pgfpathlineto{\pgfqpoint{0.957512in}{0.470908in}}%
\pgfpathlineto{\pgfqpoint{0.958995in}{0.469079in}}%
\pgfpathlineto{\pgfqpoint{0.985693in}{0.469079in}}%
\pgfpathlineto{\pgfqpoint{0.986434in}{0.490505in}}%
\pgfpathlineto{\pgfqpoint{0.990142in}{0.810413in}}%
\pgfpathlineto{\pgfqpoint{0.993850in}{0.932266in}}%
\pgfpathlineto{\pgfqpoint{0.997558in}{0.978768in}}%
\pgfpathlineto{\pgfqpoint{1.001266in}{0.996547in}}%
\pgfpathlineto{\pgfqpoint{1.004974in}{1.003358in}}%
\pgfpathlineto{\pgfqpoint{1.009424in}{1.006256in}}%
\pgfpathlineto{\pgfqpoint{1.016099in}{1.007364in}}%
\pgfpathlineto{\pgfqpoint{1.039830in}{1.007607in}}%
\pgfpathlineto{\pgfqpoint{1.907505in}{1.007608in}}%
\pgfpathlineto{\pgfqpoint{1.913437in}{0.909751in}}%
\pgfpathlineto{\pgfqpoint{1.919370in}{0.847524in}}%
\pgfpathlineto{\pgfqpoint{1.925303in}{0.805907in}}%
\pgfpathlineto{\pgfqpoint{1.931978in}{0.774317in}}%
\pgfpathlineto{\pgfqpoint{1.938652in}{0.753537in}}%
\pgfpathlineto{\pgfqpoint{1.944585in}{0.741545in}}%
\pgfpathlineto{\pgfqpoint{1.950518in}{0.734118in}}%
\pgfpathlineto{\pgfqpoint{1.956450in}{0.730195in}}%
\pgfpathlineto{\pgfqpoint{1.962383in}{0.728976in}}%
\pgfpathlineto{\pgfqpoint{1.969058in}{0.730072in}}%
\pgfpathlineto{\pgfqpoint{1.977215in}{0.734030in}}%
\pgfpathlineto{\pgfqpoint{1.987598in}{0.741930in}}%
\pgfpathlineto{\pgfqpoint{2.001688in}{0.755758in}}%
\pgfpathlineto{\pgfqpoint{2.020970in}{0.777507in}}%
\pgfpathlineto{\pgfqpoint{2.031352in}{0.816987in}}%
\pgfpathlineto{\pgfqpoint{2.041735in}{0.847167in}}%
\pgfpathlineto{\pgfqpoint{2.052117in}{0.870737in}}%
\pgfpathlineto{\pgfqpoint{2.063241in}{0.890726in}}%
\pgfpathlineto{\pgfqpoint{2.075107in}{0.907697in}}%
\pgfpathlineto{\pgfqpoint{2.112187in}{0.953066in}}%
\pgfpathlineto{\pgfqpoint{2.124053in}{0.963312in}}%
\pgfpathlineto{\pgfqpoint{2.138143in}{0.972539in}}%
\pgfpathlineto{\pgfqpoint{2.155200in}{0.980959in}}%
\pgfpathlineto{\pgfqpoint{2.171515in}{0.988698in}}%
\pgfpathlineto{\pgfqpoint{2.249383in}{1.158356in}}%
\pgfpathlineto{\pgfqpoint{2.274598in}{1.219810in}}%
\pgfpathlineto{\pgfqpoint{2.294621in}{1.274501in}}%
\pgfpathlineto{\pgfqpoint{2.310936in}{1.325413in}}%
\pgfpathlineto{\pgfqpoint{2.322060in}{1.366395in}}%
\pgfpathlineto{\pgfqpoint{2.331701in}{1.411403in}}%
\pgfpathlineto{\pgfqpoint{2.340601in}{1.464358in}}%
\pgfpathlineto{\pgfqpoint{2.353208in}{1.542614in}}%
\pgfpathlineto{\pgfqpoint{2.354691in}{1.546136in}}%
\pgfpathlineto{\pgfqpoint{2.366557in}{1.546136in}}%
\pgfpathlineto{\pgfqpoint{2.368040in}{1.489286in}}%
\pgfpathlineto{\pgfqpoint{2.373231in}{1.288535in}}%
\pgfpathlineto{\pgfqpoint{2.378422in}{1.171774in}}%
\pgfpathlineto{\pgfqpoint{2.383613in}{1.103731in}}%
\pgfpathlineto{\pgfqpoint{2.388805in}{1.064002in}}%
\pgfpathlineto{\pgfqpoint{2.393996in}{1.040760in}}%
\pgfpathlineto{\pgfqpoint{2.399187in}{1.027135in}}%
\pgfpathlineto{\pgfqpoint{2.404378in}{1.019132in}}%
\pgfpathlineto{\pgfqpoint{2.410311in}{1.013931in}}%
\pgfpathlineto{\pgfqpoint{2.417727in}{1.010605in}}%
\pgfpathlineto{\pgfqpoint{2.428110in}{1.008669in}}%
\pgfpathlineto{\pgfqpoint{2.447391in}{1.007765in}}%
\pgfpathlineto{\pgfqpoint{2.531934in}{1.007608in}}%
\pgfpathlineto{\pgfqpoint{3.651754in}{1.007608in}}%
\pgfpathlineto{\pgfqpoint{3.651754in}{1.007608in}}%
\pgfusepath{stroke}%
\end{pgfscope}%
\begin{pgfscope}%
\pgfpathrectangle{\pgfqpoint{0.538581in}{0.415000in}}{\pgfqpoint{3.261419in}{1.185000in}}%
\pgfusepath{clip}%
\pgfsetrectcap%
\pgfsetroundjoin%
\pgfsetlinewidth{1.505625pt}%
\definecolor{currentstroke}{rgb}{1.000000,0.498039,0.054902}%
\pgfsetstrokecolor{currentstroke}%
\pgfsetdash{}{0pt}%
\pgfpathmoveto{\pgfqpoint{0.686827in}{1.007608in}}%
\pgfpathlineto{\pgfqpoint{1.423980in}{1.007608in}}%
\pgfpathlineto{\pgfqpoint{1.425463in}{0.981594in}}%
\pgfpathlineto{\pgfqpoint{1.434362in}{0.775089in}}%
\pgfpathlineto{\pgfqpoint{1.441037in}{0.907603in}}%
\pgfpathlineto{\pgfqpoint{1.448453in}{1.010508in}}%
\pgfpathlineto{\pgfqpoint{1.455869in}{1.084580in}}%
\pgfpathlineto{\pgfqpoint{1.464026in}{1.144813in}}%
\pgfpathlineto{\pgfqpoint{1.472926in}{1.193835in}}%
\pgfpathlineto{\pgfqpoint{1.481825in}{1.231143in}}%
\pgfpathlineto{\pgfqpoint{1.491466in}{1.262469in}}%
\pgfpathlineto{\pgfqpoint{1.501848in}{1.288777in}}%
\pgfpathlineto{\pgfqpoint{1.512972in}{1.310899in}}%
\pgfpathlineto{\pgfqpoint{1.524838in}{1.329538in}}%
\pgfpathlineto{\pgfqpoint{1.537445in}{1.345274in}}%
\pgfpathlineto{\pgfqpoint{1.551535in}{1.359244in}}%
\pgfpathlineto{\pgfqpoint{1.567109in}{1.371433in}}%
\pgfpathlineto{\pgfqpoint{1.584166in}{1.381917in}}%
\pgfpathlineto{\pgfqpoint{1.603448in}{1.391125in}}%
\pgfpathlineto{\pgfqpoint{1.625696in}{1.399197in}}%
\pgfpathlineto{\pgfqpoint{1.650910in}{1.405916in}}%
\pgfpathlineto{\pgfqpoint{1.679833in}{1.411258in}}%
\pgfpathlineto{\pgfqpoint{1.712463in}{1.414944in}}%
\pgfpathlineto{\pgfqpoint{1.747319in}{1.416593in}}%
\pgfpathlineto{\pgfqpoint{1.781432in}{1.416030in}}%
\pgfpathlineto{\pgfqpoint{1.813321in}{1.413340in}}%
\pgfpathlineto{\pgfqpoint{1.840760in}{1.408898in}}%
\pgfpathlineto{\pgfqpoint{1.864492in}{1.402859in}}%
\pgfpathlineto{\pgfqpoint{1.884515in}{1.395487in}}%
\pgfpathlineto{\pgfqpoint{1.901572in}{1.386810in}}%
\pgfpathlineto{\pgfqpoint{1.915662in}{1.377228in}}%
\pgfpathlineto{\pgfqpoint{1.928270in}{1.365984in}}%
\pgfpathlineto{\pgfqpoint{1.939394in}{1.353053in}}%
\pgfpathlineto{\pgfqpoint{1.949034in}{1.338575in}}%
\pgfpathlineto{\pgfqpoint{1.957934in}{1.321341in}}%
\pgfpathlineto{\pgfqpoint{1.966091in}{1.300837in}}%
\pgfpathlineto{\pgfqpoint{1.973507in}{1.276502in}}%
\pgfpathlineto{\pgfqpoint{1.980923in}{1.244069in}}%
\pgfpathlineto{\pgfqpoint{1.987598in}{1.204294in}}%
\pgfpathlineto{\pgfqpoint{1.993531in}{1.155773in}}%
\pgfpathlineto{\pgfqpoint{1.999463in}{1.087327in}}%
\pgfpathlineto{\pgfqpoint{2.004655in}{1.000466in}}%
\pgfpathlineto{\pgfqpoint{2.009846in}{0.869601in}}%
\pgfpathlineto{\pgfqpoint{2.014295in}{0.692180in}}%
\pgfpathlineto{\pgfqpoint{2.019487in}{0.469079in}}%
\pgfpathlineto{\pgfqpoint{2.462223in}{0.469079in}}%
\pgfpathlineto{\pgfqpoint{2.463707in}{0.532063in}}%
\pgfpathlineto{\pgfqpoint{2.468156in}{0.722469in}}%
\pgfpathlineto{\pgfqpoint{2.472606in}{0.836367in}}%
\pgfpathlineto{\pgfqpoint{2.477055in}{0.904605in}}%
\pgfpathlineto{\pgfqpoint{2.482247in}{0.950569in}}%
\pgfpathlineto{\pgfqpoint{2.487438in}{0.975954in}}%
\pgfpathlineto{\pgfqpoint{2.492629in}{0.990004in}}%
\pgfpathlineto{\pgfqpoint{2.497820in}{0.997796in}}%
\pgfpathlineto{\pgfqpoint{2.503753in}{1.002565in}}%
\pgfpathlineto{\pgfqpoint{2.511169in}{1.005404in}}%
\pgfpathlineto{\pgfqpoint{2.522293in}{1.006966in}}%
\pgfpathlineto{\pgfqpoint{2.546766in}{1.007564in}}%
\pgfpathlineto{\pgfqpoint{2.798169in}{1.007608in}}%
\pgfpathlineto{\pgfqpoint{3.651754in}{1.007608in}}%
\pgfpathlineto{\pgfqpoint{3.651754in}{1.007608in}}%
\pgfusepath{stroke}%
\end{pgfscope}%
\begin{pgfscope}%
\pgfpathrectangle{\pgfqpoint{0.538581in}{0.415000in}}{\pgfqpoint{3.261419in}{1.185000in}}%
\pgfusepath{clip}%
\pgfsetbuttcap%
\pgfsetroundjoin%
\pgfsetlinewidth{1.505625pt}%
\definecolor{currentstroke}{rgb}{0.172549,0.627451,0.172549}%
\pgfsetstrokecolor{currentstroke}%
\pgfsetdash{{1.500000pt}{2.475000pt}}{0.000000pt}%
\pgfpathmoveto{\pgfqpoint{0.686827in}{1.232055in}}%
\pgfpathlineto{\pgfqpoint{0.720941in}{1.206522in}}%
\pgfpathlineto{\pgfqpoint{0.765437in}{1.170203in}}%
\pgfpathlineto{\pgfqpoint{0.897442in}{1.060599in}}%
\pgfpathlineto{\pgfqpoint{0.907825in}{1.052765in}}%
\pgfpathlineto{\pgfqpoint{0.908566in}{1.117894in}}%
\pgfpathlineto{\pgfqpoint{0.909308in}{1.117152in}}%
\pgfpathlineto{\pgfqpoint{0.957512in}{1.070538in}}%
\pgfpathlineto{\pgfqpoint{0.993850in}{1.038361in}}%
\pgfpathlineto{\pgfqpoint{1.025739in}{1.012976in}}%
\pgfpathlineto{\pgfqpoint{1.055403in}{0.992170in}}%
\pgfpathlineto{\pgfqpoint{1.082843in}{0.975568in}}%
\pgfpathlineto{\pgfqpoint{1.109540in}{0.961982in}}%
\pgfpathlineto{\pgfqpoint{1.122148in}{0.956465in}}%
\pgfpathlineto{\pgfqpoint{1.122889in}{1.037778in}}%
\pgfpathlineto{\pgfqpoint{1.123631in}{1.037243in}}%
\pgfpathlineto{\pgfqpoint{1.157003in}{1.014575in}}%
\pgfpathlineto{\pgfqpoint{1.187409in}{0.996548in}}%
\pgfpathlineto{\pgfqpoint{1.216331in}{0.981943in}}%
\pgfpathlineto{\pgfqpoint{1.244512in}{0.970205in}}%
\pgfpathlineto{\pgfqpoint{1.255636in}{0.966258in}}%
\pgfpathlineto{\pgfqpoint{1.256378in}{1.061373in}}%
\pgfpathlineto{\pgfqpoint{1.257119in}{1.060859in}}%
\pgfpathlineto{\pgfqpoint{1.294941in}{1.036006in}}%
\pgfpathlineto{\pgfqpoint{1.328313in}{1.016635in}}%
\pgfpathlineto{\pgfqpoint{1.340920in}{1.010023in}}%
\pgfpathlineto{\pgfqpoint{1.341662in}{1.114980in}}%
\pgfpathlineto{\pgfqpoint{1.342404in}{1.114315in}}%
\pgfpathlineto{\pgfqpoint{1.386900in}{1.075944in}}%
\pgfpathlineto{\pgfqpoint{1.394316in}{1.069906in}}%
\pgfpathlineto{\pgfqpoint{1.395057in}{1.179925in}}%
\pgfpathlineto{\pgfqpoint{1.395799in}{1.179030in}}%
\pgfpathlineto{\pgfqpoint{1.423980in}{1.145576in}}%
\pgfpathlineto{\pgfqpoint{1.424721in}{1.255483in}}%
\pgfpathlineto{\pgfqpoint{1.425463in}{1.254327in}}%
\pgfpathlineto{\pgfqpoint{1.432879in}{1.242804in}}%
\pgfpathlineto{\pgfqpoint{1.433621in}{1.295502in}}%
\pgfpathlineto{\pgfqpoint{1.434362in}{1.294213in}}%
\pgfpathlineto{\pgfqpoint{1.483308in}{1.211352in}}%
\pgfpathlineto{\pgfqpoint{1.518905in}{1.155415in}}%
\pgfpathlineto{\pgfqpoint{1.550052in}{1.110522in}}%
\pgfpathlineto{\pgfqpoint{1.578975in}{1.072606in}}%
\pgfpathlineto{\pgfqpoint{1.607156in}{1.039238in}}%
\pgfpathlineto{\pgfqpoint{1.635337in}{1.009288in}}%
\pgfpathlineto{\pgfqpoint{1.664259in}{0.981826in}}%
\pgfpathlineto{\pgfqpoint{1.694665in}{0.956083in}}%
\pgfpathlineto{\pgfqpoint{1.728037in}{0.930832in}}%
\pgfpathlineto{\pgfqpoint{1.768083in}{0.903484in}}%
\pgfpathlineto{\pgfqpoint{1.908988in}{0.809996in}}%
\pgfpathlineto{\pgfqpoint{1.939394in}{0.785567in}}%
\pgfpathlineto{\pgfqpoint{1.966833in}{0.760712in}}%
\pgfpathlineto{\pgfqpoint{1.992047in}{0.734911in}}%
\pgfpathlineto{\pgfqpoint{2.015779in}{0.707511in}}%
\pgfpathlineto{\pgfqpoint{2.038768in}{0.677622in}}%
\pgfpathlineto{\pgfqpoint{2.061758in}{0.644032in}}%
\pgfpathlineto{\pgfqpoint{2.084748in}{0.606453in}}%
\pgfpathlineto{\pgfqpoint{2.109221in}{0.562017in}}%
\pgfpathlineto{\pgfqpoint{2.136660in}{0.507359in}}%
\pgfpathlineto{\pgfqpoint{2.155942in}{0.469079in}}%
\pgfpathlineto{\pgfqpoint{2.494854in}{0.469438in}}%
\pgfpathlineto{\pgfqpoint{2.511169in}{0.566584in}}%
\pgfpathlineto{\pgfqpoint{2.528226in}{0.682111in}}%
\pgfpathlineto{\pgfqpoint{2.551216in}{0.854854in}}%
\pgfpathlineto{\pgfqpoint{2.580880in}{1.075312in}}%
\pgfpathlineto{\pgfqpoint{2.596453in}{1.174322in}}%
\pgfpathlineto{\pgfqpoint{2.610544in}{1.249917in}}%
\pgfpathlineto{\pgfqpoint{2.623893in}{1.309751in}}%
\pgfpathlineto{\pgfqpoint{2.637983in}{1.362535in}}%
\pgfpathlineto{\pgfqpoint{2.653557in}{1.411944in}}%
\pgfpathlineto{\pgfqpoint{2.678030in}{1.480695in}}%
\pgfpathlineto{\pgfqpoint{2.698795in}{1.541592in}}%
\pgfpathlineto{\pgfqpoint{2.701019in}{1.546136in}}%
\pgfpathlineto{\pgfqpoint{3.544221in}{1.546136in}}%
\pgfpathlineto{\pgfqpoint{3.547929in}{1.496028in}}%
\pgfpathlineto{\pgfqpoint{3.556087in}{1.357061in}}%
\pgfpathlineto{\pgfqpoint{3.563503in}{1.187495in}}%
\pgfpathlineto{\pgfqpoint{3.570177in}{0.978074in}}%
\pgfpathlineto{\pgfqpoint{3.576110in}{0.722716in}}%
\pgfpathlineto{\pgfqpoint{3.582043in}{0.469079in}}%
\pgfpathlineto{\pgfqpoint{3.651754in}{0.469079in}}%
\pgfpathlineto{\pgfqpoint{3.651754in}{0.469079in}}%
\pgfusepath{stroke}%
\end{pgfscope}%
\begin{pgfscope}%
\pgfpathrectangle{\pgfqpoint{0.538581in}{0.415000in}}{\pgfqpoint{3.261419in}{1.185000in}}%
\pgfusepath{clip}%
\pgfsetrectcap%
\pgfsetroundjoin%
\pgfsetlinewidth{1.505625pt}%
\definecolor{currentstroke}{rgb}{0.000000,0.000000,0.000000}%
\pgfsetstrokecolor{currentstroke}%
\pgfsetdash{}{0pt}%
\pgfpathmoveto{\pgfqpoint{0.686827in}{0.469079in}}%
\pgfpathlineto{\pgfqpoint{3.651754in}{0.469079in}}%
\pgfpathlineto{\pgfqpoint{3.651754in}{0.469079in}}%
\pgfusepath{stroke}%
\end{pgfscope}%
\begin{pgfscope}%
\pgfpathrectangle{\pgfqpoint{0.538581in}{0.415000in}}{\pgfqpoint{3.261419in}{1.185000in}}%
\pgfusepath{clip}%
\pgfsetrectcap%
\pgfsetroundjoin%
\pgfsetlinewidth{1.505625pt}%
\definecolor{currentstroke}{rgb}{0.000000,0.000000,0.000000}%
\pgfsetstrokecolor{currentstroke}%
\pgfsetdash{}{0pt}%
\pgfpathmoveto{\pgfqpoint{0.686827in}{1.546136in}}%
\pgfpathlineto{\pgfqpoint{3.651754in}{1.546136in}}%
\pgfpathlineto{\pgfqpoint{3.651754in}{1.546136in}}%
\pgfusepath{stroke}%
\end{pgfscope}%
\begin{pgfscope}%
\pgfsetrectcap%
\pgfsetmiterjoin%
\pgfsetlinewidth{0.803000pt}%
\definecolor{currentstroke}{rgb}{0.000000,0.000000,0.000000}%
\pgfsetstrokecolor{currentstroke}%
\pgfsetdash{}{0pt}%
\pgfpathmoveto{\pgfqpoint{0.538581in}{0.415000in}}%
\pgfpathlineto{\pgfqpoint{0.538581in}{1.600000in}}%
\pgfusepath{stroke}%
\end{pgfscope}%
\begin{pgfscope}%
\pgfsetrectcap%
\pgfsetmiterjoin%
\pgfsetlinewidth{0.803000pt}%
\definecolor{currentstroke}{rgb}{0.000000,0.000000,0.000000}%
\pgfsetstrokecolor{currentstroke}%
\pgfsetdash{}{0pt}%
\pgfpathmoveto{\pgfqpoint{3.800000in}{0.415000in}}%
\pgfpathlineto{\pgfqpoint{3.800000in}{1.600000in}}%
\pgfusepath{stroke}%
\end{pgfscope}%
\begin{pgfscope}%
\pgfsetrectcap%
\pgfsetmiterjoin%
\pgfsetlinewidth{0.803000pt}%
\definecolor{currentstroke}{rgb}{0.000000,0.000000,0.000000}%
\pgfsetstrokecolor{currentstroke}%
\pgfsetdash{}{0pt}%
\pgfpathmoveto{\pgfqpoint{0.538581in}{0.415000in}}%
\pgfpathlineto{\pgfqpoint{3.800000in}{0.415000in}}%
\pgfusepath{stroke}%
\end{pgfscope}%
\begin{pgfscope}%
\pgfsetrectcap%
\pgfsetmiterjoin%
\pgfsetlinewidth{0.803000pt}%
\definecolor{currentstroke}{rgb}{0.000000,0.000000,0.000000}%
\pgfsetstrokecolor{currentstroke}%
\pgfsetdash{}{0pt}%
\pgfpathmoveto{\pgfqpoint{0.538581in}{1.600000in}}%
\pgfpathlineto{\pgfqpoint{3.800000in}{1.600000in}}%
\pgfusepath{stroke}%
\end{pgfscope}%
\end{pgfpicture}%
\makeatother%
\endgroup%

%% file: imgs/sim/20_veh_min_d.pgf
\begingroup%
\makeatletter%
\begin{pgfpicture}%
\pgfpathrectangle{\pgfpointorigin}{\pgfqpoint{3.800000in}{1.600000in}}%
\pgfusepath{use as bounding box, clip}%
\begin{pgfscope}%
\pgfsetbuttcap%
\pgfsetmiterjoin%
\definecolor{currentfill}{rgb}{1.000000,1.000000,1.000000}%
\pgfsetfillcolor{currentfill}%
\pgfsetlinewidth{0.000000pt}%
\definecolor{currentstroke}{rgb}{1.000000,1.000000,1.000000}%
\pgfsetstrokecolor{currentstroke}%
\pgfsetdash{}{0pt}%
\pgfpathmoveto{\pgfqpoint{0.000000in}{0.000000in}}%
\pgfpathlineto{\pgfqpoint{3.800000in}{0.000000in}}%
\pgfpathlineto{\pgfqpoint{3.800000in}{1.600000in}}%
\pgfpathlineto{\pgfqpoint{0.000000in}{1.600000in}}%
\pgfpathclose%
\pgfusepath{fill}%
\end{pgfscope}%
\begin{pgfscope}%
\pgfsetbuttcap%
\pgfsetmiterjoin%
\definecolor{currentfill}{rgb}{1.000000,1.000000,1.000000}%
\pgfsetfillcolor{currentfill}%
\pgfsetlinewidth{0.000000pt}%
\definecolor{currentstroke}{rgb}{0.000000,0.000000,0.000000}%
\pgfsetstrokecolor{currentstroke}%
\pgfsetstrokeopacity{0.000000}%
\pgfsetdash{}{0pt}%
\pgfpathmoveto{\pgfqpoint{0.430556in}{0.415000in}}%
\pgfpathlineto{\pgfqpoint{3.800000in}{0.415000in}}%
\pgfpathlineto{\pgfqpoint{3.800000in}{1.600000in}}%
\pgfpathlineto{\pgfqpoint{0.430556in}{1.600000in}}%
\pgfpathclose%
\pgfusepath{fill}%
\end{pgfscope}%
\begin{pgfscope}%
\pgfsetbuttcap%
\pgfsetroundjoin%
\definecolor{currentfill}{rgb}{0.000000,0.000000,0.000000}%
\pgfsetfillcolor{currentfill}%
\pgfsetlinewidth{0.803000pt}%
\definecolor{currentstroke}{rgb}{0.000000,0.000000,0.000000}%
\pgfsetstrokecolor{currentstroke}%
\pgfsetdash{}{0pt}%
\pgfsys@defobject{currentmarker}{\pgfqpoint{0.000000in}{-0.048611in}}{\pgfqpoint{0.000000in}{0.000000in}}{%
\pgfpathmoveto{\pgfqpoint{0.000000in}{0.000000in}}%
\pgfpathlineto{\pgfqpoint{0.000000in}{-0.048611in}}%
\pgfusepath{stroke,fill}%
}%
\begin{pgfscope}%
\pgfsys@transformshift{0.582946in}{0.415000in}%
\pgfsys@useobject{currentmarker}{}%
\end{pgfscope}%
\end{pgfscope}%
\begin{pgfscope}%
\definecolor{textcolor}{rgb}{0.000000,0.000000,0.000000}%
\pgfsetstrokecolor{textcolor}%
\pgfsetfillcolor{textcolor}%
\pgftext[x=0.582946in,y=0.317777in,,top]{\color{textcolor}\rmfamily\fontsize{10.000000}{12.000000}\selectfont \(\displaystyle {0}\)}%
\end{pgfscope}%
\begin{pgfscope}%
\pgfsetbuttcap%
\pgfsetroundjoin%
\definecolor{currentfill}{rgb}{0.000000,0.000000,0.000000}%
\pgfsetfillcolor{currentfill}%
\pgfsetlinewidth{0.803000pt}%
\definecolor{currentstroke}{rgb}{0.000000,0.000000,0.000000}%
\pgfsetstrokecolor{currentstroke}%
\pgfsetdash{}{0pt}%
\pgfsys@defobject{currentmarker}{\pgfqpoint{0.000000in}{-0.048611in}}{\pgfqpoint{0.000000in}{0.000000in}}{%
\pgfpathmoveto{\pgfqpoint{0.000000in}{0.000000in}}%
\pgfpathlineto{\pgfqpoint{0.000000in}{-0.048611in}}%
\pgfusepath{stroke,fill}%
}%
\begin{pgfscope}%
\pgfsys@transformshift{1.349112in}{0.415000in}%
\pgfsys@useobject{currentmarker}{}%
\end{pgfscope}%
\end{pgfscope}%
\begin{pgfscope}%
\definecolor{textcolor}{rgb}{0.000000,0.000000,0.000000}%
\pgfsetstrokecolor{textcolor}%
\pgfsetfillcolor{textcolor}%
\pgftext[x=1.349112in,y=0.317777in,,top]{\color{textcolor}\rmfamily\fontsize{10.000000}{12.000000}\selectfont \(\displaystyle {5}\)}%
\end{pgfscope}%
\begin{pgfscope}%
\pgfsetbuttcap%
\pgfsetroundjoin%
\definecolor{currentfill}{rgb}{0.000000,0.000000,0.000000}%
\pgfsetfillcolor{currentfill}%
\pgfsetlinewidth{0.803000pt}%
\definecolor{currentstroke}{rgb}{0.000000,0.000000,0.000000}%
\pgfsetstrokecolor{currentstroke}%
\pgfsetdash{}{0pt}%
\pgfsys@defobject{currentmarker}{\pgfqpoint{0.000000in}{-0.048611in}}{\pgfqpoint{0.000000in}{0.000000in}}{%
\pgfpathmoveto{\pgfqpoint{0.000000in}{0.000000in}}%
\pgfpathlineto{\pgfqpoint{0.000000in}{-0.048611in}}%
\pgfusepath{stroke,fill}%
}%
\begin{pgfscope}%
\pgfsys@transformshift{2.115278in}{0.415000in}%
\pgfsys@useobject{currentmarker}{}%
\end{pgfscope}%
\end{pgfscope}%
\begin{pgfscope}%
\definecolor{textcolor}{rgb}{0.000000,0.000000,0.000000}%
\pgfsetstrokecolor{textcolor}%
\pgfsetfillcolor{textcolor}%
\pgftext[x=2.115278in,y=0.317777in,,top]{\color{textcolor}\rmfamily\fontsize{10.000000}{12.000000}\selectfont \(\displaystyle {10}\)}%
\end{pgfscope}%
\begin{pgfscope}%
\pgfsetbuttcap%
\pgfsetroundjoin%
\definecolor{currentfill}{rgb}{0.000000,0.000000,0.000000}%
\pgfsetfillcolor{currentfill}%
\pgfsetlinewidth{0.803000pt}%
\definecolor{currentstroke}{rgb}{0.000000,0.000000,0.000000}%
\pgfsetstrokecolor{currentstroke}%
\pgfsetdash{}{0pt}%
\pgfsys@defobject{currentmarker}{\pgfqpoint{0.000000in}{-0.048611in}}{\pgfqpoint{0.000000in}{0.000000in}}{%
\pgfpathmoveto{\pgfqpoint{0.000000in}{0.000000in}}%
\pgfpathlineto{\pgfqpoint{0.000000in}{-0.048611in}}%
\pgfusepath{stroke,fill}%
}%
\begin{pgfscope}%
\pgfsys@transformshift{2.881444in}{0.415000in}%
\pgfsys@useobject{currentmarker}{}%
\end{pgfscope}%
\end{pgfscope}%
\begin{pgfscope}%
\definecolor{textcolor}{rgb}{0.000000,0.000000,0.000000}%
\pgfsetstrokecolor{textcolor}%
\pgfsetfillcolor{textcolor}%
\pgftext[x=2.881444in,y=0.317777in,,top]{\color{textcolor}\rmfamily\fontsize{10.000000}{12.000000}\selectfont \(\displaystyle {15}\)}%
\end{pgfscope}%
\begin{pgfscope}%
\pgfsetbuttcap%
\pgfsetroundjoin%
\definecolor{currentfill}{rgb}{0.000000,0.000000,0.000000}%
\pgfsetfillcolor{currentfill}%
\pgfsetlinewidth{0.803000pt}%
\definecolor{currentstroke}{rgb}{0.000000,0.000000,0.000000}%
\pgfsetstrokecolor{currentstroke}%
\pgfsetdash{}{0pt}%
\pgfsys@defobject{currentmarker}{\pgfqpoint{0.000000in}{-0.048611in}}{\pgfqpoint{0.000000in}{0.000000in}}{%
\pgfpathmoveto{\pgfqpoint{0.000000in}{0.000000in}}%
\pgfpathlineto{\pgfqpoint{0.000000in}{-0.048611in}}%
\pgfusepath{stroke,fill}%
}%
\begin{pgfscope}%
\pgfsys@transformshift{3.647610in}{0.415000in}%
\pgfsys@useobject{currentmarker}{}%
\end{pgfscope}%
\end{pgfscope}%
\begin{pgfscope}%
\definecolor{textcolor}{rgb}{0.000000,0.000000,0.000000}%
\pgfsetstrokecolor{textcolor}%
\pgfsetfillcolor{textcolor}%
\pgftext[x=3.647610in,y=0.317777in,,top]{\color{textcolor}\rmfamily\fontsize{10.000000}{12.000000}\selectfont \(\displaystyle {20}\)}%
\end{pgfscope}%
\begin{pgfscope}%
\definecolor{textcolor}{rgb}{0.000000,0.000000,0.000000}%
\pgfsetstrokecolor{textcolor}%
\pgfsetfillcolor{textcolor}%
\pgftext[x=2.115278in,y=0.138889in,,top]{\color{textcolor}\rmfamily\fontsize{10.000000}{12.000000}\selectfont time (seconds)}%
\end{pgfscope}%
\begin{pgfscope}%
\pgfsetbuttcap%
\pgfsetroundjoin%
\definecolor{currentfill}{rgb}{0.000000,0.000000,0.000000}%
\pgfsetfillcolor{currentfill}%
\pgfsetlinewidth{0.803000pt}%
\definecolor{currentstroke}{rgb}{0.000000,0.000000,0.000000}%
\pgfsetstrokecolor{currentstroke}%
\pgfsetdash{}{0pt}%
\pgfsys@defobject{currentmarker}{\pgfqpoint{-0.048611in}{0.000000in}}{\pgfqpoint{-0.000000in}{0.000000in}}{%
\pgfpathmoveto{\pgfqpoint{-0.000000in}{0.000000in}}%
\pgfpathlineto{\pgfqpoint{-0.048611in}{0.000000in}}%
\pgfusepath{stroke,fill}%
}%
\begin{pgfscope}%
\pgfsys@transformshift{0.430556in}{0.749727in}%
\pgfsys@useobject{currentmarker}{}%
\end{pgfscope}%
\end{pgfscope}%
\begin{pgfscope}%
\definecolor{textcolor}{rgb}{0.000000,0.000000,0.000000}%
\pgfsetstrokecolor{textcolor}%
\pgfsetfillcolor{textcolor}%
\pgftext[x=0.194444in, y=0.701533in, left, base]{\color{textcolor}\rmfamily\fontsize{10.000000}{12.000000}\selectfont \(\displaystyle {20}\)}%
\end{pgfscope}%
\begin{pgfscope}%
\pgfsetbuttcap%
\pgfsetroundjoin%
\definecolor{currentfill}{rgb}{0.000000,0.000000,0.000000}%
\pgfsetfillcolor{currentfill}%
\pgfsetlinewidth{0.803000pt}%
\definecolor{currentstroke}{rgb}{0.000000,0.000000,0.000000}%
\pgfsetstrokecolor{currentstroke}%
\pgfsetdash{}{0pt}%
\pgfsys@defobject{currentmarker}{\pgfqpoint{-0.048611in}{0.000000in}}{\pgfqpoint{-0.000000in}{0.000000in}}{%
\pgfpathmoveto{\pgfqpoint{-0.000000in}{0.000000in}}%
\pgfpathlineto{\pgfqpoint{-0.048611in}{0.000000in}}%
\pgfusepath{stroke,fill}%
}%
\begin{pgfscope}%
\pgfsys@transformshift{0.430556in}{1.124213in}%
\pgfsys@useobject{currentmarker}{}%
\end{pgfscope}%
\end{pgfscope}%
\begin{pgfscope}%
\definecolor{textcolor}{rgb}{0.000000,0.000000,0.000000}%
\pgfsetstrokecolor{textcolor}%
\pgfsetfillcolor{textcolor}%
\pgftext[x=0.194444in, y=1.076018in, left, base]{\color{textcolor}\rmfamily\fontsize{10.000000}{12.000000}\selectfont \(\displaystyle {40}\)}%
\end{pgfscope}%
\begin{pgfscope}%
\pgfsetbuttcap%
\pgfsetroundjoin%
\definecolor{currentfill}{rgb}{0.000000,0.000000,0.000000}%
\pgfsetfillcolor{currentfill}%
\pgfsetlinewidth{0.803000pt}%
\definecolor{currentstroke}{rgb}{0.000000,0.000000,0.000000}%
\pgfsetstrokecolor{currentstroke}%
\pgfsetdash{}{0pt}%
\pgfsys@defobject{currentmarker}{\pgfqpoint{-0.048611in}{0.000000in}}{\pgfqpoint{-0.000000in}{0.000000in}}{%
\pgfpathmoveto{\pgfqpoint{-0.000000in}{0.000000in}}%
\pgfpathlineto{\pgfqpoint{-0.048611in}{0.000000in}}%
\pgfusepath{stroke,fill}%
}%
\begin{pgfscope}%
\pgfsys@transformshift{0.430556in}{1.498698in}%
\pgfsys@useobject{currentmarker}{}%
\end{pgfscope}%
\end{pgfscope}%
\begin{pgfscope}%
\definecolor{textcolor}{rgb}{0.000000,0.000000,0.000000}%
\pgfsetstrokecolor{textcolor}%
\pgfsetfillcolor{textcolor}%
\pgftext[x=0.194444in, y=1.450504in, left, base]{\color{textcolor}\rmfamily\fontsize{10.000000}{12.000000}\selectfont \(\displaystyle {60}\)}%
\end{pgfscope}%
\begin{pgfscope}%
\definecolor{textcolor}{rgb}{0.000000,0.000000,0.000000}%
\pgfsetstrokecolor{textcolor}%
\pgfsetfillcolor{textcolor}%
\pgftext[x=0.138889in,y=1.007500in,,bottom,rotate=90.000000]{\color{textcolor}\rmfamily\fontsize{10.000000}{12.000000}\selectfont vehicle dist (m)}%
\end{pgfscope}%
\begin{pgfscope}%
\pgfpathrectangle{\pgfqpoint{0.430556in}{0.415000in}}{\pgfqpoint{3.369444in}{1.185000in}}%
\pgfusepath{clip}%
\pgfsetbuttcap%
\pgfsetroundjoin%
\pgfsetlinewidth{1.505625pt}%
\definecolor{currentstroke}{rgb}{0.121569,0.466667,0.705882}%
\pgfsetstrokecolor{currentstroke}%
\pgfsetdash{{5.550000pt}{2.400000pt}}{0.000000pt}%
\pgfpathmoveto{\pgfqpoint{0.583712in}{1.546058in}}%
\pgfpathlineto{\pgfqpoint{0.666458in}{1.472871in}}%
\pgfpathlineto{\pgfqpoint{0.757632in}{1.389204in}}%
\pgfpathlineto{\pgfqpoint{0.784448in}{1.361907in}}%
\pgfpathlineto{\pgfqpoint{0.808199in}{1.334476in}}%
\pgfpathlineto{\pgfqpoint{0.831184in}{1.304502in}}%
\pgfpathlineto{\pgfqpoint{0.854169in}{1.270883in}}%
\pgfpathlineto{\pgfqpoint{0.877920in}{1.232144in}}%
\pgfpathlineto{\pgfqpoint{0.903203in}{1.186435in}}%
\pgfpathlineto{\pgfqpoint{0.955303in}{1.086224in}}%
\pgfpathlineto{\pgfqpoint{1.060267in}{0.884176in}}%
\pgfpathlineto{\pgfqpoint{1.110834in}{0.791789in}}%
\pgfpathlineto{\pgfqpoint{1.149143in}{0.726125in}}%
\pgfpathlineto{\pgfqpoint{1.179023in}{0.679003in}}%
\pgfpathlineto{\pgfqpoint{1.203540in}{0.644298in}}%
\pgfpathlineto{\pgfqpoint{1.223461in}{0.619734in}}%
\pgfpathlineto{\pgfqpoint{1.238018in}{0.605058in}}%
\pgfpathlineto{\pgfqpoint{1.254107in}{0.592269in}}%
\pgfpathlineto{\pgfqpoint{1.268665in}{0.583453in}}%
\pgfpathlineto{\pgfqpoint{1.282456in}{0.577557in}}%
\pgfpathlineto{\pgfqpoint{1.296247in}{0.574015in}}%
\pgfpathlineto{\pgfqpoint{1.309271in}{0.572858in}}%
\pgfpathlineto{\pgfqpoint{1.323062in}{0.573817in}}%
\pgfpathlineto{\pgfqpoint{1.338386in}{0.577174in}}%
\pgfpathlineto{\pgfqpoint{1.356007in}{0.583462in}}%
\pgfpathlineto{\pgfqpoint{1.377460in}{0.593686in}}%
\pgfpathlineto{\pgfqpoint{1.406574in}{0.610256in}}%
\pgfpathlineto{\pgfqpoint{1.497748in}{0.663718in}}%
\pgfpathlineto{\pgfqpoint{1.526862in}{0.677609in}}%
\pgfpathlineto{\pgfqpoint{1.535290in}{0.680829in}}%
\pgfpathlineto{\pgfqpoint{1.547549in}{0.670941in}}%
\pgfpathlineto{\pgfqpoint{1.559041in}{0.664487in}}%
\pgfpathlineto{\pgfqpoint{1.569768in}{0.660884in}}%
\pgfpathlineto{\pgfqpoint{1.575131in}{0.658843in}}%
\pgfpathlineto{\pgfqpoint{1.618036in}{0.633065in}}%
\pgfpathlineto{\pgfqpoint{1.651747in}{0.615363in}}%
\pgfpathlineto{\pgfqpoint{1.684693in}{0.600597in}}%
\pgfpathlineto{\pgfqpoint{1.705379in}{0.590945in}}%
\pgfpathlineto{\pgfqpoint{1.743687in}{0.572679in}}%
\pgfpathlineto{\pgfqpoint{1.779697in}{0.557961in}}%
\pgfpathlineto{\pgfqpoint{1.813408in}{0.546533in}}%
\pgfpathlineto{\pgfqpoint{1.844821in}{0.538176in}}%
\pgfpathlineto{\pgfqpoint{1.870105in}{0.533664in}}%
\pgfpathlineto{\pgfqpoint{1.894622in}{0.531560in}}%
\pgfpathlineto{\pgfqpoint{1.919139in}{0.531698in}}%
\pgfpathlineto{\pgfqpoint{1.944423in}{0.534071in}}%
\pgfpathlineto{\pgfqpoint{1.971239in}{0.538834in}}%
\pgfpathlineto{\pgfqpoint{2.001119in}{0.546490in}}%
\pgfpathlineto{\pgfqpoint{2.034830in}{0.555639in}}%
\pgfpathlineto{\pgfqpoint{2.057049in}{0.535770in}}%
\pgfpathlineto{\pgfqpoint{2.075437in}{0.522203in}}%
\pgfpathlineto{\pgfqpoint{2.089994in}{0.514006in}}%
\pgfpathlineto{\pgfqpoint{2.103019in}{0.509201in}}%
\pgfpathlineto{\pgfqpoint{2.116044in}{0.506920in}}%
\pgfpathlineto{\pgfqpoint{2.133666in}{0.506143in}}%
\pgfpathlineto{\pgfqpoint{2.151288in}{0.507527in}}%
\pgfpathlineto{\pgfqpoint{2.166611in}{0.510754in}}%
\pgfpathlineto{\pgfqpoint{2.181934in}{0.516416in}}%
\pgfpathlineto{\pgfqpoint{2.198790in}{0.525145in}}%
\pgfpathlineto{\pgfqpoint{2.214879in}{0.534109in}}%
\pgfpathlineto{\pgfqpoint{2.255486in}{0.517829in}}%
\pgfpathlineto{\pgfqpoint{2.289198in}{0.506528in}}%
\pgfpathlineto{\pgfqpoint{2.316779in}{0.499707in}}%
\pgfpathlineto{\pgfqpoint{2.338232in}{0.496512in}}%
\pgfpathlineto{\pgfqpoint{2.354322in}{0.496289in}}%
\pgfpathlineto{\pgfqpoint{2.440898in}{0.504968in}}%
\pgfpathlineto{\pgfqpoint{2.491465in}{0.512431in}}%
\pgfpathlineto{\pgfqpoint{2.545863in}{0.522799in}}%
\pgfpathlineto{\pgfqpoint{2.597962in}{0.534755in}}%
\pgfpathlineto{\pgfqpoint{2.620947in}{0.542858in}}%
\pgfpathlineto{\pgfqpoint{2.646231in}{0.554161in}}%
\pgfpathlineto{\pgfqpoint{2.675345in}{0.569719in}}%
\pgfpathlineto{\pgfqpoint{2.709056in}{0.590379in}}%
\pgfpathlineto{\pgfqpoint{2.748897in}{0.617498in}}%
\pgfpathlineto{\pgfqpoint{2.797932in}{0.653680in}}%
\pgfpathlineto{\pgfqpoint{2.859991in}{0.702387in}}%
\pgfpathlineto{\pgfqpoint{2.940438in}{0.768509in}}%
\pgfpathlineto{\pgfqpoint{3.048468in}{0.860340in}}%
\pgfpathlineto{\pgfqpoint{3.200169in}{0.992407in}}%
\pgfpathlineto{\pgfqpoint{3.422357in}{1.189007in}}%
\pgfpathlineto{\pgfqpoint{3.646843in}{1.389468in}}%
\pgfpathlineto{\pgfqpoint{3.646843in}{1.389468in}}%
\pgfusepath{stroke}%
\end{pgfscope}%
\begin{pgfscope}%
\pgfpathrectangle{\pgfqpoint{0.430556in}{0.415000in}}{\pgfqpoint{3.369444in}{1.185000in}}%
\pgfusepath{clip}%
\pgfsetrectcap%
\pgfsetroundjoin%
\pgfsetlinewidth{1.505625pt}%
\definecolor{currentstroke}{rgb}{1.000000,0.498039,0.054902}%
\pgfsetstrokecolor{currentstroke}%
\pgfsetdash{}{0pt}%
\pgfpathmoveto{\pgfqpoint{0.583712in}{1.545990in}}%
\pgfpathlineto{\pgfqpoint{1.373629in}{0.792461in}}%
\pgfpathlineto{\pgfqpoint{1.401211in}{0.771429in}}%
\pgfpathlineto{\pgfqpoint{1.444117in}{0.741744in}}%
\pgfpathlineto{\pgfqpoint{1.513838in}{0.696467in}}%
\pgfpathlineto{\pgfqpoint{1.605778in}{0.639403in}}%
\pgfpathlineto{\pgfqpoint{1.686225in}{0.591940in}}%
\pgfpathlineto{\pgfqpoint{1.746752in}{0.558675in}}%
\pgfpathlineto{\pgfqpoint{1.794254in}{0.535005in}}%
\pgfpathlineto{\pgfqpoint{1.832563in}{0.518289in}}%
\pgfpathlineto{\pgfqpoint{1.865508in}{0.506276in}}%
\pgfpathlineto{\pgfqpoint{1.894622in}{0.498003in}}%
\pgfpathlineto{\pgfqpoint{1.921438in}{0.492705in}}%
\pgfpathlineto{\pgfqpoint{1.947488in}{0.489878in}}%
\pgfpathlineto{\pgfqpoint{1.977368in}{0.489065in}}%
\pgfpathlineto{\pgfqpoint{2.008015in}{0.490113in}}%
\pgfpathlineto{\pgfqpoint{2.039427in}{0.493457in}}%
\pgfpathlineto{\pgfqpoint{2.071606in}{0.499163in}}%
\pgfpathlineto{\pgfqpoint{2.105318in}{0.507411in}}%
\pgfpathlineto{\pgfqpoint{2.141327in}{0.518513in}}%
\pgfpathlineto{\pgfqpoint{2.180402in}{0.532937in}}%
\pgfpathlineto{\pgfqpoint{2.243228in}{0.558773in}}%
\pgfpathlineto{\pgfqpoint{2.324441in}{0.595150in}}%
\pgfpathlineto{\pgfqpoint{2.512152in}{0.682119in}}%
\pgfpathlineto{\pgfqpoint{2.641634in}{0.745491in}}%
\pgfpathlineto{\pgfqpoint{2.757325in}{0.805061in}}%
\pgfpathlineto{\pgfqpoint{2.903663in}{0.883023in}}%
\pgfpathlineto{\pgfqpoint{3.094438in}{0.987307in}}%
\pgfpathlineto{\pgfqpoint{3.350337in}{1.129870in}}%
\pgfpathlineto{\pgfqpoint{3.646843in}{1.297168in}}%
\pgfpathlineto{\pgfqpoint{3.646843in}{1.297168in}}%
\pgfusepath{stroke}%
\end{pgfscope}%
\begin{pgfscope}%
\pgfpathrectangle{\pgfqpoint{0.430556in}{0.415000in}}{\pgfqpoint{3.369444in}{1.185000in}}%
\pgfusepath{clip}%
\pgfsetbuttcap%
\pgfsetroundjoin%
\pgfsetlinewidth{1.505625pt}%
\definecolor{currentstroke}{rgb}{0.172549,0.627451,0.172549}%
\pgfsetstrokecolor{currentstroke}%
\pgfsetdash{{1.500000pt}{2.475000pt}}{0.000000pt}%
\pgfpathmoveto{\pgfqpoint{0.583712in}{1.546136in}}%
\pgfpathlineto{\pgfqpoint{1.009700in}{1.221816in}}%
\pgfpathlineto{\pgfqpoint{1.376694in}{0.944933in}}%
\pgfpathlineto{\pgfqpoint{1.496982in}{0.857026in}}%
\pgfpathlineto{\pgfqpoint{1.614972in}{0.773540in}}%
\pgfpathlineto{\pgfqpoint{1.721469in}{0.700807in}}%
\pgfpathlineto{\pgfqpoint{1.808811in}{0.643719in}}%
\pgfpathlineto{\pgfqpoint{1.878533in}{0.600702in}}%
\pgfpathlineto{\pgfqpoint{1.934463in}{0.568724in}}%
\pgfpathlineto{\pgfqpoint{1.980433in}{0.544940in}}%
\pgfpathlineto{\pgfqpoint{2.018741in}{0.527543in}}%
\pgfpathlineto{\pgfqpoint{2.051686in}{0.514916in}}%
\pgfpathlineto{\pgfqpoint{2.081567in}{0.505801in}}%
\pgfpathlineto{\pgfqpoint{2.109149in}{0.499721in}}%
\pgfpathlineto{\pgfqpoint{2.135198in}{0.496290in}}%
\pgfpathlineto{\pgfqpoint{2.159715in}{0.495301in}}%
\pgfpathlineto{\pgfqpoint{2.184233in}{0.496542in}}%
\pgfpathlineto{\pgfqpoint{2.209516in}{0.500089in}}%
\pgfpathlineto{\pgfqpoint{2.236332in}{0.506152in}}%
\pgfpathlineto{\pgfqpoint{2.265446in}{0.515050in}}%
\pgfpathlineto{\pgfqpoint{2.298391in}{0.527483in}}%
\pgfpathlineto{\pgfqpoint{2.337466in}{0.544701in}}%
\pgfpathlineto{\pgfqpoint{2.384968in}{0.568184in}}%
\pgfpathlineto{\pgfqpoint{2.446262in}{0.601126in}}%
\pgfpathlineto{\pgfqpoint{2.516749in}{0.641458in}}%
\pgfpathlineto{\pgfqpoint{2.573445in}{0.676430in}}%
\pgfpathlineto{\pgfqpoint{2.629375in}{0.713596in}}%
\pgfpathlineto{\pgfqpoint{2.692201in}{0.758084in}}%
\pgfpathlineto{\pgfqpoint{2.777245in}{0.821281in}}%
\pgfpathlineto{\pgfqpoint{3.146537in}{1.098938in}}%
\pgfpathlineto{\pgfqpoint{3.239243in}{1.164241in}}%
\pgfpathlineto{\pgfqpoint{3.325054in}{1.221999in}}%
\pgfpathlineto{\pgfqpoint{3.405501in}{1.273494in}}%
\pgfpathlineto{\pgfqpoint{3.482118in}{1.319913in}}%
\pgfpathlineto{\pgfqpoint{3.556436in}{1.362340in}}%
\pgfpathlineto{\pgfqpoint{3.615431in}{1.396485in}}%
\pgfpathlineto{\pgfqpoint{3.646843in}{1.415958in}}%
\pgfpathlineto{\pgfqpoint{3.646843in}{1.415958in}}%
\pgfusepath{stroke}%
\end{pgfscope}%
\begin{pgfscope}%
\pgfpathrectangle{\pgfqpoint{0.430556in}{0.415000in}}{\pgfqpoint{3.369444in}{1.185000in}}%
\pgfusepath{clip}%
\pgfsetrectcap%
\pgfsetroundjoin%
\pgfsetlinewidth{1.505625pt}%
\definecolor{currentstroke}{rgb}{0.000000,0.000000,0.000000}%
\pgfsetstrokecolor{currentstroke}%
\pgfsetdash{}{0pt}%
\pgfpathmoveto{\pgfqpoint{0.583712in}{0.468863in}}%
\pgfpathlineto{\pgfqpoint{3.646843in}{0.468863in}}%
\pgfpathlineto{\pgfqpoint{3.646843in}{0.468863in}}%
\pgfusepath{stroke}%
\end{pgfscope}%
\begin{pgfscope}%
\pgfsetrectcap%
\pgfsetmiterjoin%
\pgfsetlinewidth{0.803000pt}%
\definecolor{currentstroke}{rgb}{0.000000,0.000000,0.000000}%
\pgfsetstrokecolor{currentstroke}%
\pgfsetdash{}{0pt}%
\pgfpathmoveto{\pgfqpoint{0.430556in}{0.415000in}}%
\pgfpathlineto{\pgfqpoint{0.430556in}{1.600000in}}%
\pgfusepath{stroke}%
\end{pgfscope}%
\begin{pgfscope}%
\pgfsetrectcap%
\pgfsetmiterjoin%
\pgfsetlinewidth{0.803000pt}%
\definecolor{currentstroke}{rgb}{0.000000,0.000000,0.000000}%
\pgfsetstrokecolor{currentstroke}%
\pgfsetdash{}{0pt}%
\pgfpathmoveto{\pgfqpoint{3.800000in}{0.415000in}}%
\pgfpathlineto{\pgfqpoint{3.800000in}{1.600000in}}%
\pgfusepath{stroke}%
\end{pgfscope}%
\begin{pgfscope}%
\pgfsetrectcap%
\pgfsetmiterjoin%
\pgfsetlinewidth{0.803000pt}%
\definecolor{currentstroke}{rgb}{0.000000,0.000000,0.000000}%
\pgfsetstrokecolor{currentstroke}%
\pgfsetdash{}{0pt}%
\pgfpathmoveto{\pgfqpoint{0.430556in}{0.415000in}}%
\pgfpathlineto{\pgfqpoint{3.800000in}{0.415000in}}%
\pgfusepath{stroke}%
\end{pgfscope}%
\begin{pgfscope}%
\pgfsetrectcap%
\pgfsetmiterjoin%
\pgfsetlinewidth{0.803000pt}%
\definecolor{currentstroke}{rgb}{0.000000,0.000000,0.000000}%
\pgfsetstrokecolor{currentstroke}%
\pgfsetdash{}{0pt}%
\pgfpathmoveto{\pgfqpoint{0.430556in}{1.600000in}}%
\pgfpathlineto{\pgfqpoint{3.800000in}{1.600000in}}%
\pgfusepath{stroke}%
\end{pgfscope}%
\end{pgfpicture}%
\makeatother%
\endgroup%

%% file: imgs/sim/20_veh1_path.pgf
\begingroup%
\makeatletter%
\begin{pgfpicture}%
\pgfpathrectangle{\pgfpointorigin}{\pgfqpoint{3.800000in}{1.600000in}}%
\pgfusepath{use as bounding box, clip}%
\begin{pgfscope}%
\pgfsetbuttcap%
\pgfsetmiterjoin%
\definecolor{currentfill}{rgb}{1.000000,1.000000,1.000000}%
\pgfsetfillcolor{currentfill}%
\pgfsetlinewidth{0.000000pt}%
\definecolor{currentstroke}{rgb}{1.000000,1.000000,1.000000}%
\pgfsetstrokecolor{currentstroke}%
\pgfsetdash{}{0pt}%
\pgfpathmoveto{\pgfqpoint{0.000000in}{0.000000in}}%
\pgfpathlineto{\pgfqpoint{3.800000in}{0.000000in}}%
\pgfpathlineto{\pgfqpoint{3.800000in}{1.600000in}}%
\pgfpathlineto{\pgfqpoint{0.000000in}{1.600000in}}%
\pgfpathclose%
\pgfusepath{fill}%
\end{pgfscope}%
\begin{pgfscope}%
\pgfsetbuttcap%
\pgfsetmiterjoin%
\definecolor{currentfill}{rgb}{1.000000,1.000000,1.000000}%
\pgfsetfillcolor{currentfill}%
\pgfsetlinewidth{0.000000pt}%
\definecolor{currentstroke}{rgb}{0.000000,0.000000,0.000000}%
\pgfsetstrokecolor{currentstroke}%
\pgfsetstrokeopacity{0.000000}%
\pgfsetdash{}{0pt}%
\pgfpathmoveto{\pgfqpoint{0.538581in}{0.415000in}}%
\pgfpathlineto{\pgfqpoint{3.699063in}{0.415000in}}%
\pgfpathlineto{\pgfqpoint{3.699063in}{1.600000in}}%
\pgfpathlineto{\pgfqpoint{0.538581in}{1.600000in}}%
\pgfpathclose%
\pgfusepath{fill}%
\end{pgfscope}%
\begin{pgfscope}%
\pgfsetbuttcap%
\pgfsetroundjoin%
\definecolor{currentfill}{rgb}{0.000000,0.000000,0.000000}%
\pgfsetfillcolor{currentfill}%
\pgfsetlinewidth{0.803000pt}%
\definecolor{currentstroke}{rgb}{0.000000,0.000000,0.000000}%
\pgfsetstrokecolor{currentstroke}%
\pgfsetdash{}{0pt}%
\pgfsys@defobject{currentmarker}{\pgfqpoint{0.000000in}{-0.048611in}}{\pgfqpoint{0.000000in}{0.000000in}}{%
\pgfpathmoveto{\pgfqpoint{0.000000in}{0.000000in}}%
\pgfpathlineto{\pgfqpoint{0.000000in}{-0.048611in}}%
\pgfusepath{stroke,fill}%
}%
\begin{pgfscope}%
\pgfsys@transformshift{0.682239in}{0.415000in}%
\pgfsys@useobject{currentmarker}{}%
\end{pgfscope}%
\end{pgfscope}%
\begin{pgfscope}%
\definecolor{textcolor}{rgb}{0.000000,0.000000,0.000000}%
\pgfsetstrokecolor{textcolor}%
\pgfsetfillcolor{textcolor}%
\pgftext[x=0.682239in,y=0.317777in,,top]{\color{textcolor}\rmfamily\fontsize{10.000000}{12.000000}\selectfont \(\displaystyle {-200}\)}%
\end{pgfscope}%
\begin{pgfscope}%
\pgfsetbuttcap%
\pgfsetroundjoin%
\definecolor{currentfill}{rgb}{0.000000,0.000000,0.000000}%
\pgfsetfillcolor{currentfill}%
\pgfsetlinewidth{0.803000pt}%
\definecolor{currentstroke}{rgb}{0.000000,0.000000,0.000000}%
\pgfsetstrokecolor{currentstroke}%
\pgfsetdash{}{0pt}%
\pgfsys@defobject{currentmarker}{\pgfqpoint{0.000000in}{-0.048611in}}{\pgfqpoint{0.000000in}{0.000000in}}{%
\pgfpathmoveto{\pgfqpoint{0.000000in}{0.000000in}}%
\pgfpathlineto{\pgfqpoint{0.000000in}{-0.048611in}}%
\pgfusepath{stroke,fill}%
}%
\begin{pgfscope}%
\pgfsys@transformshift{1.435637in}{0.415000in}%
\pgfsys@useobject{currentmarker}{}%
\end{pgfscope}%
\end{pgfscope}%
\begin{pgfscope}%
\definecolor{textcolor}{rgb}{0.000000,0.000000,0.000000}%
\pgfsetstrokecolor{textcolor}%
\pgfsetfillcolor{textcolor}%
\pgftext[x=1.435637in,y=0.317777in,,top]{\color{textcolor}\rmfamily\fontsize{10.000000}{12.000000}\selectfont \(\displaystyle {-100}\)}%
\end{pgfscope}%
\begin{pgfscope}%
\pgfsetbuttcap%
\pgfsetroundjoin%
\definecolor{currentfill}{rgb}{0.000000,0.000000,0.000000}%
\pgfsetfillcolor{currentfill}%
\pgfsetlinewidth{0.803000pt}%
\definecolor{currentstroke}{rgb}{0.000000,0.000000,0.000000}%
\pgfsetstrokecolor{currentstroke}%
\pgfsetdash{}{0pt}%
\pgfsys@defobject{currentmarker}{\pgfqpoint{0.000000in}{-0.048611in}}{\pgfqpoint{0.000000in}{0.000000in}}{%
\pgfpathmoveto{\pgfqpoint{0.000000in}{0.000000in}}%
\pgfpathlineto{\pgfqpoint{0.000000in}{-0.048611in}}%
\pgfusepath{stroke,fill}%
}%
\begin{pgfscope}%
\pgfsys@transformshift{2.189036in}{0.415000in}%
\pgfsys@useobject{currentmarker}{}%
\end{pgfscope}%
\end{pgfscope}%
\begin{pgfscope}%
\definecolor{textcolor}{rgb}{0.000000,0.000000,0.000000}%
\pgfsetstrokecolor{textcolor}%
\pgfsetfillcolor{textcolor}%
\pgftext[x=2.189036in,y=0.317777in,,top]{\color{textcolor}\rmfamily\fontsize{10.000000}{12.000000}\selectfont \(\displaystyle {0}\)}%
\end{pgfscope}%
\begin{pgfscope}%
\pgfsetbuttcap%
\pgfsetroundjoin%
\definecolor{currentfill}{rgb}{0.000000,0.000000,0.000000}%
\pgfsetfillcolor{currentfill}%
\pgfsetlinewidth{0.803000pt}%
\definecolor{currentstroke}{rgb}{0.000000,0.000000,0.000000}%
\pgfsetstrokecolor{currentstroke}%
\pgfsetdash{}{0pt}%
\pgfsys@defobject{currentmarker}{\pgfqpoint{0.000000in}{-0.048611in}}{\pgfqpoint{0.000000in}{0.000000in}}{%
\pgfpathmoveto{\pgfqpoint{0.000000in}{0.000000in}}%
\pgfpathlineto{\pgfqpoint{0.000000in}{-0.048611in}}%
\pgfusepath{stroke,fill}%
}%
\begin{pgfscope}%
\pgfsys@transformshift{2.942434in}{0.415000in}%
\pgfsys@useobject{currentmarker}{}%
\end{pgfscope}%
\end{pgfscope}%
\begin{pgfscope}%
\definecolor{textcolor}{rgb}{0.000000,0.000000,0.000000}%
\pgfsetstrokecolor{textcolor}%
\pgfsetfillcolor{textcolor}%
\pgftext[x=2.942434in,y=0.317777in,,top]{\color{textcolor}\rmfamily\fontsize{10.000000}{12.000000}\selectfont \(\displaystyle {100}\)}%
\end{pgfscope}%
\begin{pgfscope}%
\pgfsetbuttcap%
\pgfsetroundjoin%
\definecolor{currentfill}{rgb}{0.000000,0.000000,0.000000}%
\pgfsetfillcolor{currentfill}%
\pgfsetlinewidth{0.803000pt}%
\definecolor{currentstroke}{rgb}{0.000000,0.000000,0.000000}%
\pgfsetstrokecolor{currentstroke}%
\pgfsetdash{}{0pt}%
\pgfsys@defobject{currentmarker}{\pgfqpoint{0.000000in}{-0.048611in}}{\pgfqpoint{0.000000in}{0.000000in}}{%
\pgfpathmoveto{\pgfqpoint{0.000000in}{0.000000in}}%
\pgfpathlineto{\pgfqpoint{0.000000in}{-0.048611in}}%
\pgfusepath{stroke,fill}%
}%
\begin{pgfscope}%
\pgfsys@transformshift{3.695833in}{0.415000in}%
\pgfsys@useobject{currentmarker}{}%
\end{pgfscope}%
\end{pgfscope}%
\begin{pgfscope}%
\definecolor{textcolor}{rgb}{0.000000,0.000000,0.000000}%
\pgfsetstrokecolor{textcolor}%
\pgfsetfillcolor{textcolor}%
\pgftext[x=3.695833in,y=0.317777in,,top]{\color{textcolor}\rmfamily\fontsize{10.000000}{12.000000}\selectfont \(\displaystyle {200}\)}%
\end{pgfscope}%
\begin{pgfscope}%
\definecolor{textcolor}{rgb}{0.000000,0.000000,0.000000}%
\pgfsetstrokecolor{textcolor}%
\pgfsetfillcolor{textcolor}%
\pgftext[x=2.118822in,y=0.138889in,,top]{\color{textcolor}\rmfamily\fontsize{10.000000}{12.000000}\selectfont \(\displaystyle x\) (meters)}%
\end{pgfscope}%
\begin{pgfscope}%
\pgfsetbuttcap%
\pgfsetroundjoin%
\definecolor{currentfill}{rgb}{0.000000,0.000000,0.000000}%
\pgfsetfillcolor{currentfill}%
\pgfsetlinewidth{0.803000pt}%
\definecolor{currentstroke}{rgb}{0.000000,0.000000,0.000000}%
\pgfsetstrokecolor{currentstroke}%
\pgfsetdash{}{0pt}%
\pgfsys@defobject{currentmarker}{\pgfqpoint{-0.048611in}{0.000000in}}{\pgfqpoint{-0.000000in}{0.000000in}}{%
\pgfpathmoveto{\pgfqpoint{-0.000000in}{0.000000in}}%
\pgfpathlineto{\pgfqpoint{-0.048611in}{0.000000in}}%
\pgfusepath{stroke,fill}%
}%
\begin{pgfscope}%
\pgfsys@transformshift{0.538581in}{0.592160in}%
\pgfsys@useobject{currentmarker}{}%
\end{pgfscope}%
\end{pgfscope}%
\begin{pgfscope}%
\definecolor{textcolor}{rgb}{0.000000,0.000000,0.000000}%
\pgfsetstrokecolor{textcolor}%
\pgfsetfillcolor{textcolor}%
\pgftext[x=0.194444in, y=0.543965in, left, base]{\color{textcolor}\rmfamily\fontsize{10.000000}{12.000000}\selectfont \(\displaystyle {-20}\)}%
\end{pgfscope}%
\begin{pgfscope}%
\pgfsetbuttcap%
\pgfsetroundjoin%
\definecolor{currentfill}{rgb}{0.000000,0.000000,0.000000}%
\pgfsetfillcolor{currentfill}%
\pgfsetlinewidth{0.803000pt}%
\definecolor{currentstroke}{rgb}{0.000000,0.000000,0.000000}%
\pgfsetstrokecolor{currentstroke}%
\pgfsetdash{}{0pt}%
\pgfsys@defobject{currentmarker}{\pgfqpoint{-0.048611in}{0.000000in}}{\pgfqpoint{-0.000000in}{0.000000in}}{%
\pgfpathmoveto{\pgfqpoint{-0.000000in}{0.000000in}}%
\pgfpathlineto{\pgfqpoint{-0.048611in}{0.000000in}}%
\pgfusepath{stroke,fill}%
}%
\begin{pgfscope}%
\pgfsys@transformshift{0.538581in}{1.002116in}%
\pgfsys@useobject{currentmarker}{}%
\end{pgfscope}%
\end{pgfscope}%
\begin{pgfscope}%
\definecolor{textcolor}{rgb}{0.000000,0.000000,0.000000}%
\pgfsetstrokecolor{textcolor}%
\pgfsetfillcolor{textcolor}%
\pgftext[x=0.371914in, y=0.953921in, left, base]{\color{textcolor}\rmfamily\fontsize{10.000000}{12.000000}\selectfont \(\displaystyle {0}\)}%
\end{pgfscope}%
\begin{pgfscope}%
\pgfsetbuttcap%
\pgfsetroundjoin%
\definecolor{currentfill}{rgb}{0.000000,0.000000,0.000000}%
\pgfsetfillcolor{currentfill}%
\pgfsetlinewidth{0.803000pt}%
\definecolor{currentstroke}{rgb}{0.000000,0.000000,0.000000}%
\pgfsetstrokecolor{currentstroke}%
\pgfsetdash{}{0pt}%
\pgfsys@defobject{currentmarker}{\pgfqpoint{-0.048611in}{0.000000in}}{\pgfqpoint{-0.000000in}{0.000000in}}{%
\pgfpathmoveto{\pgfqpoint{-0.000000in}{0.000000in}}%
\pgfpathlineto{\pgfqpoint{-0.048611in}{0.000000in}}%
\pgfusepath{stroke,fill}%
}%
\begin{pgfscope}%
\pgfsys@transformshift{0.538581in}{1.412071in}%
\pgfsys@useobject{currentmarker}{}%
\end{pgfscope}%
\end{pgfscope}%
\begin{pgfscope}%
\definecolor{textcolor}{rgb}{0.000000,0.000000,0.000000}%
\pgfsetstrokecolor{textcolor}%
\pgfsetfillcolor{textcolor}%
\pgftext[x=0.302469in, y=1.363877in, left, base]{\color{textcolor}\rmfamily\fontsize{10.000000}{12.000000}\selectfont \(\displaystyle {20}\)}%
\end{pgfscope}%
\begin{pgfscope}%
\definecolor{textcolor}{rgb}{0.000000,0.000000,0.000000}%
\pgfsetstrokecolor{textcolor}%
\pgfsetfillcolor{textcolor}%
\pgftext[x=0.138889in,y=1.007500in,,bottom,rotate=90.000000]{\color{textcolor}\rmfamily\fontsize{10.000000}{12.000000}\selectfont \(\displaystyle y\) (meters)}%
\end{pgfscope}%
\begin{pgfscope}%
\pgfpathrectangle{\pgfqpoint{0.538581in}{0.415000in}}{\pgfqpoint{3.160483in}{1.185000in}}%
\pgfusepath{clip}%
\pgfsetbuttcap%
\pgfsetroundjoin%
\pgfsetlinewidth{1.505625pt}%
\definecolor{currentstroke}{rgb}{0.121569,0.466667,0.705882}%
\pgfsetstrokecolor{currentstroke}%
\pgfsetdash{{5.550000pt}{2.400000pt}}{0.000000pt}%
\pgfpathmoveto{\pgfqpoint{0.682239in}{1.002116in}}%
\pgfpathlineto{\pgfqpoint{0.710644in}{1.036407in}}%
\pgfpathlineto{\pgfqpoint{0.739501in}{1.067781in}}%
\pgfpathlineto{\pgfqpoint{0.767877in}{1.095378in}}%
\pgfpathlineto{\pgfqpoint{0.795700in}{1.119424in}}%
\pgfpathlineto{\pgfqpoint{0.823816in}{1.140792in}}%
\pgfpathlineto{\pgfqpoint{0.851272in}{1.158899in}}%
\pgfpathlineto{\pgfqpoint{0.878939in}{1.174453in}}%
\pgfpathlineto{\pgfqpoint{0.905853in}{1.187045in}}%
\pgfpathlineto{\pgfqpoint{0.932905in}{1.197222in}}%
\pgfpathlineto{\pgfqpoint{0.959128in}{1.204746in}}%
\pgfpathlineto{\pgfqpoint{0.985425in}{1.210000in}}%
\pgfpathlineto{\pgfqpoint{1.011770in}{1.212979in}}%
\pgfpathlineto{\pgfqpoint{1.038136in}{1.213680in}}%
\pgfpathlineto{\pgfqpoint{1.066382in}{1.212070in}}%
\pgfpathlineto{\pgfqpoint{2.199824in}{1.121440in}}%
\pgfpathlineto{\pgfqpoint{2.233714in}{1.117007in}}%
\pgfpathlineto{\pgfqpoint{2.265960in}{1.110495in}}%
\pgfpathlineto{\pgfqpoint{2.299357in}{1.101442in}}%
\pgfpathlineto{\pgfqpoint{2.341139in}{1.087602in}}%
\pgfpathlineto{\pgfqpoint{2.401846in}{1.064917in}}%
\pgfpathlineto{\pgfqpoint{2.501960in}{1.027343in}}%
\pgfpathlineto{\pgfqpoint{2.533122in}{1.018439in}}%
\pgfpathlineto{\pgfqpoint{2.558112in}{1.013514in}}%
\pgfpathlineto{\pgfqpoint{2.579006in}{1.011629in}}%
\pgfpathlineto{\pgfqpoint{2.628164in}{1.011015in}}%
\pgfpathlineto{\pgfqpoint{3.555405in}{1.003286in}}%
\pgfpathlineto{\pgfqpoint{3.555405in}{1.003286in}}%
\pgfusepath{stroke}%
\end{pgfscope}%
\begin{pgfscope}%
\pgfpathrectangle{\pgfqpoint{0.538581in}{0.415000in}}{\pgfqpoint{3.160483in}{1.185000in}}%
\pgfusepath{clip}%
\pgfsetrectcap%
\pgfsetroundjoin%
\pgfsetlinewidth{1.505625pt}%
\definecolor{currentstroke}{rgb}{1.000000,0.498039,0.054902}%
\pgfsetstrokecolor{currentstroke}%
\pgfsetdash{}{0pt}%
\pgfpathmoveto{\pgfqpoint{0.682239in}{1.002116in}}%
\pgfpathlineto{\pgfqpoint{1.716233in}{1.003171in}}%
\pgfpathlineto{\pgfqpoint{1.744149in}{1.006812in}}%
\pgfpathlineto{\pgfqpoint{1.770746in}{1.012549in}}%
\pgfpathlineto{\pgfqpoint{1.796267in}{1.020332in}}%
\pgfpathlineto{\pgfqpoint{1.821438in}{1.030333in}}%
\pgfpathlineto{\pgfqpoint{1.846301in}{1.042586in}}%
\pgfpathlineto{\pgfqpoint{1.871467in}{1.057492in}}%
\pgfpathlineto{\pgfqpoint{1.896323in}{1.074783in}}%
\pgfpathlineto{\pgfqpoint{1.921441in}{1.094945in}}%
\pgfpathlineto{\pgfqpoint{1.946781in}{1.118126in}}%
\pgfpathlineto{\pgfqpoint{1.972858in}{1.145069in}}%
\pgfpathlineto{\pgfqpoint{1.999041in}{1.175385in}}%
\pgfpathlineto{\pgfqpoint{2.025819in}{1.209874in}}%
\pgfpathlineto{\pgfqpoint{2.053116in}{1.248702in}}%
\pgfpathlineto{\pgfqpoint{2.082970in}{1.295179in}}%
\pgfpathlineto{\pgfqpoint{2.131105in}{1.371005in}}%
\pgfpathlineto{\pgfqpoint{2.154728in}{1.402705in}}%
\pgfpathlineto{\pgfqpoint{2.178040in}{1.430314in}}%
\pgfpathlineto{\pgfqpoint{2.201544in}{1.454722in}}%
\pgfpathlineto{\pgfqpoint{2.224601in}{1.475537in}}%
\pgfpathlineto{\pgfqpoint{2.247782in}{1.493523in}}%
\pgfpathlineto{\pgfqpoint{2.271100in}{1.508806in}}%
\pgfpathlineto{\pgfqpoint{2.294256in}{1.521329in}}%
\pgfpathlineto{\pgfqpoint{2.316806in}{1.531047in}}%
\pgfpathlineto{\pgfqpoint{2.339341in}{1.538347in}}%
\pgfpathlineto{\pgfqpoint{2.361138in}{1.543133in}}%
\pgfpathlineto{\pgfqpoint{2.382800in}{1.545673in}}%
\pgfpathlineto{\pgfqpoint{2.404270in}{1.546003in}}%
\pgfpathlineto{\pgfqpoint{2.426128in}{1.544079in}}%
\pgfpathlineto{\pgfqpoint{2.447662in}{1.539917in}}%
\pgfpathlineto{\pgfqpoint{2.468815in}{1.533591in}}%
\pgfpathlineto{\pgfqpoint{2.492553in}{1.524040in}}%
\pgfpathlineto{\pgfqpoint{2.777611in}{1.400420in}}%
\pgfpathlineto{\pgfqpoint{3.376904in}{1.140460in}}%
\pgfpathlineto{\pgfqpoint{3.376904in}{1.140460in}}%
\pgfusepath{stroke}%
\end{pgfscope}%
\begin{pgfscope}%
\pgfpathrectangle{\pgfqpoint{0.538581in}{0.415000in}}{\pgfqpoint{3.160483in}{1.185000in}}%
\pgfusepath{clip}%
\pgfsetbuttcap%
\pgfsetroundjoin%
\pgfsetlinewidth{1.505625pt}%
\definecolor{currentstroke}{rgb}{0.172549,0.627451,0.172549}%
\pgfsetstrokecolor{currentstroke}%
\pgfsetdash{{1.500000pt}{2.475000pt}}{0.000000pt}%
\pgfpathmoveto{\pgfqpoint{0.682239in}{1.002116in}}%
\pgfpathlineto{\pgfqpoint{0.719152in}{1.003237in}}%
\pgfpathlineto{\pgfqpoint{0.758308in}{1.006664in}}%
\pgfpathlineto{\pgfqpoint{0.801947in}{1.012761in}}%
\pgfpathlineto{\pgfqpoint{0.852307in}{1.022118in}}%
\pgfpathlineto{\pgfqpoint{0.916131in}{1.036366in}}%
\pgfpathlineto{\pgfqpoint{0.976130in}{1.051953in}}%
\pgfpathlineto{\pgfqpoint{1.332222in}{1.148682in}}%
\pgfpathlineto{\pgfqpoint{1.401147in}{1.168570in}}%
\pgfpathlineto{\pgfqpoint{1.443763in}{1.182962in}}%
\pgfpathlineto{\pgfqpoint{1.478036in}{1.196930in}}%
\pgfpathlineto{\pgfqpoint{1.517380in}{1.215471in}}%
\pgfpathlineto{\pgfqpoint{1.565462in}{1.240729in}}%
\pgfpathlineto{\pgfqpoint{1.638494in}{1.281940in}}%
\pgfpathlineto{\pgfqpoint{1.743271in}{1.340728in}}%
\pgfpathlineto{\pgfqpoint{1.800272in}{1.370046in}}%
\pgfpathlineto{\pgfqpoint{1.848546in}{1.392442in}}%
\pgfpathlineto{\pgfqpoint{1.891756in}{1.410119in}}%
\pgfpathlineto{\pgfqpoint{1.931359in}{1.423995in}}%
\pgfpathlineto{\pgfqpoint{1.968069in}{1.434565in}}%
\pgfpathlineto{\pgfqpoint{2.002611in}{1.442207in}}%
\pgfpathlineto{\pgfqpoint{2.034204in}{1.446977in}}%
\pgfpathlineto{\pgfqpoint{2.064327in}{1.449287in}}%
\pgfpathlineto{\pgfqpoint{2.092202in}{1.449231in}}%
\pgfpathlineto{\pgfqpoint{2.118557in}{1.446994in}}%
\pgfpathlineto{\pgfqpoint{2.143368in}{1.442714in}}%
\pgfpathlineto{\pgfqpoint{2.167361in}{1.436337in}}%
\pgfpathlineto{\pgfqpoint{2.191255in}{1.427614in}}%
\pgfpathlineto{\pgfqpoint{2.215016in}{1.416538in}}%
\pgfpathlineto{\pgfqpoint{2.239348in}{1.402666in}}%
\pgfpathlineto{\pgfqpoint{2.263472in}{1.386326in}}%
\pgfpathlineto{\pgfqpoint{2.288075in}{1.366934in}}%
\pgfpathlineto{\pgfqpoint{2.313094in}{1.344290in}}%
\pgfpathlineto{\pgfqpoint{2.337763in}{1.318962in}}%
\pgfpathlineto{\pgfqpoint{2.362731in}{1.290153in}}%
\pgfpathlineto{\pgfqpoint{2.387920in}{1.257678in}}%
\pgfpathlineto{\pgfqpoint{2.413244in}{1.221363in}}%
\pgfpathlineto{\pgfqpoint{2.438612in}{1.181040in}}%
\pgfpathlineto{\pgfqpoint{2.463924in}{1.136554in}}%
\pgfpathlineto{\pgfqpoint{2.489072in}{1.087766in}}%
\pgfpathlineto{\pgfqpoint{2.517579in}{1.027237in}}%
\pgfpathlineto{\pgfqpoint{2.611818in}{0.822815in}}%
\pgfpathlineto{\pgfqpoint{2.639638in}{0.770110in}}%
\pgfpathlineto{\pgfqpoint{2.665765in}{0.725294in}}%
\pgfpathlineto{\pgfqpoint{2.691966in}{0.684800in}}%
\pgfpathlineto{\pgfqpoint{2.718137in}{0.648472in}}%
\pgfpathlineto{\pgfqpoint{2.744180in}{0.616138in}}%
\pgfpathlineto{\pgfqpoint{2.770008in}{0.587617in}}%
\pgfpathlineto{\pgfqpoint{2.795538in}{0.562721in}}%
\pgfpathlineto{\pgfqpoint{2.820697in}{0.541255in}}%
\pgfpathlineto{\pgfqpoint{2.845418in}{0.523020in}}%
\pgfpathlineto{\pgfqpoint{2.869643in}{0.507815in}}%
\pgfpathlineto{\pgfqpoint{2.894059in}{0.495087in}}%
\pgfpathlineto{\pgfqpoint{2.917887in}{0.485128in}}%
\pgfpathlineto{\pgfqpoint{2.941832in}{0.477529in}}%
\pgfpathlineto{\pgfqpoint{2.965111in}{0.472426in}}%
\pgfpathlineto{\pgfqpoint{2.988441in}{0.469553in}}%
\pgfpathlineto{\pgfqpoint{3.011794in}{0.468914in}}%
\pgfpathlineto{\pgfqpoint{3.035141in}{0.470510in}}%
\pgfpathlineto{\pgfqpoint{3.058453in}{0.474338in}}%
\pgfpathlineto{\pgfqpoint{3.081701in}{0.480393in}}%
\pgfpathlineto{\pgfqpoint{3.105601in}{0.488973in}}%
\pgfpathlineto{\pgfqpoint{3.129371in}{0.499907in}}%
\pgfpathlineto{\pgfqpoint{3.152980in}{0.513181in}}%
\pgfpathlineto{\pgfqpoint{3.177124in}{0.529302in}}%
\pgfpathlineto{\pgfqpoint{3.201750in}{0.548472in}}%
\pgfpathlineto{\pgfqpoint{3.226797in}{0.570888in}}%
\pgfpathlineto{\pgfqpoint{3.251497in}{0.595992in}}%
\pgfpathlineto{\pgfqpoint{3.276500in}{0.624576in}}%
\pgfpathlineto{\pgfqpoint{3.301728in}{0.656822in}}%
\pgfpathlineto{\pgfqpoint{3.327097in}{0.692907in}}%
\pgfpathlineto{\pgfqpoint{3.352514in}{0.733000in}}%
\pgfpathlineto{\pgfqpoint{3.377880in}{0.777257in}}%
\pgfpathlineto{\pgfqpoint{3.403086in}{0.825821in}}%
\pgfpathlineto{\pgfqpoint{3.428018in}{0.878820in}}%
\pgfpathlineto{\pgfqpoint{3.452560in}{0.936344in}}%
\pgfpathlineto{\pgfqpoint{3.497281in}{1.043154in}}%
\pgfpathlineto{\pgfqpoint{3.522319in}{1.095782in}}%
\pgfpathlineto{\pgfqpoint{3.527802in}{1.106633in}}%
\pgfpathlineto{\pgfqpoint{3.527802in}{1.106633in}}%
\pgfusepath{stroke}%
\end{pgfscope}%
\begin{pgfscope}%
\pgfsetrectcap%
\pgfsetmiterjoin%
\pgfsetlinewidth{0.803000pt}%
\definecolor{currentstroke}{rgb}{0.000000,0.000000,0.000000}%
\pgfsetstrokecolor{currentstroke}%
\pgfsetdash{}{0pt}%
\pgfpathmoveto{\pgfqpoint{0.538581in}{0.415000in}}%
\pgfpathlineto{\pgfqpoint{0.538581in}{1.600000in}}%
\pgfusepath{stroke}%
\end{pgfscope}%
\begin{pgfscope}%
\pgfsetrectcap%
\pgfsetmiterjoin%
\pgfsetlinewidth{0.803000pt}%
\definecolor{currentstroke}{rgb}{0.000000,0.000000,0.000000}%
\pgfsetstrokecolor{currentstroke}%
\pgfsetdash{}{0pt}%
\pgfpathmoveto{\pgfqpoint{3.699063in}{0.415000in}}%
\pgfpathlineto{\pgfqpoint{3.699063in}{1.600000in}}%
\pgfusepath{stroke}%
\end{pgfscope}%
\begin{pgfscope}%
\pgfsetrectcap%
\pgfsetmiterjoin%
\pgfsetlinewidth{0.803000pt}%
\definecolor{currentstroke}{rgb}{0.000000,0.000000,0.000000}%
\pgfsetstrokecolor{currentstroke}%
\pgfsetdash{}{0pt}%
\pgfpathmoveto{\pgfqpoint{0.538581in}{0.415000in}}%
\pgfpathlineto{\pgfqpoint{3.699063in}{0.415000in}}%
\pgfusepath{stroke}%
\end{pgfscope}%
\begin{pgfscope}%
\pgfsetrectcap%
\pgfsetmiterjoin%
\pgfsetlinewidth{0.803000pt}%
\definecolor{currentstroke}{rgb}{0.000000,0.000000,0.000000}%
\pgfsetstrokecolor{currentstroke}%
\pgfsetdash{}{0pt}%
\pgfpathmoveto{\pgfqpoint{0.538581in}{1.600000in}}%
\pgfpathlineto{\pgfqpoint{3.699063in}{1.600000in}}%
\pgfusepath{stroke}%
\end{pgfscope}%
\end{pgfpicture}%
\makeatother%
\endgroup%

%% file: conclusion.tex
\section{CONCLUSION}
\label{sec_conclusion}

In this paper we have
examined method for ensuring
a system with constrained inputs can
be safe for all future times.
The main result is a general method for constructing
a barrier function given a safety constraint,
system dynamics with actuator limits, and an evasive function specified
by a safety engineer.
We then apply this method to show
how collision avoidance for two UAVs can be ensured
for all future times.
The result is then extended to the case of 
collision avoidance for arbitrarily many UAVs
by considering how to ensure that arbitrarily
many safety objectives can be satisfied simultaneously.
In the case of arbitrarily many UAVs, network constraints
may limit the message throughput so we provide
a reformulation of the algorithm that requires
less message passing while still ensuring 
that vehicles will stay safe.
The final result is demonstrated in a simulation
of 20 UAVs where the vehicles are on a collision course.
However, due to the role of the barrier function in ensuring
safety, all twenty vehicles nevertheless maintain safe distances
from each other and then able to reach their assigned waypoints.

%% file: appendix.tex
\begin{appendix_env}

\appendixSection{An Analysis of The Role of $\delta$ in The Continuous Differentiability of $h_{turn}$}
\label{app_delta_cont_diff}

Note that (\ref{eq_min_h_turn}) is not necessarily differentiable when
$A_2 = 0$ since $A_2$ results from a square root performed in phasor addition.
Thus, in this section, we consider how to ensure $A_2$ is continuously differentiable
to ensure $h$ in (\ref{eq_min_h_turn}) is continuously differentiable.
Consider (\ref{eq_min_h_turn}) in phasor form
\begin{IEEEeqnarray}{rCl}
A_1  -D_s^2+ A_2e^{j\Theta} 
&=& A_1 -D_s^2 +\sigma A_3e^{j(\theta_{1,0} - \pi / 2)}
    + A_3e^{j(\theta_{2,0} + \pi / 2)} + \delta e^{j(\theta_{1,0} - \pi / 2)} \nonumber\\
&&+ \sigma A_4e^{j(\theta_{1,0} - \pi)}
    + A_4e^{j\theta_{2,0}} + \delta e^{j(\theta_{1,0}-\pi)} \nonumber\\
&=& A_1 -D_s^2+ A_5 e^{j\Theta_5} + A_6 e^{j\Theta_6}
\end{IEEEeqnarray}
where $A_3 = 2\Delta b_0 \VW$, $A_4 = 2\Delta c_0 \VW$,
$A_5 e^{j\Theta_5} = 
\sigma A_3e^{j(\theta_{1,0} - \pi / 2)}
    + A_3e^{j(\theta_{2,0} + \pi / 2)}  + \delta e^{j(\theta_{1,0} - \pi / 2)}$,
and
$A_6 e^{j\Theta_6} = 
\sigma A_4e^{j(\theta_{1,0} - \pi)}
    + A_4e^{j\theta_{2,0}} + \delta e^{j(\theta_{1,0} - \pi)}$.
Notice that $\Theta_5 - \Theta_6 = \pi / 2$.
In other words, $A_2$ is zero only when
both $A_5$ and $A_6$ are zero.
For $\delta = 0$, $A_5$ and $A_6$ are both
zero on the set $Z_1\subseteq \mathcal{D}$ where
$\theta_{1,0} = \theta_{2,0}$ or $\theta_{1,0} = \theta_{2,0} + \pi$.
Although $Z_1$ is a zero measure set,
we note that for $\delta > 0$ that
$A_2$ is zero on a set $Z_2 \subset Z_1$
where $Z_2$ is the restriction of $Z_1$
to a specific set of positions which we now specify.

\begin{case}\emph{Vehicles Start in Opposite Directions.}
Suppose $\theta_{1,0} = \theta_{2,0} + \pi$. Then
$A_5 = 0$ when $\delta = -(1 + \sigma)A_3 = -2(1 + \sigma)\Delta b_0 \VW$.
Similarly,
$A_6 = 0$ when $\delta = -(1 + \sigma)A_4 = -2(1 + \sigma)\Delta c_0 \VW$.
Suppose $\delta$ is fixed. 
Then $A_2 = 0$
when
$-\frac{\delta}{2(1+\sigma)\VW} = \Delta b_0 = p_{1,x_0} - p_{2,x_0} + r(1 + \sigma)\sin\theta_{2,0}$
and
$-\frac{\delta}{2(1+\sigma)\VW} = \Delta c_0 = p_{1,y_0} - p_{2,y_0} - r(1 + \sigma)\cos\theta_{2,0}$.
\end{case}

\begin{case}\emph{Vehicles Start in the Same Direction.}
Suppose $\theta_{1,0} = \theta_{2,0}$. Then
$A_5 = 0$ when $\delta = (1 - \sigma)A_3$.
Similarly,
$A_6 = 0$ when $\delta = (1 - \sigma)A_4$.
For $\sigma=1$, let $\delta > 0$ to ensure $A_5$ and $A_6$ are not simultaneously 0.
For $0 < \sigma < 1$, a similar analysis to the previous case
implies $A_2 = 0$ when
when
$-\frac{\delta}{2(1-\sigma)\VW} = \Delta b_0 = p_{1,x_0} + p_{2,x_0} - r(1 - \sigma)\sin\theta_{2,0}$
and
$-\frac{\delta}{2(1+\sigma)\VW} = \Delta c_0 = p_{1,y_0} - p_{2,y_0} - r(1 + \sigma)\cos\theta_{2,0}$.
\end{case}

\appendixSection{An Analysis of the Continuous Differentiability of $h_{straight}$}
\label{app_cont_diff_of_h_straight}

From (\ref{eq_min_h_straight}) we expand terms to get 
\begin{equation}
h(x) = \inf_{\tau\in[0,\infty)} c(x) + b(x)\tau + a(x)\tau^2
\label{eq_h_straight_quadratic_form}
\end{equation}
where $c(x) = \Delta x^2 + \Delta y^2 + \Delta z^2 - D_s^2$,
$b(x) = 2(\Delta x \Delta C + \Delta y \Delta S)$,
$a(x) = \Delta C^2 + \Delta S^2$,
$\Delta x = p_{1,x_0} - p_{2,x_0}$,
$\Delta y = p_{1,y_0} - p_{2,y_0}$,
$\Delta z = p_{1,z_0} - p_{2,z_0}$,
$\Delta C = v_1\cos\theta_1 - v_2\cos\theta_2$,
$\Delta S = v_1\sin\theta_1 - v_2\sin\theta_2$.
We also note that $a(x) > 0$ since
\begin{IEEEeqnarray*}{rCl}
a(x)
&=& (v_1\cos\theta_1 - v_2\cos\theta_2)^2 + (v_1\sin\theta_1 - v_2\sin\theta_2)^2 \\
&=& v_1^2 + v_2^2 - 2v_1v_2\cos(\theta_1 - \theta_2) \\
&=& v_1^2 + v_2^2 - 2v_1v_2 + 2v_1v_2 - 2v_1v_2\cos(\theta_1 - \theta_2) \\
&=& (v_1 - v_2)^2 + 2v_1v_2(1 - \cos(\theta_1 - \theta_2))\\
&>& 0
\end{IEEEeqnarray*}
since $v_1 \ne v_2$ and $v_1$ and $v_2$ are positive.
Then
$\tau_{min}(x) = -b(x) / 2a(x)$ is well defined.
Then $h$ has a minimum at $\tau_{nonneg,min} = \max(0,\tau_{min}(x))$.

For $\tau_{nonneg,min}(x) > 0$,
$h$ is continuously differentiable 
because $c, b,\tau_{min}$, and $a$ are continuously
differentiable. 
Consider now when $\tau_{nonneg,min}(x) = 0$.
We verify that $\pf{h(x)}{x} = \pf{c(x)}{x}$
for either the case of $\tau_{min} = 0$ or $\tau_{min} = -b(x)/2a(x)$.
In the first case,
$h(x) = c(x)$ and $\pf{h(x)}{x} = \pf{c(x)}{x}$.
In the second case, 
$h(x) = c(x) + b(x)\tau_{min} + a(x)\tau_{min}^2$
and 
\begin{IEEEeqnarray*}{rCl}
\pf{h(x)}{x} &=& \pf{c(x)}{x} + \pf{b(x)}{x}\tau_{min}(x) + b(x) \pf{\tau_{min}(x)}{x} + \pf{a(x)}{x}\tau_{min}(x) + 2a(x)\tau_{min}\pf{\tau_{min}(x)}{x} \\
&=& \pf{c(x)}{x}
\end{IEEEeqnarray*}
because in this case $b(x)$ and $\tau_{min}(x)$ are 0.

\appendixSection{Funding Sources}

The work of Eric Squires was supported by the University System of Georgia's Tuition Assistance Program.
The work by Magnus Egerstedt and Pietro Pierpaoli was 
supported by Grant No. ARL DCIST CRA W911NF-17-2-0181 by the US Army Research Lab.
The work of Samuel Coogan and Rohit Konda was 
supported by the Air Force Office of Scientific Research under grant number FA9550-19-1-0015.
\end{appendix_env}